\documentclass{article}

\PassOptionsToPackage{numbers, compress, sort}{natbib}
\usepackage[final]{neurips_2024}

\usepackage[utf8]{inputenc} 
\usepackage[T1]{fontenc}    
\usepackage{hyperref}       
\usepackage{url}            
\usepackage{booktabs}       
\usepackage{amsfonts}       
\usepackage{nicefrac}       
\usepackage{microtype}      
\usepackage{xcolor}         
\usepackage{footnote}
\usepackage{longtable}
\usepackage{fancyvrb}
\usepackage{selectp}
\usepackage[pdftex]{graphicx}
\usepackage[acronym]{glossaries}
\newacronym{nas}{NAS}{neural architecture search}
\newacronym{hpo}{HPO}{hyperparameter optimization}
\usepackage{wrapfig}
\usepackage{xcolor}
\usepackage{amsmath}
\usepackage[bottom]{footmisc}
\usepackage{mathtools}

\usepackage[utf8]{inputenc} 
\usepackage[T1]{fontenc}    
\usepackage{url}            
\usepackage{booktabs}       
\usepackage{amsfonts}       
\usepackage{nicefrac}       
\usepackage{microtype}      
\usepackage{wrapfig}
\usepackage[nameinlink,capitalise,noabbrev]{cleveref} 
\usepackage{listings}
\usepackage{lipsum}                     
\usepackage{xargs}
\usepackage[inline]{enumitem}
\usepackage{subcaption}
\usepackage[colorinlistoftodos,prependcaption,textsize=tiny]{todonotes}

\newcommandx{\unsure}[2][1=]{\todo[linecolor=red,backgroundcolor=red!25,bordercolor=red,#1]{#2}}
\newcommandx{\change}[2][1=]{\todo[linecolor=blue,backgroundcolor=blue!25,bordercolor=blue,#1]{#2}}
\newcommandx{\info}[2][1=]{\todo[linecolor=OliveGreen,backgroundcolor=OliveGreen!25,bordercolor=OliveGreen,#1]{#2}}
\newcommandx{\improvement}[2][1=]{\todo[linecolor=Plum,backgroundcolor=Plum!25,bordercolor=Plum,#1]{#2}}
\newcommandx{\thiswillnotshow}[2][1=]{\todo[disable,#1]{#2}}
\newcommand{\github}{\url{https://github.com/automl/HW-GPT-Bench/}}
\usepackage{url}
\usepackage{microtype}
\usepackage{booktabs}
\usepackage{csquotes}
\usepackage{adjustbox}
\usepackage{pdflscape}

\usepackage{comment}
\usepackage[utf8]{inputenc} 
\usepackage[T1]{fontenc}    
\usepackage{url}            
\usepackage{booktabs}       
\usepackage{amsfonts}       
\usepackage{nicefrac}       
\usepackage{microtype}      
\usepackage{siunitx}

\usepackage{multicol}
\usepackage{mathtools}
\usepackage{caption}
\usepackage{float}
\usepackage{wrapfig}

\usepackage{multirow}
\usepackage{mathrsfs}
\usepackage{newfloat}
\usepackage{relsize}
\usepackage{enumitem}
\usepackage{pifont}
\usepackage{bm}
\usepackage{caption}
\usepackage{floatflt}

\newcommand{\improvementimgnet}[1]{{xx}}
\newcommand{\paramsfactor}[1]{{35}}
\newcommand{\flopsfactor}[1]{{xx}}

\newcommand{\xaxis}[1]{{error}}
\newcommand{\Xaxis}[1]{{Error}}
\hypersetup{
    colorlinks=true,
    linkcolor=blue,
    filecolor=magenta,      
    urlcolor=cyan,
    citecolor=blue,
    pdftitle={HW-Aware-GPT-Bench},
    }

\definecolor{codegreen}{rgb}{0,0.6,0}
\definecolor{codegray}{rgb}{0.5,0.5,0.5}
\definecolor{codepurple}{rgb}{0.58,0,0.82}
\definecolor{backcolour}{HTML}{F2F2F2}

\title{HW-GPT-Bench: Hardware-Aware Architecture Benchmark for Language Models}

\author{Rhea Sanjay Sukthanker$^1$\hspace{4mm} Arber Zela$^1$\hspace{1.5mm}\hspace{4mm} Benedikt Staffler$^{3}$ \vspace{1mm} \\ \textbf{Aaron Klein$^4$\hspace{4mm} Lennart Purucker$^1$\hspace{4mm} Jörg K. H. Franke$^1$\hspace{5mm} Frank Hutter$^{2,1}$} \vspace{2mm} \\
 $^1$University of Freiburg,
 $^2$ELLIS Institute Tübingen,\\
 $^3$Bosch Center for Artificial Intelligence (BCAI), Germany,\\
 $^4$ScaDS.AI, Leipzig,\\
 $\texttt{\{sukthank, zelaa, frankej, fh\}@cs.uni-freiburg.de}$\\
 $\texttt{benediktsebastian.staffler@de.bosch.com}$
}

\begin{document}
\maketitle

\begin{abstract}
The increasing size of language models necessitates a thorough analysis across multiple dimensions to assess trade-offs among crucial hardware metrics such as latency, energy consumption, GPU memory usage, and performance. Identifying optimal model configurations under specific hardware constraints is becoming essential but remains challenging due to the computational load of exhaustive training and evaluation on multiple devices. To address this, we introduce HW-GPT-Bench, a hardware-aware benchmark that utilizes surrogate predictions to approximate various hardware metrics across 13 devices of architectures in the GPT-2 family, with architectures containing up to 1.55B parameters. Our surrogates, via calibrated predictions and reliable uncertainty estimates, faithfully model the heteroscedastic noise inherent in the energy and latency measurements. To estimate perplexity, we employ weight-sharing techniques from Neural Architecture Search (NAS), inheriting pretrained weights from the largest GPT-2 model. Finally, we demonstrate the utility of HW-GPT-Bench by simulating optimization trajectories of various multi-objective optimization algorithms in just a few seconds.
\end{abstract}
\section{Introduction}
\label{sec:introduction}

\emph{Language models (LMs)} based on the transformer architectures~\citep{vaswani-neurips17a} mark the current state-of-the-art \citep{minaee2024large} in most natural language understanding tasks, including text summarization, question-answering and language generation. This has led to a surge in research, with models \citep{shoeybi2019megatron,brown2020language,chowdhery2023palm} and training data \citep{hoffmann2022training,longpre2023flan} growing in size. Consequently, inference costs have also risen significantly, making it often challenging to deploy these models in practice. For instance, ChatGPT utilizes over half a million kilowatt-hours of electricity daily, a consumption sufficient to handle approximately two hundred million requests. This energy usage is comparable to that of around 180,000 U.S. households, each consuming approximately twenty-nine kilowatt-hours \citep{forbesaisustainability}. 

There is a natural trade-off (Pareto frontier) between latency and performance of LLMs. While techniques such as KV-Cache optimization \citep{waddington2013kv} and pruning \citep{wang2019structured,zafrir2021prune} have been used to improve inference efficiency, they do not explicitly balance performance and latency. Hence, discovering the inference-optimal frontier of language models is a multi-objective optimization problem~\citep{deb2016multi}, where we are interested in the Pareto set of architectures, i.e. the set of all dominating architectures that have a lower loss value on at least one objective and no higher loss value on any other objective.

Neural architecture search (NAS) \citep{elsken-jmlr19a} is a powerful framework to derive Pareto-optimal neural network architectures in an automated data-driven way.
However, as pointed out by \citet{wan2023efficient}, training a single language model can require millions of GPU hours, making the use of simple multi-objective NAS strategies, such as NSGA-II~\citep{lu-gecco19a}, that need to train multiple architectures from scratch, impractical. To foster the development of more efficient NAS methods, surrogate~\citep{zela-iclr22a,yan-neurips21a,dou2022ea,Cai2020Once-for-All:} and tabular~\citep{ying-icml19a,tu2022bench,li2021hw,dong-iclr20a} NAS benchmarks have been proposed -- particularly for convolutional networks and image classification tasks. These benchmarks have significantly aided the development of search algorithms to replace manual heuristics. Surrogate benchmarks such as Once-for-all~\citep{Cai2020Once-for-All:} or HAT~\citep{wang2020hat} follow the idea of two-stage weight-sharing based NAS \citep{bender-icml18a}, which trains a single \emph{supernet} subsuming a large number of architectures into a single model, followed by a gradient-free search to select the Pareto optimal \emph{sub-networks}. While some benchmarks focus on natural language understanding tasks, such as machine translation~\citep{wang2020hat} and speech recognition~\citep{mehrotra-iclr21a}, the efficacy of these techniques does not directly transfer to architectures for causal language modelling problems and across various hardware metrics (e.g., FLOPS, latency) and devices (e.g., CPUs, GPUs). Therefore, a hardware-aware benchmark for evaluating multi-objective NAS methods is crucial for advancing the design of inference-optimal LM architectures.
\begin{figure}[t]
\centering
\includegraphics[width=\linewidth]{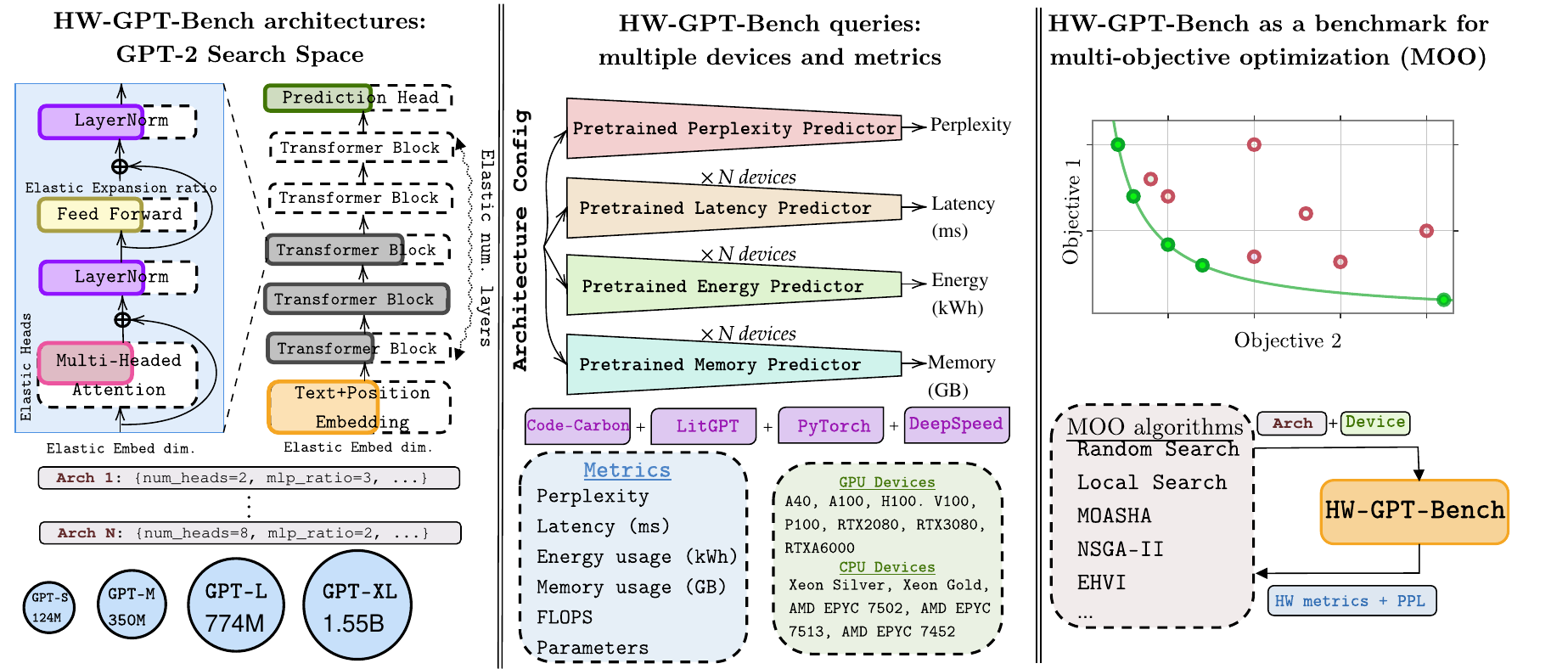}
    \caption{\textbf{HW-GPT-Bench Overview}. Illustration of the search space (\textit{left}), hardware devices and metrics (\textit{middle}) and multi-objective algorithms (\textit{right}) used in the HW-GPT-Bench framework.}
    \label{fig:overview}
\vspace{-4mm}
\end{figure}
In this paper, we introduce \textbf{HW-GPT-Bench} (see Figure~\ref{fig:overview} for an overview), a hardware-aware LM benchmark, based on the GPT-2~\citep{radford2019language} architecture, for multi-objective NAS across 13 devices, 5 hardware metrics, and 4 model scales. Our contributions include:
\begin{itemize}[leftmargin=*]
    \item \textbf{Benchmark Creation} (Section \ref{sec:creationofbench}): Establishing a benchmark across small, medium, and large model scales with surrogate supernets, performance predictors, and hardware metric predictors.
    \item \textbf{Faithful Latency and Energy Estimates} (Section \ref{sec:creationofbench}): Contrary to previous in works in the NAS literature, we use surrogate predictors that provide calibrated predictions and faithfully model the uncertainties inherent in latency and energy profiling. 
    \item \textbf{Metric Interaction Analysis and Algorithm Evaluation} (Sections \ref{sec:analysis} and \ref{sec:experiments}): Studying interaction effects between different hardware and performance metrics, importance of architectural choices and evaluations of various multi-objective optimization algorithms, providing out-of-the-box baselines for future development. 
\end{itemize}
 We provide an open-source API~\footnote{Code at: \github}, for latency and perplexity predictors, the supernetwork weights and different baselines studied, making integration of new methods into this benchmark straightforward. Through HW-GPT-Bench, we aim to accelerate research in hardware-aware NAS for language models, ultimately advancing the development of efficient and high-performing language model architectures.

\section{Related work}
\label{sec:relworkandbg}

\textbf{Structural Pruning, Computational Efficiency, and Neural Architecture Search.}
Network pruning \citep{liang2021pruning}, a model compression technique, reduces model complexity with minimal performance loss. Empirical studies \citep{michel2019sixteen,sajjad2023effect,voita2019analyzing,kim2024shortened} have examined the impact of pruning different layers and modules in pretrained transformers on validation loss. Techniques such as KV-Caching \citep{pope2023efficiently} and Quantization \citep{bai2022towards,ma2024era,tao2022compression,chee2024quip} improve inference time, memory footprint, and energy usage. These methods complement structural pruning and can be incorporated for further speedups. \emph{Structured pruning} removes structured parameter blocks from pretrained models, while \emph{unstructured pruning} introduces sparsity by zeroing out unimportant weights. Adaptive pruning methods \citep{fan2024not,an2024fluctuation}, which prune based on task difficulty, have also been proposed. Recently, \citet{klein2023structural} and \citet{sarah2024llama} used neural architecture search (NAS) \citep{white-arxiv23a} for automated structural pruning of pretrained LLMs. Similarly \citet{munoz2024shears} and \citet{munoz2024lonas} study, starting from pretrained models and parameter-efficient finetuning for NAS.
Training multiple architectures from scratch is computationally expensive \citep{zoph2016neural,real2019regularized,so2019evolved}, so efficient NAS methods employ one-shot models (or supernetworks) \citep{bender-cvpr20a,liu-iclr19a,Cai2020Once-for-All:} as performance proxies by inheriting weights and fine-tuning individual architectures. In our benchmark, we follow this procedure to create perplexity proxies from the largest GPT-2~\citep{radford2019language} model. As a second stage, many multi-objective NAS methods \citep{Cai2020Once-for-All:,wang2020hat} run gradient-free search routines efficiently, such as evolutionary strategies, to optimize performance and hardware metrics. Recently, \citet{sukthanker2024multi} proposed a method that generates the entire Pareto set in a single stage based on preference vectors of objectives and hardware type.

\textbf{(Hardware-aware) NAS Benchmarks.} 
NAS benchmarks, both tabular and surrogate, emerged due to the high computational costs and reproducibility challenges in developing and evaluating NAS algorithms~\citep{ying-icml19a,dong-iclr20a,zela-iclr20b,zela-iclr22a,mehrotra-iclr21a}. \emph{Tabular} benchmarks~\citep{ying-icml19a,dong-iclr20a,duan-cvpr21a,mehrotra-iclr21a} evaluate all architectures in the search space upfront, but this becomes infeasible as search spaces and training times grow. To address this, \emph{surrogate} benchmarks~\citep{zela-iclr22a,yan-neurips21a} use model-based performance predictors, overcoming the limitations of tabular benchmarks and providing more realistic search space evaluations~\citep{zela-iclr22a}. One-shot models~\citep{Cai2020Once-for-All:,chen-iccv21b,wang2020hat,guo-eccv20a} can also act as surrogates by inheriting weights and evaluating performance on validation sets.
Most NAS benchmarks focus on convolutional spaces and computer vision tasks~\citep{dong-iclr20a,duan-cvpr21a,ying-icml19a,mehta-iclr22a}, with some targeting natural language processing (NLP)~\citep{wang2020hat,klyuchnikov2022bench} and speech recognition\citep{mehrotra-iclr21a}. Given the rising computational costs of training, deploying, and searching models via multi-objective NAS, extended tabular NAS benchmarks now include hardware-specific metrics like FLOPS, on-device latency, and energy consumption~\citep{wang2020hat,lee2021hardware,li2021hw,dou2022ea,bakhtiarifard2024ec}. Unlike these benchmarks, HW-GPT-Bench focuses on language modeling with decoder-only transformers~\citep{vaswani-neurips17a,radford2019language}. Additionally, our surrogates offer calibrated predictions by modeling the intrinsic heteroscedastic noise in latency and energy usage, rather than relying on single measurements~\citep{li2021hw,lee2021hardware,dou2022ea}.

\section{HW-GPT-Bench Design Choices}
\label{sec:creationofbench}

In this section we provide details on design choices , such as the architecture search space, data collection procedure, performance and hardware metrics, as well as the surrogate model types.


\subsection{Architecture Search Space}

To construct our architecture search space, we pick the GPT-2 \citep{radford2019language} language model, which is an autoregressive decoder-only transformer~\citep{vaswani-neurips17a} composed by three primary components (see Figure~\ref{fig:overview}, \textit{left}): \begin{enumerate*}[label=(\roman*)]
\item \textbf{Embedding layers} that map input tokens to learnable representations and encode their position;
\item \textbf{Transformer blocks} stacked multiple times;
\item a \textbf{Prediction head} that predicts the next token for a given sequence.
\end{enumerate*} 
Moreover, each of the transformer blocks consist of: \begin{enumerate*}[label=(\alph*)]\item a \textbf{Causal Self-Attention Block} that weights the significance of different input tokens \item a \textbf{MLP block} containing two layers that project the input to a higher dimension and back to a lower one. We denote the ratio of the higher projection dimension to the transformer dimension as MLP ratio.\end{enumerate*} 
In addition, we apply the following enhancements to the original architecture:
\begin{itemize}[leftmargin=*]
    \item \textbf{Rotary positional embeddings (RoPE)} \citep{su2024roformer}: A form of position embedding that captures absolute positional details using rotation matrices while seamlessly integrating explicit relative positional relationships into the self-attention mechanism. Importantly, RoPE offers several advantages, including adaptability to sequences of varying lengths, diminishing token interactions over greater relative distances, and the ability to enhance linear self-attention via relative positional encoding.
    \item \textbf{Parallel residual}: Following PaLM \citep{chowdhery2023palm,wang2021gpt}, in contrast to the standard serialized formulation, we use a parallel formulation in each transformer block. Specifically, if $x$ is the input to the block, the standard and parallel formulations can be written as:
    \begin{align*}
        y_{serialized} = x + \mathtt{MLP}(\mathtt{LayerNorm}(x + \mathtt{Attention}(\mathtt{LayerNorm}(x)))
        \end{align*}
        \vspace{-5mm}
    \begin{align*}
        y_{parallel} = x + \mathtt{MLP}(\mathtt{LayerNorm}(x)) + \mathtt{Attention}(\mathtt{LayerNorm}(x))
    \end{align*}
    As reported in PaLM \citep{chowdhery2023palm}, the parallel formulation is faster at larger scales as the MLP and attention input matrix multiplications can be fused.
\end{itemize}

\textbf{Architectural choices.}  Consider a search space $\mathcal{S} = \mathcal{D}_e \times \mathcal{D}_l \times \mathcal{D}_h \times \mathcal{D}_b \times \mathcal{D}_m$, obtained by parameterizing the building blocks of the transformer architecture, where $\mathcal{D}_e := \{e_1, e_2, e_3\}$, $\mathcal{D}_l := \{{l_1}, {l_2}, {l_3}\}$, $\mathcal{D}_h := \{{h_1}, {h_2}, {h_3}\}$, $\mathcal{D}_b := \{\mathtt{On}, \mathtt{Off}\}$ and $\mathcal{D}_m := \{m_1, m_2, m_3\}$ correspond to the set of embedding dimension choices, number of layers, number of heads, choice of setting the bias in linear layers and the MLP ratio choices, respectively. Furthermore, we choose the MLP ratio and the number of heads on a per-layer basis, amounting to a search space size of $\sim 10^{36}$ possible architectures. 
We represent architecture configurations as a list of integers $s = \{e, l, h^1, \cdots, h^l, m^1,\cdots,m^l,b \}$, where $e\in \mathcal{D}_e$, $l\in \mathcal{D}_l$, $h^l\in \mathcal{D}_h$, $m^l\in \mathcal{D}_m$ and $b\in \mathcal{D}_b$. $h^l$ and $m^l$ denote the number of heads and MLP ratio of layer $l$, respectively.
Given a set of $m$ metrics (objectives) $\mathcal{Y} = \{y_m \in \mathcal{R}^m: y = f(s), s \in \mathcal{S}\}$ e.g.: latency, perplexity, energy, memory consumption etc., a NAS algorithm searches for the (Pareto) optimal architectures, evaluated using these metrics, from the space $\mathcal{S}$.

\begin{table}[!t]
\centering 
\caption{HW-GPT-Bench search space. We pretrain 7 supernetworks with different sizes and search space strides: GPT-S, -M and -L, -S-wide, -M-wide, -L-wide, -XL-wide. On each of them we parameterize the dimensinality of the embedding layer, number of stacked layers (transformers blocks), number of self-attention heads and MLP ratio for every active layer, as well as if the bias is on or off.}
\vspace{1mm}
\label{tab:search_space}
\resizebox{\linewidth}{!}{
\begin{tabular}{lccccccc}
\toprule
\textbf{Supernet Type} & \textbf{Embedding Dim.}   & \textbf{Layer No.} & \textbf{Head No.} & \textbf{MLP Ratio} & \textbf{Bias}   & \textbf{No. of Archs} & \textbf{Supernet Size}   \\
\midrule
\textbf{GPT-S}        & \texttt{{[}192, 384, 768{]}} & \texttt{{[}10, 11, 12{]}}    & \texttt{{[}4, 8, 12{]}}     & \texttt{{[}2, 3, 4{]}}       & \texttt{{[}On, Off{]}} & \texttt{{$\sim 10^{12}$}} & 124M \\
\textbf{GPT-M}         & \texttt{{[}256, 512, 1024{]}} & \texttt{{[}22, 23, 24{]}}    & \texttt{{[}8, 12, 16{]}}    & \texttt{{[}2, 3, 4{]}}       & \texttt{{[}On, Off{]}} & \texttt{{$\sim 10^{24}$}}  & 350M\\
\textbf{GPT-L}        & \texttt{{[}320, 640, 1280{]}}  & \texttt{{[}34, 35, 36{]}}    & \texttt{{[}8,  16, 20{]}}  & \texttt{{[}2, 3, 4{]}}       & \texttt{{[}On, Off{]}} & \texttt{{$\sim 10^{36}$}}& 774M\\
\textbf{GPT-S-wide}        & \texttt{{[}192, 384, 768{]}} & \texttt{{[}3, 6, 12{]}}    & \texttt{{[}3, 6, 12{]}}     & \texttt{{[}1, 2, 4{]}}       & \texttt{{[}On, Off{]}} & \texttt{{$\sim 10^{12}$}} & 124M \\
\textbf{GPT-M-wide}         & \texttt{{[}256, 512, 1024{]}} & \texttt{{[}6, 12, 24{]}}    & \texttt{{[}4, 8, 16{]}}    & \texttt{{[}1, 2, 4{]}}       & \texttt{{[}On, Off{]}} & \texttt{{$\sim 10^{24}$}}  & 350M\\
\textbf{GPT-L-wide}        & \texttt{{[}320, 640, 1280{]}}  & \texttt{{[}9, 18, 36{]}}    & \texttt{{[}5,  10, 20{]}}  & \texttt{{[}1, 2, 4{]}}       & \texttt{{[}On, Off{]}} & \texttt{{$\sim 10^{36}$}}& 774M\\
\textbf{GPT-XL-wide}        & \texttt{{[}400,800,1600{]}} & \texttt{{[}12,24,48{]}}    & \texttt{{[}6, 12, 25{]}}     & \texttt{{[}1, 2, 4{]}}       & \texttt{{[}On, Off{]}} & \texttt{{$\sim 10^{48}$}} & 1.55B \\
\bottomrule
\end{tabular}
}
\vspace{-3ex}
\end{table}

\textbf{Four Transformer scales.} 
Based on the values we assign to the choices in every architectural block, we can obtain arbitrary number of search spaces. In HW-GPT-Bench, we construct 7 such spaces, namely, GPT-S, GPT-M, GPT-L, GPT-S-wide, GPT-M-wide, GPT-L-wide and GPT-XL-wide as defined in Table \ref{tab:search_space}, with the largest model containing 1.55B parameters. Note that in every search space, the \emph{supernetwork} is the largest possible model, e.g. in GPT-S that would be $s = \{784, 12, 12,\cdots,4,\cdots, \mathtt{On}\}$.

\subsection{Dataset Collection}
\label{subsec:dataset_collection}


\begin{wrapfigure}[14]{r}{0.4\textwidth}
\vspace{-5mm}
    \centering
    \includegraphics[width=0.99\linewidth]{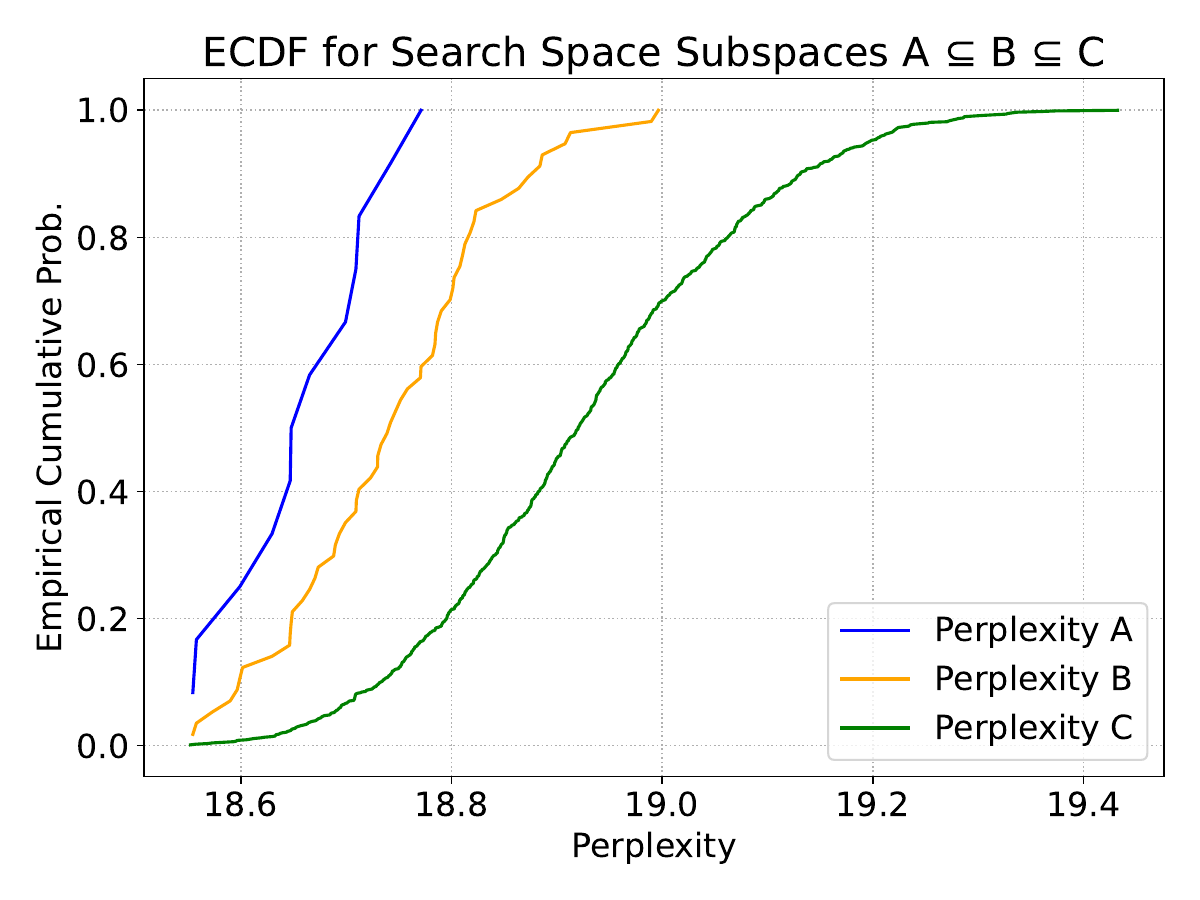}
    \caption{Empirical Cumulative distribution of different search space subspaces.}
    \label{fig:ecdf_ss}
\vspace{-4mm}
\end{wrapfigure}
Building a tabular benchmark for our search spaces with cardinality ranging from $\sim 10^{12}$ to $\sim 10^{48}$ is infeasible even for objectives such as latency or energy usage that are faster to measure than performance.
Therefore, following \citet{zela-iclr22a}, \textbf{we sample 10000 unique architectures uniformly at random} from each of the search spaces (GPT-S, -M, -L, -S-wide, -M-wide, -L-wide, -XL-wide), and use observations from these architectures to train our hardware and performance surrogates. 

\textbf{Performance data.} We evaluate the \textit{perplexity and accuracy} of the sampled architectures at every scale, by inheriting the weights corresponding to a particular architecture from the \emph{supernetwork}, which subsumes all possible architectures in a single network (see Section~\ref{sec:relworkandbg}). Since architectures index the same supernetwork to access their weights, all their individual weights are entangled \citep{Cai2020Once-for-All:,chen-iccv21b,wang2020hat}.
%
Various strategies exist to pretrain the supernetwork, such as random sampling, sandwich scheme, and pre-defined structured sampling \citep{klein2023structural,yu2018slimmable}. Following its effectiveness, as shown by \cite{klein2023structural}, we employ the \emph{sandwich scheme}, that at every mini-batch training iteration samples the largest, the smallest, and two random architectures from the search space. 
Similar to \cite{klein2023structural}, the weights of different sub-networks are tightly coupled with each other and the memory footprint of the supernetwork is the same as the largest network in the space. This allows for extremely efficient training of about $10^{48}$ (for GPT-XL-wide) architectures by updating multiple architectures that share weights simultaneously. We train the supernetwork with the standard language modeling loss function, which is the average negative log likelihood of all tokens without label smoothing. 
We use the OpenWebText \footnote{\url{https://skylion007.github.io/OpenWebTextCorpus/}} dataset, split to train and test sets, for training the supernetwork and evaluating individual architectures' perplexity, respectively. We refer the reader to Appendix~\ref{sec:training_details} for more details on the training pipeline and used hyperparameters. In Figure~\ref{fig:ecdf_ss} we plot the \emph{empirical cumulative distribution} of the 10k architectures evaluated using the pretrained supernetwork weights on the validation set. The green curve represents random sampling in a space of fixed embedding dimension of 1280 in the GPT-L space, the orange curve with the number of layers fixed to 36 as well, and the blue curve
with the average number of heads and MLP ratio across layers greater than 16 and 3, respectively (with embedding dimension fixed to 1280 and number of layers to 36). From the cumulative distribution we can see that there is a lot to gain from searching for the right architectural choices instead of randomly sampling. 

\textbf{Hardware metrics and devices.} 
In addition to perplexity, we also collect the following hardware-related metrics: \begin{enumerate*}[label=(\roman*)] \item \textbf{number of parameters} \item \textbf{FLOPS} (Floating Point Operations)\item on-device \textbf{latency} (in ms) \item on-device \textbf{energy consumption} (in kWh) \item \textbf{memory footprint} (in GB) \end{enumerate*}.
We compute latencies and energies for all 10k sampled architectures on a variety of GPU and CPU types:
\begin{itemize}[leftmargin=*]
    \item[-] \textbf{GPU devices:} NVidia RTX A6000, RTX 2080Ti, RTX 3080Ti, P100, A40, A100 and H100.
    \item[-] \textbf{CPU devices:} Intel Xeon Silver, Xeon Gold, and AMD EPYC 7513, 7452, 7502.
\end{itemize}
For details on hardware specifications refer to Appendix~\ref{sec:hardwares}.
To profile the energy usage, we use CodeCarbon\footnote{\url{https://codecarbon.io/}} on CPU devices and Nvidia's visual profiling tool\footnote{\url{https://docs.nvidia.com/cuda/profiler-users-guide/index.html}} on GPUs.
For latency profiling we use the native PyTorch profiler\footnote{\url{https://pytorch.org/tutorials/recipes/recipes/profiler_recipe.html}} on both GPUs and CPUs. For FLOPs, we use the DeepSpeed library\footnote{\url{https://github.com/microsoft/DeepSpeed}}. 
Furthermore, due to the high intrinsic measurement noise for energy and latencies, to get robust estimates, \emph{we collect up to 10 observations per architecture for latency and up to 50 observations per architecture for energy} on GPUs. We then use these observations to incorporate the \emph{aleatoric} uncertainty into the surrogates (see Section~\ref{subsec:surrogates}) and estimate the noisy latency and energy distributions more reliably. In all evaluations of every metric, we used a batch size of 8, 4, 1 and 1 for GPT-S, -M, -L and -XL scales respectively, and a sequence length of 1024.
\begin{figure}[!t]
\centering
\begin{subfigure}{\textwidth}
    \centering
    \includegraphics[width=\textwidth]{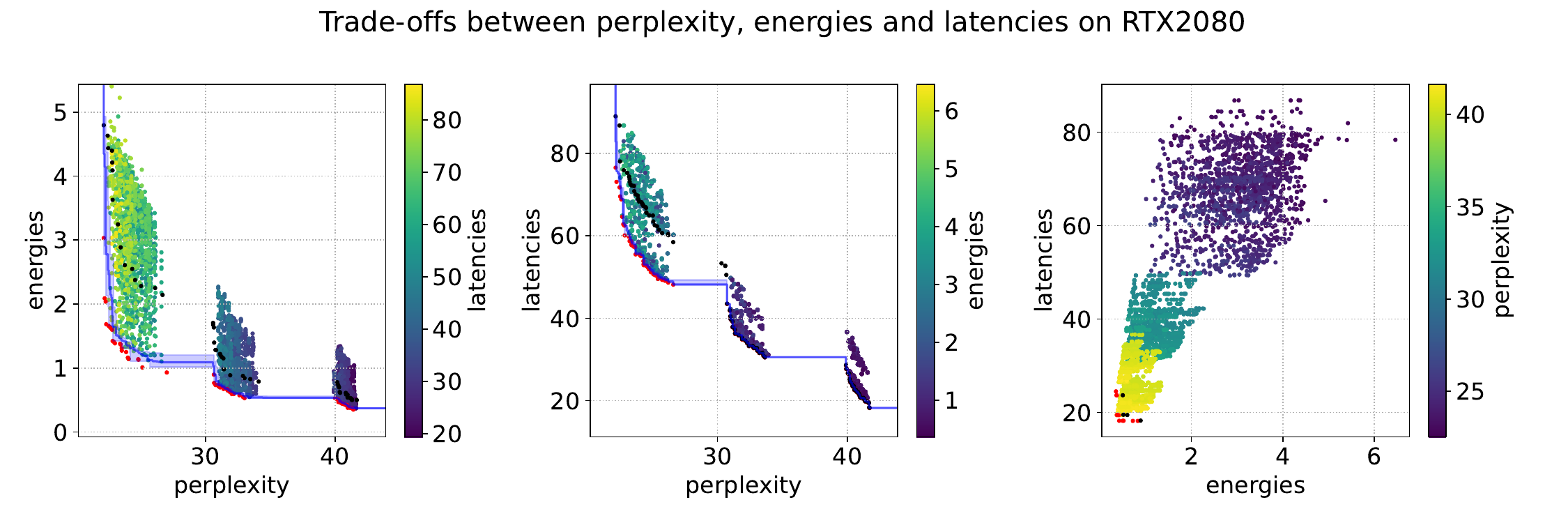}
    \caption{Latency vs. Perplexity vs. Energy for GPT-S on RTX 2080Ti GPU}
    \label{fig:energy_perplexity}
\end{subfigure}

\vspace{1ex} 

\begin{subfigure}{\textwidth}
    \centering
    \includegraphics[width=\textwidth]{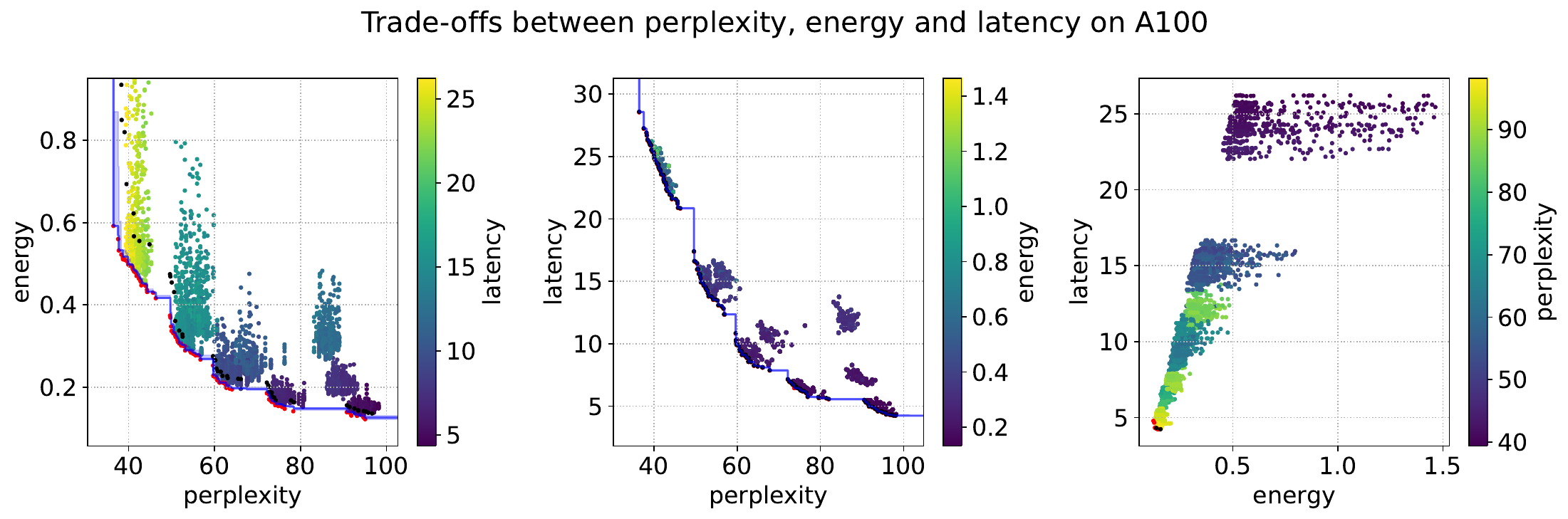}
    \caption{Latency vs. Perplexity vs. Energy for GPT-S-wide on A100 GPU}
    \label{fig:latency_perplexity}
\end{subfigure}

\caption{Trade-offs between Energy, Latency, and Perplexity across architectures for different search spaces. The blue curve represents the Pareto front obtained by randomly sampling an observation, while the best and worst possible Pareto fronts (red and black markers, respectively) are obtained by using the best and worst measured value, respectively, for latencies and energies.}
\label{fig:tradeoffs}
\vspace{-3ex}
\end{figure}

In Figure~\ref{fig:tradeoffs} we show the computed ground-truth perplexity from the trained supernet, latency and energy usage values of all 10000 architectures. We can clearly see the \emph{heteroscedastic noise in the latency and energy measurements}, with an increasing variance as the model perplexity improves. In the same figure, we show the Pareto fronts obtained by randomly sampling an observation (blue line) -- 1 out of 10 for latency and 1 out of 50 for energy -- while the best and worst possible Pareto fronts (red and black markers, respectively) are obtained by using the best and worst measured value, respectively. These results show that the high observation noise in the data can potentially affect the optimization trajectories of multi-objective algorithms, hence resulting in different Pareto fronts.

\subsection{Surrogate Models}
\label{subsec:surrogates}

\textbf{Perplexity and Memory Usage Surrogates.} 
After pretraining the supernetwork, evaluating thousands of architectures on the test set can still be relatively expensive. To this end, similar to \citet{Cai2020Once-for-All:}, we train a MLP surrogate model on 80\% of the collected datapoints, obtained by evaluating the supernetwork, to predict the perplexity given the architecture encoding as input. We also train a MLP surrogate to predict GPU memory usage.
Evaluations on the unseen 2000 architectures, yielded a rank correlation of $> 0.90$ for every metric. Refer to Appendix~\ref{sec:surrogates_ppl} for more details on the MLP architecture and training hyperparameters.


\begin{figure}[t]
\centering
\begin{minipage}{\linewidth}
  \centering
  \includegraphics[width=0.32\textwidth]{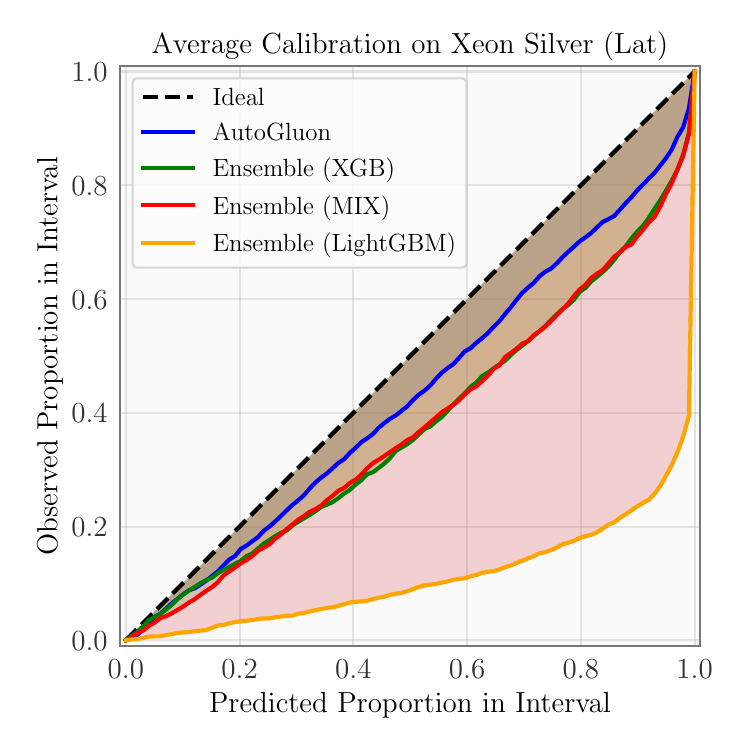}
  \includegraphics[width=0.32\textwidth]{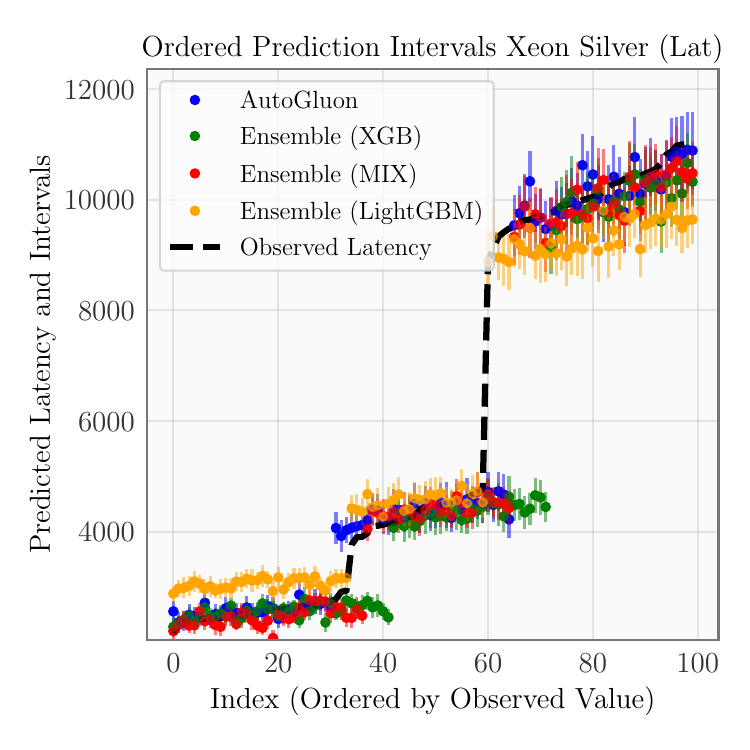}
  \includegraphics[width=0.32\textwidth]{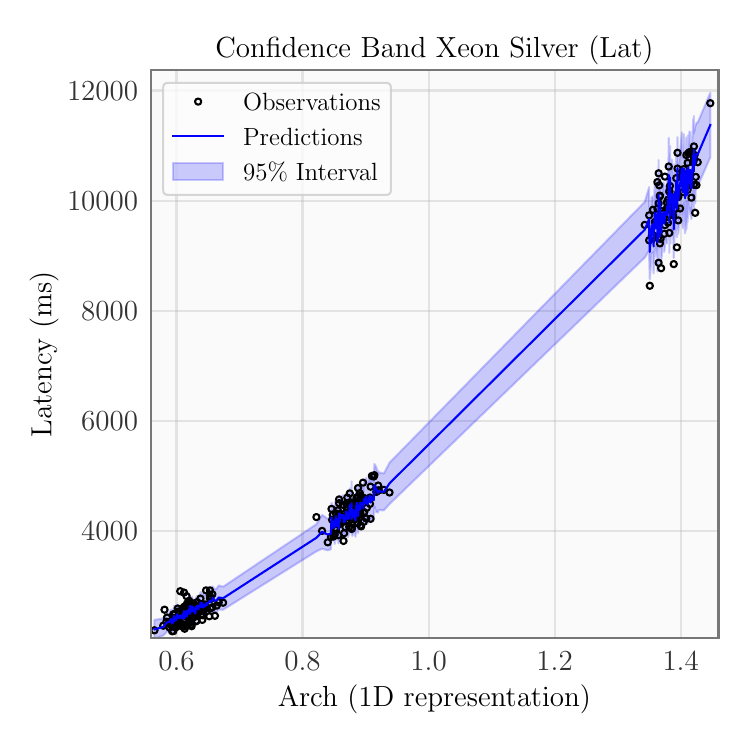}
\end{minipage}
\begin{minipage}{\linewidth}
  \centering
  \includegraphics[width=0.32\textwidth]{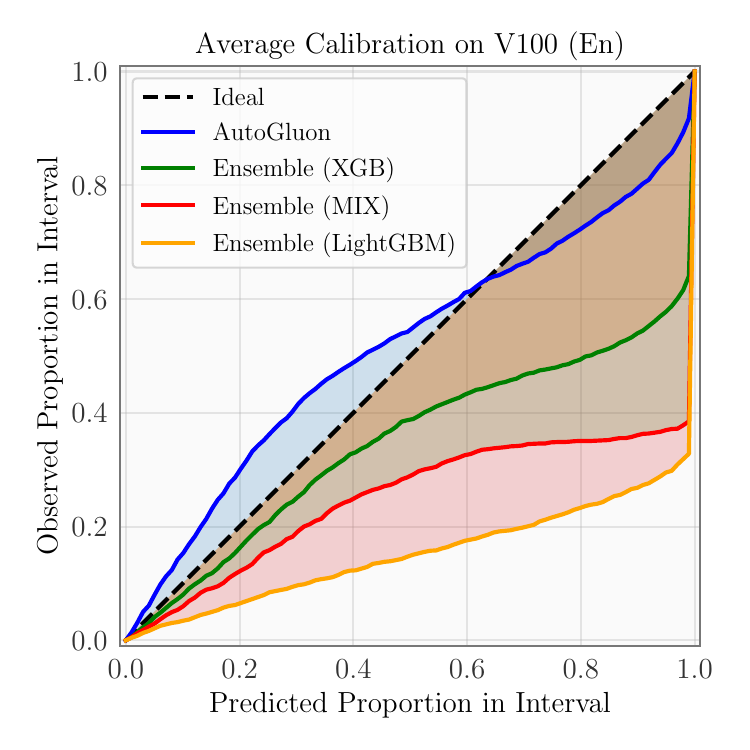}
  \includegraphics[width=0.32\textwidth]{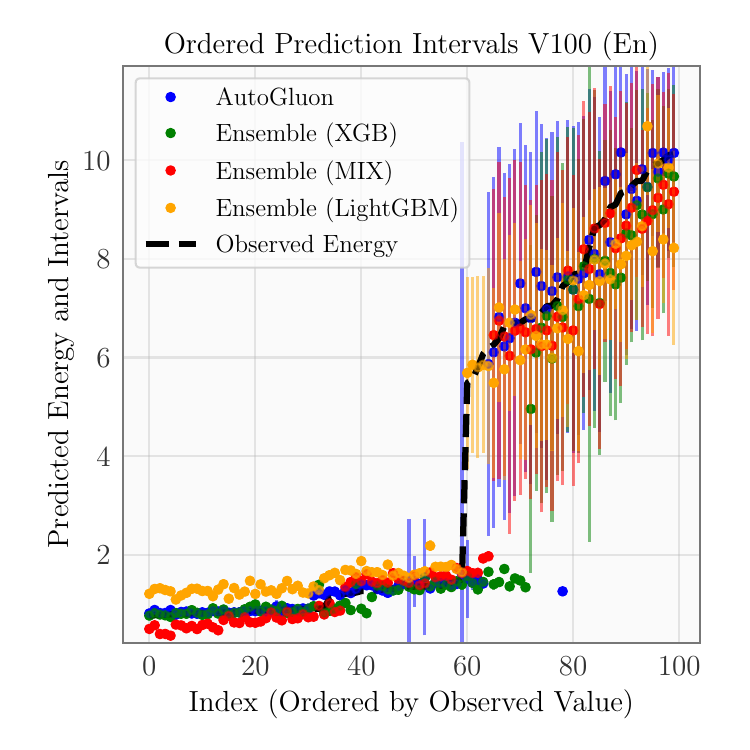}
  \includegraphics[width=0.32\textwidth]{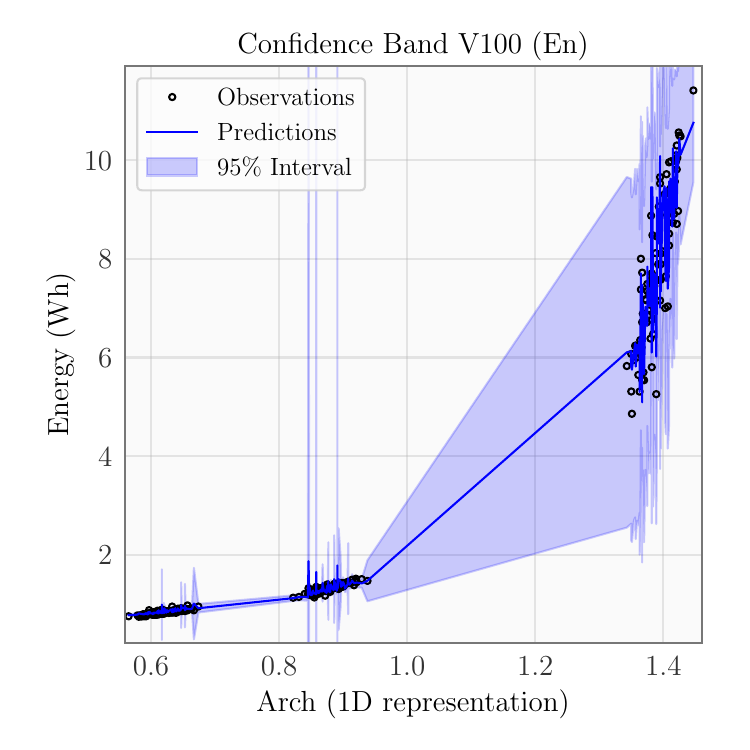}
\end{minipage}
\caption{Calibration area, Prediction Intervals and Confidence Bounds for different surrogate types on Xeon Silver CPU (Latency) and V100 (Energy). The rightmost plots show only the predictions and confidence bands of AutoGluon.}
\label{fig:uct_plots}
\vspace{-3ex}
\end{figure}

\textbf{Energy and Latency Surrogates.} 
From our initial collection of energy and latency observations for different architectures, we observe that on-device latencies and energies tend to be \emph{very noisy}, and the median of observations is insufficient to capture this noisy distribution (see Figure~\ref{fig:tradeoffs}). 
Moreover, we empirically observe that the distribution of energies and latencies is often normally distributed with a few outliers. 
A reliable surrogate model in such a case should not only be performant in terms of accuracy or ranking of the architectures, but on \emph{uncertainty quantification and calibration} as well.
As our surrogate model for predicting per-device latency and energy, we choose \textbf{AutoGluon}~\citep{erickson-arxiv20a}, the state-of-the-art automated machine learning system for tabular data \citep{gijsbers2024amlb} that has been shown to outperform (ensembles of) individual model types \citep{salinas2023tabrepo}. Specifically, we train two AutoGluon models on the 80\% split of the sampled architectures to predict the first and second moments of the latency and energy distributions. AutoGluon builds stacking ensembles~\cite {wolpert1992stacked} to further enhance performance while using a portfolio \citep{feurer-jmlr22a,salinas2023tabrepo} of linear models, neural networks, and decision tree-based models to be robust to outliers and performant across diverse data distributions. 
To analyze different model choices, we compare AutoGluon to LightGBM~\citep{ke-neurips17a} and XGBoost~\citep{chen-kdd16a}, state-of-the-art tabular regression models \citep{grinsztajn2022tree,salinas2023tabrepo,mcelfresh2024neural} as baselines. 
To enhance performance, for both baselines, we ensemble various configurations of LightGBM and XGBoost, and estimate the first and second moments using the individual baselearners' predictions. 
In addition, we also evaluate an ensemble mix of scitkit-learn's~\citep{scikit-learn} Linear Regression, Ridge Regression, and Random Forest \citep{breiman2001random}. We refer the reader to Appendix~\ref{sec:surrogates_lat_en} for more details on the surrogate models.

After fitting AutoGluon and the baselines and computing evaluations on the testing data points, we compute various performance and calibration metrics from the \emph{Uncertainty Toolbox}\footnote{\url{https://github.com/uncertainty-toolbox/uncertainty-toolbox}}~\citep{tran2020methods, chung2020beyond} to quantitatively compare AutoGluon to the other baselines. 
In Table~\ref{tab:surrogatemetrics}, we report accuracy metrics, such as mean absolute error (MAE), Spearman rank correlation, etc., between predicted mean and true mean of observations (e.g. mean of the 50 energy observations per architecture), averaged across architectures in the test set. Furthermore, we also compute various calibration metrics, such as average calibration error.
In summary, from the results in Table~\ref{tab:surrogatemetrics} we can conclude that: \textbf{AutoGluon is the best model that provides both accurate and calibrated predictions, as well as reliable uncertainty estimates for energy usage and latency}. Refer to Appendix~\ref{sec:metrics_app} for more details on these evaluation metrics.

\begin{table}[t]
\centering
\caption{Various performance metrics of surrogates evaluated to predict H100 Latency and RTX2080 Energy. The arrow on the side of the metric name indicates if lower or higher is better.}
\label{tab:surrogatemetrics}
\resizebox{\textwidth}{!}{%
\begin{tabular}{lclclclclclclclclcl} 
\toprule
\multirow{2}{*}{\textbf{Surrogate}} & \multicolumn{12}{c}{\textbf{Accuracy}}                                                                                                                                                                                                                                                                                                   & \multicolumn{6}{c}{\textbf{Calibration}}                                                                                                                      \\ 
\cmidrule(lr){2-13}\cmidrule(lr){14-19}
                                    & \multicolumn{2}{c}{\textbf{MAE} $\downarrow$}                   & \multicolumn{2}{c}{\textbf{RMSE} $\downarrow$}                               & \multicolumn{2}{c}{\textbf{MDAE} $\downarrow$}                  & \multicolumn{2}{c}{\textbf{MARPD} $\downarrow$}                 & \multicolumn{2}{c}{\textbf{R\textsuperscript{2}} $\uparrow$}  & \multicolumn{2}{c}{\textbf{Corr.} $\uparrow$}                  & \multicolumn{2}{c}{\textbf{RMS Cal.} $\downarrow$}               & \multicolumn{2}{c}{\textbf{MA Cal.} $\downarrow$}                & \multicolumn{2}{c}{\textbf{Miscal. Area} $\downarrow$}            \\ 
\midrule
\multicolumn{1}{c}{}                & \textbf{H100}  & \multicolumn{1}{c}{\textbf{2080}} & \textbf{H100}               & \multicolumn{1}{c}{\textbf{2080}} & \textbf{H100}  & \multicolumn{1}{c}{\textbf{2080}} & \textbf{H100}  & \multicolumn{1}{c}{\textbf{2080}} & \textbf{H100}  & \multicolumn{1}{c}{\textbf{2080}} & \textbf{H100}  & \multicolumn{1}{c}{\textbf{2080}} & \textbf{H100}  & \multicolumn{1}{c}{\textbf{2080}} & \textbf{H100}  & \multicolumn{1}{c}{\textbf{2080}} & \textbf{H100}  & \multicolumn{1}{c}{\textbf{2080}}  \\ 
\cmidrule(lr){2-3} \cmidrule(lr){4-5}\cmidrule(lr){6-7} \cmidrule(lr){8-9}\cmidrule(lr){10-11} \cmidrule(lr){12-13}\cmidrule(lr){14-15} \cmidrule(lr){16-17}\cmidrule(lr){18-19}
\textbf{AutoGluon}                  & \textbf{0.153} & \textbf{1.80}                    & \textbf{0.211}              & \textbf{2.576}                    & \textbf{0.111} & 1.121                             & \textbf{0.153} & \textbf{10.677}                   & \textbf{0.999} & \textbf{0.904}                    & \textbf{0.999} & \textbf{0.950}                    & \textbf{0.223} & \textbf{0.244}                    & \textbf{0.198} & \textbf{0.217}                    & \textbf{0.199} & \textbf{0.220}                     \\
\textbf{Ensemble (Mix)}              & 0.569          & 1.830                             & 0.785                       & 2.621                             & 0.413          & \textbf{1.092}                    & 0.565          & 10.920                            & 0.999          & 0.900                             & 0.999          & 0.949                             & 0.472          & 0.298                             & 0.411          & 0.264                             & 0.415          & 0.267                              \\
\textbf{Ensemble (XGB)}              & 0.620          & 1.832                             & 0.827 & 2.628                             & 0.475          & 1.154                             & 0.629          & 10.919                            & 0.990          & 0.899                             & 0.990          & 0.948                             & 0.481          & 0.286                             & 0.417          & 0.251                             & 0.421          & 0.254                              \\
\textbf{Ensemble (LGB)}         & 0.361          & 2.094                             & 0.411                       & 2.922                             & 0.379          & 1.415                             & 0.384          & 13.140                            & 0.970          & 0.875                             & 0.999          & 0.947                             & 0.559          & 0.347                             & 0.481          & 0.304                             & 0.486          & 0.308                              \\
\bottomrule
\end{tabular}
}
\vspace{-3ex}
\end{table}
Figure~\ref{fig:uct_plots} shows the \emph{average calibration plot} (left), \emph{prediction interval plot} (middle) and \emph{predicted mean and 95\% confidence interval} (right) on two devices and two metrics. We can see from the rightmost plot that AutoGluon reliably predicts the mean of the observations and has calibrated uncertainty estimates (left plot: calibrated models have calibration curves that approach the ideal diagonal line). Following \citet{tran2020methods}, we utilized the standard deviation of predictions from each surrogate model to generate Gaussian random variables for each test point. We then evaluated the residuals' adherence to their respective Gaussian random variables. Consequently, models considered "well-calibrated" had residuals that formed Gaussian distributions with standard deviations closely matching the model's predicted standard deviations. For more details on these plots see Appendix~\ref{sec:metrics_app}.

\section{Analysis and Interpretability on HW-GPT-Bench}
\label{sec:analysis}

\paragraph{Correlations between hardware metrics.}
We now use our collected data to examine the relationships between performance and hardware metrics 
in order to gain insights on how these different metrics are correlated with each other and across devices.
Interestingly, we observe that at smaller scales (Figure \ref{fig:correlation_scales_s} in the Appendix), the energy and latency are highly correlated (assesed via the Kendall-$\tau$ correlation coefficient between the ground truth metrics across devices) with FLOPS, often used as a proxy for device-agnostic latency measurements. However, the correlation coefficient become progressively lower at larger model scales as shown in Figures \ref{fig:correlation_scales_m} and \ref{fig:correlation_scales_l} of the Appendix, mainly due to higher energy and latency measurement noise. 
As expected, per-device latency and energy are highly correlated, whilst latency/energy and perplexity are anti-correlated (see Figure~\ref{fig:tradeoffs}). 
\paragraph{Importance of Architectural Choices.}
\begin{figure}[t]
    \centering
        \includegraphics[width=0.32\textwidth]{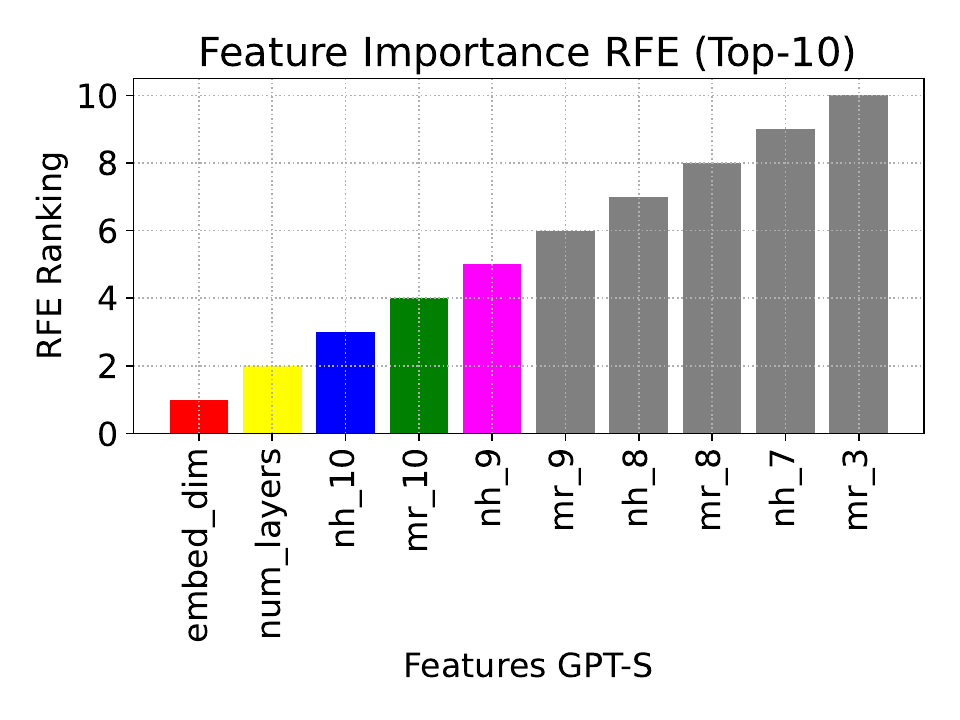}
        \includegraphics[width=0.32\textwidth]{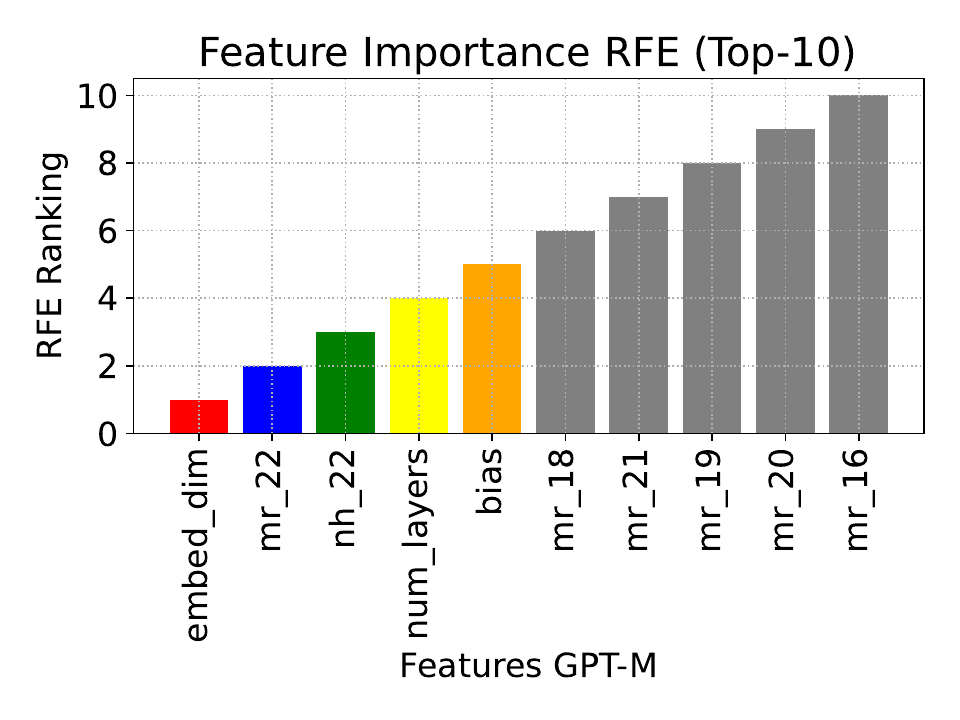}
        \includegraphics[width=0.32\textwidth]{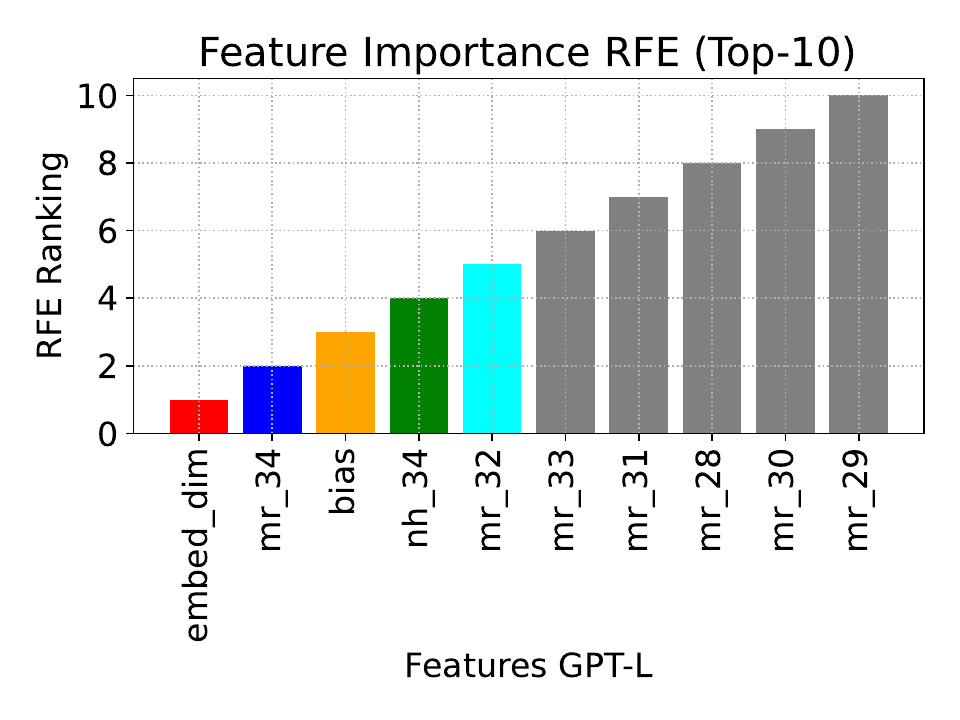}
    \caption{Feature ranking of architecture dimensions at different scales (lower rank is better). The embedding dimension (in red) is most important across scales and the number of layers (in yellow) is more important at smaller scales. MLP ratio (mr) and Number of heads (nh) at layer $N-1$ is important across different scales (depicted in blue and green).}
    \label{fig:importance}
\end{figure}

For the collected perplexity data at each scale, to model the function of perplexity dependent on the architectural choices, we assume a power law of the form, for analysis purposes:
\begin{equation*}
    y = \mathcal{C} \times l^{\alpha} \times e^{\beta} \times \left( \frac{\sum_{i=1}^l h^i}{l} \right)^{\gamma} \times \left( \frac{\sum_{i=1}^l m^i}{l} \right)^{\delta} \times (b+1)^\sigma,
\end{equation*}
where $\mathcal{C},\alpha, \beta, \gamma, \delta, \sigma$ are data-dependent constants, $l$ is the number of layers, $e$ is the embedding dimension, $m^i$ and $h^i$ are the MLP ratio and number of heads on layer $i$, respectively, and b is the bias.
After fitting the power law on the collected 10000 pairs (architecture, perplexity), we obtain the following estimated coefficients: 
\begin{align*}
\text{GPT-S:} & \quad y = 646.234 \cdot l^{-0.226} \cdot e^{-0.371} \cdot m^{-0.100} \cdot h^{-0.076} \cdot b^{-0.001}, \\
\text{GPT-M:} & \quad y = 404.456 \cdot l^{-0.104} \cdot e^{-0.343} \cdot m^{-0.091} \cdot h^{-0.049} \cdot b^{-0.005}, \\
\text{GPT-L:} & \quad y = 280.757 \cdot l^{-0.073} \cdot e^{-0.309} \cdot m^{-0.088} \cdot h^{-0.051} \cdot b^{-0.005}. \\
\end{align*}

\begin{figure}[b]
\centering
\begin{minipage}{.24\linewidth}
  \centering
\includegraphics[width=\linewidth]{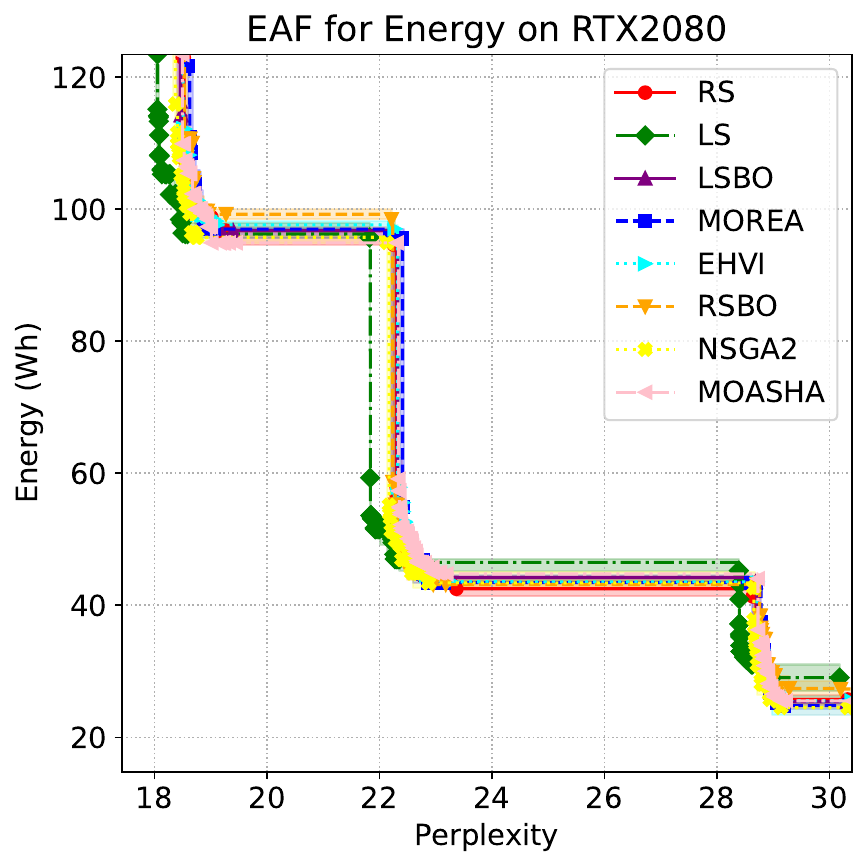}
\end{minipage}
\begin{minipage}{.24\linewidth}
  \centering
\includegraphics[width=\linewidth]{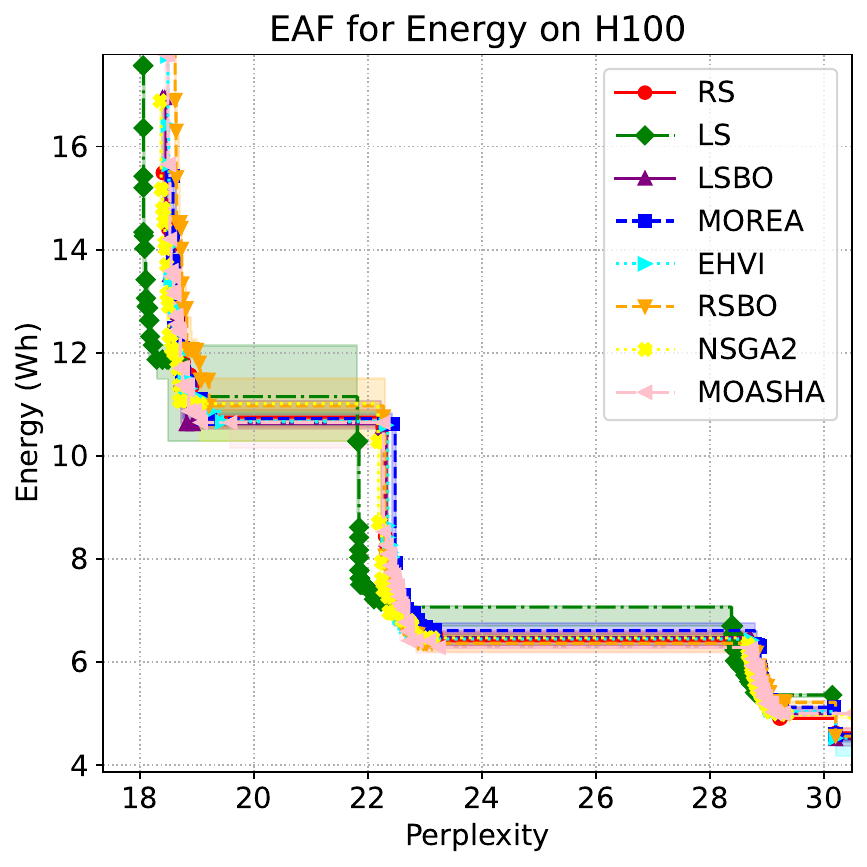}
\end{minipage}
\begin{minipage}{.24\linewidth}
  \centering
\includegraphics[width=\linewidth]{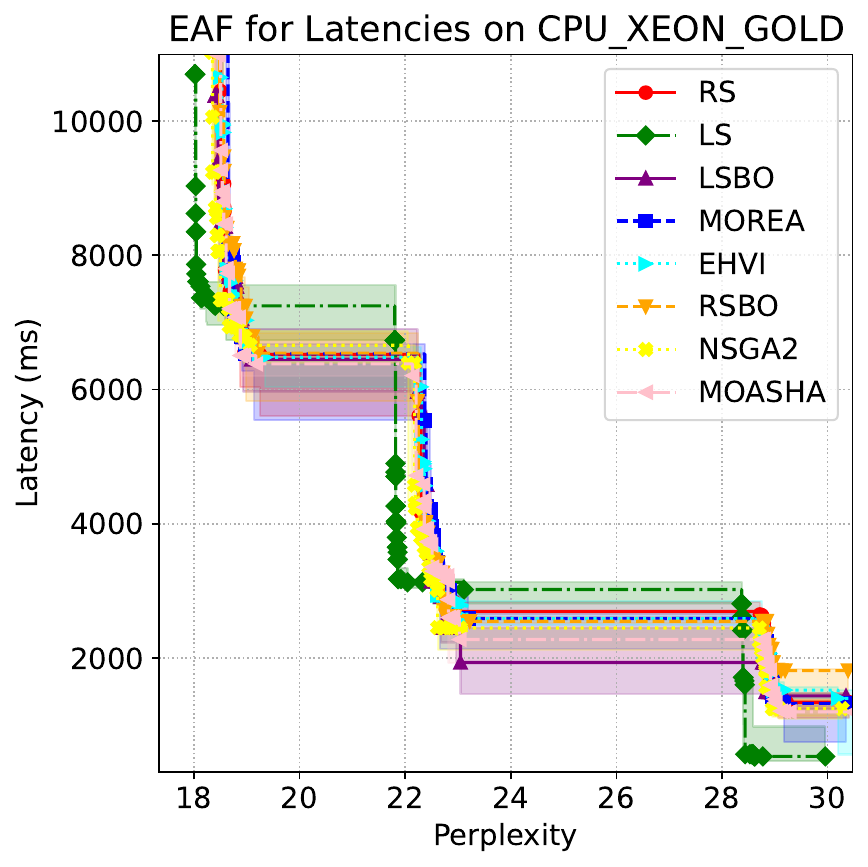}
\end{minipage}
\begin{minipage}{.24\linewidth}
  \centering
\includegraphics[width=\linewidth]{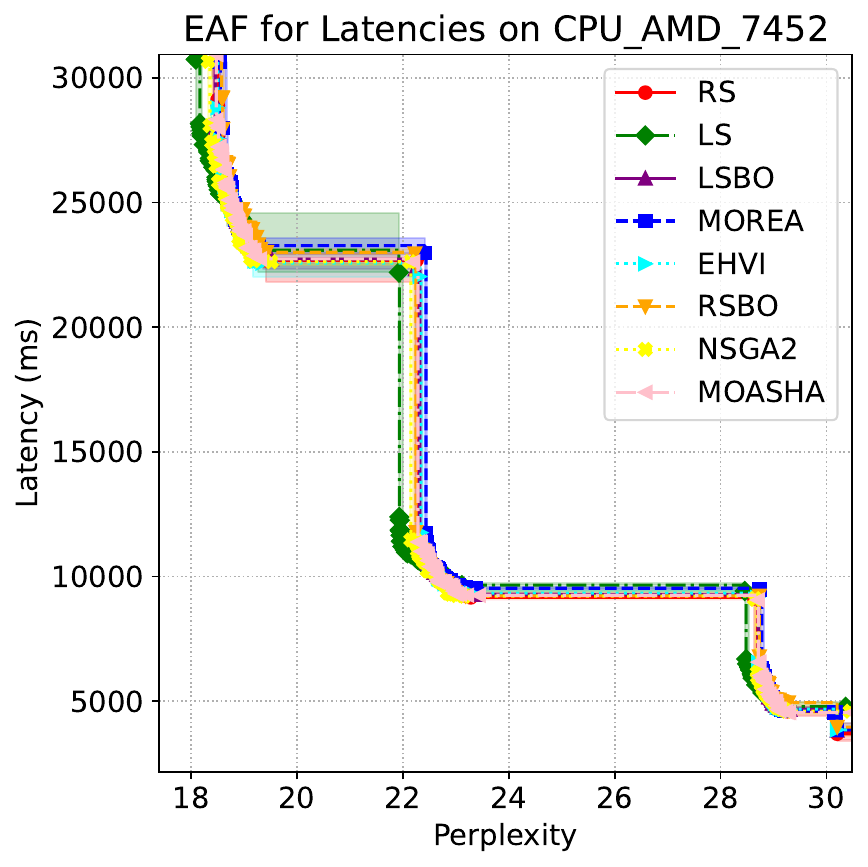}
\end{minipage}

\caption{Different multi-objective NAS baselines on RTX2080Ti and H100 energy, Xeon Gold CPU Latency and CPU AMD 7452 Latency for GPT-L}
\label{fig:baselines}
\vspace{-4mm}
\end{figure}

An ordinary-least-squares (OLS) fit on the log-transformed data indicates that all search dimensions are statistically significant with p-value $< 0.001$, with embedding dimension being the most important architecture parameter in the search space. The number of layers $l$ and MLP ratio become increasingly less important at larger scales and could potentially be pruned without significantly impacting perplexity.
The importance of bias stays more or less constant across scales, while the MLP ratio is more important than the number of heads, indicating that a significant number of heads are possibly redundant and are amenable to pruning. 
In Figure \ref{fig:importance}, we study the ranking upon applying Recursive Feature Elimination (RFE) \citep{chen2007enhanced} to the architecture features and present the 10 top-ranked features.
We observe that embedding dimension and number of layers are important across different model scales. Furthermore, the MLP ratio and the number of heads chosen in the transformer's later layers (e.g., layer $N-1$ and layer $N-2$) are more important than their choices in earlier layers. We present additional results in Appendix~\ref{sec:importance_app}.


\section{HW-GPT as a Benchmark for Multi-objective Optimization}
\label{sec:experiments}

\begin{wrapfigure}[18]{R}{.45\textwidth}
    \vspace{-4ex}
    \centering
    \includegraphics[width=.97\linewidth]{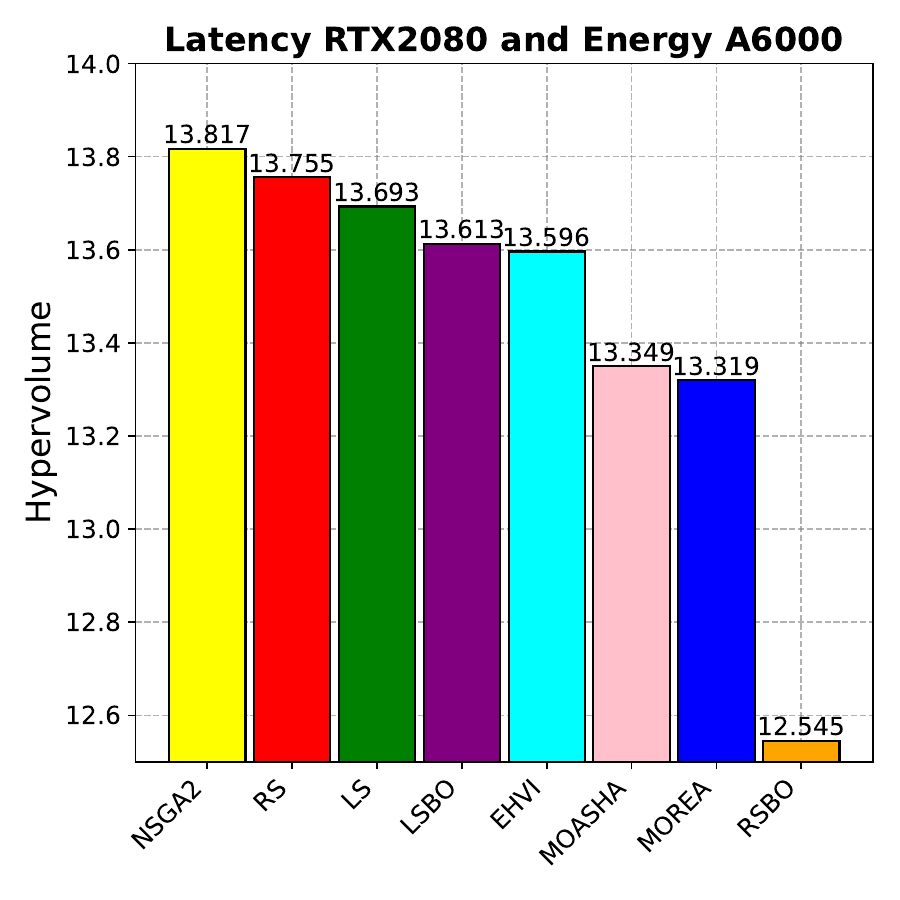}
    \vspace{-2ex}
    \caption{HV of MO-NAS methods on 3 objectives (on GPT-S), namely, perplexity, RTX 2080Ti latency and A6000 energy usage.}
    \label{fig:3d_hv_barplot}
\end{wrapfigure}
In this section, we showcase how HW-GPT-Bench can be used as a benchmark for evaluating multi-objective NAS algorithms. 

\textbf{Multi-objective NAS algorithms.}
State-of-the-art multi-objective NAS (MO-NAS) methods aim to identify Pareto-optimal configurations that balance performance and efficiency.
HW-GPT-Bench, enables simulations of their optimization trajectories \emph{in just a few minutes} using the predictions from our surrogate models. We simulate multiple runs of the following MO-NAS methods implemented in Syne Tune \citep{salinas2022syne}: \begin{enumerate*}[label=(\roman*)]\item Random Search (RS) \item Multi-objective Regularized Evolution (MOREA)~\citep{real2019regularized} \item Non-dominated Sorting Genetic Algorithm II (NSGA-II)~\citep{deb2002fast} \item Local Search (LS) \item Multi-objective Asynchronous Successive Halving (MOASHA)~\citep{schmucker-arxiv21a} \item Bayesian Optimization with Random Scalarizations (RSBO) and \item Linear Scalarizations (LSBO)~\citep{paria2020flexible} \item Expected Hypervolume Improvement (EHVI)~\citep{daulton2020differentiable}\end{enumerate*}. Refer to Appendix~\ref{sec:algos_full} for a more detailed description of them.

\begin{wrapfigure}[18]{R}{.35\textwidth}
    \centering
    \vspace{-3ex}
    \includegraphics[width=.99\linewidth]{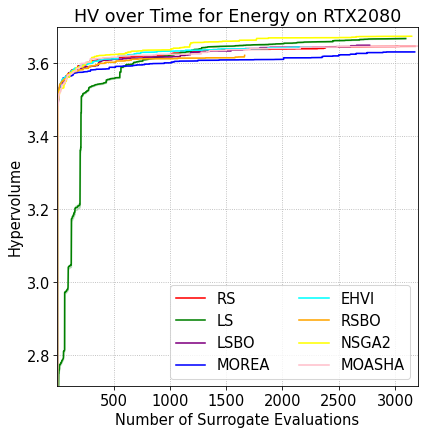}
    \vspace{-2ex}
    \caption{HV of blackbox optimizers over 200 surrogate evaluations on HW-GPT-Bench for RTX2080 Energy. }
    \label{fig:hv_over_time}
\end{wrapfigure}
\subsection{Experiments with 2 objectives}
We run all MO-NAS methods for a fixed (simulated) time budget of 16 CPU hours. We repeat each run 4 times to account for the noise in the latency and energy predictions. In Figure \ref{fig:hv_over_time} shows the hypervolume indicator over number of surrogate evaluations. Notably, EHVI and NSGA-II achieve a higher hypervolume under smaller budgets compared to Local Search (LS) and Random Search (RS), underscoring the efficiency of model-based optimization algorithms in navigating the search space and identify Pareto-optimal architectures.
To aggregate the resulting Pareto fronts from our multiple runs, we use the Empirical Attainment Function (EAF)~\citep{eaf}, which represents a coherent way to capture the uncertainty in the multi-objective metric space (see Appendix~\ref{sec:eaf_app} for details). We show the results of all methods in Figure \ref{fig:baselines}. We can see that LS and NSGA-II generally yield more favorable trade-offs between the objectives, achieving lower perplexity for a given energy, latency or memory consumption. In contrast, Random Search (RS) shows wider variability, with worst case scenarios containing solutions in the Pareto set with relatively high energy usage and low perplexity. 
Appendix~\ref{sec:2obj_app} contains the rest of the experimental results across devices and metrics.

\subsection{Experiments with more than 2 objectives}
\label{subsec:three_obj_exp}

All the methods we run on HW-GPT-Bench on 2 objectives in the section above, are extendable to more than 2 objectives. HW-GPT-Bench enables optimizing for different hardware metrics, not necessarily measured on the same hardware device. Here, we showcase this by optimizing simultaneously for perplexity, latency on RTX 2080Ti and energy usage on the A6000 GPU. We picked these objectives due to their relatively low correlation, as we can see in Figure \ref{fig:correlation_scales_s}. We run the baselines for 16 hrs on a single CPU and compute the hypervolume~\footnote{computed with respect to the empirical nadir point for the 3 objectives}, which we show in Figure~\ref{fig:3d_hv_barplot} (see Appendix~\ref{sec:3obj_app} for the other results.). Notably, NSGA-II achieves good trade-offs also in the 3-objective case, however, the Bayesian optimization methods perform worse than RS and LS here.

With this set of baselines and evaluations, we hope that HW-GPT-Bench will serve as a de-facto testbed for prototyping and benchmarking future hardware-aware optimization methods.


\begin{figure}[t]
\centering
\begin{lstlisting}[caption={A minimal example of the HW-GPT-Bench API.},label=fig:one-line,language=Python,numbers=none,basicstyle=\ttfamily\tiny]
from hwgpt.api import HWGPT 
api = HWGPT(search_space="s") # initialize API
random_arch = api.sample_arch() # sample random arch
api.set_arch(random_arch) # set  arch
results = api.query() # query all
energy = api.query(metric="energy") # query energy
rtx2080 = api.query(device="rtx2080") # query device
# query perplexity based on mlp predictor
perplexity_mlp = api.query(metric="perplexity",predictor="mlp")
# query perplexity based on supernet
perplexity_supernet = api.query(metric="perplexity",predictor="supernet")
# run baseline and plot EAF
nsga2_results = api.run_baseline(method="nsga2", device="rtx2080", metrics=["energy","perplexity"],ppl_predictor="mlp") 
# plot Pareto-front
api.plot_eaf(nsga2_results)
\end{lstlisting}
\end{figure}

\subsection{HW-GPT-Bench API}
We provide a minimal API for users, that enables loading the benchmark and querying all or specific hardware and performance metrics across devices, and different search spaces with just a few lines of code.

Users can also select which perplexity surrogate to use -- either the supernetwork itself or the MLP performance predictor -- ensuring flexibility in performance assessment. Furthermore, the API supports the execution of multi-objective baselines, enabling rigorous benchmarking and comparative analysis. See code snippet~\ref{fig:one-line} for a simple example. Additional examples can be found in Appendix \ref{sec:moreexamples}.


\section{Conclusions, Broader Impact and Implications}
\label{sec:conclusion}
We introduce HW-GPT-Bench, a hardware-aware surrogate benchmark for evaluating language models across various \textbf{hardware devices}, \textbf{metrics}, and \textbf{scales} on a single CPU \emph{in just a few seconds}. By enabling efficient exploration of multi-objective NAS algorithms to achieve Pareto-optimal configurations across multiple hardware metrics and devices, our work has several broader societal implications:
\begin{enumerate*}[label=(\roman*)]
\item \textbf{Energy Efficiency and Environmental Impact} - It promotes the development of energy-efficient language models, mitigating the environmental cost of large-scale AI systems and enhancing sustainability;
\item \textbf{Enhanced Research and Development} - It accelerates research in NAS and structural pruning, leading to more energy-efficient architectures;
\item \textbf{Accessibility and Democratization of AI} - Resource-efficient language models enable innovation for users and organizations with limited resources;
\item \textbf{Economic Benefits} - Optimizing for hardware efficiency reduces training and deployment costs, benefiting industries reliant on extensive language model querying and improving user experience.
\end{enumerate*}

Overall, HW-GPT-Bench addresses critical challenges in developing and deploying algorithms that enhance the resource efficiency of language models, providing a more sustainable, accessible, and reliable benchmarking framework. It underscores the importance of considering hardware efficiency constraints alongside performance in advancing language models.

\newpage
\section*{Acknowledgments}
This research was partially supported by the following sources:
TAILOR, a project funded by EU Horizon 2020 research and innovation programme under GA No
952215; the Deutsche Forschungsgemeinschaft (DFG, German Research Foundation) under grant number 417962828 and 499552394, SFB 1597 (SmallData);
the European Research Council (ERC) Consolidator Grant “Deep Learning
2.0” (grant no. 101045765). Robert Bosch GmbH is acknowledged for financial support. The authors acknowledge support from ELLIS and ELIZA. Funded by
the European Union. Views and opinions expressed are however those of the author(s) only and do
not necessarily reflect those of the European Union or the ERC. Neither the European Union nor the
ERC can be held responsible for them. Aaron Klein acknowledges the financial support by the Federal Ministry of Education and Research of Germany and by Sächsische Staatsministerium für Wissenschaft, Kultur und Tourismus in the programme Center of Excellence for AI-research „Center for Scalable Data Analytics and Artificial Intelligence Dresden/Leipzig", project identification number: ScaDS.AI. Frank Hutter acknowledges the financial support of the Hector Foundation.
\begin{center}\includegraphics[width=0.3\textwidth]{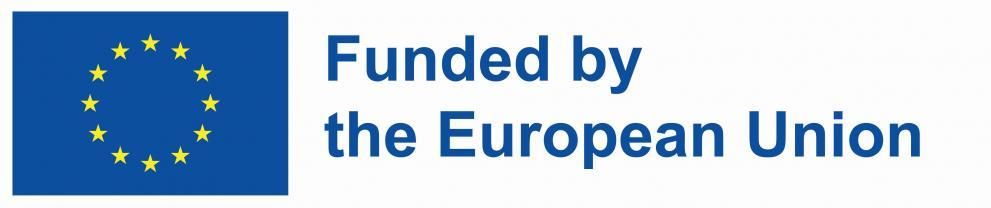}\end{center}. 


\bibliography{lib,proc,shortproc,strings,shortstrings}

\begin{thebibliography}{89}
\providecommand{\natexlab}[1]{#1}
\providecommand{\url}[1]{\texttt{#1}}
\expandafter\ifx\csname urlstyle\endcsname\relax
  \providecommand{\doi}[1]{doi: #1}\else
  \providecommand{\doi}{doi: \begingroup \urlstyle{rm}\Url}\fi

\bibitem[icl(2020)]{iclr20}
\emph{Proceedings of the International Conference on Learning Representations ({ICLR}'20)}, 2020.
\newblock Published online: \url{iclr.cc}.

\bibitem[icl(2022)]{iclr22}
\emph{Proceedings of the International Conference on Learning Representations ({ICLR}'22)}, 2022.
\newblock Published online: \url{iclr.cc}.

\bibitem[An et~al.(2024)An, Zhao, Yu, Tang, and Wang]{an2024fluctuation}
Y.~An, X.~Zhao, T.~Yu, M.~Tang, and J.~Wang.
\newblock Fluctuation-based adaptive structured pruning for large language models.
\newblock In \emph{Proceedings of the AAAI Conference on Artificial Intelligence}, volume~38, pages 10865--10873, 2024.

\bibitem[Bai et~al.(2022)Bai, Hou, Shang, Jiang, King, and Lyu]{bai2022towards}
H.~Bai, L.~Hou, L.~Shang, X.~Jiang, I.~King, and M.~R. Lyu.
\newblock Towards efficient post-training quantization of pre-trained language models.
\newblock \emph{Advances in Neural Information Processing Systems}, 35:\penalty0 1405--1418, 2022.

\bibitem[Bakhtiarifard et~al.(2024)Bakhtiarifard, Igel, and Selvan]{bakhtiarifard2024ec}
P.~Bakhtiarifard, C.~Igel, and R.~Selvan.
\newblock Ec-nas: Energy consumption aware tabular benchmarks for neural architecture search.
\newblock In \emph{ICASSP 2024-2024 IEEE International Conference on Acoustics, Speech and Signal Processing (ICASSP)}, pages 5660--5664. IEEE, 2024.

\bibitem[Bender et~al.(2018)Bender, Kindermans, Zoph, Vasudevan, and Le]{bender-icml18a}
G.~Bender, P-J. Kindermans, B.~Zoph, V.~Vasudevan, and Q.~Le.
\newblock Understanding and simplifying one-shot architecture search.
\newblock In J.~Dy and A.~Krause, editors, \emph{Proceedings of the 35th International Conference on Machine Learning ({ICML}'18)}, volume~80. Proceedings of Machine Learning Research, 2018.

\bibitem[Bender et~al.(2020)Bender, Liu, Chen, Chu, Cheng, Kindermans, and Le]{bender-cvpr20a}
G.~Bender, H.~Liu, B.~Chen, G.~Chu, Sh. Cheng, P.~Kindermans, and Q.~V. Le.
\newblock Can weight sharing outperform random architecture search? an investigation with tunas.
\newblock In \emph{Proceedings of the International Conference on Computer Vision and Pattern Recognition ({CVPR}'20)}, 2020.

\bibitem[Breiman(2001)]{breiman2001random}
L.~Breiman.
\newblock Random forests.
\newblock \emph{Machine learning}, 45:\penalty0 5--32, 2001.

\bibitem[Brown et~al.(2020)Brown, Mann, Ryder, Subbiah, Kaplan, Dhariwal, Neelakantan, Shyam, Sastry, Askell, et~al.]{brown2020language}
T.~Brown, B.~Mann, N.~Ryder, M.~Subbiah, J.~D. Kaplan, P.~Dhariwal, A.~Neelakantan, P.~Shyam, G.~Sastry, A.~Askell, et~al.
\newblock Language models are few-shot learners.
\newblock \emph{Advances in neural information processing systems}, 33:\penalty0 1877--1901, 2020.

\bibitem[Chen et~al.(2021)Chen, Peng, Fu, and Ling]{chen-iccv21b}
M.~Chen, H.~Peng, J.~Fu, and H.~Ling.
\newblock Autoformer: Searching transformers for visual recognition.
\newblock In \emph{Proceedings of the 24nd IEEE/CVF International Conference on Computer Vision ({ICCV}'21)}, pages 12270--12280. cvfandieee, 2021.

\bibitem[Chen and Guestrin(2016)]{chen-kdd16a}
T.~Chen and C.~Guestrin.
\newblock {XGBoost}: {A} scalable tree boosting system.
\newblock In B.~Krishnapuram, M.~Shah, A.~Smola, C.~Aggarwal, D.~Shen, and R.~Rastogi, editors, \emph{Proceedings of the 22nd {ACM} {SIGKDD} International Conference on Knowledge Discovery and Data Mining ({KDD}'16)}, pages 785--794, 2016.

\bibitem[Chen and j.~C.~Jeong(2007)]{chen2007enhanced}
X.~Chen and j.~C.~Jeong.
\newblock Enhanced recursive feature elimination.
\newblock In \emph{Sixth international conference on machine learning and applications (ICMLA 2007)}, pages 429--435. IEEE, 2007.

\bibitem[Chowdhery et~al.(2023)Chowdhery, Narang, Devlin, Bosma, Mishra, Barham, Chung, Won, Sutton, Gehrmann, et~al.]{chowdhery2023palm}
A.~Chowdhery, S.~Narang, J.~Devlin, M.~Bosma, G.~Mishra, A.~Robertsand~P. Barham, H.~Chung, W.~Hyung Won, C.~Sutton, S.~Gehrmann, et~al.
\newblock Palm: Scaling language modeling with pathways.
\newblock \emph{Journal of Machine Learning Research}, 24\penalty0 (240):\penalty0 1--113, 2023.

\bibitem[Chung et~al.(2020)Chung, Neiswanger, Char, and Schneider]{chung2020beyond}
Y.~Chung, W.~Neiswanger, I.~Char, and J.~Schneider.
\newblock Beyond pinball loss: Quantile methods for calibrated uncertainty quantification.
\newblock \emph{arXiv preprint arXiv:2011.09588}, 2020.

\bibitem[Daultonl et~al.(2020)Daultonl, Balandat, and Bakshy]{daulton2020differentiable}
S.~Daultonl, M.~Balandat, and E.~Bakshy.
\newblock Differentiable expected hypervolume improvement for parallel multi-objective bayesian optimization.
\newblock \emph{Advances in Neural Information Processing Systems}, 33:\penalty0 9851--9864, 2020.

\bibitem[Deb et~al.(2002)Deb, Pratap, Agarwal, and Meyarivan]{deb2002fast}
K.~Deb, A.~Pratap, S.~Agarwal, and T.~Meyarivan.
\newblock A fast and elitist multiobjective genetic algorithm: Nsga-ii.
\newblock \emph{IEEE transactions on evolutionary computation}, 6\penalty0 (2):\penalty0 182--197, 2002.

\bibitem[Deb et~al.(2016)Deb, Sindhya, and Hakanen]{deb2016multi}
K.~Deb, K.~Sindhya, and J.~Hakanen.
\newblock Multi-objective optimization.
\newblock In \emph{Decision sciences}, pages 161--200. CRC Press, 2016.

\bibitem[Dong and Yang(2020)]{dong-iclr20a}
X.~Dong and Y.~Yang.
\newblock {NAS-Bench-201}: Extending the scope of reproducible {N}eural {A}rchitecture {S}earch.
\newblock In \emph{Proceedings of the International Conference on Learning Representations ({ICLR}'20)} \citet{iclr20}.
\newblock Published online: \url{iclr.cc}.

\bibitem[Dou et~al.(2022)Dou, Jiang, Zhao, and Li]{dou2022ea}
S.~Dou, X.~Jiang, C.~R. Zhao, and D.~Li.
\newblock Ea-has-bench: Energy-aware hyperparameter and architecture search benchmark.
\newblock In \emph{The Eleventh International Conference on Learning Representations}, 2022.

\bibitem[Duan et~al.(2021)Duan, Chen, H, Chen, Liang, Zhang, and Li]{duan-cvpr21a}
Y.~Duan, X.~Chen, Xu~H, Z.~Chen, X.~Liang, T.~Zhang, and Z.~Li.
\newblock {TransNAS-Bench-101}: {I}mproving {T}ransferability and {G}eneralizability of {Cross-Task} {Neural Architecture Search}.
\newblock In \emph{Proceedings of the International Conference on Computer Vision and Pattern Recognition ({CVPR}'21)}, pages 5251--5260, 2021.

\bibitem[Elsken et~al.(2019)Elsken, Metzen, and Hutter]{elsken-jmlr19a}
T.~Elsken, J.~Metzen, and F.~Hutter.
\newblock Neural {A}rchitecture {S}earch: A survey.
\newblock \emph{jmlr}, 20\penalty0 (55):\penalty0 1--21, 2019.

\bibitem[Erickson et~al.(2020)Erickson, Mueller, Shirkov, Zhang, Larroy, Li, and Smola]{erickson-arxiv20a}
N.~Erickson, J.~Mueller, A.~Shirkov, H.~Zhang, P.~Larroy, M.~Li, and A.~Smola.
\newblock Autogluon-tabular: Robust and accurate automl for structured data.
\newblock \emph{arXiv:2003.06505 [stat.ML]}, 2020.

\bibitem[Fan et~al.(2024)Fan, Jiang, Li, Meng, Han, Shang, Sun, Wang, and Wang]{fan2024not}
S.~Fan, X.~Jiang, X.~Li, X.~Meng, P.~Han, S.~Shang, A.~Sun, Y.~Wang, and Z.~Wang.
\newblock Not all layers of llms are necessary during inference.
\newblock \emph{arXiv preprint arXiv:2403.02181}, 2024.

\bibitem[Feurer et~al.(2022)Feurer, Eggensperger, Falkner, Lindauer, and Hutter]{feurer-jmlr22a}
M.~Feurer, K.~Eggensperger, S.~Falkner, M.~Lindauer, and F.~Hutter.
\newblock {Auto-Sklearn} 2.0: Hands-free automl via meta-learning.
\newblock \emph{jmlr}, 23\penalty0 (261):\penalty0 1--61, 2022.

\bibitem[Fonseca and Fleming(1996)]{fonseca1996}
Carlos~M. Fonseca and Peter~J. Fleming.
\newblock On the performance assessment and comparison of stochastic multiobjective optimizers.
\newblock In \emph{Proceedings of the 4th International Conference on Parallel Problem Solving from Nature}, PPSN IV, page 584–593, Berlin, Heidelberg, 1996. Springer-Verlag.
\newblock ISBN 354061723X.

\bibitem[Gijsbers et~al.(2024)Gijsbers, Bueno, LP, Coors, LeDell, Thomas, Bischl, and Vanschoren]{gijsbers2024amlb}
P.~Gijsbers, M.~Bueno, LP, S.~Coors, E.~LeDell, S.~Poirierand~J. Thomas, B.~Bischl, and J.~Vanschoren.
\newblock Amlb: an automl benchmark.
\newblock \emph{Journal of Machine Learning Research}, 25\penalty0 (101):\penalty0 1--65, 2024.

\bibitem[Gordon(2024)]{forbesaisustainability}
C.~Gordon.
\newblock Forbes ai sustainability havoc.
\newblock \href{https://www.forbes.com/sites/cindygordon/2024/03/12/chatgpt-and-generative-ai-innovations-are-creating-sustainability-havoc/}{Forbes AI ChatGPT}, 2024.

\bibitem[Grinsztajn et~al.(2022)Grinsztajn, Oyallon, and Varoquaux]{grinsztajn2022tree}
L.~Grinsztajn, E.~Oyallon, and G.~Varoquaux.
\newblock Why do tree-based models still outperform deep learning on typical tabular data?
\newblock \emph{Advances in neural information processing systems}, 35:\penalty0 507--520, 2022.

\bibitem[Guo et~al.(2020)Guo, Zhang, Mu, Heng, Liu, Wei, and Sun]{guo-eccv20a}
Z.~Guo, X.~Zhang, H.~Mu, W.~Heng, Z.~Liu, Y.~Wei, and J.~Sun.
\newblock Single path one-shot neural architecture search with uniform sampling.
\newblock In A.~Vedaldi, H.~Bischof, T.~Brox, and J.~Frahm, editors, \emph{16th European Conference on Computer Vision ({ECCV}'20)}, pages 544--560. Springer, 2020.

\bibitem[Guyon et~al.(2017)Guyon, von Luxburg, Bengio, Wallach, Fergus, Vishwanathan, and Garnett]{nips17}
I.~Guyon, U.~von Luxburg, S.~Bengio, H.~Wallach, R.~Fergus, S.~Vishwanathan, and R.~Garnett, editors.
\newblock \emph{Proceedings of the 31st International Conference on Advances in Neural Information Processing Systems ({N}eur{IPS}'17)}, 2017.

\bibitem[Han et~al.(2020)Han, Gan, Tianzhe, Zhang, and Song]{Cai2020Once-for-All:}
C.~Han, G.~Chuang Gan, W.~Tianzhe, Z.~Zhekai Zhang, and H.~Song.
\newblock Once-for-all: Train one network and specialize it for efficient deployment.
\newblock In \emph{International Conference on Learning Representations}, 2020.

\bibitem[hee et~al.(2024)hee, Cai, Kuleshov, and Sa]{chee2024quip}
J.~hee, Y.~Cai, V.~Kuleshov, and C.~M.~De Sa.
\newblock Quip: 2-bit quantization of large language models with guarantees.
\newblock \emph{Advances in Neural Information Processing Systems}, 36, 2024.

\bibitem[Hoffmann et~al.(2022)Hoffmann, Borgeaud, Mensch, Buchatskaya, Cai, Rutherford, Casas, Hendricks, Welbl, Clark, et~al.]{hoffmann2022training}
J.~Hoffmann, S.~Borgeaud, A.~Mensch, E.~Buchatskaya, T.~Cai, E.~Rutherford, D.~Casas, L.~Hendricks, J.~Welbl, A.~Clark, et~al.
\newblock Training compute-optimal large language models.
\newblock \emph{arXiv preprint arXiv:2203.15556}, 2022.

\bibitem[Ke et~al.(2017)Ke, Meng, Finley, Wang, Chen, Ma, Ye, and Liu]{ke-neurips17a}
G.~Ke, Q.~Meng, T.~Finley, T.~Wang, W.~Chen, W.~Ma, Q.~Ye, and T.-Y. Liu.
\newblock Lightgbm: A highly efficient gradient boosting decision tree.
\newblock In  \citet{nips17}.

\bibitem[Kim et~al.(2024)Kim, Kim, Kim, Castells, hoi, Shin, and Song]{kim2024shortened}
B.~Kim, G.~Kim, T.~Kim, T.~Castells, S.~hoi, J.~Shin, and H.~Song.
\newblock Shortened llama: A simple depth pruning for large language models.
\newblock \emph{arXiv preprint arXiv:2402.02834}, 2024.

\bibitem[Klein et~al.(2024)Klein, Golebiowski, Ma, Perrone, and Archambeau]{klein2023structural}
A.~Klein, J.~Golebiowski, X.~Ma, V.~Perrone, and C.~Archambeau.
\newblock Structural pruning of pre-trained language models via neural architecture search.
\newblock \emph{arXiv:2405.02267 [cs.LG]}, 2024.

\bibitem[Klyuchnikov et~al.(2022)Klyuchnikov, Trofimov, Artemova, Salnikov, Fedorov, Filippov, and Burnaev]{klyuchnikov2022bench}
N.~Klyuchnikov, I.~Trofimov, E.~Artemova, M.~Salnikov, M.~Fedorov, A.~Filippov, and E.~Burnaev.
\newblock Nas-bench-nlp: neural architecture search benchmark for natural language processing.
\newblock \emph{IEEE Access}, 10:\penalty0 45736--45747, 2022.

\bibitem[Lee et~al.(2021)Lee, Lee, Chong, and Hwang]{lee2021hardware}
H.~Lee, S.~Lee, S.~Chong, and S.~J. Hwang.
\newblock Hardware-adaptive efficient latency prediction for nas via meta-learning.
\newblock \emph{Advances in Neural Information Processing Systems}, 34:\penalty0 27016--27028, 2021.

\bibitem[Li et~al.(2021)Li, Yu, Fu, Zhang, Zhao, You, Yu, Wang, and Lin]{li2021hw}
C.~Li, Z.~Yu, Y.~Fu, Y.~Zhang, Y.~Zhao, H.~You, Q.~Yu, Y.~Wang, and Y.~Lin.
\newblock Hw-nas-bench: Hardware-aware neural architecture search benchmark.
\newblock In \emph{The 9th International Conference on Learning Representations 2021 (ICLR 2021)}, 2021.

\bibitem[Li and Talwalkar(2020)]{li2020random}
L.~Li and A.~Talwalkar.
\newblock Random search and reproducibility for neural architecture search.
\newblock In \emph{Uncertainty in artificial intelligence}, pages 367--377. PMLR, 2020.

\bibitem[Li et~al.(2020)Li, Jamieson, Rostamizadeh, Gonina, Ben{-}tzur, Hardt, Recht, and Talwalkar]{li-mlsys20a}
L.~Li, K.~Jamieson, A.~Rostamizadeh, E.~Gonina, J.~Ben{-}tzur, M.~Hardt, B.~Recht, and A.~Talwalkar.
\newblock A system for massively parallel hyperparameter tuning.
\newblock In I.~Dhillon, D.~Papailiopoulos, and V.~Sze, editors, \emph{Proceedings of Machine Learning and Systems 2}, volume~2, 2020.

\bibitem[Liang et~al.(2021)Liang, Glossner, Wang, Shi, and Zhang]{liang2021pruning}
T.~Liang, J.~Glossner, L.~Wang, S.~Shi, and X.~Zhang.
\newblock Pruning and quantization for deep neural network acceleration: A survey.
\newblock \emph{Neurocomputing}, 461:\penalty0 370--403, 2021.

\bibitem[Liu et~al.(2019)Liu, Simonyan, and Yang]{liu-iclr19a}
H.~Liu, K.~Simonyan, and Y.~Yang.
\newblock {DARTS}: Differentiable architecture search.
\newblock In \emph{Proceedings of the International Conference on Learning Representations ({ICLR}'19)}, 2019.
\newblock Published online: \url{iclr.cc}.

\bibitem[Longpre et~al.(2023)Longpre, Hou, Vu, Webson, H.Chung, Hyung, Tay, Zhou, Le, Zoph, Wei, et~al.]{longpre2023flan}
S.~Longpre, L.~Hou, T.~Vu, A.~Webson, H.Chung, W.~Hyung, Y.~Tay, D.~Zhou, Q.~V. Le, B.~Zoph, J.~Wei, et~al.
\newblock The flan collection: Designing data and methods for effective instruction tuning.
\newblock In \emph{International Conference on Machine Learning}, pages 22631--22648. PMLR, 2023.

\bibitem[Lu et~al.(2019)Lu, Whalen, Boddeti, Dhebar, Deb, Goodman, and Banzhaf]{lu-gecco19a}
Z.~Lu, I.~Whalen, V.~Boddeti, Y.~Dhebar, K.~Deb, E.~Goodman, and W.~Banzhaf.
\newblock Nsga-net: Neural architecture search using multi-objective genetic algorithm.
\newblock In M.~{López-Ibáñez}, editor, \emph{Proceedings of the Genetic and Evolutionary Computation Conference ({GECCO}'19)}, page 419–427, 2019.

\bibitem[Ma et~al.(2024)Ma, Wang, Ma, Wang, Wang, Huang, Dong, Wang, Xue, and Wei]{ma2024era}
S.~Ma, H.~Wang, L.~Ma, L.~Wang, W.~Wang, S.~Huang, L.~Dong, R.~Wang, J.~Xue, and F.~Wei.
\newblock The era of 1-bit llms: All large language models are in 1.58 bits.
\newblock \emph{arXiv preprint arXiv:2402.17764}, 2024.

\bibitem[Manuel et~al.(2010)Manuel, Luis, and Thomas]{eaf}
L.~Manuel, P.~Luis, and S.~Thomas.
\newblock Exploratory analysis of stochastic local search algorithms in biobjective optimization.
\newblock In Thomas Bartz-Beielstein, Marco Chiarandini, Luís Paquete, and Mike Preuss, editors, \emph{Experimental Methods for the Analysis of Optimization Algorithms}, pages 209--222. Springer, Berlin, Germany, 2010.
\newblock \doi{10.1007/978-3-642-02538-9_9}.

\bibitem[McElfresh et~al.(2024)McElfresh, Khandagale, Valverde, C, Ramakrishnan, Goldblum, and White]{mcelfresh2024neural}
D.~McElfresh, S.~Khandagale, J.~Valverde, C.~V.~Prasad C, G.~Ramakrishnan, M.~Goldblum, and C.~White.
\newblock When do neural nets outperform boosted trees on tabular data?
\newblock \emph{Advances in Neural Information Processing Systems}, 36, 2024.

\bibitem[Mehrotra et~al.(2021)Mehrotra, Ramos, Bhattacharya, Dudziak, Vipperla, Chau, Abdelfattah, Ishtiaq, and Lane]{mehrotra-iclr21a}
A.~Mehrotra, A.~Ramos, S.~Bhattacharya, {\L}.~Dudziak, R.~Vipperla, T.~Chau, M.~Abdelfattah, S.~Ishtiaq, and N.~Lane.
\newblock {NAS}-{B}ench-{ASR}: {R}eproducible {N}eural {A}rchitecture {S}earch for {S}peech {R}ecognition.
\newblock In \emph{Proceedings of the International Conference on Learning Representations ({ICLR}'21)}, 2021.
\newblock Published online: \url{iclr.cc}.

\bibitem[Mehta et~al.(2022)Mehta, White, Zela, Krishnakumar, Zabergja, Moradian, Safari, Yu, and Hutter]{mehta-iclr22a}
Y.~Mehta, C.~White, A.~Zela, A.~Krishnakumar, G.~Zabergja, S.~Moradian, M.~Safari, K.~Yu, and F.~Hutter.
\newblock {NAS-Bench-Suite}: {NAS} evaluation is (now) surprisingly easy.
\newblock In \emph{Proceedings of the International Conference on Learning Representations ({ICLR}'22)} \citet{iclr22}.
\newblock Published online: \url{iclr.cc}.

\bibitem[Michel et~al.(2019)Michel, Levy, and Neubig]{michel2019sixteen}
P.~Michel, O.~Levy, and G.~Neubig.
\newblock Are sixteen heads really better than one?
\newblock \emph{Advances in neural information processing systems}, 32, 2019.

\bibitem[Minaee et~al.(2024)Minaee, Mikolov, Nikzad, Chenaghlu, Socher, Amatriain, and Gao]{minaee2024large}
Shervin Minaee, Tomas Mikolov, Narjes Nikzad, Meysam Chenaghlu, Richard Socher, Xavier Amatriain, and Jianfeng Gao.
\newblock Large language models: A survey.
\newblock \emph{arXiv preprint arXiv:2402.06196}, 2024.

\bibitem[Mu{\~n}oz et~al.(2024)Mu{\~n}oz, Yuan, and Jain]{munoz2024shears}
J~Pablo Mu{\~n}oz, Jinjie Yuan, and Nilesh Jain.
\newblock Shears: Unstructured sparsity with neural low-rank adapter search.
\newblock \emph{arXiv preprint arXiv:2404.10934}, 2024.

\bibitem[Munoz et~al.(2024)Munoz, Yuan, Zheng, and Jain]{munoz2024lonas}
Juan~Pablo Munoz, Jinjie Yuan, Yi~Zheng, and Nilesh Jain.
\newblock Lonas: Elastic low-rank adapters for efficient large language models.
\newblock In \emph{Proceedings of the 2024 Joint International Conference on Computational Linguistics, Language Resources and Evaluation (LREC-COLING 2024)}, pages 10760--10776, 2024.

\bibitem[Paria et~al.(2020)Paria, Kandasamy, and P{\'o}czos]{paria2020flexible}
B.~Paria, K.~Kandasamy, and B.~P{\'o}czos.
\newblock A flexible framework for multi-objective bayesian optimization using random scalarizations.
\newblock In \emph{Uncertainty in Artificial Intelligence}, pages 766--776. PMLR, 2020.

\bibitem[Pedregosa et~al.(2011)Pedregosa, Varoquaux, Gramfort, Michel, Thirion, Grisel, Blondel, Prettenhofer, Weiss, Dubourg, Vanderplas, Passos, Cournapeau, Brucher, Perrot, and Duchesnay]{scikit-learn}
F.~Pedregosa, G.~Varoquaux, A.~Gramfort, V.~Michel, B.~Thirion, O.~Grisel, M.~Blondel, P.~Prettenhofer, R.~Weiss, V.~Dubourg, J.~Vanderplas, A.~Passos, D.~Cournapeau, M.~Brucher, M.~Perrot, and E.~Duchesnay.
\newblock Scikit-learn: Machine learning in {P}ython.
\newblock \emph{JMLR}, 12:\penalty0 2825--2830, 2011.

\bibitem[Pope et~al.(2023)Pope, Douglas, Chowdhery, Devlin, Bradbury, Heek, Xiao, Agrawal, and Dean]{pope2023efficiently}
R.~Pope, S.~Douglas, A.~Chowdhery, J.~Devlin, J.~Bradbury, J.~Heek, K.~Xiao, S.~Agrawal, and J.~Dean.
\newblock Efficiently scaling transformer inference.
\newblock \emph{Proceedings of Machine Learning and Systems}, 5, 2023.

\bibitem[Radford et~al.(2019)Radford, J.~Wu, Luan, Amodei, Sutskever, et~al.]{radford2019language}
A.~Radford, L.~Rewon J.~Wu, D.~Luan, D.~Amodei, I.~Sutskever, et~al.
\newblock Language models are unsupervised multitask learners.
\newblock \emph{OpenAI blog}, 1\penalty0 (8):\penalty0 9, 2019.

\bibitem[Real et~al.(2019)Real, Aggarwal, Huang, and Le]{real2019regularized}
E.~A. Real, A.~Aggarwal, Y.~Huang, and Q.~V. Le.
\newblock Regularized evolution for image classifier architecture search.
\newblock In \emph{Proceedings of the aaai conference on artificial intelligence}, volume~33, pages 4780--4789, 2019.

\bibitem[Sajjad et~al.(2023)Sajjad, Dalvi, Durrani, and Nakov]{sajjad2023effect}
H.~Sajjad, F.~Dalvi, N.~Durrani, and P.~Nakov.
\newblock On the effect of dropping layers of pre-trained transformer models.
\newblock \emph{Computer Speech \& Language}, 77:\penalty0 101429, 2023.

\bibitem[Salinas and Erickson(2023)]{salinas2023tabrepo}
D.~Salinas and N.~Erickson.
\newblock Tabrepo: A large scale repository of tabular model evaluations and its automl applications.
\newblock \emph{arXiv preprint arXiv:2311.02971}, 2023.

\bibitem[Salinas et~al.(2022)Salinas, Seeger, Klein, Perrone, Wistuba, and Archambeau]{salinas2022syne}
D.~Salinas, M.~Seeger, A.~Klein, V.~Perrone, M.~Wistuba, and C.~Archambeau.
\newblock Syne tune: A library for large scale hyperparameter tuning and reproducible research.
\newblock In \emph{International Conference on Automated Machine Learning}, pages 16--1. PMLR, 2022.

\bibitem[Salinas et~al.(2021)Salinas, Perrone, Cruchant, and Archambeau]{Salinas2021AMP}
David Salinas, Valerio Perrone, Olivier Cruchant, and C.~Archambeau.
\newblock A multi-objective perspective on jointly tuning hardware and hyperparameters.
\newblock \emph{ArXiv}, abs/2106.05680, 2021.

\bibitem[Sarah et~al.(2024)Sarah, Sridhar, Szankin, and Sundaresan]{sarah2024llama}
Anthony Sarah, Sharath~Nittur Sridhar, Maciej Szankin, and Sairam Sundaresan.
\newblock Llama-nas: Efficient neural architecture search for large language models.
\newblock \emph{arXiv preprint arXiv:2405.18377}, 2024.

\bibitem[Schmucker et~al.(2021)Schmucker, Donini, Zafar, Salinas, and Archambeau]{schmucker-arxiv21a}
R.~Schmucker, M.~Donini, M.~Zafar, D.~Salinas, and C.~Archambeau.
\newblock Multi-objective asynchronous successive halving.
\newblock \emph{arXiv:2106.12639 {[stat.ML]}}, 2021.

\bibitem[Shoeybi et~al.(2019)Shoeybi, Patwary, Puri, LeGresley, J.~Casper, and Catanzaro]{shoeybi2019megatron}
M.~Shoeybi, M.~Patwary, R.~Puri, P.~LeGresley, Jared J.~Casper, and Bryan Catanzaro.
\newblock Megatron-lm: Training multi-billion parameter language models using model parallelism.
\newblock \emph{arXiv preprint arXiv:1909.08053}, 2019.

\bibitem[So et~al.(2019)So, Le, and Liang]{so2019evolved}
D.~So, Q.~Le, and C.~Liang.
\newblock The evolved transformer.
\newblock In \emph{International conference on machine learning}, pages 5877--5886. PMLR, 2019.

\bibitem[Su et~al.(2024)Su, Ahmed, Lu, Pan, Bo, and Liu]{su2024roformer}
J.~Su, M.~Ahmed, Y.~Lu, S.~Pan, W.~Bo, and Y.~Liu.
\newblock Roformer: Enhanced transformer with rotary position embedding.
\newblock \emph{Neurocomputing}, 568:\penalty0 127063, 2024.

\bibitem[Sukthanker et~al.(2024)Sukthanker, Zela, Staffler, Dooley, Grabocka, and Hutter]{sukthanker2024multi}
R.~S. Sukthanker, A.~Zela, B.~Staffler, S.~Dooley, J.~Grabocka, and F.~Hutter.
\newblock Multi-objective differentiable neural architecture search.
\newblock \emph{arXiv preprint arXiv:2402.18213}, 2024.

\bibitem[Tao et~al.(2022)Tao, Hou, Zhang, Shang, Jiang, Liu, Ping, and Wong]{tao2022compression}
C.~Tao, L.~Hou, W.~Zhang, L.~Shang, X.~Jiang, Q.~Liu, P.~Luo Ping, and N.~Wong.
\newblock Compression of generative pre-trained language models via quantization.
\newblock \emph{arXiv preprint arXiv:2203.10705}, 2022.

\bibitem[Tran et~al.(2020)Tran, Neiswanger, Yoon, Zhang, Xing, and Ulissi]{tran2020methods}
K.~Tran, W.~Neiswanger, J.~Yoon, Q.~Zhang, E.~Xing, and Z.~W. Ulissi.
\newblock Methods for comparing uncertainty quantifications for material property predictions.
\newblock \emph{Machine Learning: Science and Technology}, 1\penalty0 (2):\penalty0 025006, 2020.

\bibitem[Tu et~al.(2022)Tu, Roberts, Khodak, Shen, Sala, and Talwalkar]{tu2022bench}
R.~Tu, N.~Roberts, M.~Khodak, J.~Shen, F.~Sala, and A.~Talwalkar.
\newblock Nas-bench-360: Benchmarking neural architecture search on diverse tasks.
\newblock \emph{Advances in Neural Information Processing Systems}, 35:\penalty0 12380--12394, 2022.

\bibitem[Vaswani et~al.(2017)Vaswani, Shazeer, Parmar, Uszkoreit, Jones, Gomez, Kaiser, and Polosukhin]{vaswani-neurips17a}
A.~Vaswani, N.~Shazeer, N.~Parmar, J.~Uszkoreit, L.~Jones, A.~Gomez, L.~Kaiser, and I.~Polosukhin.
\newblock Attention is all you need.
\newblock In  \citet{nips17}.

\bibitem[Voita et~al.(2019)Voita, Talbot, Moiseev, Sennrich, and Titov]{voita2019analyzing}
E.~Voita, D.~Talbot, F.~Moiseev, R.~Sennrich, and I.~Titov.
\newblock Analyzing multi-head self-attention: Specialized heads do the heavy lifting, the rest can be pruned.
\newblock In \emph{Proceedings of the 57th Annual Meeting of the Association for Computational Linguistics}. Association for Computational Linguistics, 2019.

\bibitem[Waddington et~al.(2013)Waddington, Colmenares, Kuang, and Song]{waddington2013kv}
D.~Waddington, J.~Colmenares, J.~Kuang, and F.~Song.
\newblock Kv-cache: A scalable high-performance web-object cache for manycore.
\newblock In \emph{2013 IEEE/ACM 6th International Conference on Utility and Cloud Computing}, pages 123--130. IEEE, 2013.

\bibitem[Wan et~al.(2023)Wan, Wang, Liu, Alam, Zheng, Qu, Yan, Zhu, Zhang, Chowdhury, et~al.]{wan2023efficient}
Z.~Wan, X.~Wang, C.~Liu, S.~Alam, Y.~Zheng, Z.~Qu, S.~Yan, Y.~Zhu, Q.~Zhang, M.~Chowdhury, et~al.
\newblock Efficient large language models: A survey.
\newblock \emph{arXiv preprint arXiv:2312.03863}, 1, 2023.

\bibitem[Wang and Komatsuzaki(2021)]{wang2021gpt}
B.~Wang and A.~Komatsuzaki.
\newblock Gpt-j-6b: A 6 billion parameter autoregressive language model, 2021.

\bibitem[Wang et~al.(2020)Wang, Wu, Liu, Cai, Zhu, Gan, and Han]{wang2020hat}
H.~Wang, Z.~Wu, Z.~Liu, H.~Cai, L.~Zhu, C.~Gan, and S.~Han.
\newblock Hat: Hardware-aware transformers for efficient natural language processing.
\newblock In \emph{Proceedings of the 58th Annual Meeting of the Association for Computational Linguistics}, pages 7675--7688, 2020.

\bibitem[Wang et~al.(2019)Wang, Wohlwend, and Lei]{wang2019structured}
Z.~Wang, J.~Wohlwend, and T.~Lei.
\newblock Structured pruning of large language models.
\newblock \emph{arXiv preprint arXiv:1910.04732}, 2019.

\bibitem[Watanabe(2023)]{watanabe2023pareto}
S.~Watanabe.
\newblock {P}ython tool for visualizing variability of {P}areto fronts over multiple runs.
\newblock \emph{arXiv:2305.08852}, 2023.

\bibitem[White et~al.(2023)White, Safari, Sukthanker, Ru, Elsken, Zela, Dey, and Hutter]{white-arxiv23a}
C.~White, M.~Safari, R.~Sukthanker, B.~Ru, T.~Elsken, A.~Zela, D.~Dey, and F.~Hutter.
\newblock Neural architecture search: Insights from 1000 papers.
\newblock \emph{arXiv:2301.08727 [cs.LG]}, 2023.

\bibitem[Wolpert(1992)]{wolpert1992stacked}
D.~H. Wolpert.
\newblock Stacked generalization.
\newblock \emph{Neural networks}, 5\penalty0 (2):\penalty0 241--259, 1992.

\bibitem[Yan et~al.(2021)Yan, White, Savani, and Hutter]{yan-neurips21a}
S.~Yan, C.~White, Y.~Savani, and F.~Hutter.
\newblock {NAS}-bench-x11 and the power of learning curves.
\newblock In M.~Ranzato, A.~Beygelzimer, K.~Nguyen, P.~Liang, J.~Vaughan, and Y.~Dauphin, editors, \emph{Proceedings of the 35th International Conference on Advances in Neural Information Processing Systems ({N}eur{IPS}'21)}, volume~34, pages 22534--22549, 2021.

\bibitem[Ying et~al.(2019)Ying, Klein, Christiansen, Real, Murphy, and Hutter]{ying-icml19a}
C.~Ying, A.~Klein, E.~Christiansen, E.~Real, K.~Murphy, and F.~Hutter.
\newblock {NAS-Bench-101}: Towards reproducible {N}eural {A}rchitecture {S}earch.
\newblock In K.~Chaudhuri and R.~Salakhutdinov, editors, \emph{Proceedings of the 36th International Conference on Machine Learning ({ICML}'19)}, volume~97, pages 7105--7114. Proceedings of Machine Learning Research, 2019.

\bibitem[Yu et~al.(2018)Yu, Yang, Xu, Yang, and Huang]{yu2018slimmable}
J.~Yu, L.~Yang, N.~Xu, J.~Yang, and T.~Huang.
\newblock Slimmable neural networks.
\newblock In \emph{International Conference on Learning Representations}, 2018.

\bibitem[Zafrir et~al.(2021)Zafrir, Larey, Boudoukh, Shen, and Wasserblat]{zafrir2021prune}
O.~Zafrir, A.~Larey, G.~Boudoukh, H.~Shen, and M.~Wasserblat.
\newblock Prune once for all: Sparse pre-trained language models.
\newblock \emph{arXiv preprint arXiv:2111.05754}, 2021.

\bibitem[Zela et~al.(2020)Zela, Siems, and Hutter]{zela-iclr20b}
A.~Zela, J.~Siems, and F.~Hutter.
\newblock {NAS-Bench-1Shot1}: Benchmarking and dissecting {O}ne-shot {N}eural {A}rchitecture {S}earch.
\newblock In \emph{Proceedings of the International Conference on Learning Representations ({ICLR}'20)} \citet{iclr20}.
\newblock Published online: \url{iclr.cc}.

\bibitem[Zela et~al.(2022)Zela, Siems, Zimmer, Lukasik, Keuper, and Hutter]{zela-iclr22a}
A.~Zela, J.~Siems, L.~Zimmer, J.~Lukasik, M.~Keuper, and F.~Hutter.
\newblock Surrogate {NAS} benchmarks: Going beyond the limited search spaces of tabular {NAS} benchmarks.
\newblock In \emph{Proceedings of the International Conference on Learning Representations ({ICLR}'22)} \citet{iclr22}.
\newblock Published online: \url{iclr.cc}.

\bibitem[Zoph and Le(2016)]{zoph2016neural}
B.~Zoph and Q.~Le.
\newblock Neural architecture search with reinforcement learning.
\newblock In \emph{International Conference on Learning Representations}, 2016.

\end{thebibliography}
\bibliographystyle{plainnat}

\newpage
\appendix
\section{Pretraining details}
\label{sec:training_details}
We present all the training hyperparameters for the supernetwork (see Section~\ref{subsec:dataset_collection} too) at different scales in Table \ref{tab:hyperparameters}. We train GPT-S, -M and -L scales for 4, 6 and 8 days, respectively, on 4 Nvidia A100 GPUs. GPT-XL is trained for 6 days across 40 A100 GPUs. As described in Section \ref{subsec:dataset_collection}, we use the sandwich rule for the supernetwork training, which accumulates gradients from the smallest, largest and 2 random subnetworks of the supernetwork in each gradient update step. 

\begin{table}[ht]
\centering
\resizebox{.66\linewidth}{!}{%
\begin{tabular}{lc}
\toprule
\textbf{Hyperparameter} & \textbf{Value} \\ \midrule
\textbf{Trainer} &  \\ \hline
num\_gpus & 4 \\ 
gradient\_clip\_val & 1.0 \\ 
max\_steps & 800000 \\
precision & BFloat16 Mixed precision \\ 
seed & 1234 \\ \hline
\textbf{Optimizer} &  \\ \hline
optimizer & AdamW \\ 
lr & 0.000316 \\ 
weight\_decay & 0.1 \\ 
betas & [0.9, 0.95] \\ 
eps & 1.0e-09 \\ 
seed & 1234 \\ \hline
\textbf{Optimizer Parameter Grouping} &  \\ \hline
bias\_weight\_decay & False \\ 
normalization\_weight\_decay & False \\ \hline
\textbf{Scheduler} &  \\ \hline
num\_warmup\_steps & 4000 \\ 
num\_training\_steps & 800000 \\ 
decay\_factor & 0.1 \\ 
schedule & cosine \\ \hline
\textbf{Model} &  \\ \hline
block\_size & 1024 \\ 
vocab\_size & 50254 \\ 
padding\_multiple & 512 \\ 
scale\_embeddings & False \\ 
padded\_vocab\_size & 50254 \\
head\_size & 64 \\ 
lm\_head\_bias & False \\ 
attn\_dropout & 0.1 \\
resi\_dropout & 0.1 \\ 
embed\_dropout & 0.1 \\ 
shared\_attention\_norm & False \\ 
norm & LayerNorm \\
rope\_condense\_ratio & 1 \\ 
scale\_attn\_weights & True \\ 
scale\_attn\_by\_inverse\_layer\_idx & True \\ 
shared\_embedding & True \\ 
pos\_embedding & False \\ 
rel\_pos\_enc & True \\ 
rotary\_percentage & 0.5 \\
norm\_eps & 1e-5 \\ 
rope\_base & 10000 \\ 
parallel\_residual & True \\
initializer\_range & 0.02 \\ \hline
\textbf{Data} &  \\ \hline
dataset & "openwebtext" \\ 
num\_cpu\_worker & 32 \\ 
batch\_size & 32(GPT-S), 8(GPT-M), 8(GPT-L), 1(GPT-XL-wide) \\
max\_sample\_len & 1024 \\  
val\_ratio & 0.0005 \\ 
val\_split\_seed & 2357 \\ \bottomrule
\end{tabular}
}
\caption{Supernet training hyperparameters.}
\label{tab:hyperparameters}
\end{table}

\clearpage

\section{Hardware specifications and Properties}
\label{sec:hardwares}
In Table \ref{tab:hardware_specifications}, we present details of the 8 GPU devices and 5 CPU devices we used for profiling our 3 search spaces GPT-S, GPT-M and GPT-L. The devices we study capture a wide range of micro-architectures, number of cores and GPU/CPU memory. The considered GPUs capture a wide range of inference latencies and throughput \footnote{\href{https://lambdalabs.com/gpu-benchmarks}{GPU benchmark}}.
\begin{table}[H]
\centering
\caption{Hardware specifications.}
\resizebox{\textwidth}{!}{%
\begin{tabular}{lllll}
\toprule
\textbf{Platform} & \textbf{Device Name} & \textbf{Micro Architecture} & \textbf{Number of Cores} & \textbf{Memory} \\ \toprule
\multirow{8}{*}{GPU} & \href{https://www.nvidia.com/content/geforce-gtx/GEFORCE_RTX_2080_User_Guide.pdf}{\texttt{NVIDIA RTX 2080}} & \texttt{Turing}& CUDA 2944 & 8GB GDDR6 \\  
 & \href{https://www.nvidia.com/de-de/geforce/graphics-cards/30-series/rtx-3080-3080ti/}{\texttt{NVIDIA RTX 3080}} & \texttt{Ampere} & CUDA 8704 & 12GB GDDR6X \\  
 & \href{https://www.nvidia.com/content/dam/en-zz/Solutions/design-visualization/quadro-product-literature/proviz-print-nvidia-rtx-a6000-datasheet-us-nvidia-1454980-r9-web\%20(1).pdf}{\texttt{NVIDIA RTX A6000}} & \texttt{Ampere} & CUDA 10752 & 48 GB GDDR6 \\ 
 & \href{https://www.nvidia.com/content/dam/en-zz/Solutions/Data-Center/a100/pdf/nvidia-a100-datasheet-us-nvidia-1758950-r4-web.pdf}{\texttt{NVIDIA A100}} & \texttt{Ampere} & CUDA 6912 & 40GB HBM2e \\ 
 & \href{https://images.nvidia.com/content/Solutions/data-center/a40/nvidia-a40-datasheet.pdf}{\texttt{NVIDIA A40}} & \texttt{Ampere} & CUDA 10752 & 48GB GDDR6 \\
  & \href{https://www.nvidia.com/content/dam/en-zz/Solutions/Data-Center/tesla-p100/pdf/nvidia-tesla-p100-datasheet.pdf}{\texttt{NVIDIA Tesla P100}} & \texttt{Pascal} & CUDA 3584 & 12GB HBM2 \\
   & \href{https://images.nvidia.com/content/technologies/volta/pdf/volta-v100-datasheet-update-us-1165301-r5.pdf}{\texttt{NVIDIA Tesla V100}} & \texttt{Volta} & CUDA 5120  & 24GB GDDR6 \\
      & \href{https://resources.nvidia.com/en-us-tensor-core/nvidia-tensor-core-gpu-datasheet}{\texttt{NVIDIA H100}} & \texttt{Hopper} & CUDA 16896 &  80GB HBM3\\ \midrule
\multirow{5}{*}{CPU} & \href{https://example.com}{\texttt{AMD 7452}} & \texttt{Zen2} & CPU Core 32 & Cache 128 MB \\ 
& \href{https://example.com}{\texttt{AMD 7513}} & \texttt{Zen3} & CPU Core 32  & Cache 128MB \\
& \href{https://example.com}{\texttt{AMD 7502}} & \texttt{Zen2} & CPU Core 32 & Cache 128MB \\
 & \href{https://ark.intel.com/content/www/us/en/ark/products/123550/intel-xeon-silver-4114-processor-13-75m-cache-2-20-ghz.html}{\texttt{Intel Xeon Silver 4114}} & \texttt{Intel P6} & CPU Core 10 & Cache 13.75MB \\  
 & \href{https://example.com}{\texttt{Intel Xeon Gold 6242}} & \texttt{Intel P6} & CPU Core 16 & Cache 22MB \\ \bottomrule
\end{tabular}%
}
\label{tab:hardware_specifications}
\end{table}
\section{Surrogate models}
\label{sec:surrogate_app}
In this section, we provide details of the latency and energy surrogate predictors used in HW-GPT-Bench. The surrogates are trained and evaluated on 8000 and 2000 fixed architectures, respectively. 

\subsection{PPL Surrogates}
\label{sec:surrogates_ppl}

\paragraph{Data collection.} We subsample 10\% from the OpenWebText training set and use that as a validation set to compute the perplexity and accuracy metrics of architectures in the supernetwork, that inherit the trained supernetwork weights.
The time to collect the perplexity pairs for 10k architectures was 8, 16, 24 and 32 days on 4 A100 GPUs for GPT-S, -M and -L and -XL-wide, respectively.
Note that this computation is trivial to parallelize, as all architectures can be evaluated independently of each other, given the pretrained supernetwork weights. Afterwards, the ($x$, $y$) pairs, where $x$ is the one-hot encoded architecture and $y$ the corresponding perplexity value,
are used to train an MLP (Multi Layer Perceptron) using the MSE loss to predict the perplexity on unseen (test) architectures. 

\paragraph{Surrogate Architecture.} 
The MLP contains 4 linear layers with hidden dimension of 128 and output dimension of 1 (the perplexity prediction). We use ReLU activations at every layer.
The dimension of the one-hot encoding is 3 (3 choices for layers) + 3 (3 choices for embedding dimension) + max\_layers$\times$3 (3 choices for number of heads for every layer) + max\_layers$\times$3 (3 choices for MLP ratio for every layer). We train the MLP for 4000 epochs using the Adam optimizer with a learning rate of $10^{-3}$ and a batch size of 1024. 

\subsection{Hardware Surrogates}
\label{sec:surrogates_lat_en}
We now describe the different surrogates we study to model uncertainties in energy and latency prediction. Each of the surrogates is trained to predict mean and standard deviation of energy or latency predictions of a given architecture. For FLOPs, memory and number of parameters, we compute and predict a single observation per architecture. We use 8000 architectures for training our surrogates and 2000 to test the performance and calibration properties of our surrogates. 

\paragraph{Data collection.} We measure energy and latency pairs across different devices multiple times to capture the noise in the energy and latency profiling. Additionally, we also profile hardware metrics like Floating Point Operations (FLOPs), Memory consumption (Float16 and BFloat16) and number of parameters for an architecture. Specifically, for every architecture, we compute 50 observations per architecture for energies on GPU and 10 observations for energies on CPU. Similarly, we compute 10 observations per architecture for latencies on both CPUs and GPUs. We use 8 CPU cores (per GPU) for all the latency and energy evaluations. 
\paragraph{AutoGluon.}
AutoGluon simplifies the process of developing, training, and deploying machine learning models by automating key tasks such as feature engineering, model selection, and hyperparameter tuning. AutoGluon has two major components. First one is Bagging (bootstrap aggregation) on wide varieties of models like decision trees, random forests, nearest neighbors, linear models and neural networks. Second one is stacking, where predictions of models on  the first level are fed as input to models at the next level. We use an AutoGluon model per metric (latency or energy) with $\mathtt{dynamic\_stacking = False}$, $\mathtt{num\_stack\_levels = 1}$, $\mathtt{number\_of\_bag\_folds = 8}$ and $\mathtt{number\_of\_bag\_sets = 4}$. The input feature map to the model is simply a concatenation of chosen hyperparameters with choices for MLP ratio and number of heads set to -1 when a particular layer is not selected. We train two AutoGluon predictors, one to predict mean, and one to predict the standard deviation of the ground truth latency or energy measurements.

\paragraph{Ensemble (XGB).}
We use an ensemble of 27 Extreme Gradient Boosting (XGB) models with different hyperparameter combinations. Specifically, the models in the ensemble contain a cross product of $\mathtt{n\_estimator}$ choices from $\texttt{[200, 500, 800]}$, $\mathtt{depth}$ choices from $\texttt{[5,9,3]}$ and $\mathtt{learning\_rate}$ choices from $\texttt{[0.01,0.1,1.0]}$. We then fit this ensemble, and predict the mean and standard deviation values for the metric at hand by aggregating the predictions of different models using bagging. 

\paragraph{Ensemble (LightGBM).}
We use an ensemble of 36 Light Gradient Boosting Machine (LightGBM) models with different hyperparameter combinations. Specifically, the models in the ensemble contain a cross product of $\mathtt{n\_estimator}$ choices from $\texttt{[200, 500, 800]}$, $\mathtt{depth}$ choices from $\texttt{[5,9,3]}$ and $\mathtt{learning\_rate}$ choices from $\texttt{[0.01,0.1,0.001,1.0]}$. We then fit this ensemble, and predict the mean and standard deviation values for the metric at hand by aggregating the predictions of different models using bagging.

\paragraph{Ensemble (MIX).}
In addition to ensembles based on XGB and LightGBM, we also fit an ensemble containing a mixture of different machine learning models. Specifically, this ensemble contains 4 XGB and LightGBM regressors (default: \texttt{[n\_estimators = 500,max\_depth = 5 ,learning\_rate = 0.01]},$\texttt{[n\_estimators = 800,max\_depth = 3,learning\_rate = 0.1]}$, \texttt{[n\_estimators = 200,max\_depth = 9,learning\_rate = 1]}), as well as LinearRegression, Ridge and RandomForestRegressor from $\texttt{sklearn}$ with their default hyperparameters. 

\subsection{Memory Surrogate.}
We model the memory consumption using a simple MLP (similar to perplexity). The MLP has 4 layers with a hidden dimension of 128 and uses ReLU activations. The input to the MLP is the normalized feature map corresponding to the concatenation of the architecture parameters. The MLP is trained using Adam, with learning rate of $10^{-4}$ and a batch size of 1024, for 4000 epochs. 

\section{Accuracy and Calibration of Surrogate models}
\label{sec:metrics_app}

In this section we provide a more detailed description on the metrics we used to evaluate our surrogate models in Section~\ref{subsec:surrogates}, as well as additional results that complement results shown in Table~\ref{tab:surrogatemetrics} and Figure~\ref{fig:uct_plots}.

\subsection{Accuracy and Calibration Metrics}

Given a finite dataset $D = \{(x_i, y_i)^N_{i=1}\}$ and a regression model that predicts the mean value $\hat{y}$, we use standard accuracy metrics:

\textbf{Root Mean Squared Error (RMSE):} 
$RMSE = \sqrt{ \frac{1}{N} \sum_{i=1}^{N} (\hat{y_i} - y_i)^2 } $

\textbf{Mean Absolute Error (MAE):}
$ MAE = \frac{1}{N} \sum_{i=1}^{N} | \hat{y_i} - y_i | $

\textbf{Median Absolute Error (MDAE):} Median of the absolute differences between predicted and actual values, in ascending order.
$$ MDAE = med( \{| \hat{y_i} - y_i | \}_{i=1}^{N} ) $$

\textbf{Mean Absolute Percentage Relative Deviation (MARPD):} A similar metric to MAE but scaled by the sum of the absolute values of predicted and actual values. This scaling can be beneficial when dealing with a wider range of values, especially when the predicted or actual values are very close to zero.
$$MARPD = \frac{1}{N} \sum_{i=1}^{N} \left| \frac{2 \cdot (\hat{y_i} - y_i)} {|\hat{y_i}| + |y_i|} \right| \cdot 100$$

\textbf{Coefficient of Determination ($R^2$):} Tells us about the proportion of variance in the dependent variable (y) explained by the independent variable (x) through the model. Denoting by $\bar{y} = \sum_{i=1}^{N} y_i$ the mean of the ground truth data, $R^2$ is defined as:
$$R^2 = 1 - \frac{\sum_{i=1}^{N} (\hat{y_i} - y_i)^2} {\sum_{i=1}^{N} (\bar{y} - y_i)^2} $$

\textbf{Correlation Coefficient:} The Pearson correlation coefficient ($r$) measures the strength and direction of the linear relationship between predicted values ($\hat{y_i}$) and actual values ($y_i$) of the data. It ranges from -1 to 1.
$$r = \frac{\sum_{i=1}^{N} (\hat{y_i} - \bar{y})(y_i - \bar{y})} {\sqrt{\sum_{i=1}^{N} (\hat{y_i} - \bar{y})^2 \sum_{i=1}^{N} (y_i - \bar{y})^2}}$$

\textbf{Root Mean Squared Calibration Error (RMS Cal.):} A metric used to evaluate the calibration of probabilistic predictions in regression models, particularly in the context of uncertainty quantification. It measures the discrepancy between predicted and observed quantiles. 
For a model to be well-calibrated, the proportion of observations below a given predicted quantile should match the quantile value. For example, the 90th percentile prediction should contain the true outcome 90\% of the time. RMS Cal. is computed as the root mean square of the difference between the predicted quantile levels and the observed frequencies over all quantiles:
\[
\text{RMS Cal.} = \sqrt{\frac{1}{K} \sum_{k=1}^{K} \big( P(Y \leq Q_k) - q_k \big)^2},
\]
where \( Q_k \) is the k-th predicted quantile, \( q_k \) is the corresponding quantile level, and $ P(Y \leq Q_k) = \frac{1}{N} \sum_{i=1}^{N} \mathbb{I}[ y_i \leq Q_k (x_i) ] $ is the empirical frequency of the target variables \( Y = \{ y_i \}_{i=1}^N \) being less than or equal to \( Q_k \). $\mathbb{I}$ stands for the indicator function. A lower RMS Cal. indicates better calibration, reflecting that the predicted uncertainties are accurate and reliable.

\textbf{Mean Absolute Calibration Error (MA Cal.):}
MA Cal. is computed as the average absolute difference between the predicted quantile levels and the observed frequencies over all quantiles:
\[
\text{MA Cal.} = \frac{1}{K} \sum_{k=1}^{K} \left| P(Y \leq Q_k) - q_k \right|.
\]

\subsection{Additional Results on the Accuracy and Calibration of Surrogates}

\begin{figure}[t]
\centering
\begin{minipage}{.16\linewidth}
  \centering
  \includegraphics[width=\linewidth]{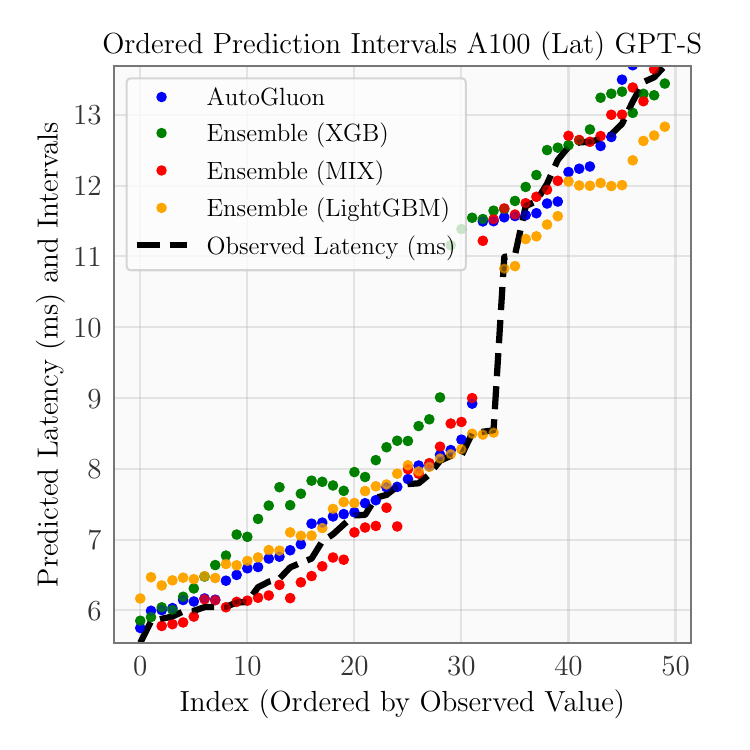}
\end{minipage}%
\begin{minipage}{.16\linewidth}
  \centering
\includegraphics[width=\linewidth]{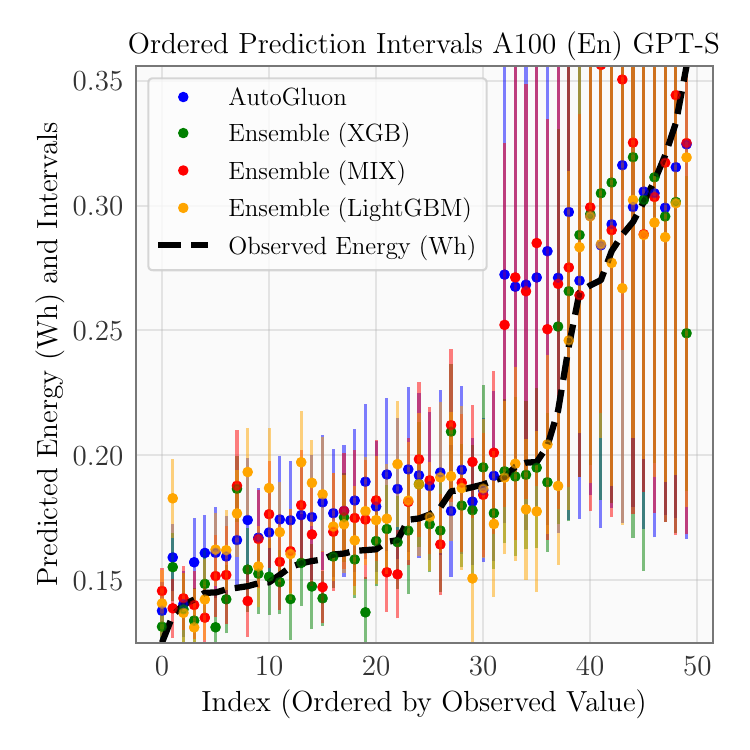}
\end{minipage}
\begin{minipage}{.16\linewidth}
  \centering
\includegraphics[width=\linewidth]{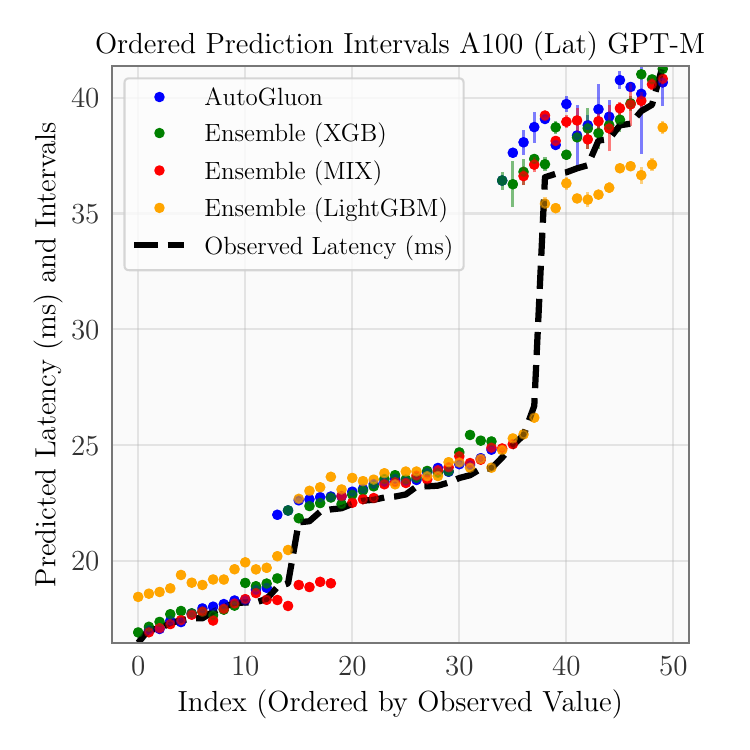}
\end{minipage}
\begin{minipage}{.16\linewidth}
  \centering
\includegraphics[width=\linewidth]{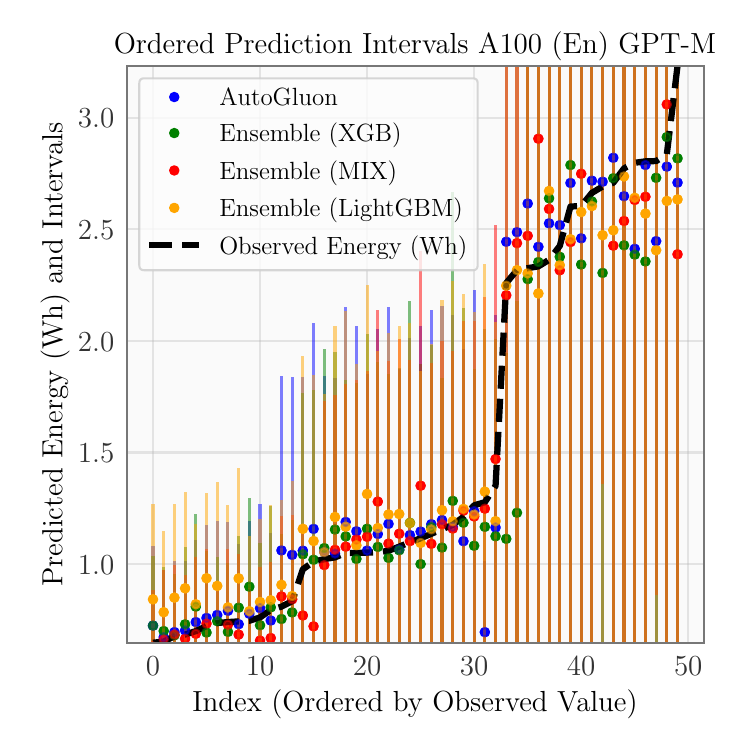}
\end{minipage}
\begin{minipage}{.16\linewidth}
  \centering
\includegraphics[width=\linewidth]{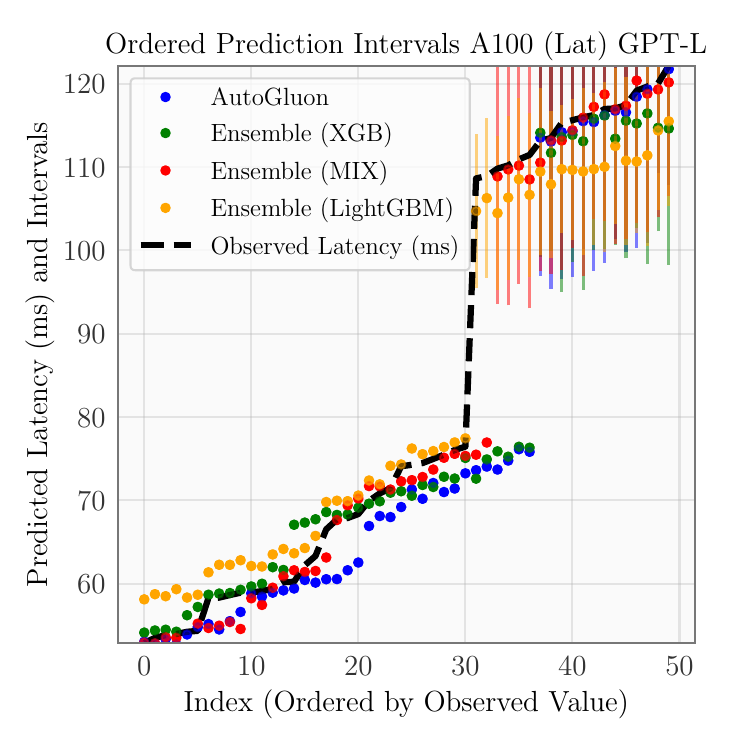}
\end{minipage}
\begin{minipage}{.16\linewidth}
  \centering
\includegraphics[width=\linewidth]{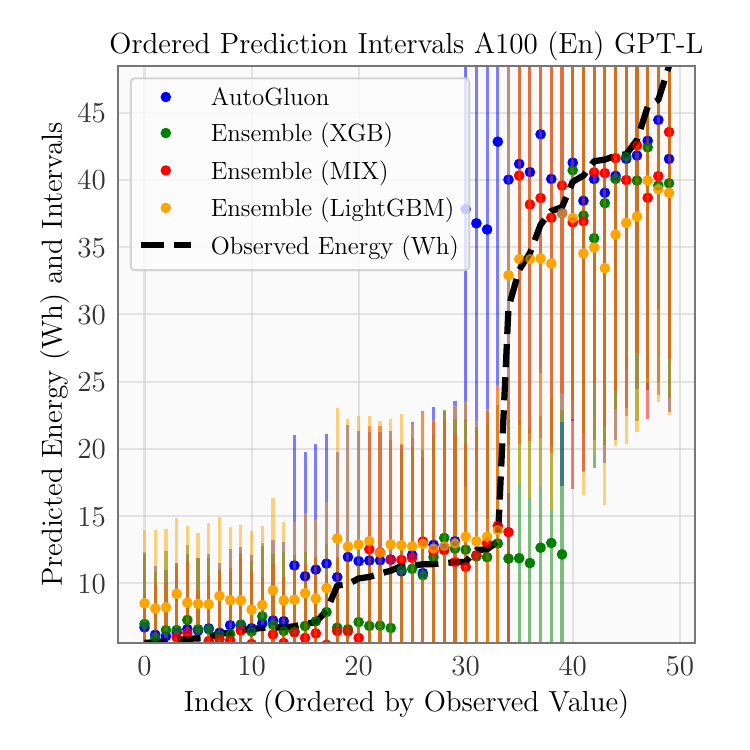}
\end{minipage}
\caption{Calibration Areas A100}
\label{fig:calib_area_a100}
\vspace{-5mm}
\end{figure}

\begin{figure}[t]
\centering
\begin{minipage}{.16\linewidth}
  \centering
  \includegraphics[width=\linewidth]{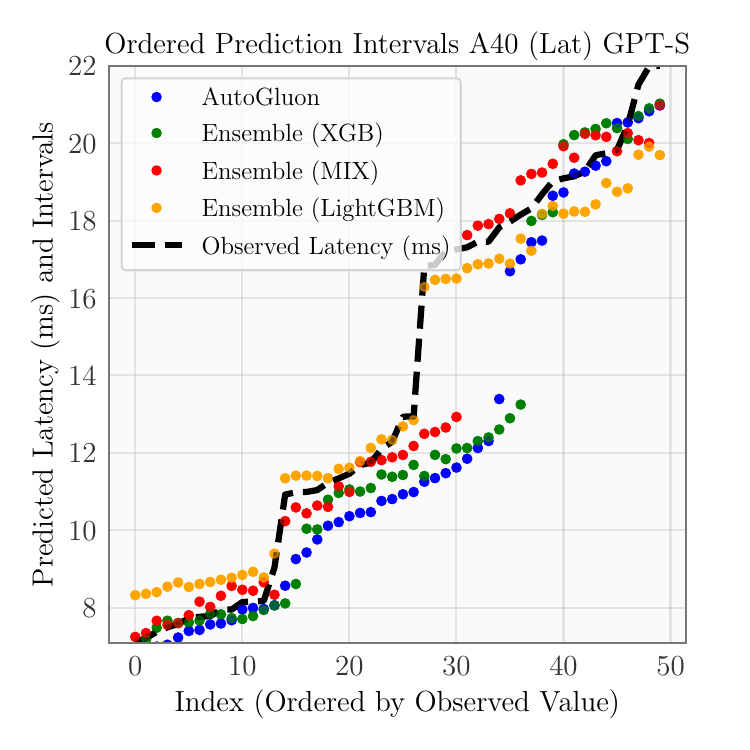}
\end{minipage}%
\begin{minipage}{.16\linewidth}
  \centering
\includegraphics[width=\linewidth]{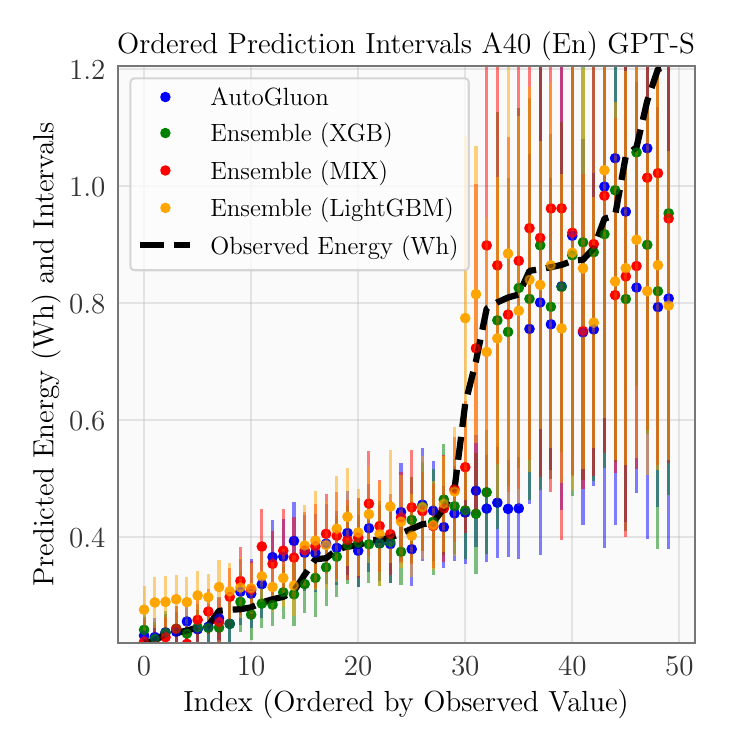}
\end{minipage}
\begin{minipage}{.16\linewidth}
  \centering
\includegraphics[width=\linewidth]{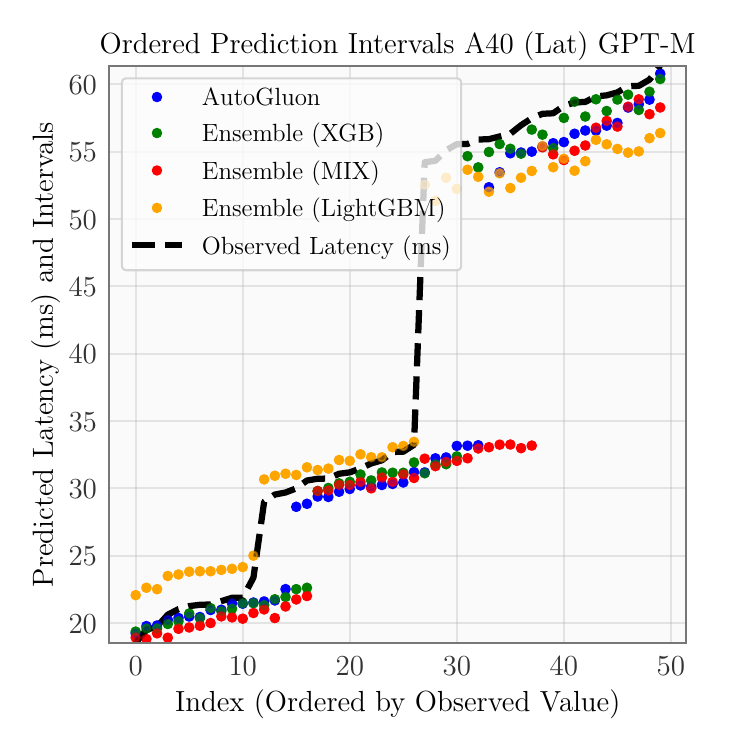}
\end{minipage}
\begin{minipage}{.16\linewidth}
  \centering
\includegraphics[width=\linewidth]{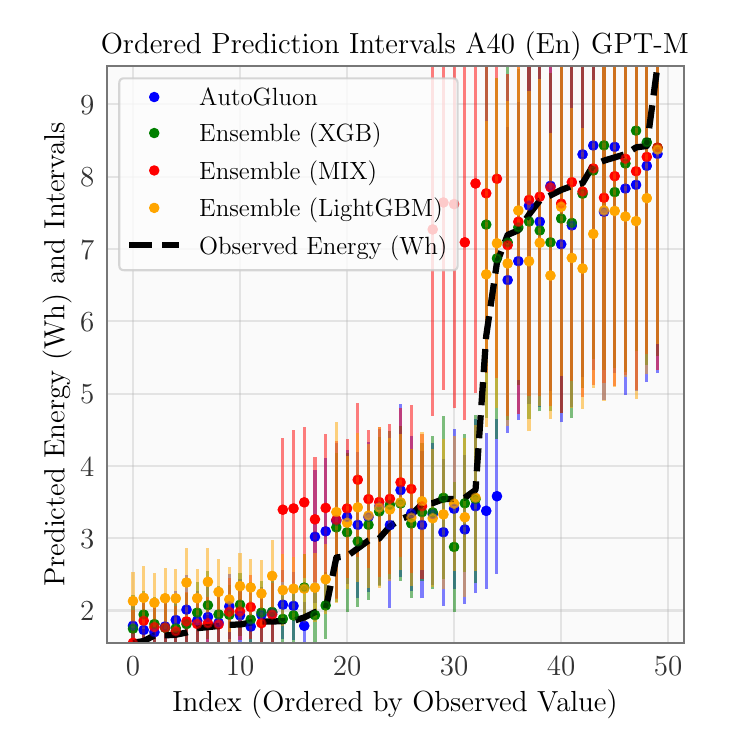}
\end{minipage}
\begin{minipage}{.16\linewidth}
  \centering
\includegraphics[width=\linewidth]{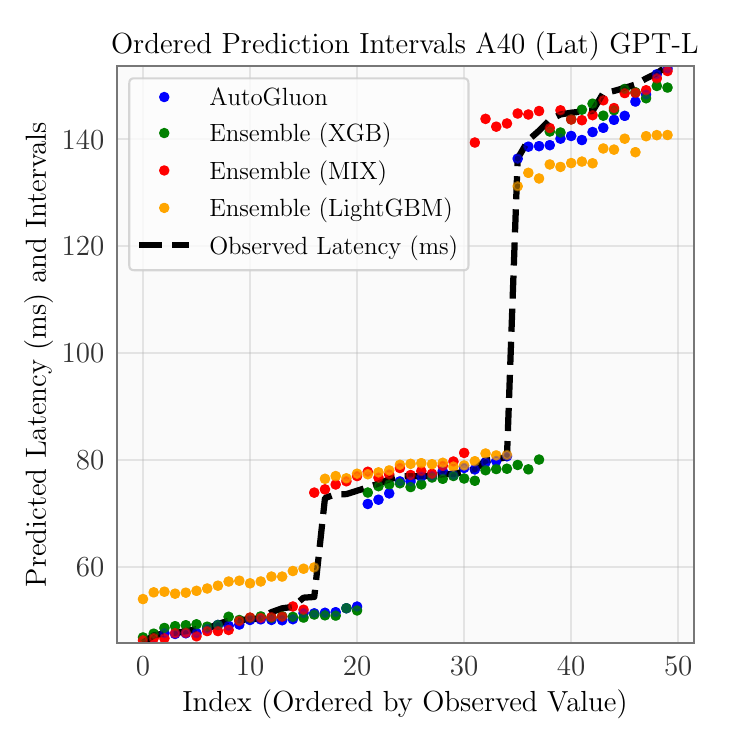}
\end{minipage}
\begin{minipage}{.16\linewidth}
  \centering
\includegraphics[width=\linewidth]{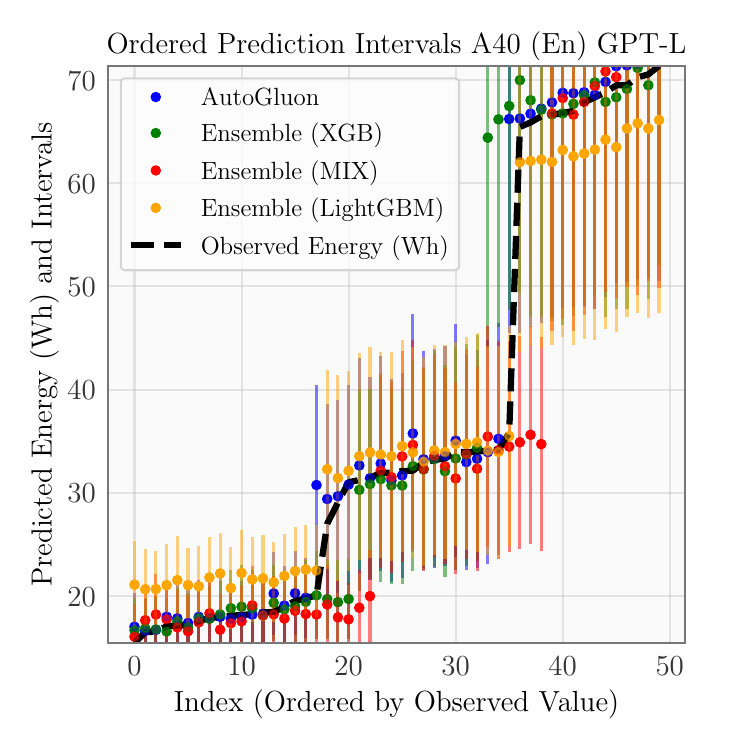}
\end{minipage}
\caption{Calibration Areas A40}
\label{fig:calib_area_a40}
\vspace{-5mm}
\end{figure}

\begin{figure}[t]
\centering
\begin{minipage}{.16\linewidth}
  \centering

  \includegraphics[width=\linewidth]{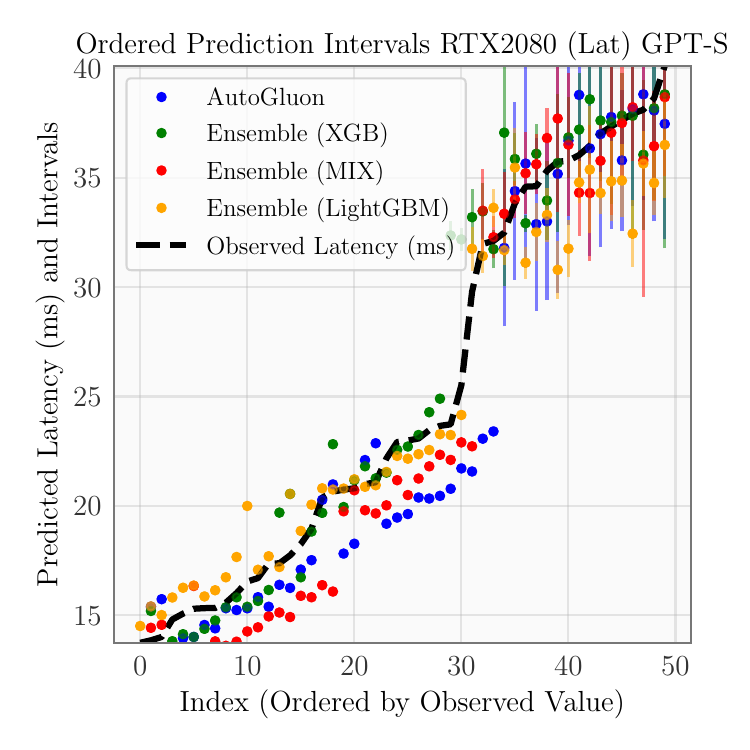}
\end{minipage}%
\begin{minipage}{.16\linewidth}
  \centering
\includegraphics[width=\linewidth]{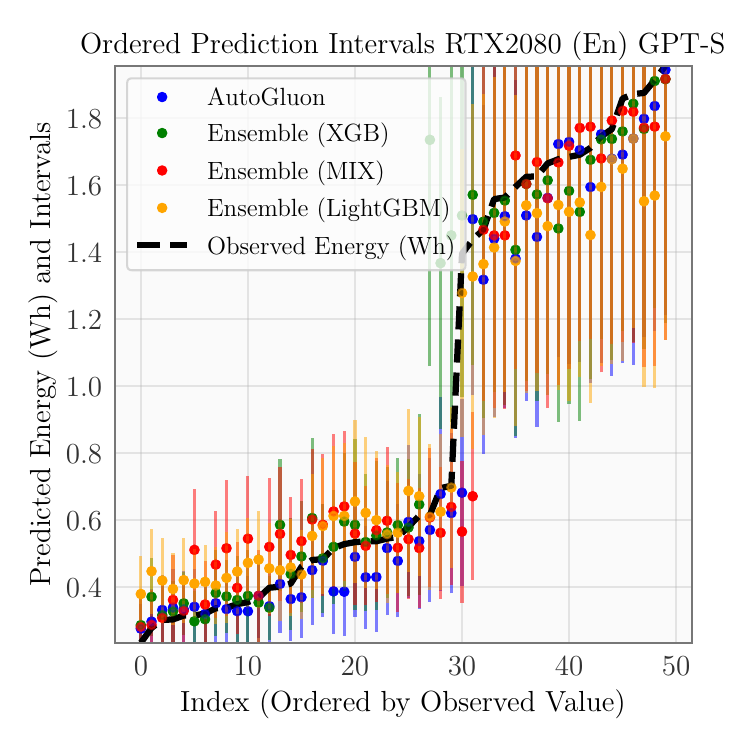}
\end{minipage}
\begin{minipage}{.16\linewidth}
  \centering
\includegraphics[width=\linewidth]{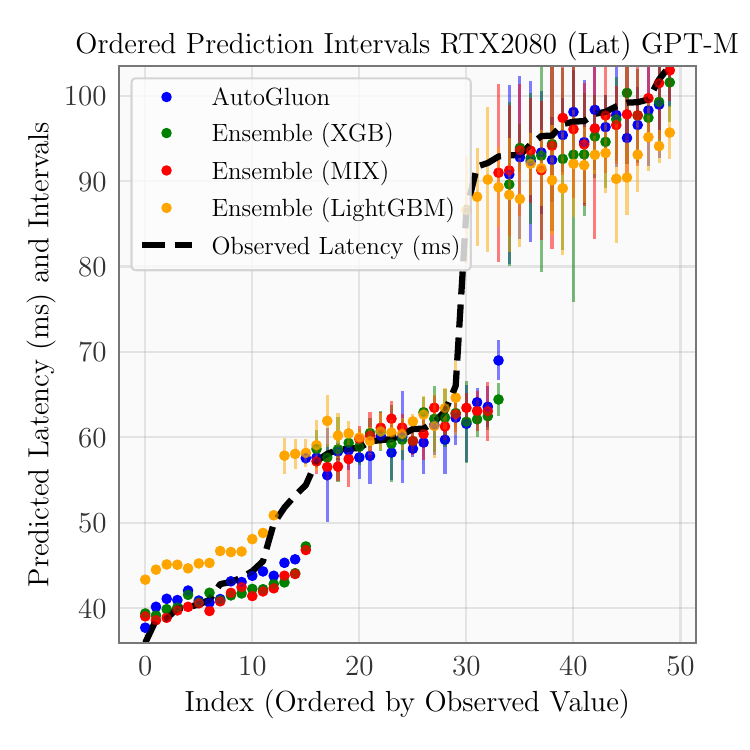}
\end{minipage}
\begin{minipage}{.16\linewidth}
  \centering
\includegraphics[width=\linewidth]{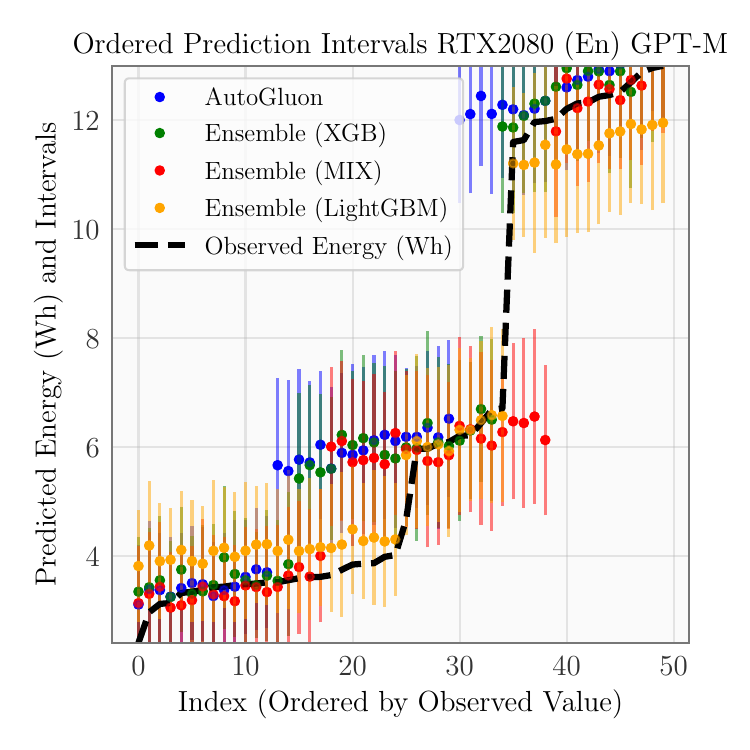}
\end{minipage}
\begin{minipage}{.16\linewidth}
  \centering
\includegraphics[width=\linewidth]{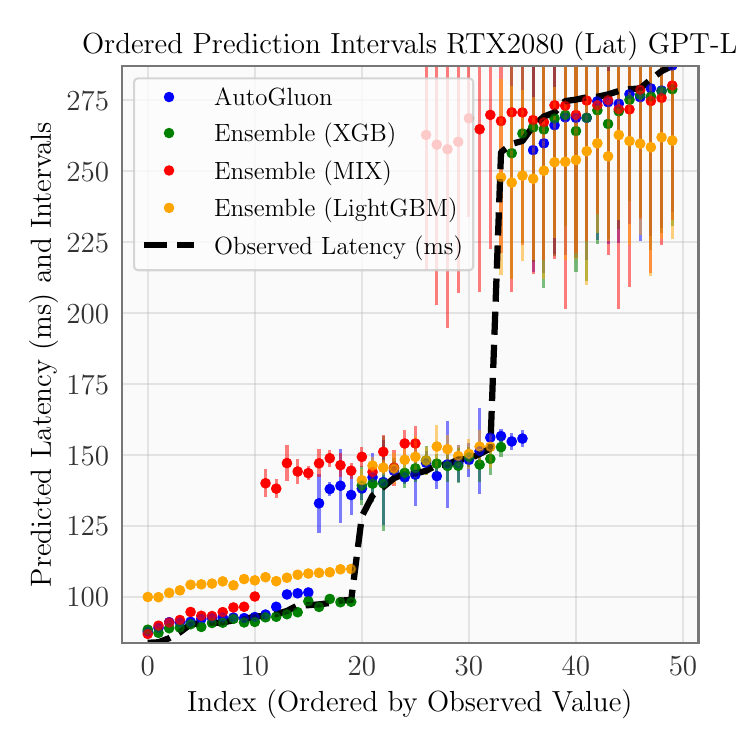}
\end{minipage}
\begin{minipage}{.16\linewidth}
  \centering
\includegraphics[width=\linewidth]{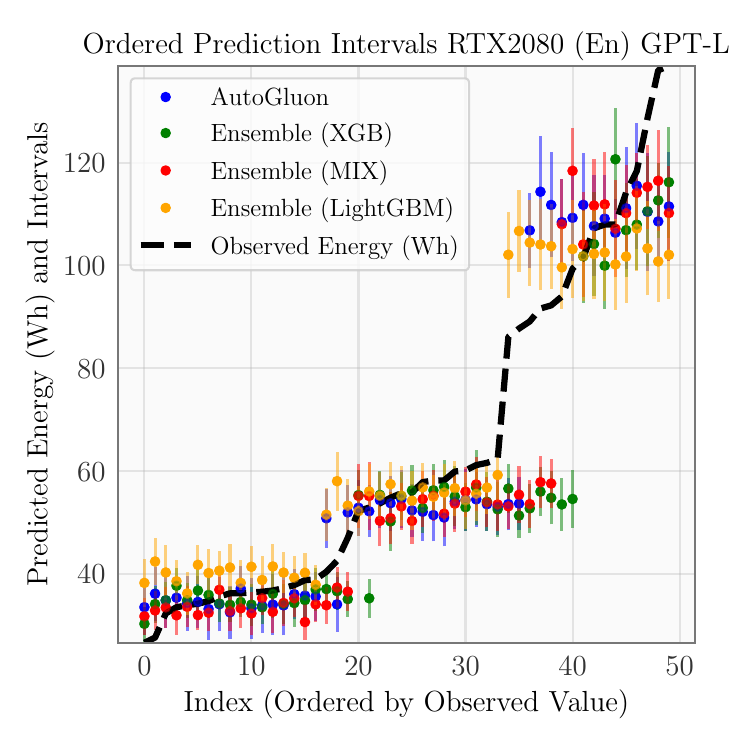}
\end{minipage}
\caption{Calibration Areas RTX2080}
\label{fig:calib_area_rtx2080}
\vspace{-5mm}
\end{figure}

\begin{figure}[t]
\centering
\begin{minipage}{.16\linewidth}
  \centering
  \includegraphics[width=\linewidth]{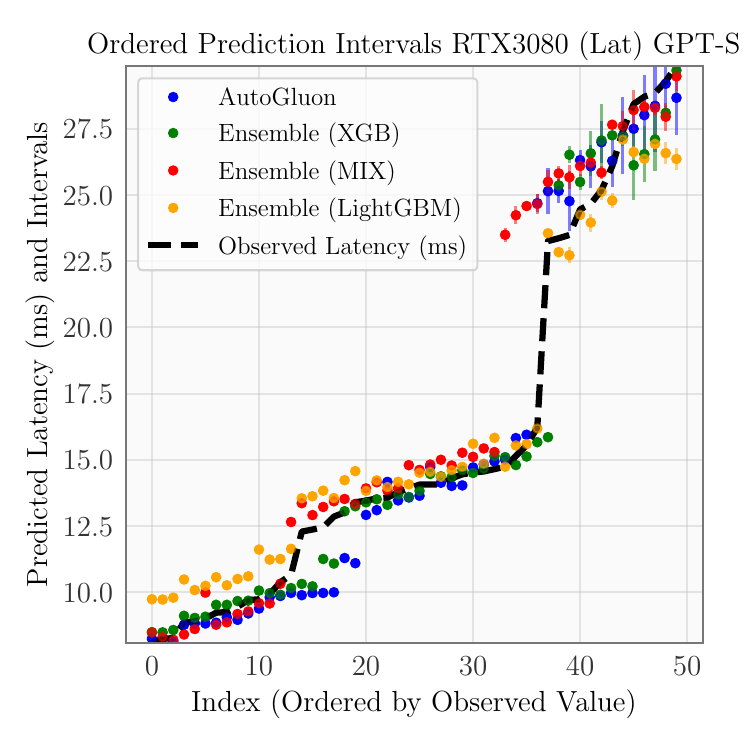}
\end{minipage}%
\begin{minipage}{.16\linewidth}
  \centering
\includegraphics[width=\linewidth]{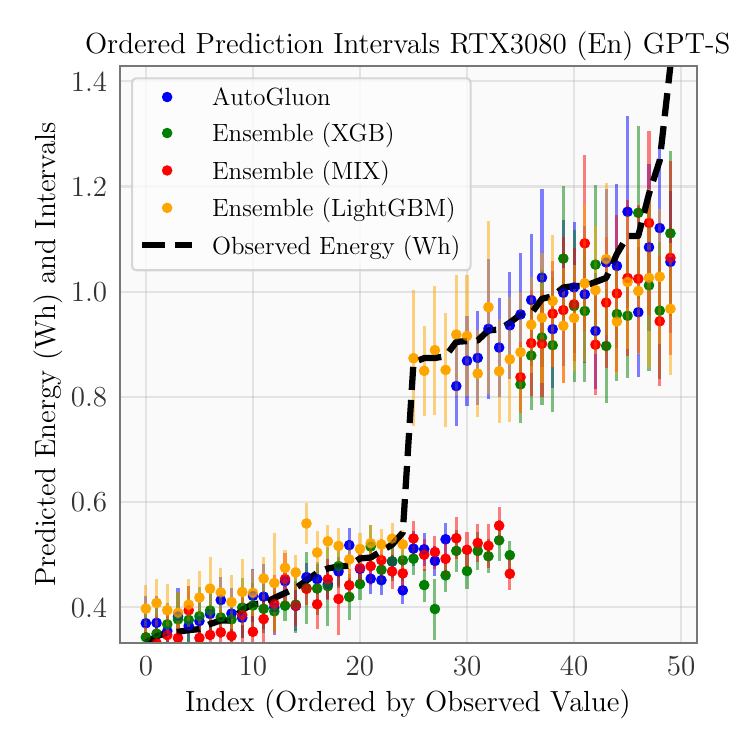}
\end{minipage}
\begin{minipage}{.16\linewidth}
  \centering
\includegraphics[width=\linewidth]{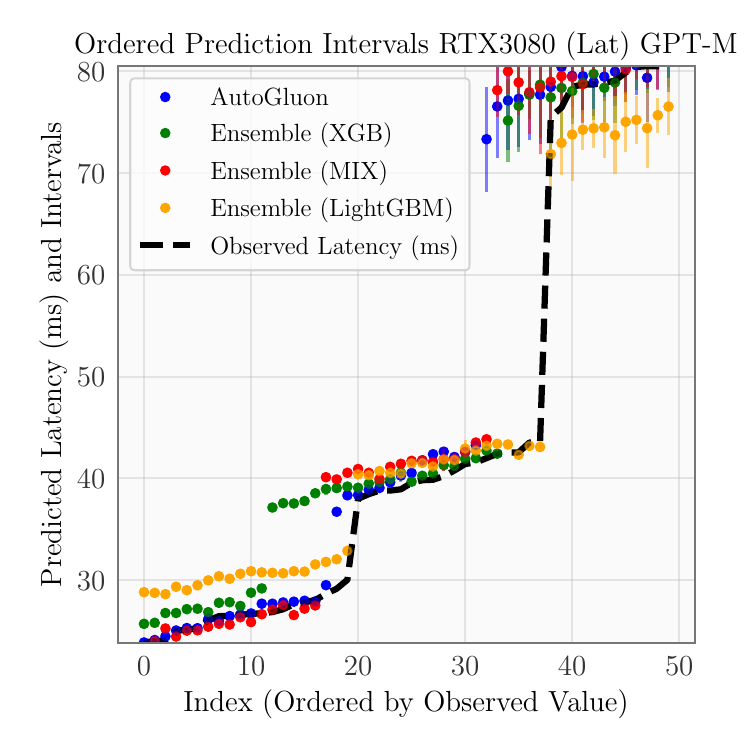}
\end{minipage}
\begin{minipage}{.16\linewidth}
  \centering
\includegraphics[width=\linewidth]{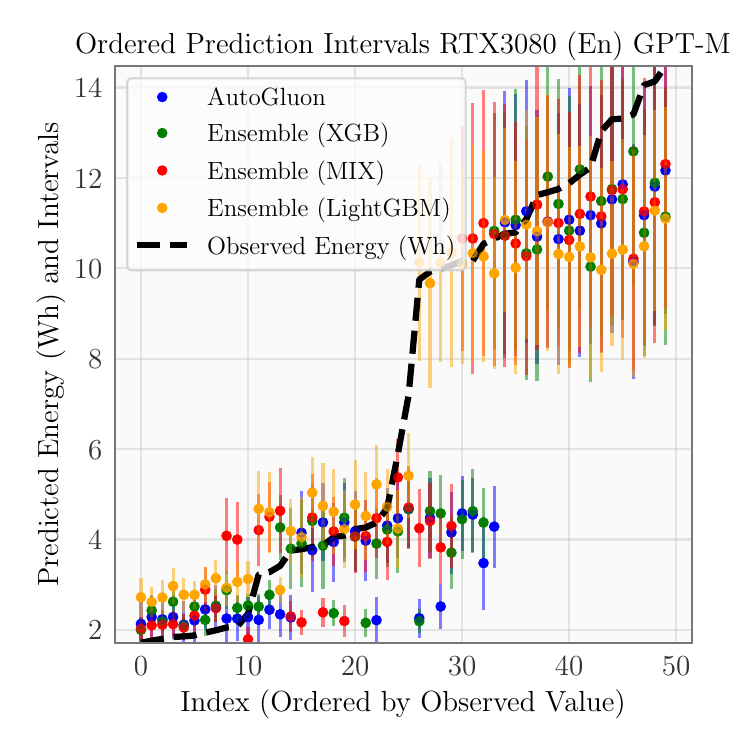}
\end{minipage}
\begin{minipage}{.16\linewidth}
  \centering
\includegraphics[width=\linewidth]{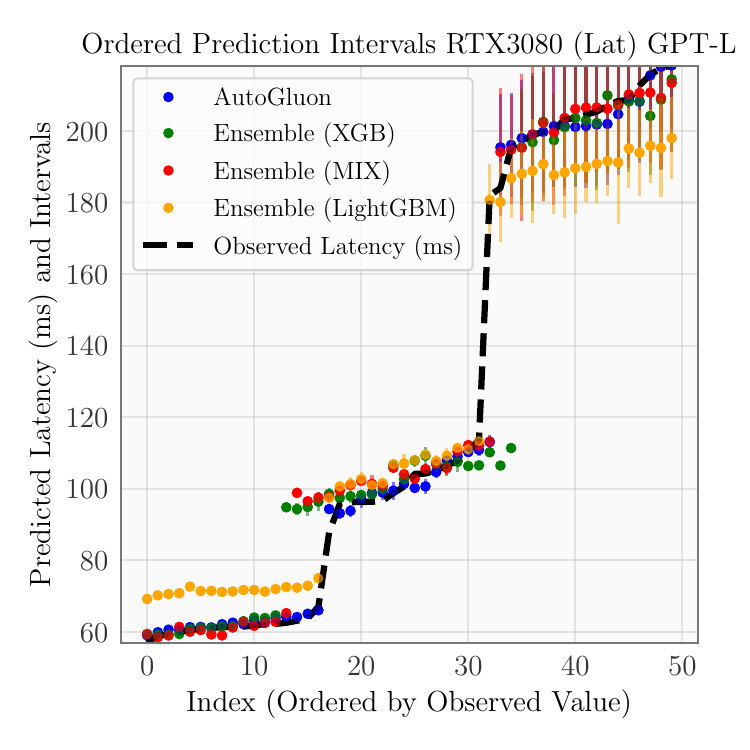}
\end{minipage}
\begin{minipage}{.16\linewidth}
  \centering
\includegraphics[width=\linewidth]{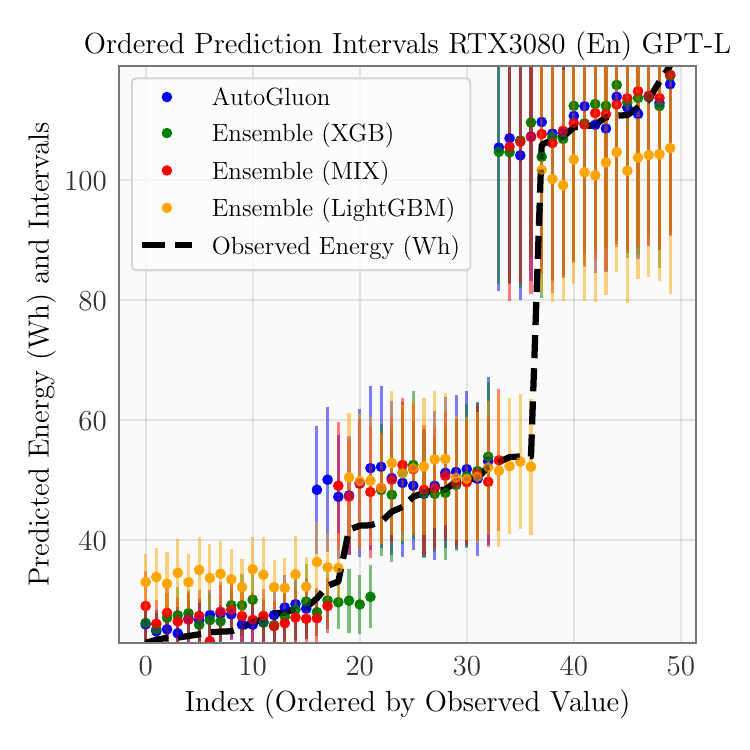}
\end{minipage}
\caption{Calibration Areas RTX3080}
\label{fig:calib_area_rtx3080}
\vspace{-5mm}
\end{figure}

\begin{figure}[t]
\centering
\begin{minipage}{.16\linewidth}
  \centering
  \includegraphics[width=\linewidth]{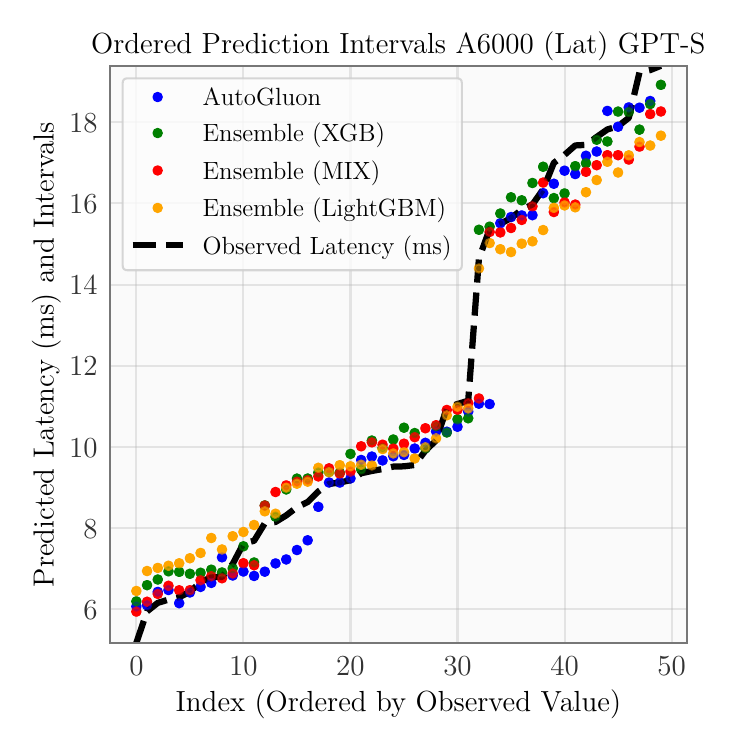}
\end{minipage}%
\begin{minipage}{.16\linewidth}
  \centering
\includegraphics[width=\linewidth]{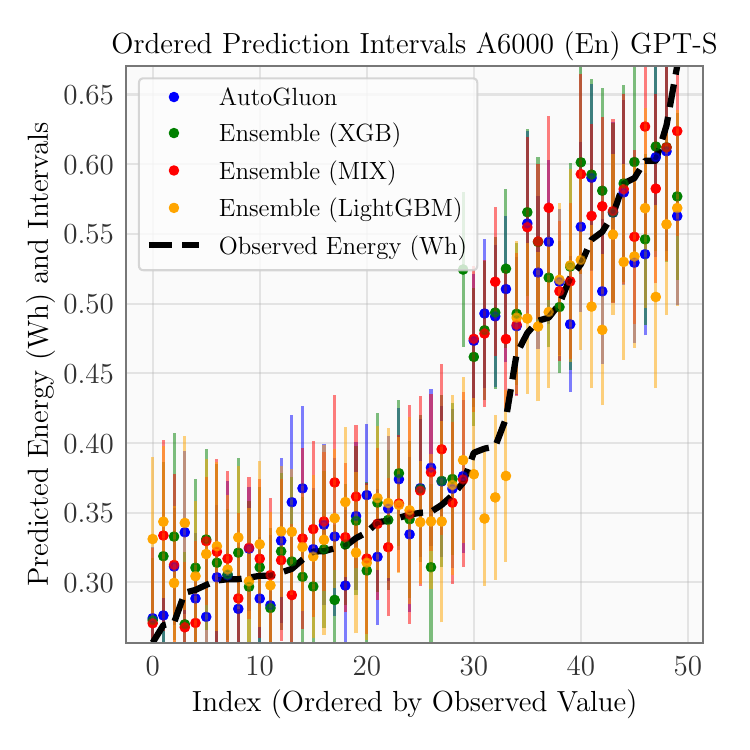}
\end{minipage}
\begin{minipage}{.16\linewidth}
  \centering
\includegraphics[width=\linewidth]{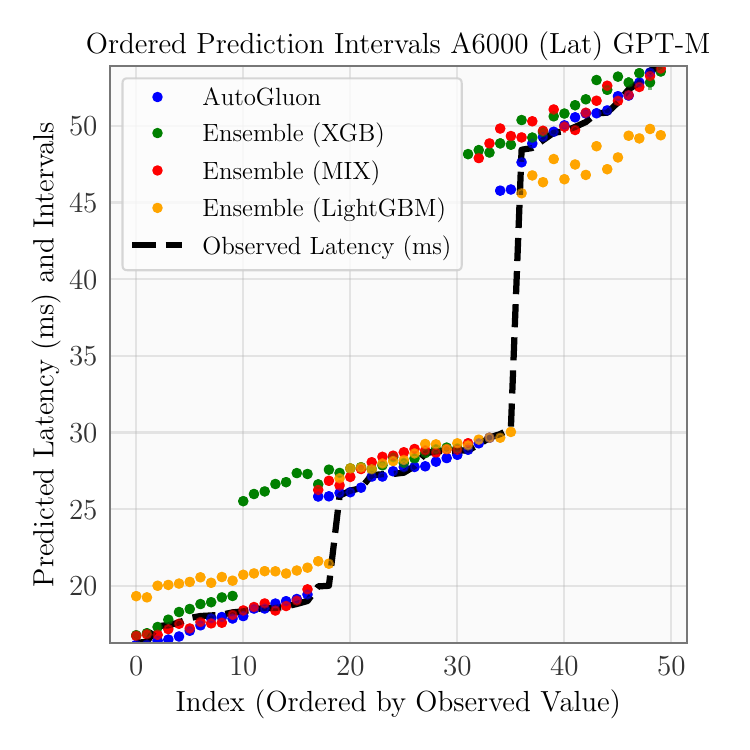}
\end{minipage}
\begin{minipage}{.16\linewidth}
  \centering
\includegraphics[width=\linewidth]{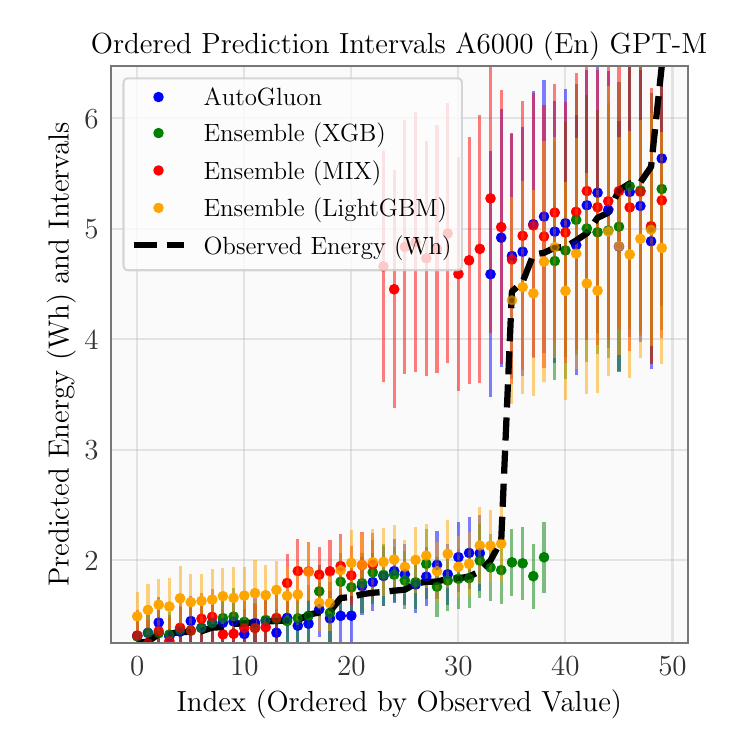}
\end{minipage}
\begin{minipage}{.16\linewidth}
  \centering
\includegraphics[width=\linewidth]{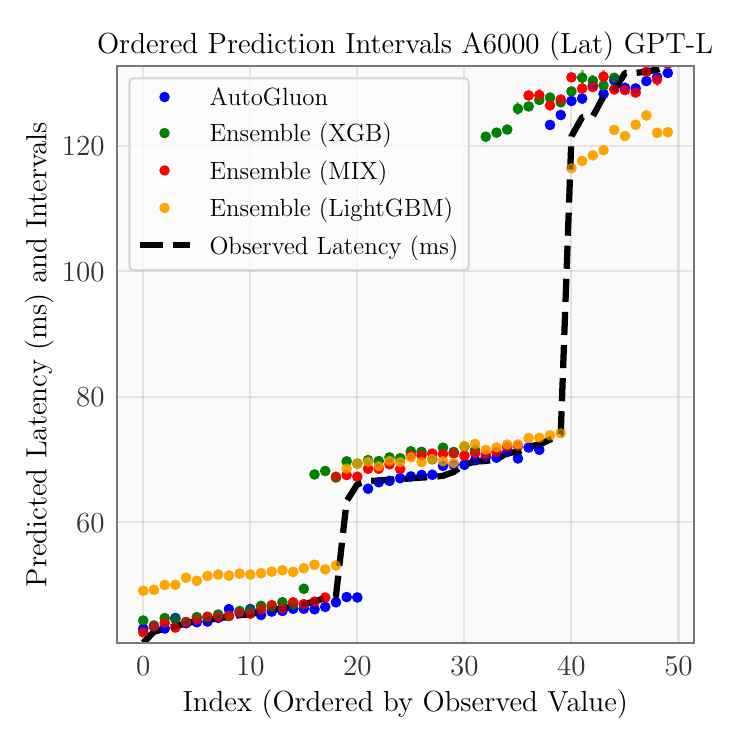}
\end{minipage}
\begin{minipage}{.16\linewidth}
  \centering
\includegraphics[width=\linewidth]{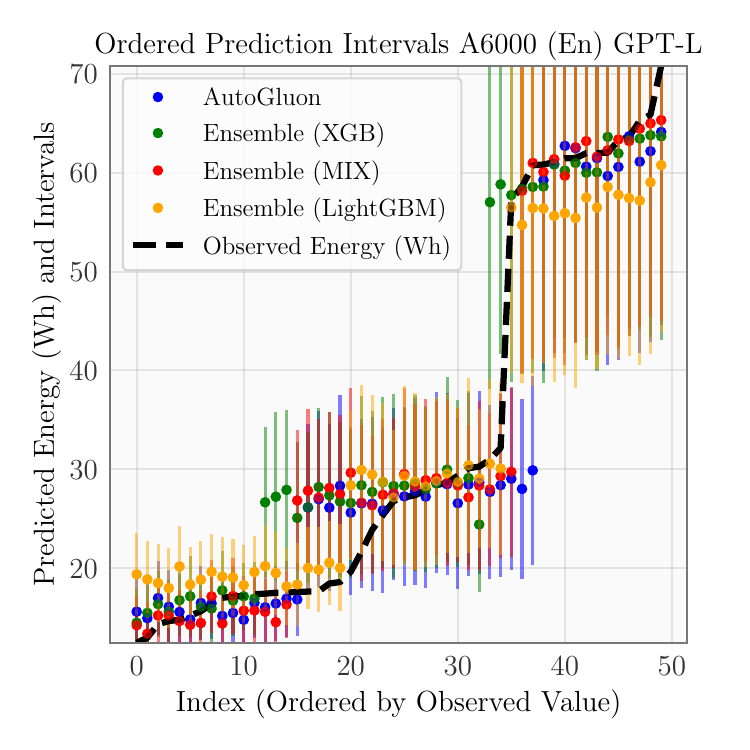}
\end{minipage}
\caption{Calibration Areas A6000}
\label{fig:calib_area_a6000}
\vspace{-5mm}
\end{figure}

\begin{figure}[t]
\centering
\begin{minipage}{.16\linewidth}
  \centering
  \includegraphics[width=\linewidth]{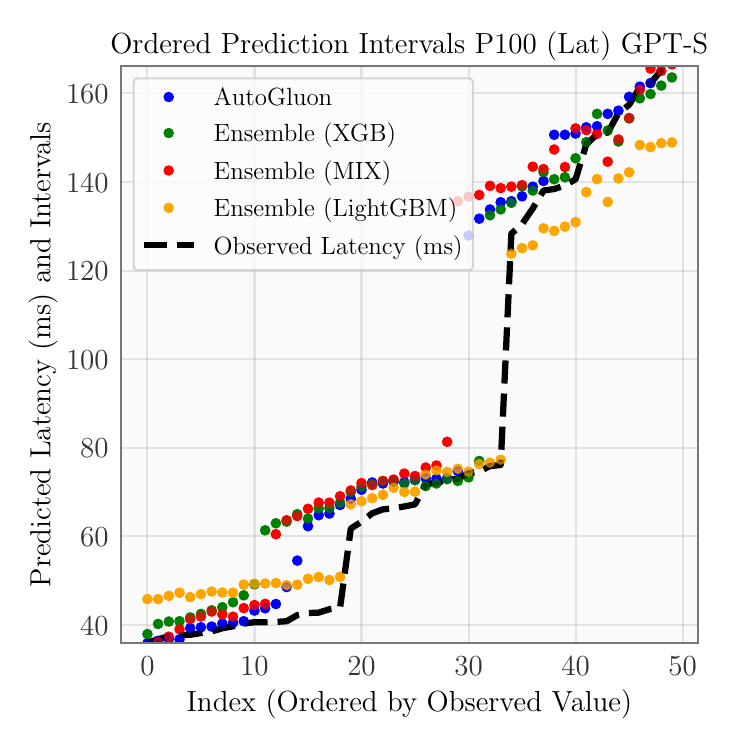}
\end{minipage}%
\begin{minipage}{.16\linewidth}
  \centering
\includegraphics[width=\linewidth]{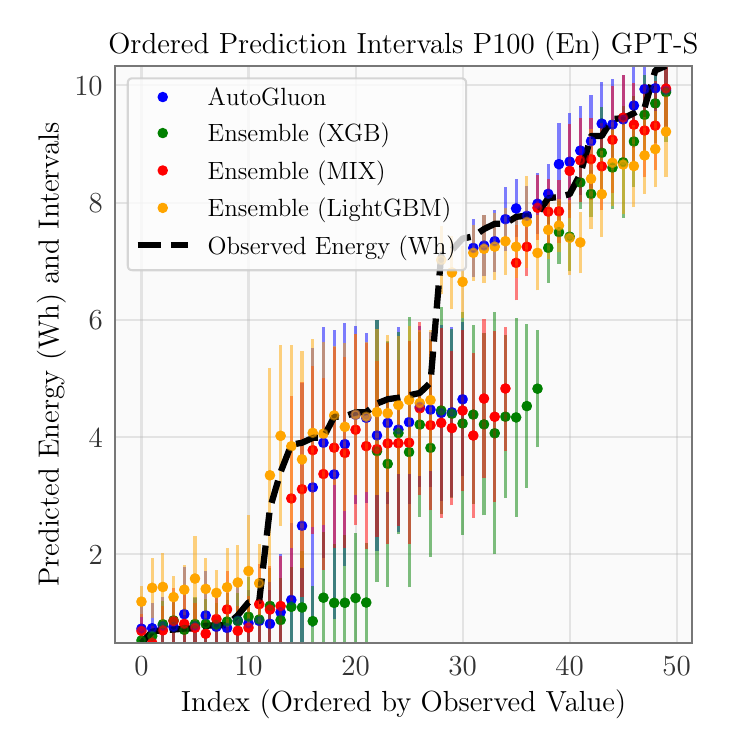}
\end{minipage}
\begin{minipage}{.16\linewidth}
  \centering
\includegraphics[width=\linewidth]{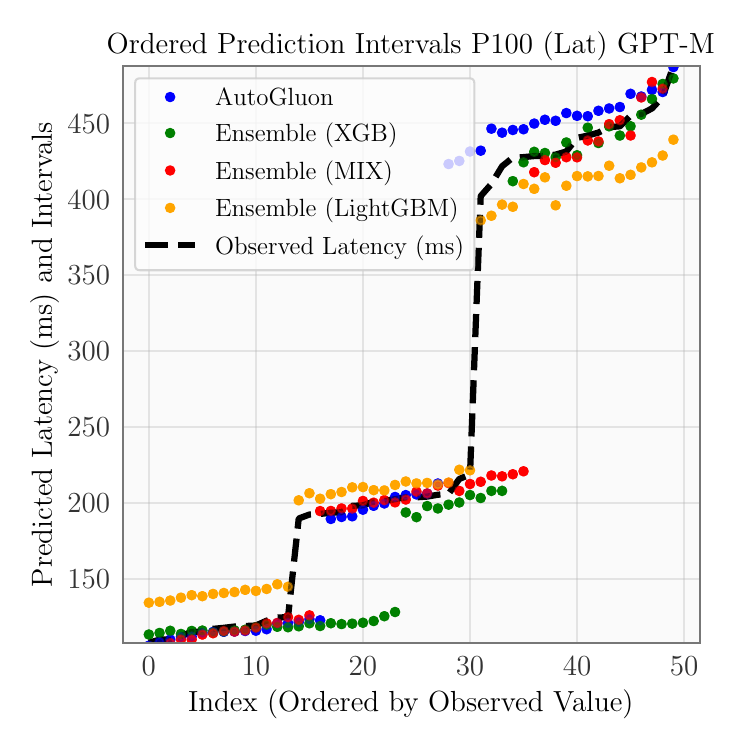}
\end{minipage}
\begin{minipage}{.16\linewidth}
  \centering
\includegraphics[width=\linewidth]{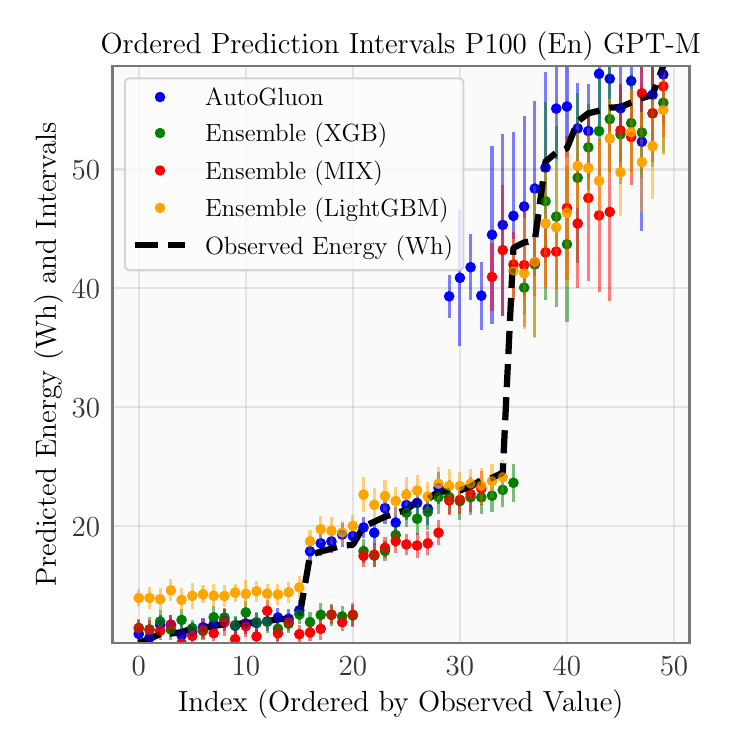}
\end{minipage}
\begin{minipage}{.16\linewidth}
  \centering
\includegraphics[width=\linewidth]{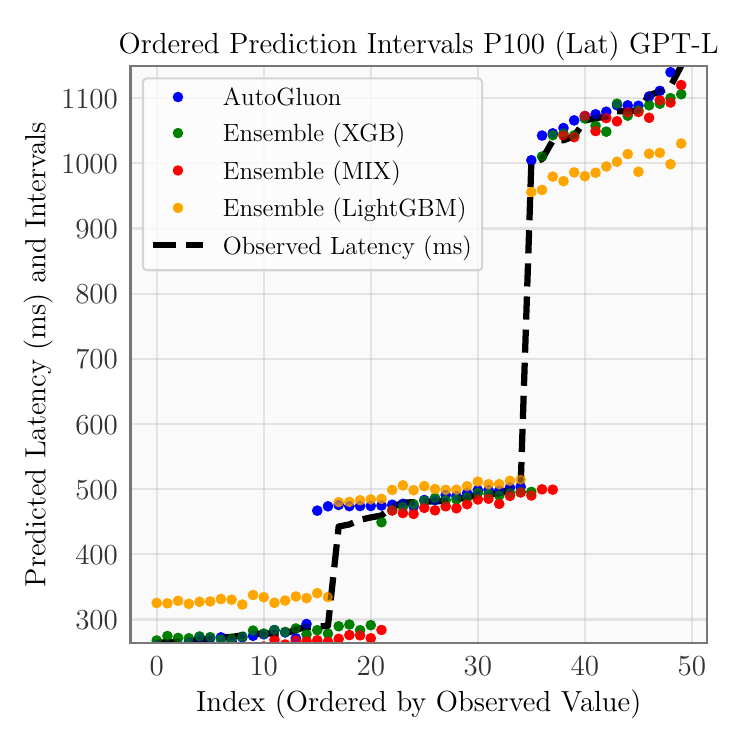}
\end{minipage}
\begin{minipage}{.16\linewidth}
  \centering
\includegraphics[width=\linewidth]{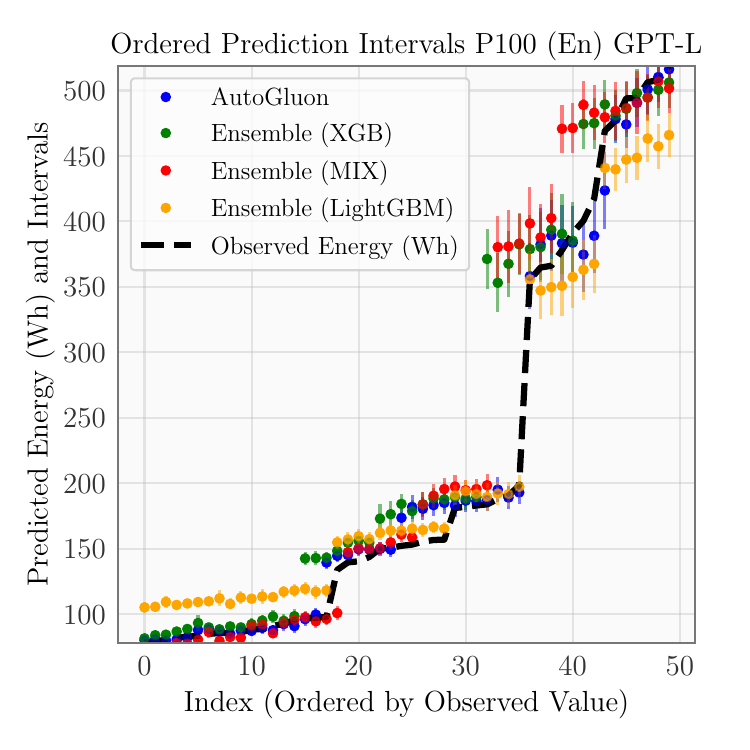}
\end{minipage}
\caption{Calibration Areas P100}
\label{fig:calib_area_p100}
\vspace{-5mm}
\end{figure}

\begin{figure}[t]
\centering
\begin{minipage}{.16\linewidth}
  \centering
  \includegraphics[width=\linewidth]{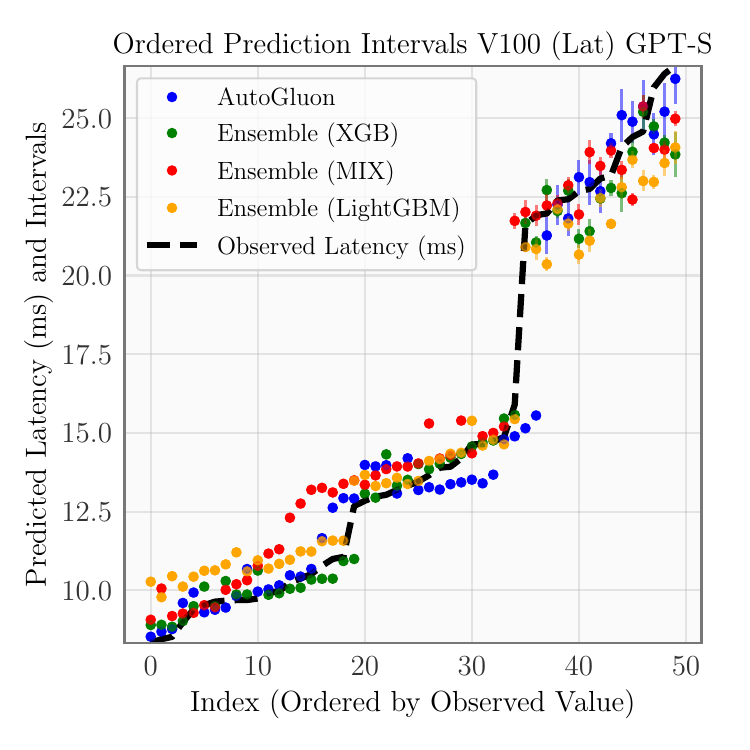}
\end{minipage}%
\begin{minipage}{.16\linewidth}
  \centering
\includegraphics[width=\linewidth]{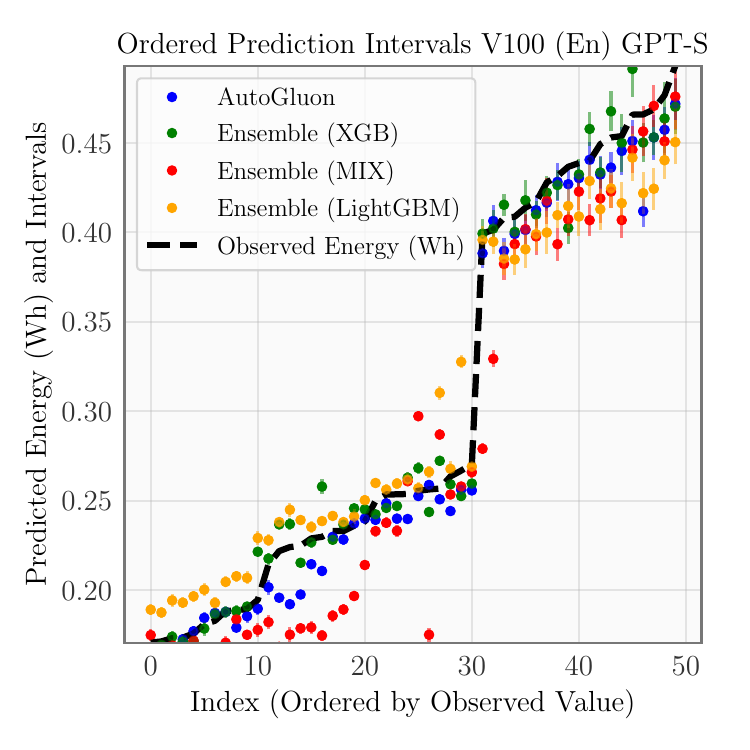}
\end{minipage}
\begin{minipage}{.16\linewidth}
  \centering
\includegraphics[width=\linewidth]{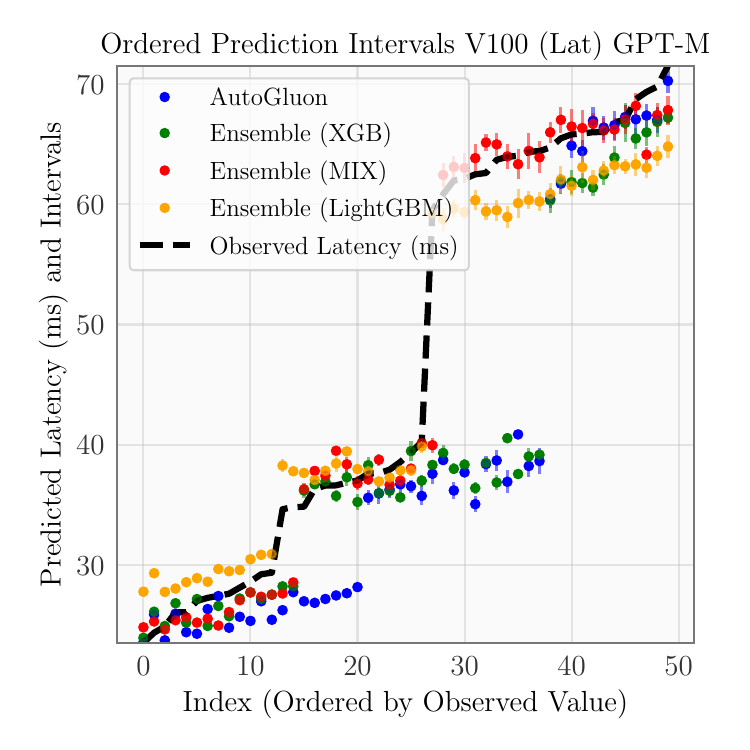}
\end{minipage}
\begin{minipage}{.16\linewidth}
  \centering
\includegraphics[width=\linewidth]{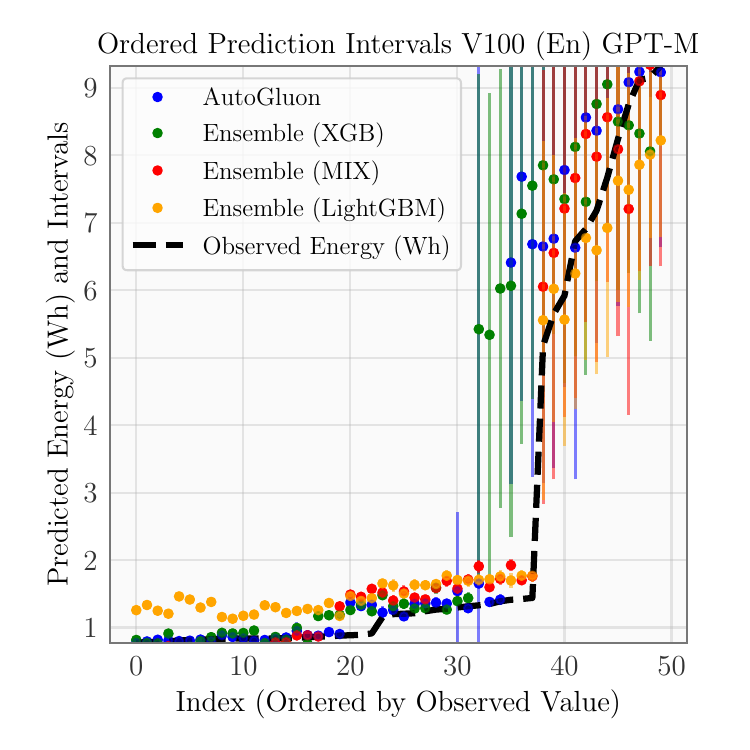}
\end{minipage}
\begin{minipage}{.16\linewidth}
  \centering
\includegraphics[width=\linewidth]{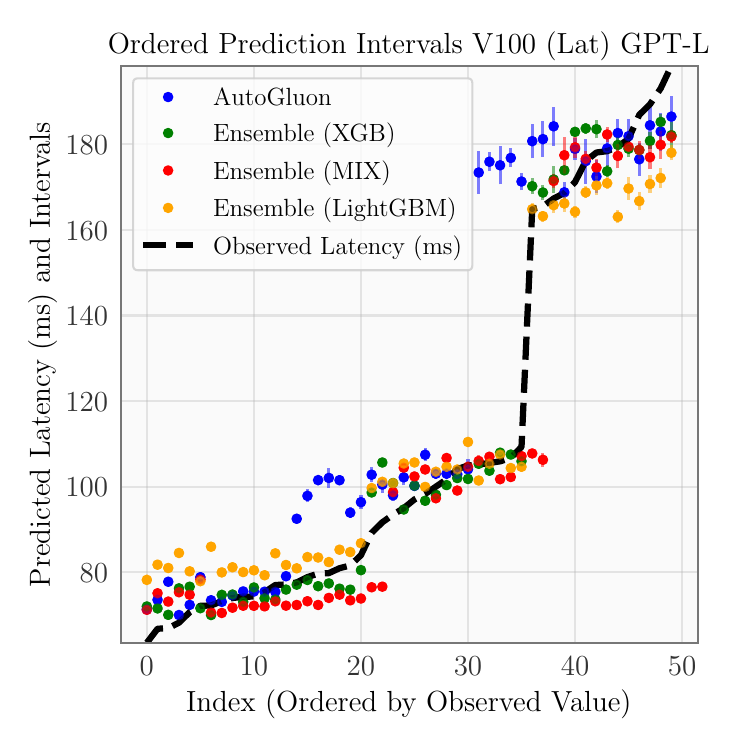}
\end{minipage}
\begin{minipage}{.16\linewidth}
  \centering
\includegraphics[width=\linewidth]{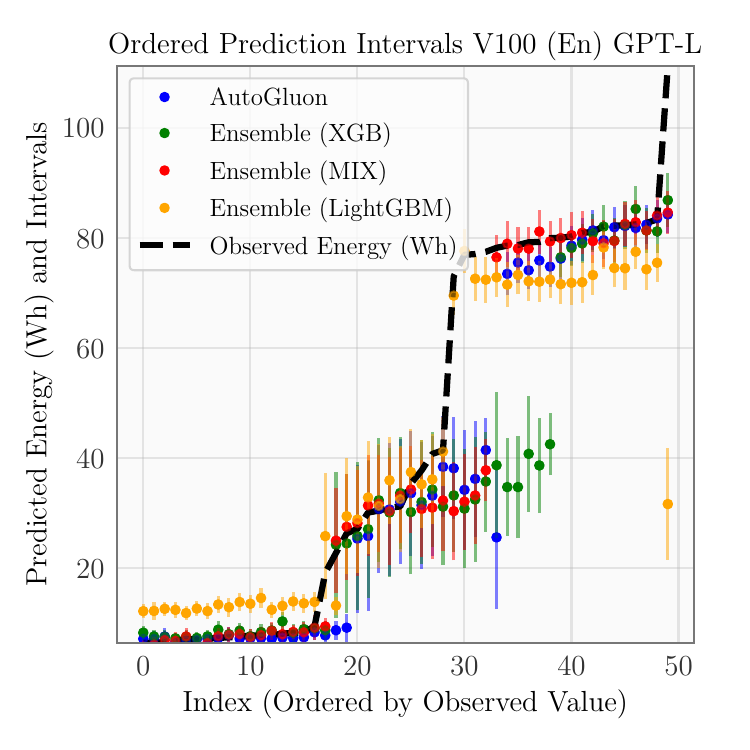}
\end{minipage}
\caption{Calibration Areas V100}
\label{fig:calib_area_v100}
\vspace{-5mm}
\end{figure}

\begin{figure}[ht]
\centering
\begin{minipage}{.16\linewidth}
  \centering
  \includegraphics[width=\linewidth]{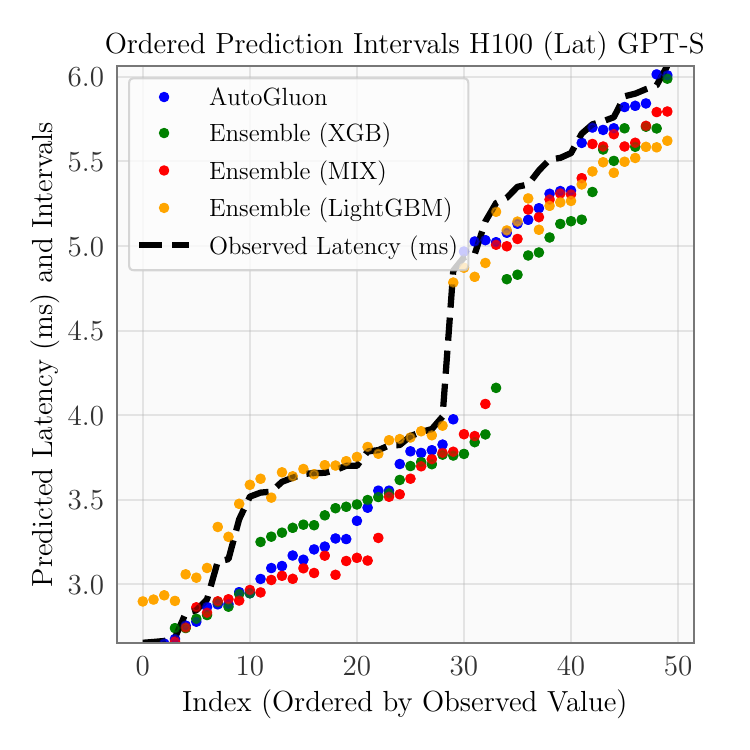}
\end{minipage}%
\begin{minipage}{.16\linewidth}
  \centering
\includegraphics[width=\linewidth]{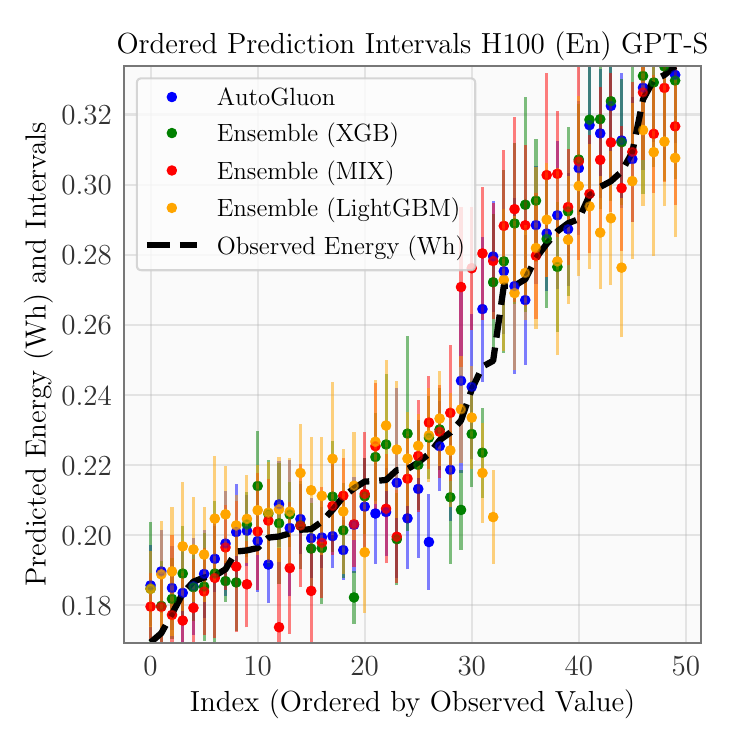}
\end{minipage}
\begin{minipage}{.16\linewidth}
  \centering
\includegraphics[width=\linewidth]{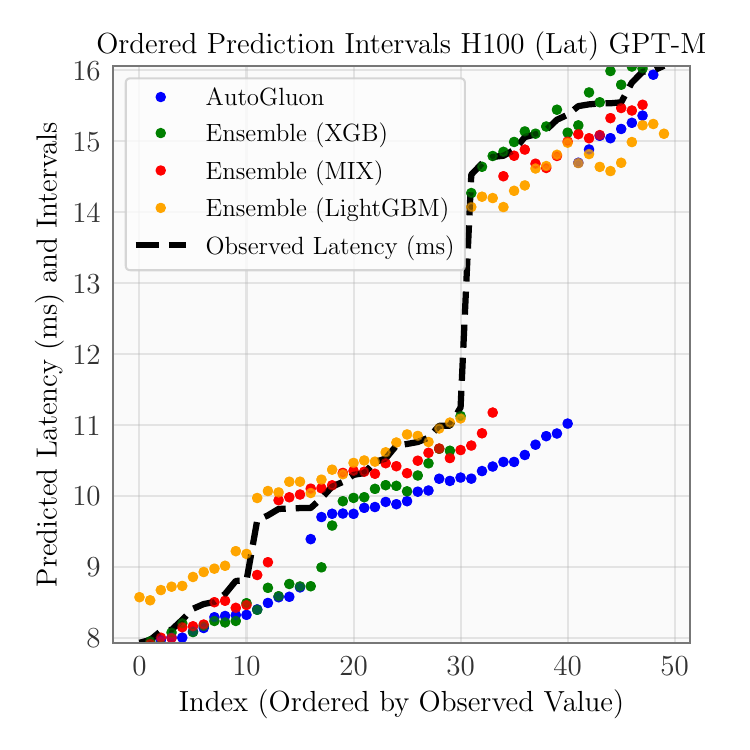}
\end{minipage}
\begin{minipage}{.16\linewidth}
  \centering
\includegraphics[width=\linewidth]{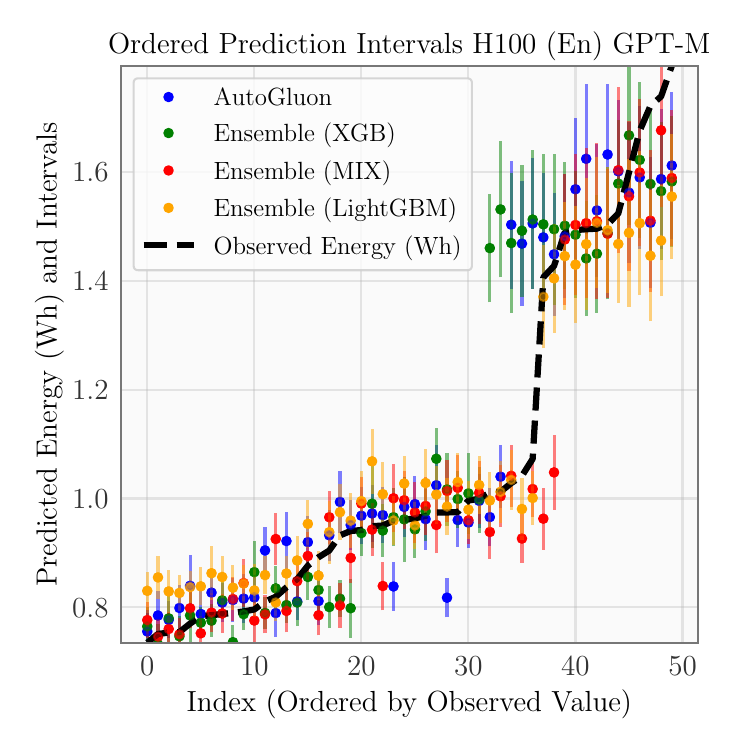}
\end{minipage}
\begin{minipage}{.16\linewidth}
  \centering
\includegraphics[width=\linewidth]{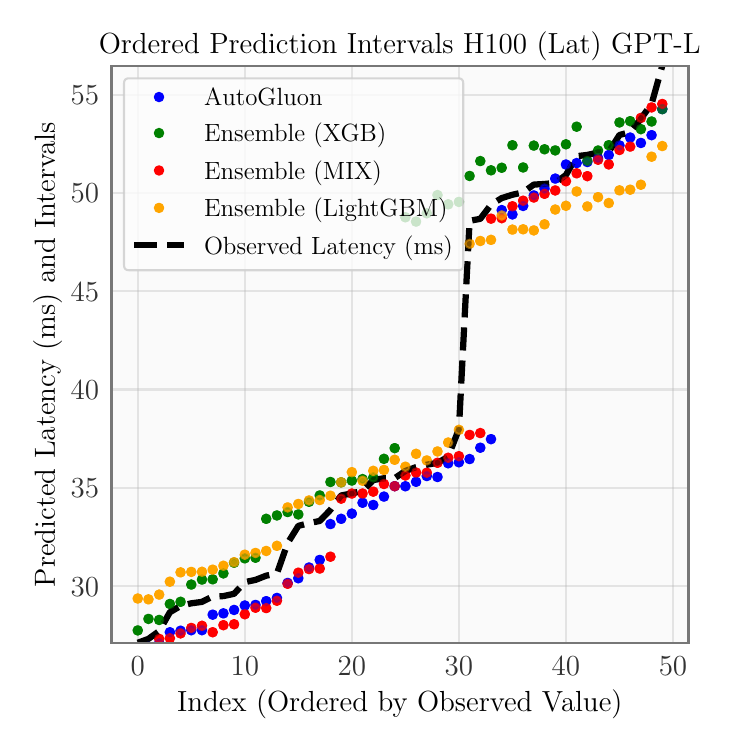}
\end{minipage}
\begin{minipage}{.16\linewidth}
  \centering
\includegraphics[width=\linewidth]{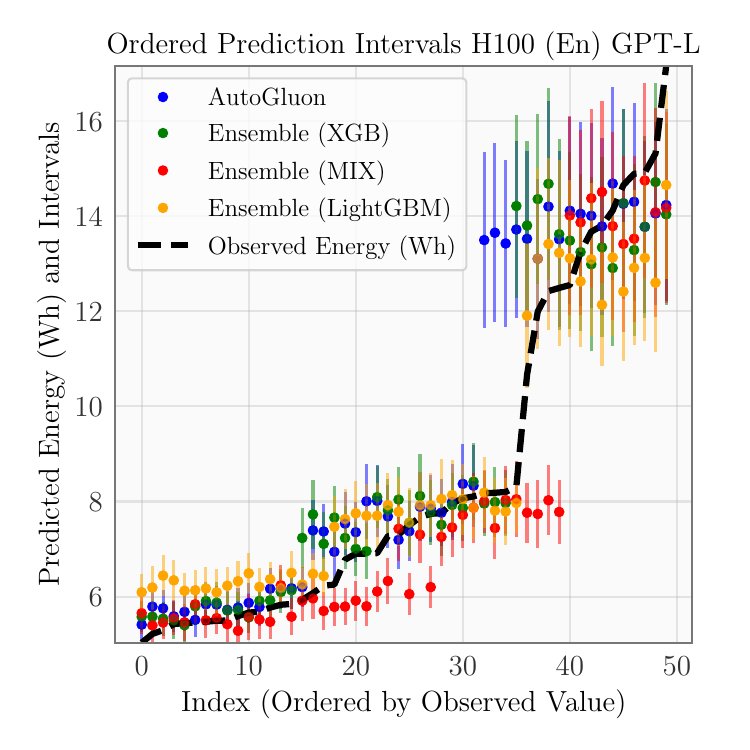}
\end{minipage}
\caption{Calibration Areas H100}
\label{fig:calib_area_h100}
\vspace{-5mm}
\end{figure}

\begin{figure}[ht]
\centering
\begin{minipage}{.16\linewidth}
  \centering
  \includegraphics[width=\linewidth]{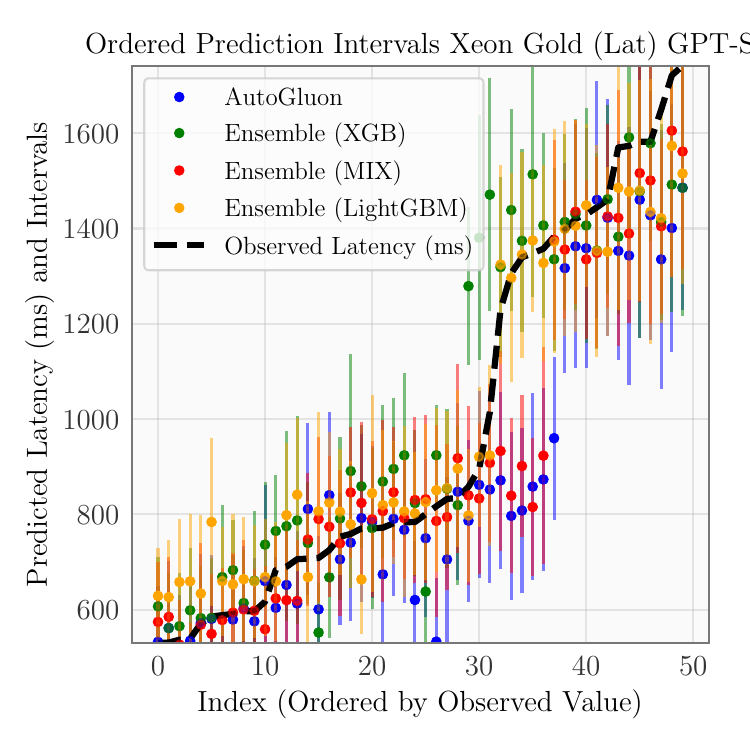}
\end{minipage}%
\begin{minipage}{.16\linewidth}
  \centering
\includegraphics[width=\linewidth]{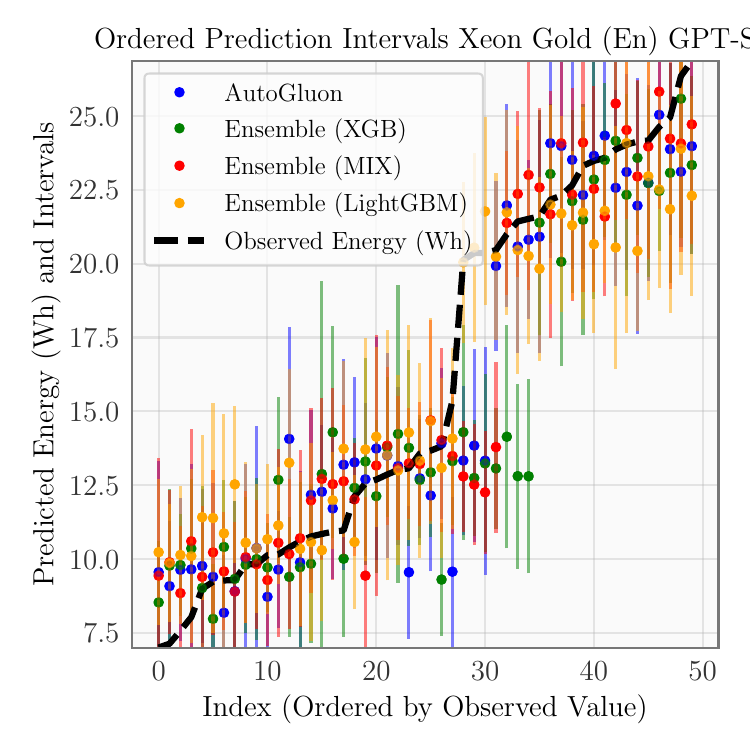}
\end{minipage}
\begin{minipage}{.16\linewidth}
  \centering
\includegraphics[width=\linewidth]{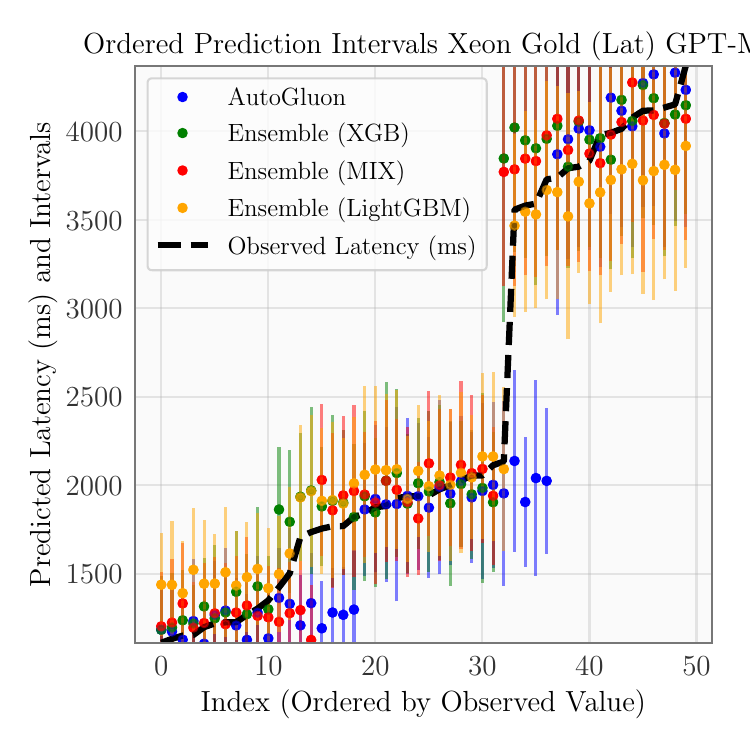}
\end{minipage}
\begin{minipage}{.16\linewidth}
  \centering
\includegraphics[width=\linewidth]{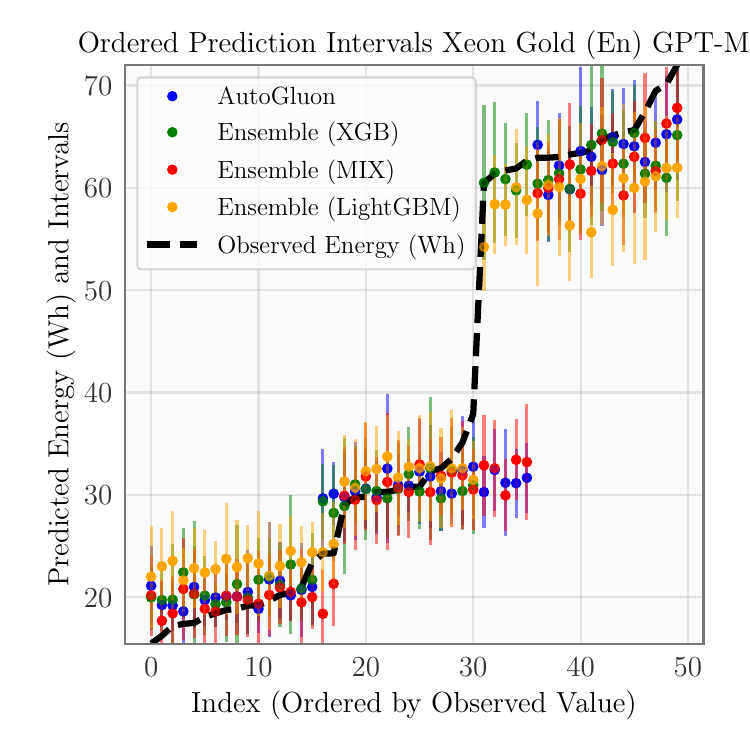}
\end{minipage}
\begin{minipage}{.16\linewidth}
  \centering
\includegraphics[width=\linewidth]{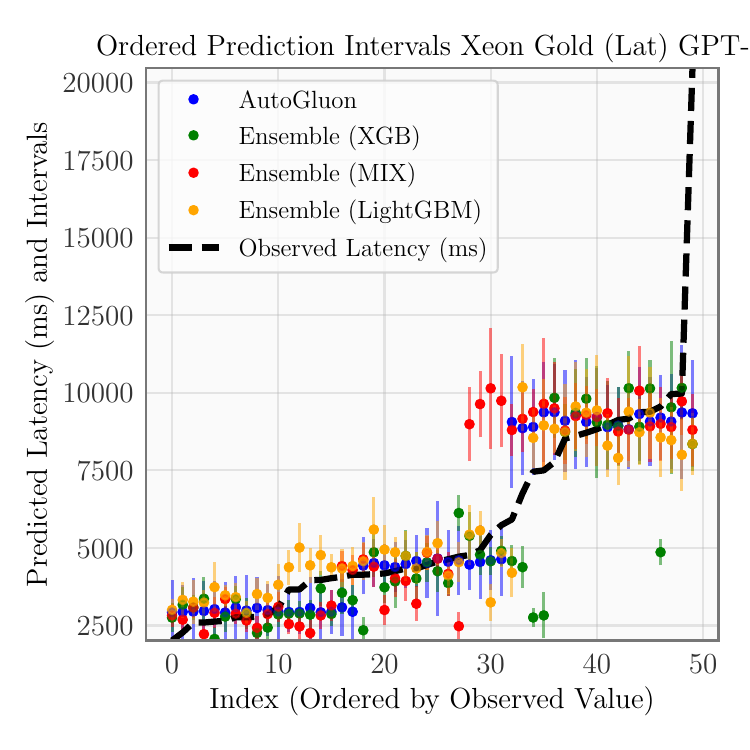}
\end{minipage}
\begin{minipage}{.16\linewidth}
  \centering
\includegraphics[width=\linewidth]{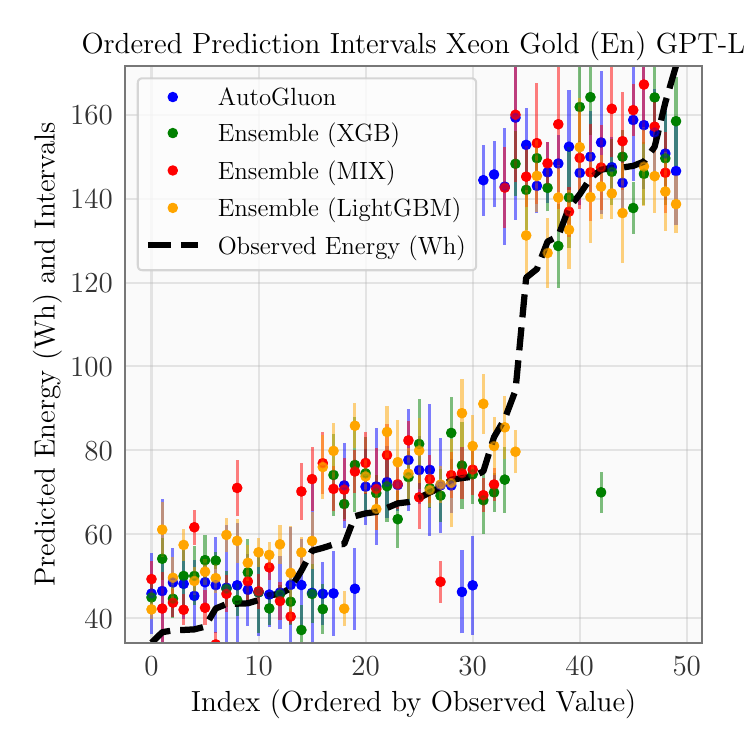}
\end{minipage}
\caption{Calibration Areas Xeon Gold}
\label{fig:calib_area_xeon_gold}
\vspace{-5mm}
\end{figure}

\begin{figure}[ht]
\centering
\begin{minipage}{.16\linewidth}
  \centering
  \includegraphics[width=\linewidth]{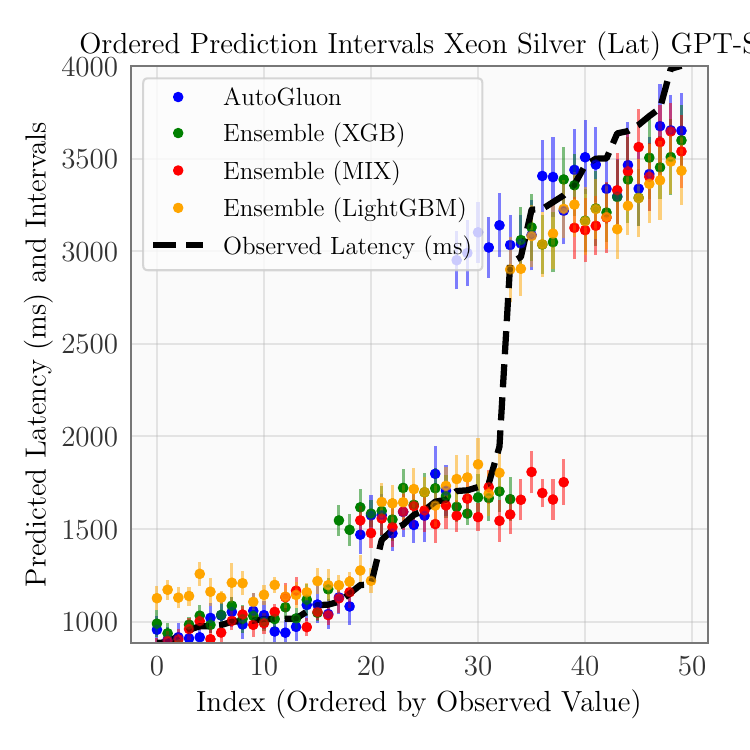}
\end{minipage}%
\begin{minipage}{.16\linewidth}
  \centering
\includegraphics[width=\linewidth]{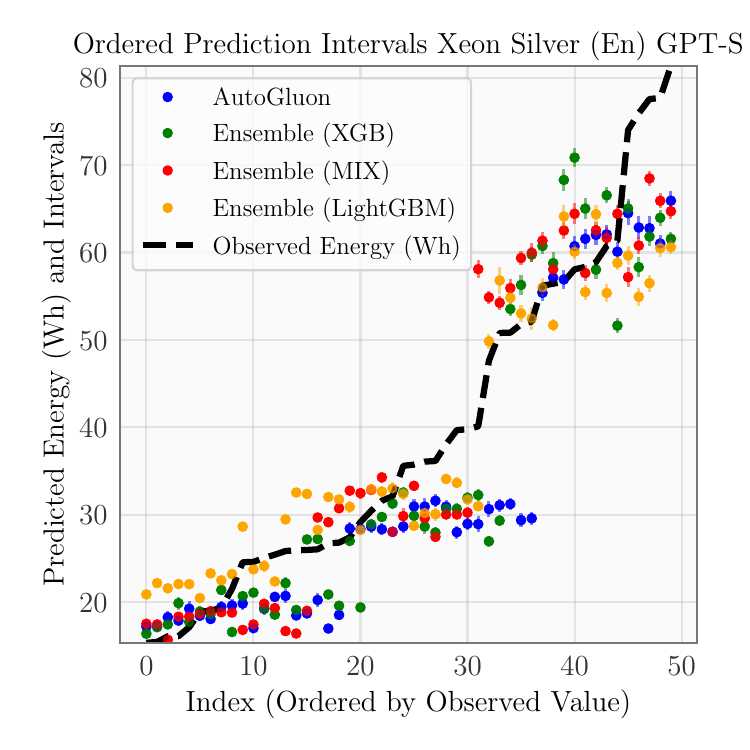}
\end{minipage}
\begin{minipage}{.16\linewidth}
  \centering
\includegraphics[width=\linewidth]{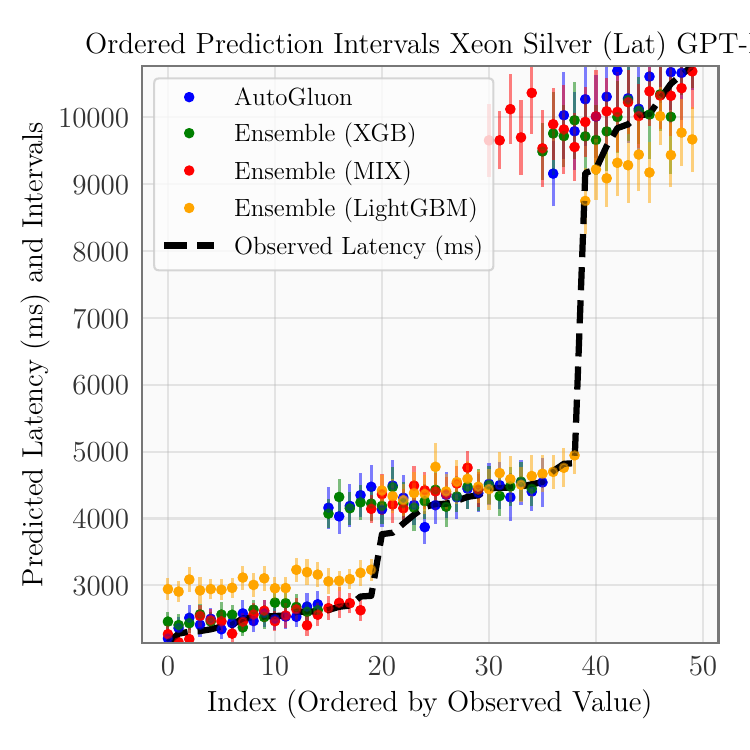}
\end{minipage}
\begin{minipage}{.16\linewidth}
  \centering
\includegraphics[width=\linewidth]{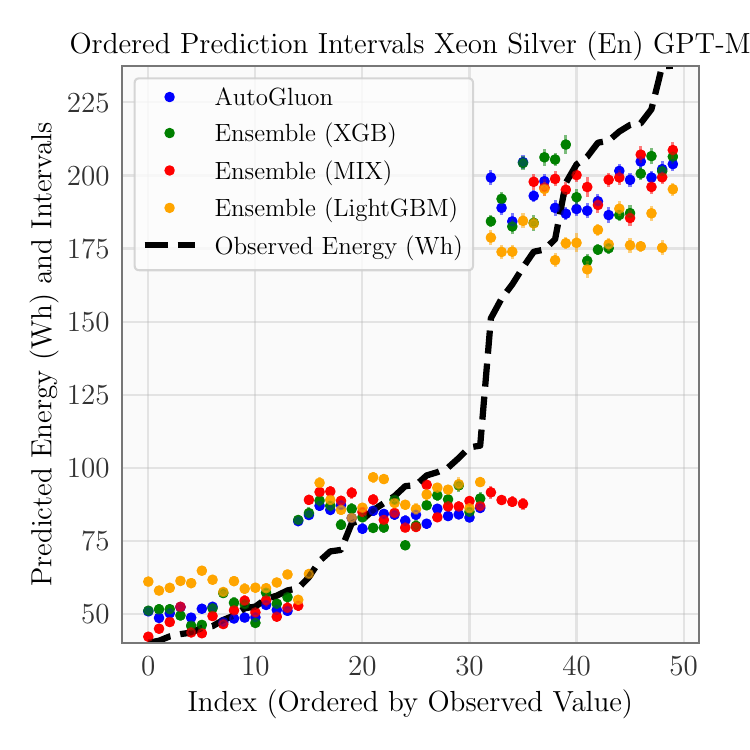}
\end{minipage}
\begin{minipage}{.16\linewidth}
  \centering
\includegraphics[width=\linewidth]{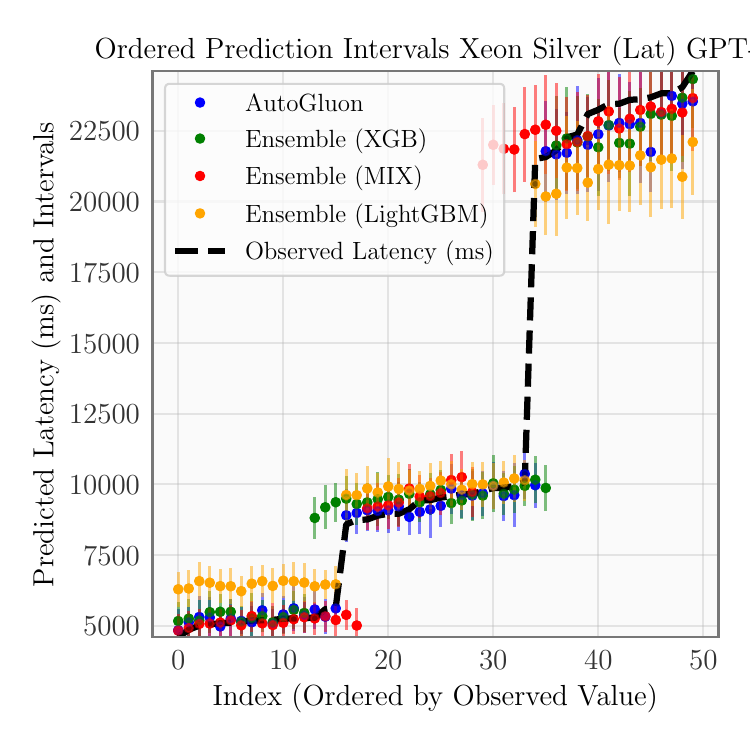}
\end{minipage}
\begin{minipage}{.16\linewidth}
  \centering
\includegraphics[width=\linewidth]{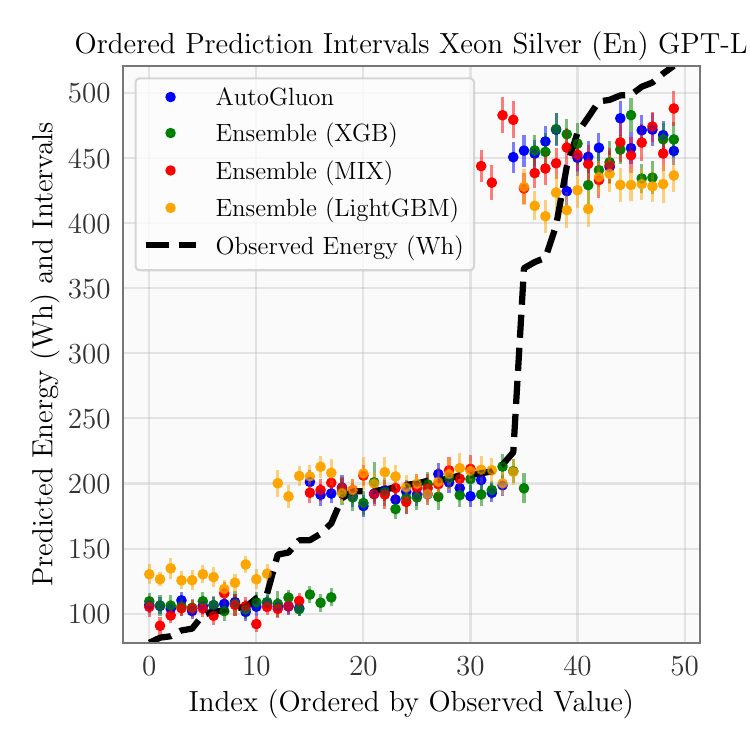}
\end{minipage}
\caption{Calibration Areas Xeon Silver}
\label{fig:calib_area_xeon_silver}
\vspace{-5mm}
\end{figure}

\begin{figure}[ht]
\centering
\begin{minipage}{.16\linewidth}
  \centering
  \includegraphics[width=\linewidth]{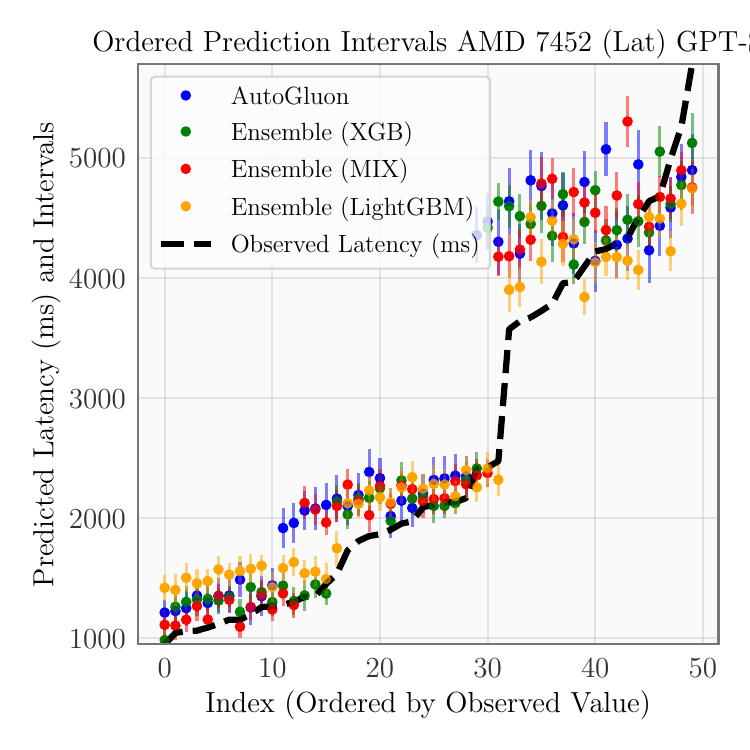}
\end{minipage}%
\begin{minipage}{.16\linewidth}
  \centering
\includegraphics[width=\linewidth]{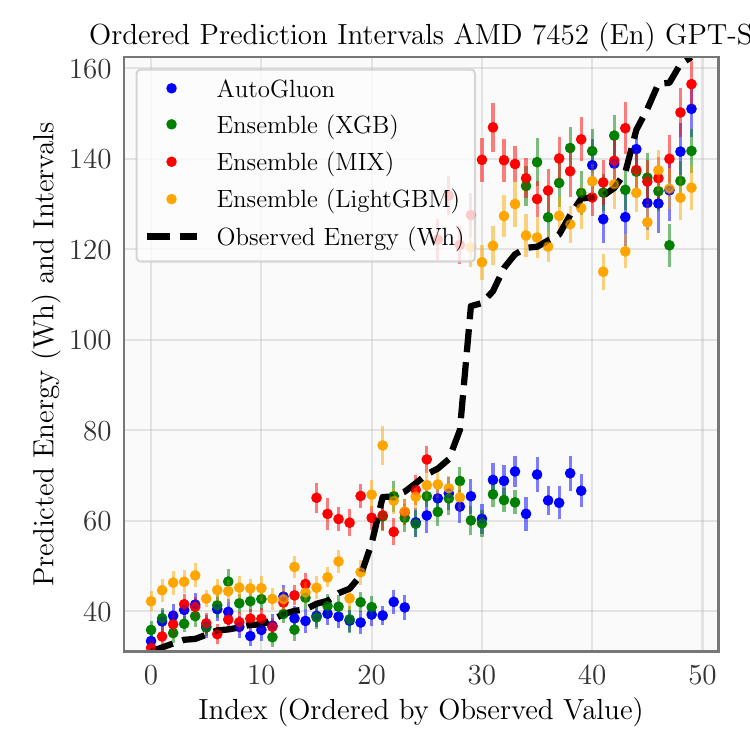}
\end{minipage}
\begin{minipage}{.16\linewidth}
  \centering
\includegraphics[width=\linewidth]{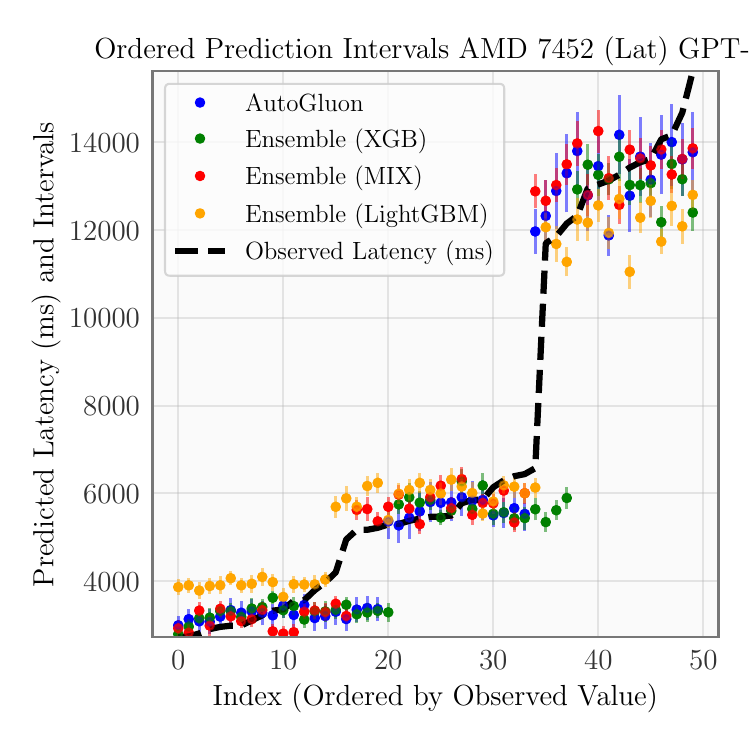}
\end{minipage}
\begin{minipage}{.16\linewidth}
  \centering
\includegraphics[width=\linewidth]{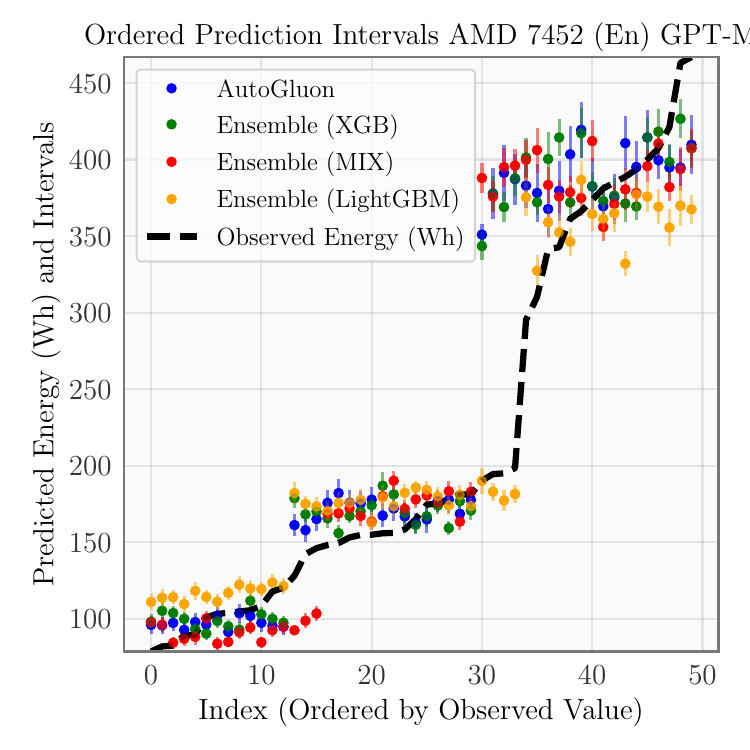}
\end{minipage}
\begin{minipage}{.16\linewidth}
  \centering
\includegraphics[width=\linewidth]{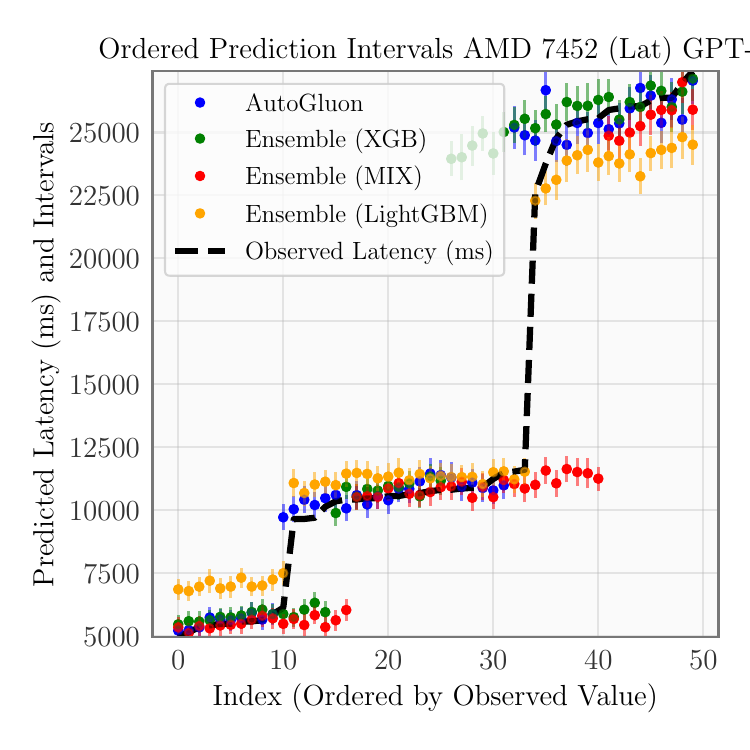}
\end{minipage}
\begin{minipage}{.16\linewidth}
  \centering
\includegraphics[width=\linewidth]{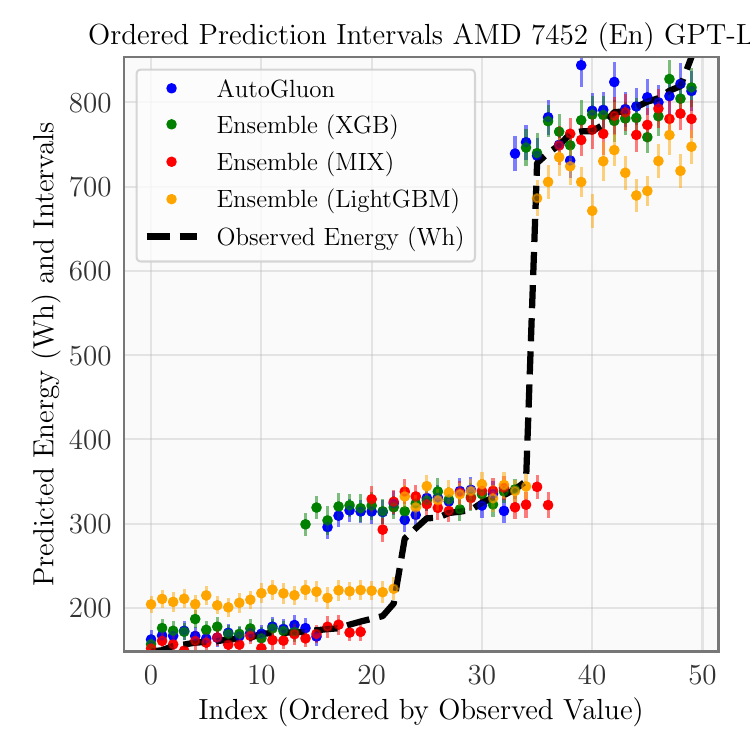}
\end{minipage}
\caption{Calibration Areas CPU AMD 7452}
\label{fig:calib_amd_7452}
\vspace{-5mm}
\end{figure}

\begin{figure}[ht]
\centering
\begin{minipage}{.16\linewidth}
  \centering
  \includegraphics[width=\linewidth]{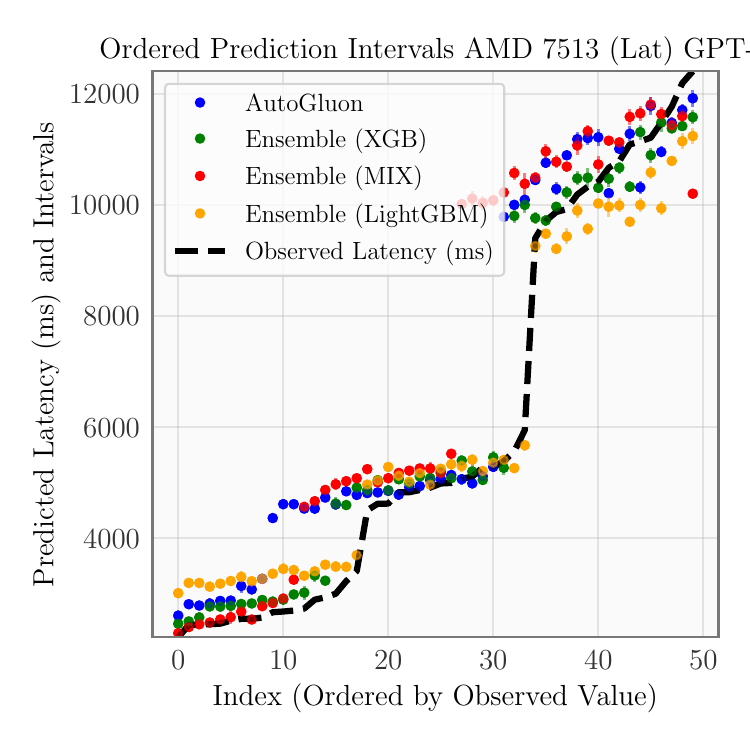}
\end{minipage}%
\begin{minipage}{.16\linewidth}
  \centering
\includegraphics[width=\linewidth]{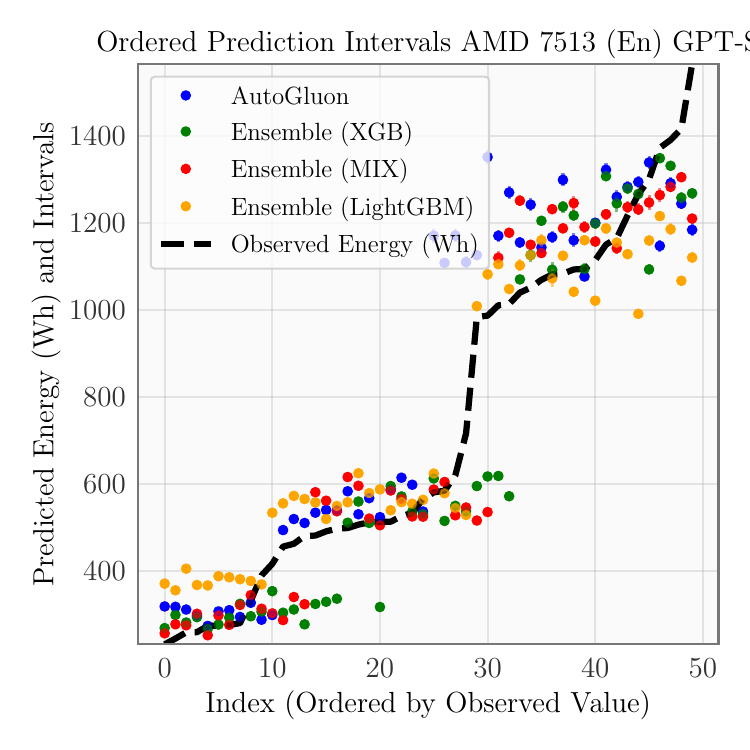}
\end{minipage}
\begin{minipage}{.16\linewidth}
  \centering
\includegraphics[width=\linewidth]{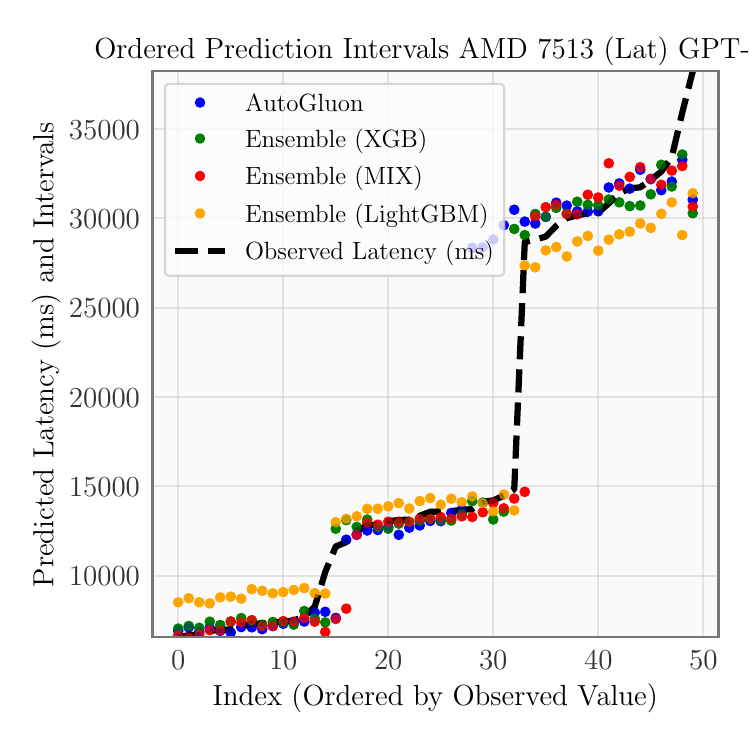}
\end{minipage}
\begin{minipage}{.16\linewidth}
  \centering
\includegraphics[width=\linewidth]{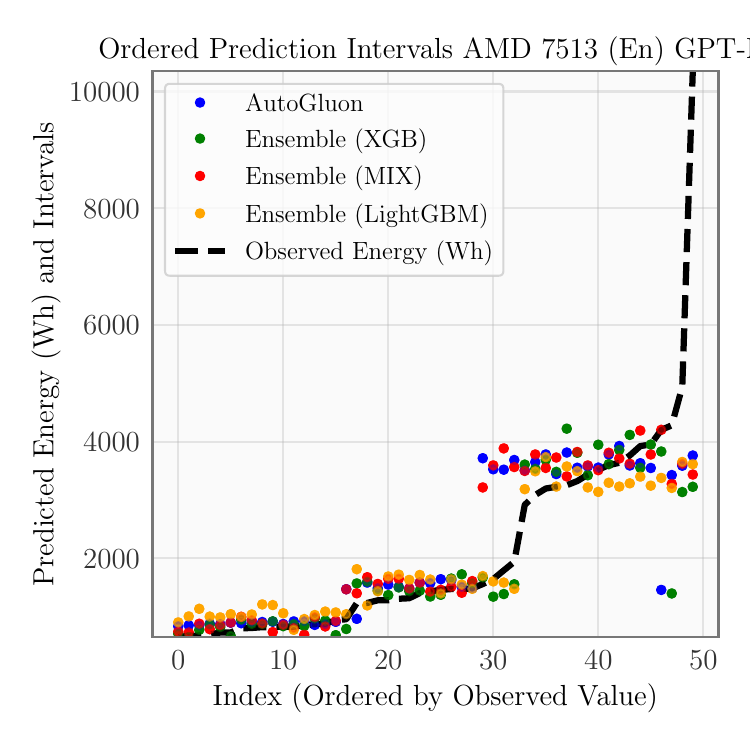}
\end{minipage}
\begin{minipage}{.16\linewidth}
  \centering
\includegraphics[width=\linewidth]{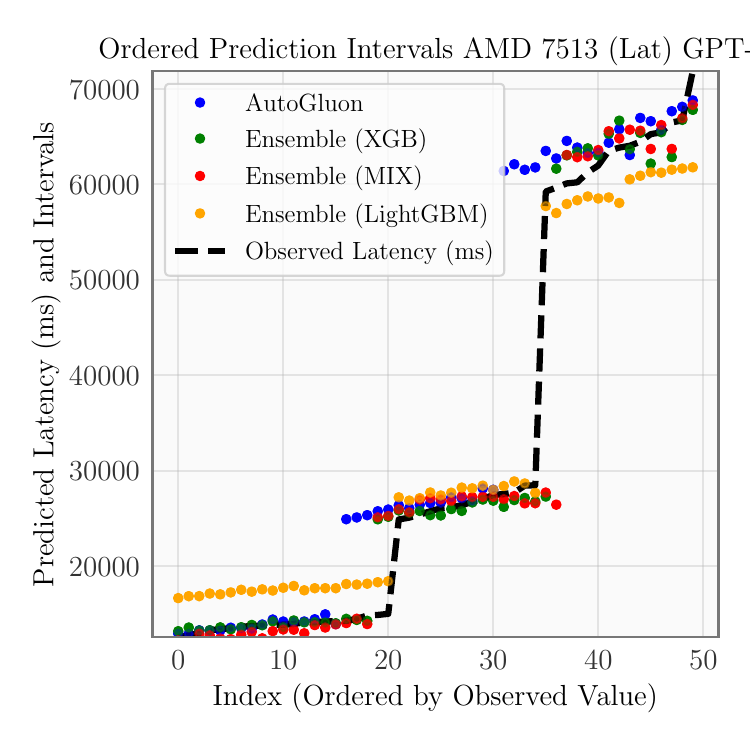}
\end{minipage}
\begin{minipage}{.16\linewidth}
  \centering
\includegraphics[width=\linewidth]{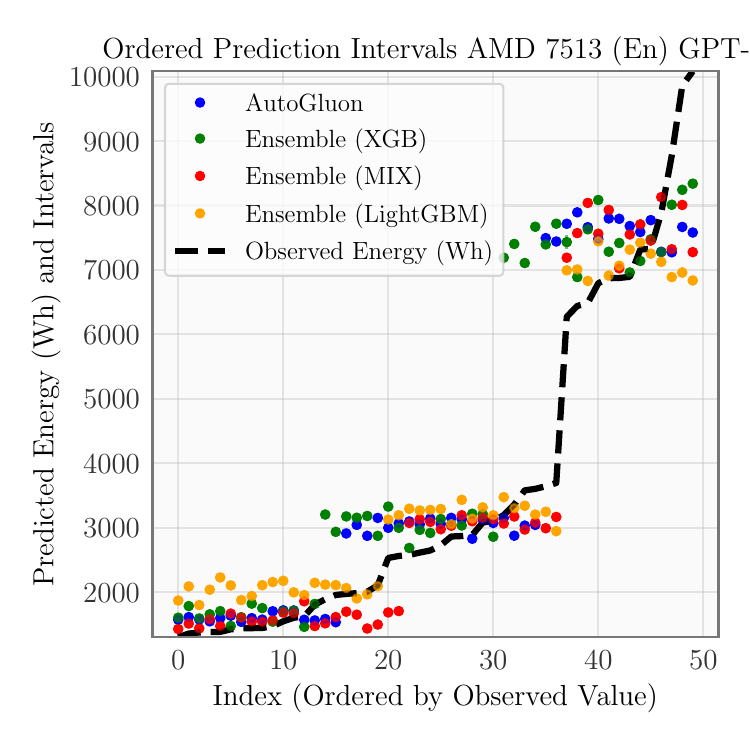}
\end{minipage}
\caption{Calibration Areas CPU AMD 7513}
\label{fig:calib_amd_7513}
\vspace{-5mm}
\end{figure}

\begin{figure}[ht]
\centering
\begin{minipage}{.16\linewidth}
  \centering
  \includegraphics[width=\linewidth]{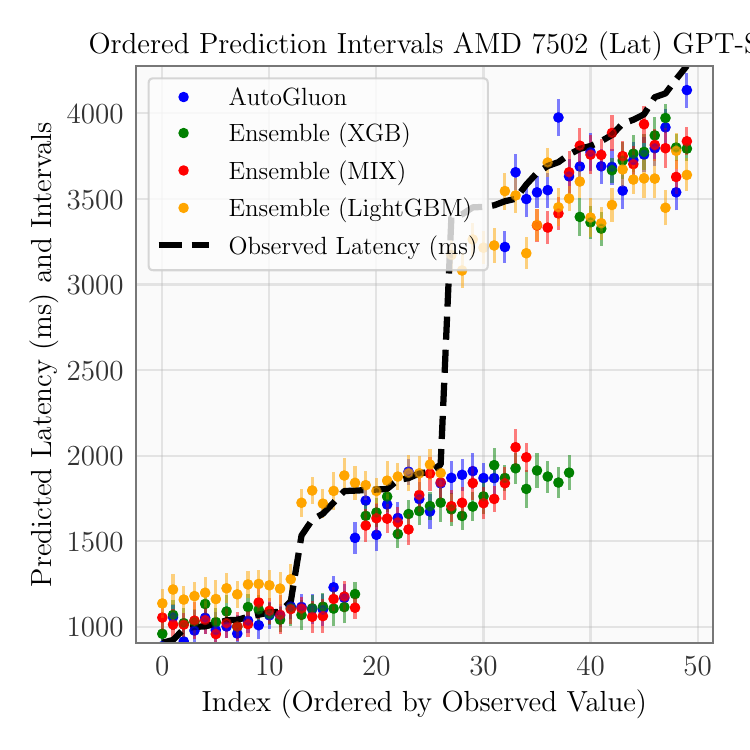}
\end{minipage}%
\begin{minipage}{.16\linewidth}
  \centering
\includegraphics[width=\linewidth]{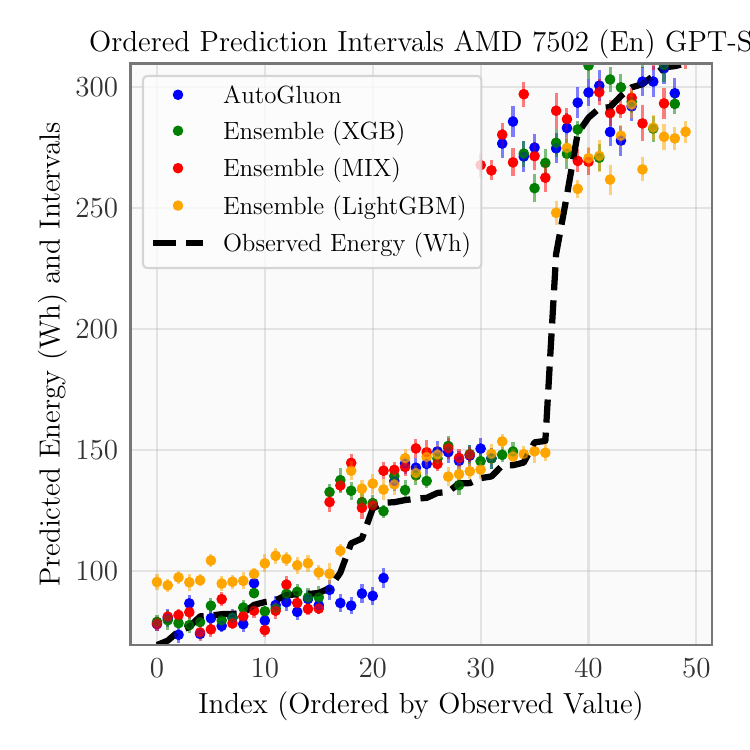}
\end{minipage}
\begin{minipage}{.16\linewidth}
  \centering
\includegraphics[width=\linewidth]{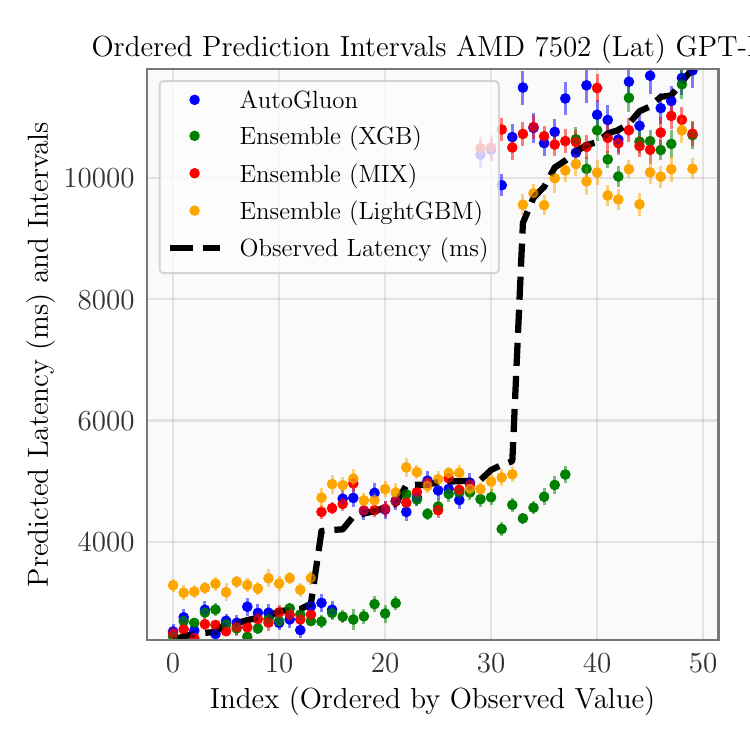}
\end{minipage}
\begin{minipage}{.16\linewidth}
  \centering
\includegraphics[width=\linewidth]{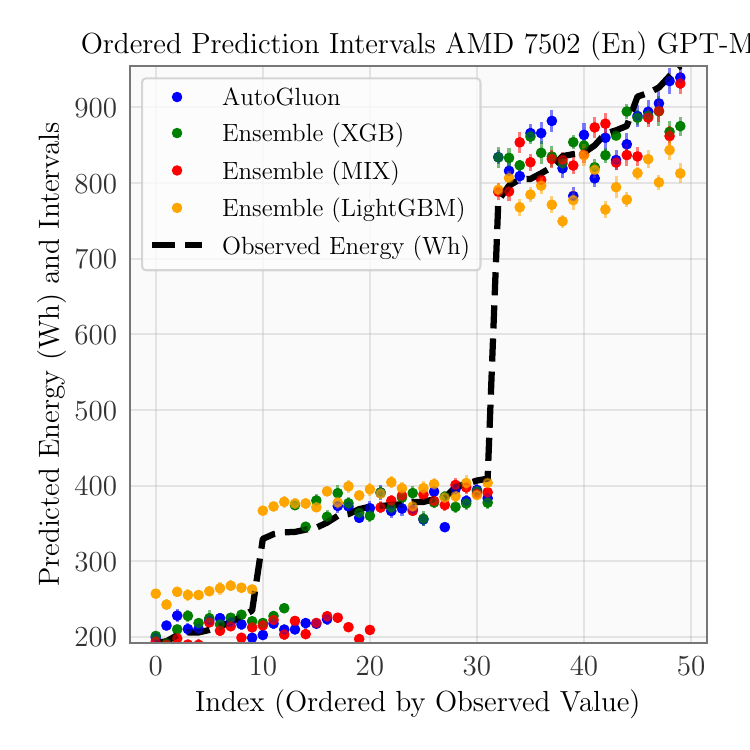}
\end{minipage}
\begin{minipage}{.16\linewidth}
  \centering
\includegraphics[width=\linewidth]{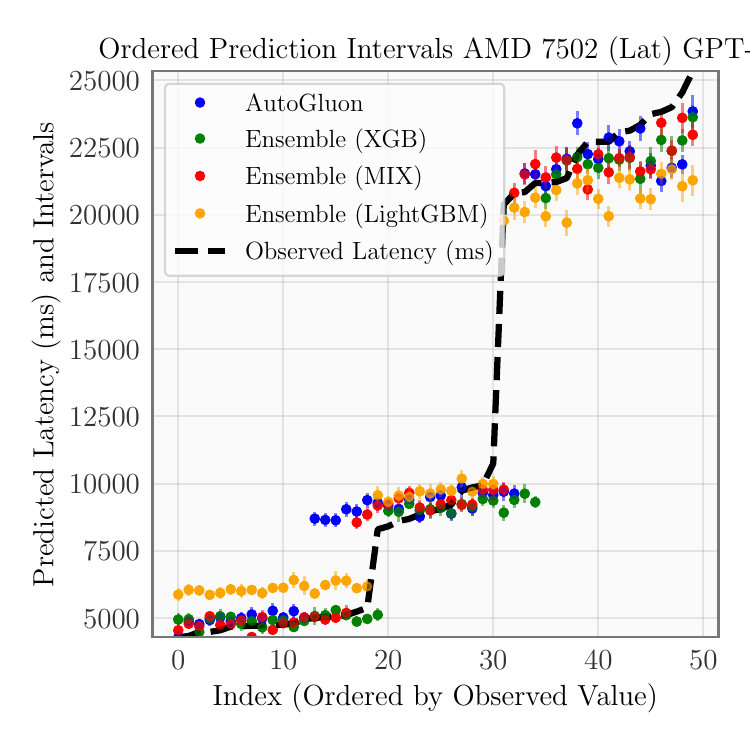}
\end{minipage}
\begin{minipage}{.16\linewidth}
  \centering
\includegraphics[width=\linewidth]{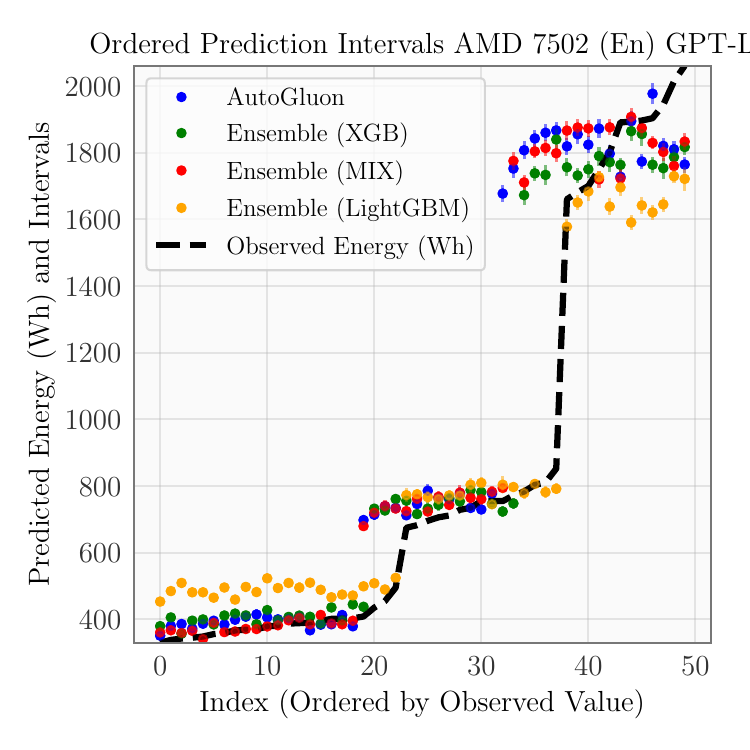}
\end{minipage}
\caption{Calibration Areas CPU AMD 7502}
\label{fig:calib_amd_7502}
\vspace{-5mm}
\end{figure}

\begin{figure}[H]

\begin{minipage}{.16\linewidth}
  \centering
  \includegraphics[width=\linewidth]{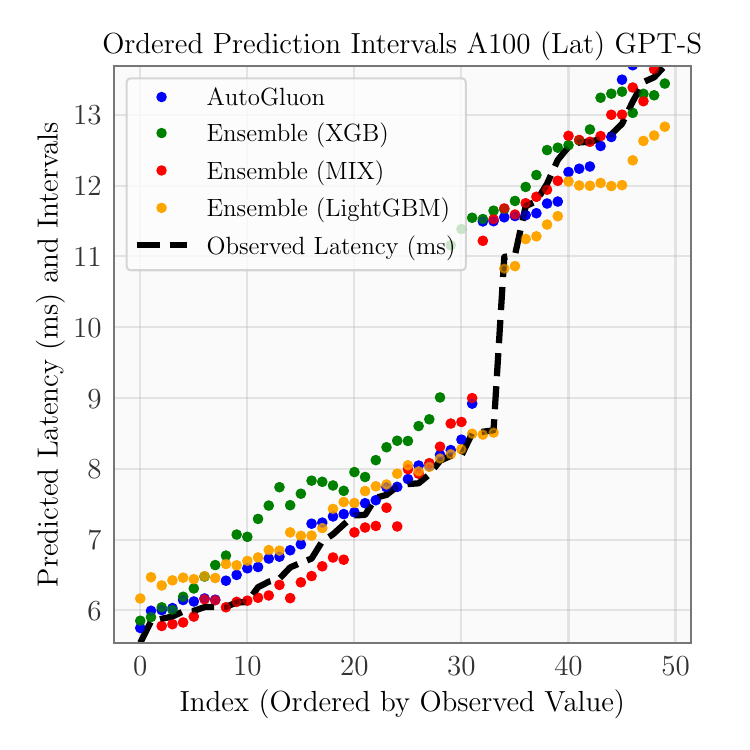}
\end{minipage}%
\begin{minipage}{.16\linewidth}
  \centering
\includegraphics[width=\linewidth]{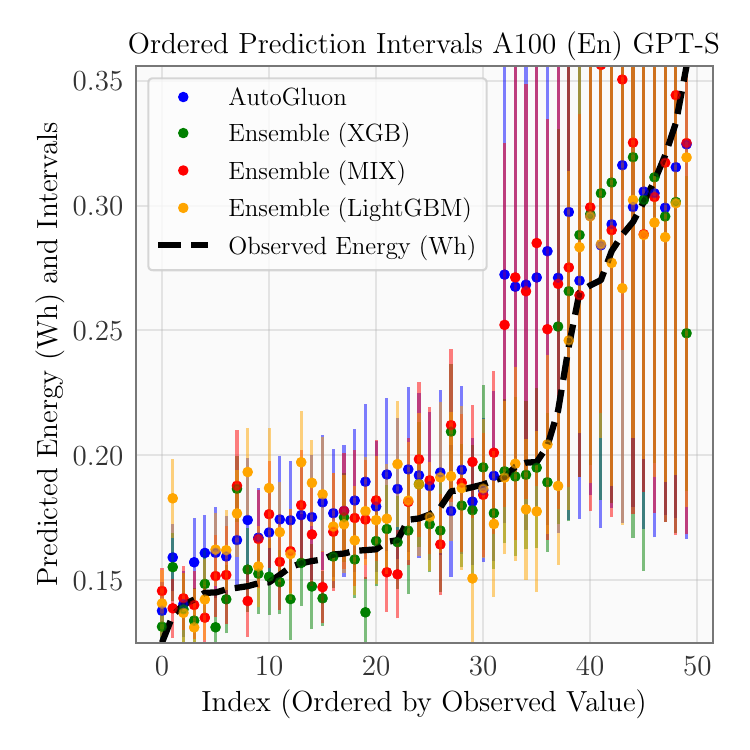}
\end{minipage}
\begin{minipage}{.16\linewidth}
  \centering
\includegraphics[width=\linewidth]{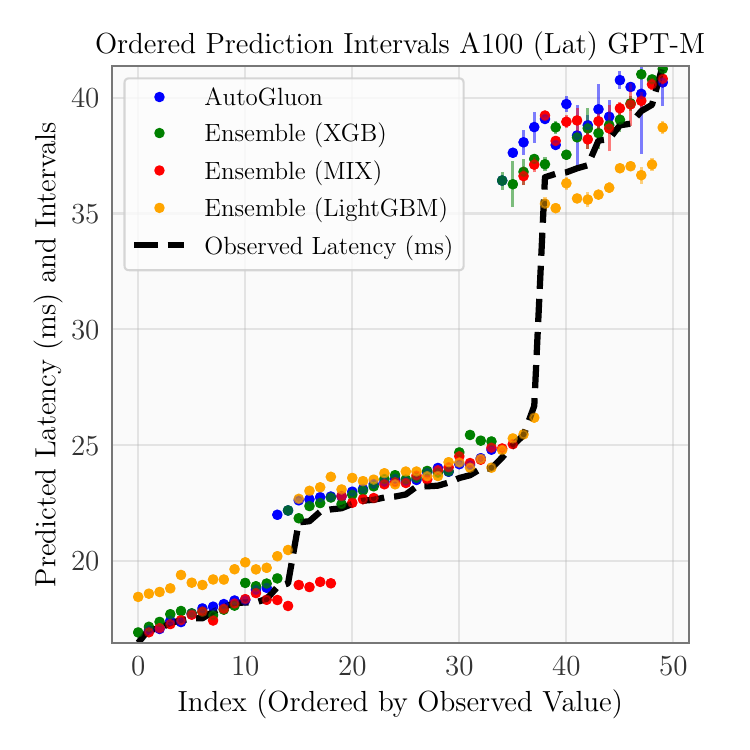}
\end{minipage}
\begin{minipage}{.16\linewidth}
  \centering
\includegraphics[width=\linewidth]{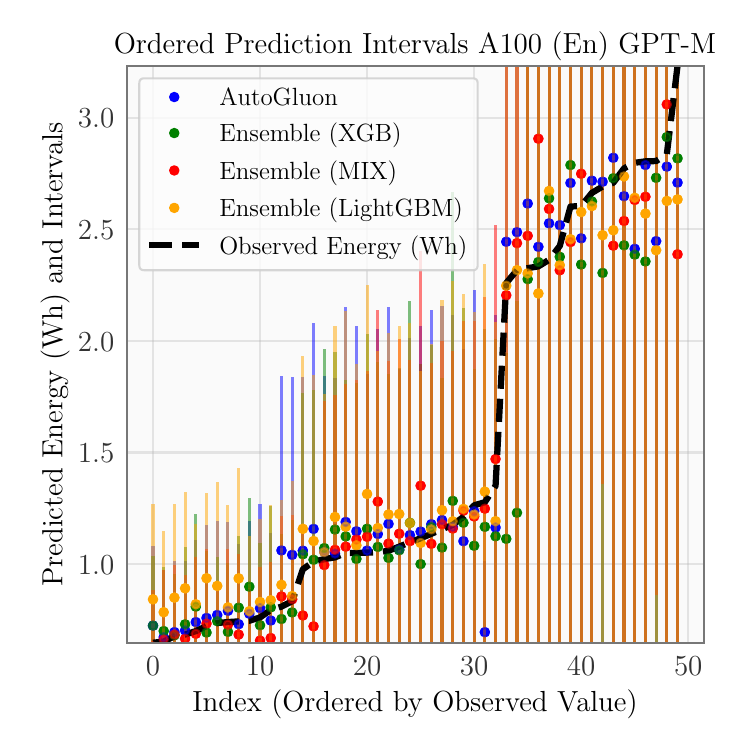}
\end{minipage}
\begin{minipage}{.16\linewidth}
  \centering
\includegraphics[width=\linewidth]{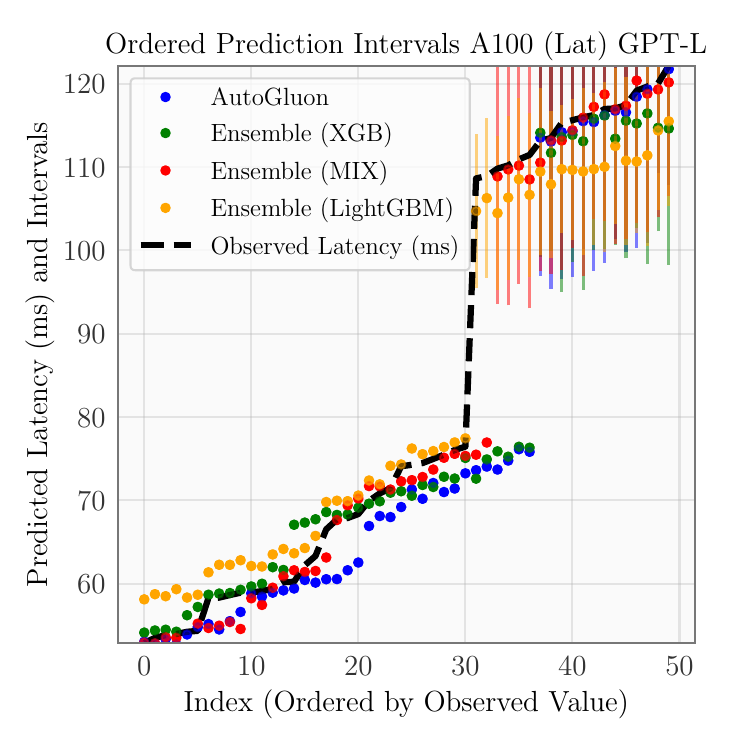}
\end{minipage}
\begin{minipage}{.16\linewidth}
  \centering
\includegraphics[width=\linewidth]{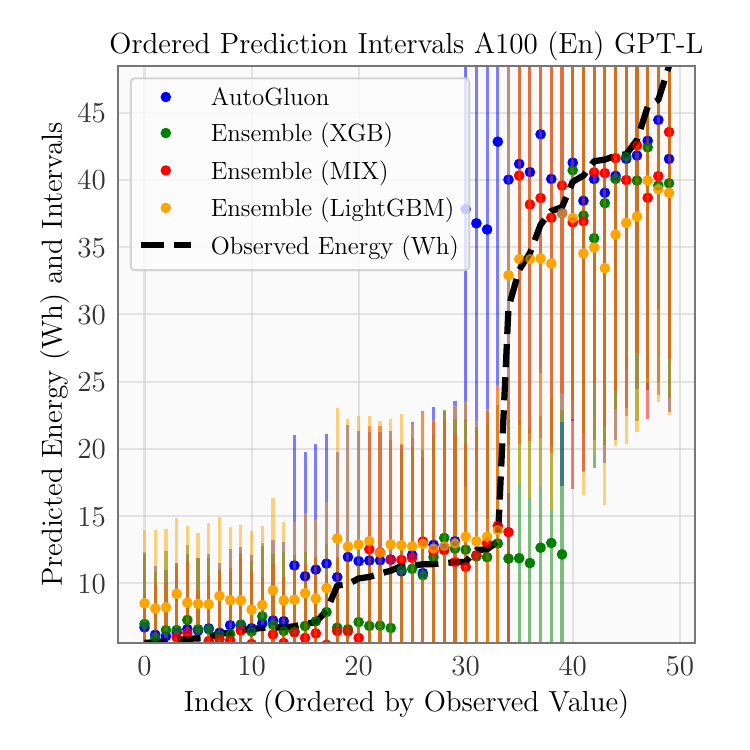}
\end{minipage}
\caption{Prediction Intervals A100}
\label{fig:interval_bounds_a100}

\end{figure}

\begin{figure}[H]

\begin{minipage}{.16\linewidth}
  \centering
  \includegraphics[width=\linewidth]{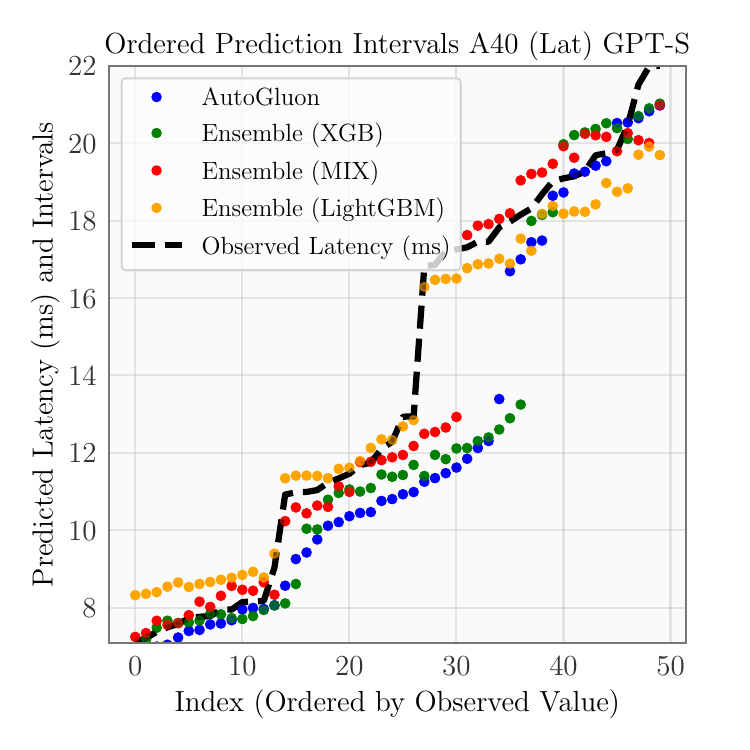}
\end{minipage}%
\begin{minipage}{.16\linewidth}
  \centering
\includegraphics[width=\linewidth]{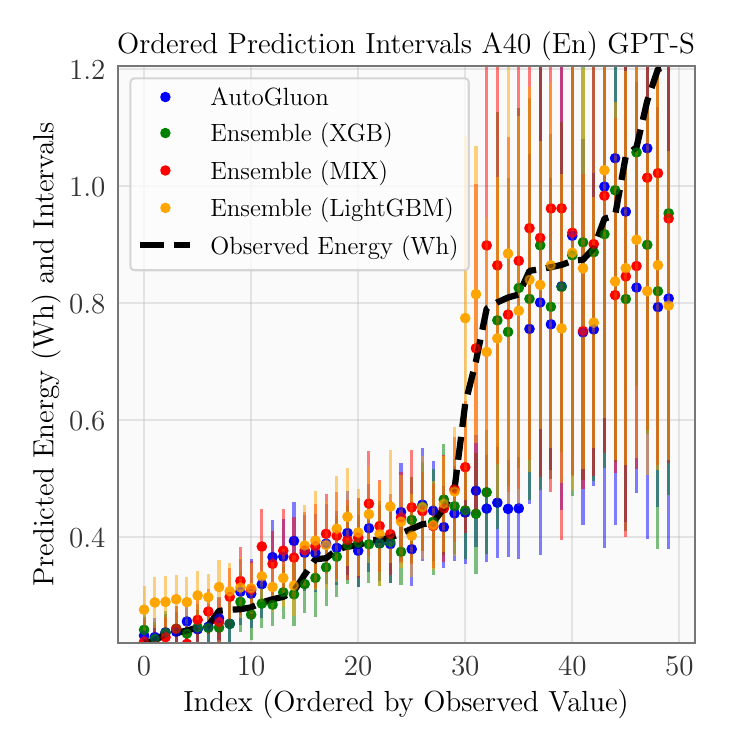}
\end{minipage}
\begin{minipage}{.16\linewidth}
  \centering
\includegraphics[width=\linewidth]{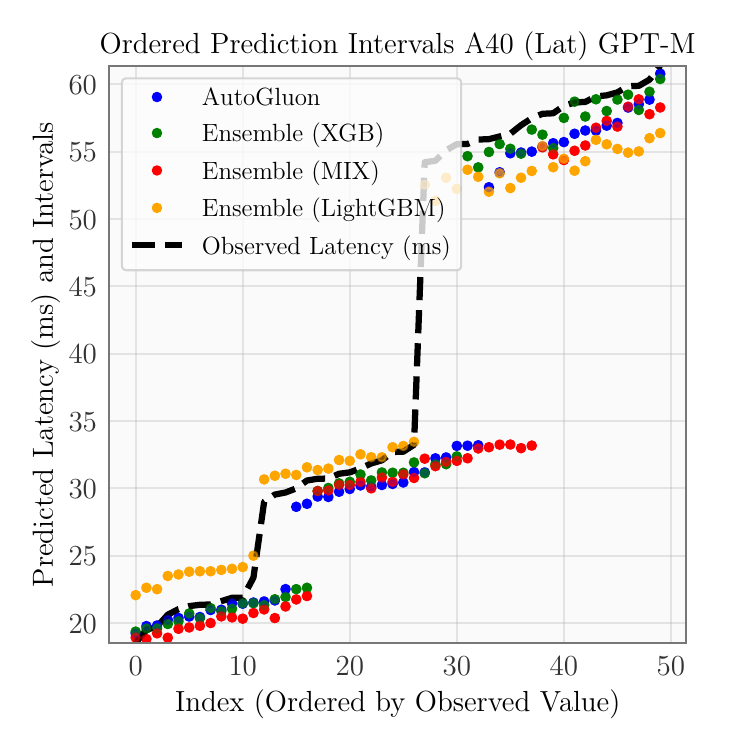}
\end{minipage}
\begin{minipage}{.16\linewidth}
  \centering
\includegraphics[width=\linewidth]{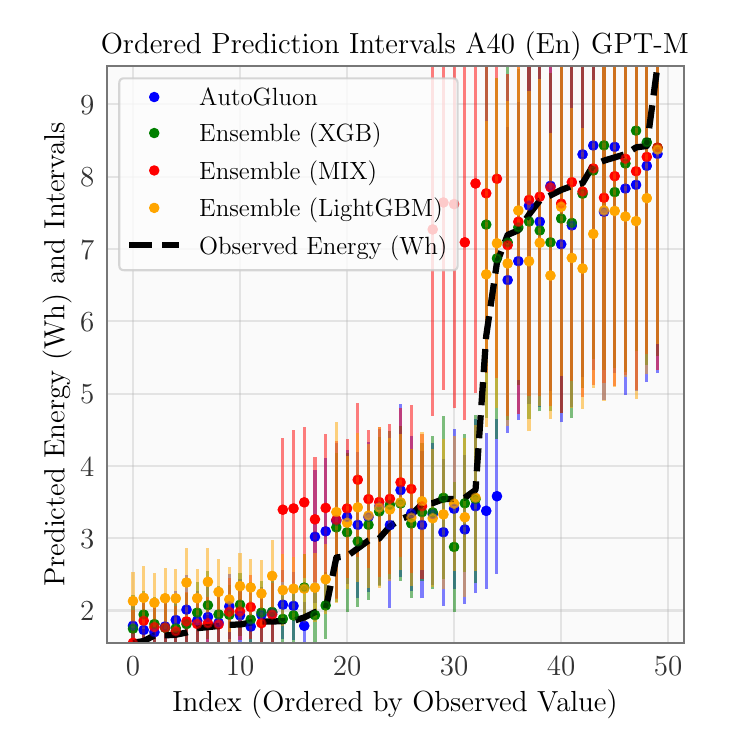}
\end{minipage}
\begin{minipage}{.16\linewidth}
  \centering
\includegraphics[width=\linewidth]{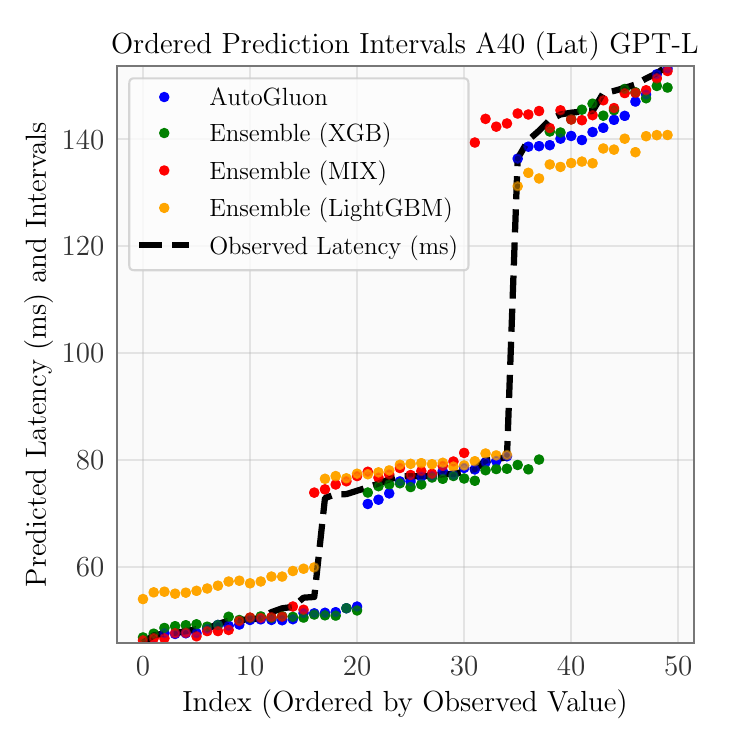}
\end{minipage}
\begin{minipage}{.16\linewidth}
  \centering
\includegraphics[width=\linewidth]{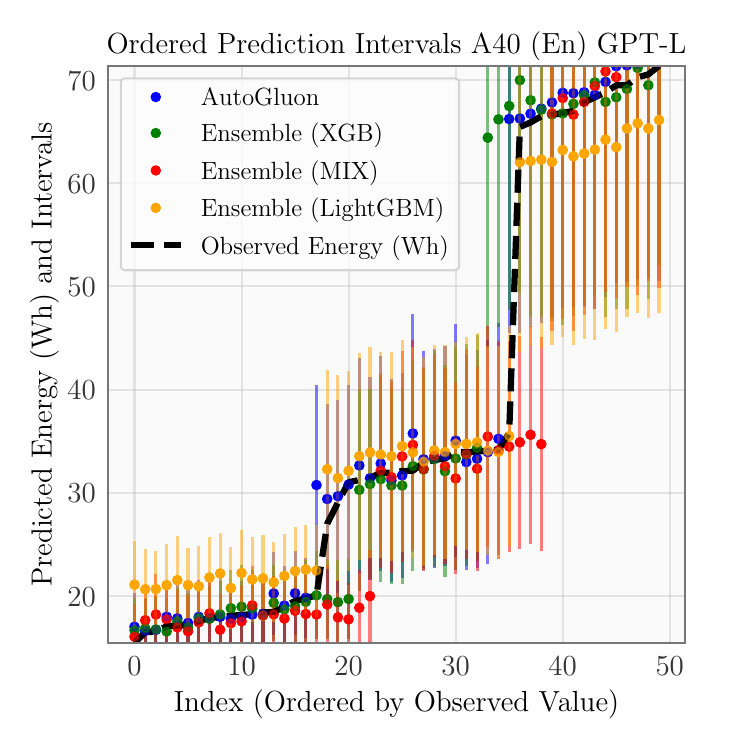}
\end{minipage}
\caption{Prediction Intervals A40}
\label{fig:interval_bounds_a40}

\end{figure}

\begin{figure}[H]

\begin{minipage}{.16\linewidth}
  \centering
  \includegraphics[width=\linewidth]{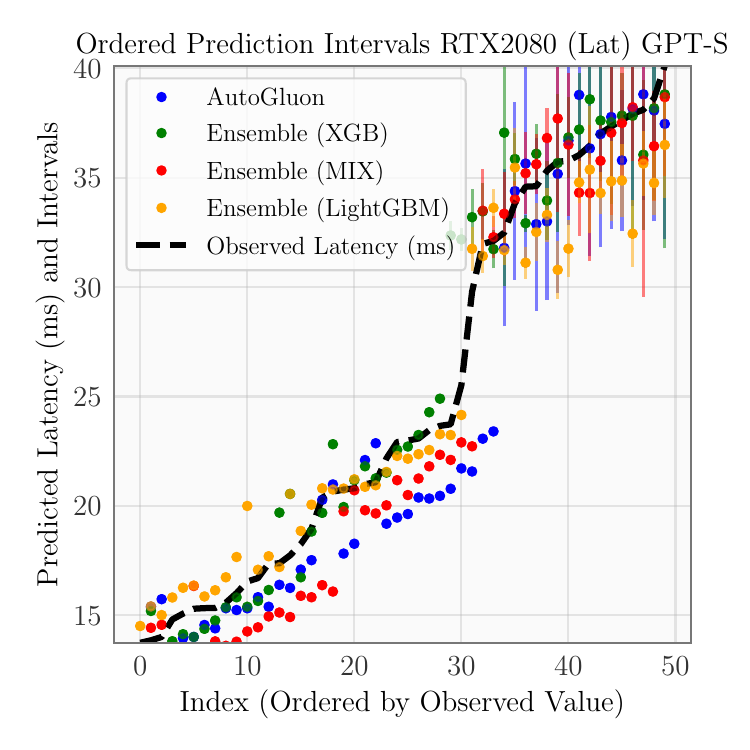}
\end{minipage}%
\begin{minipage}{.16\linewidth}
  \centering
\includegraphics[width=\linewidth]{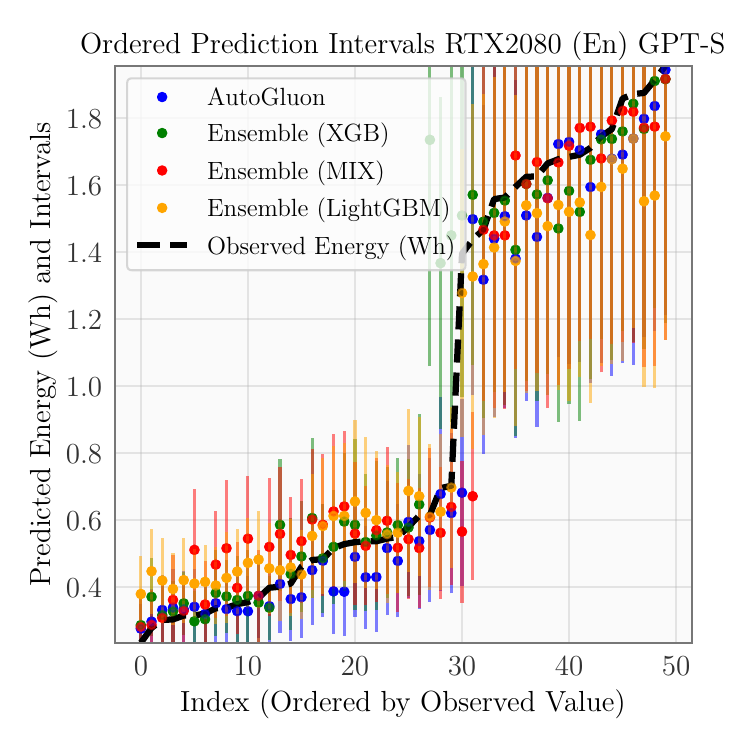}
\end{minipage}
\begin{minipage}{.16\linewidth}
  \centering
\includegraphics[width=\linewidth]{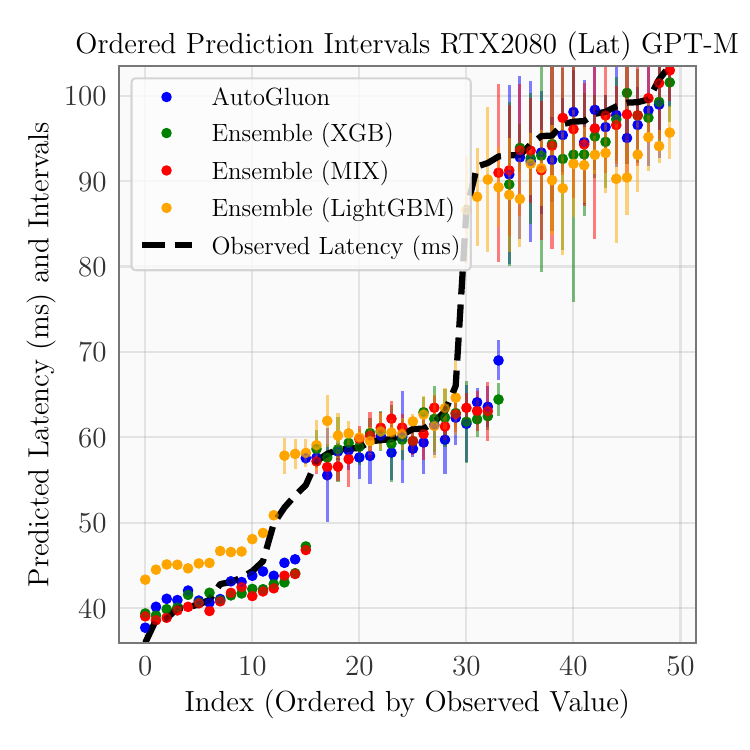}
\end{minipage}
\begin{minipage}{.16\linewidth}
  \centering
\includegraphics[width=\linewidth]{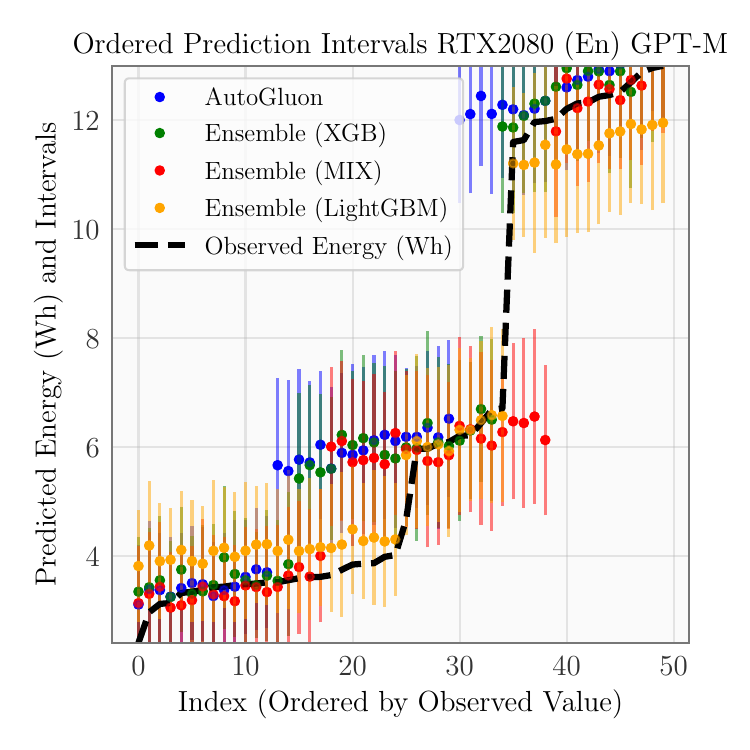}
\end{minipage}
\begin{minipage}{.16\linewidth}
  \centering
\includegraphics[width=\linewidth]{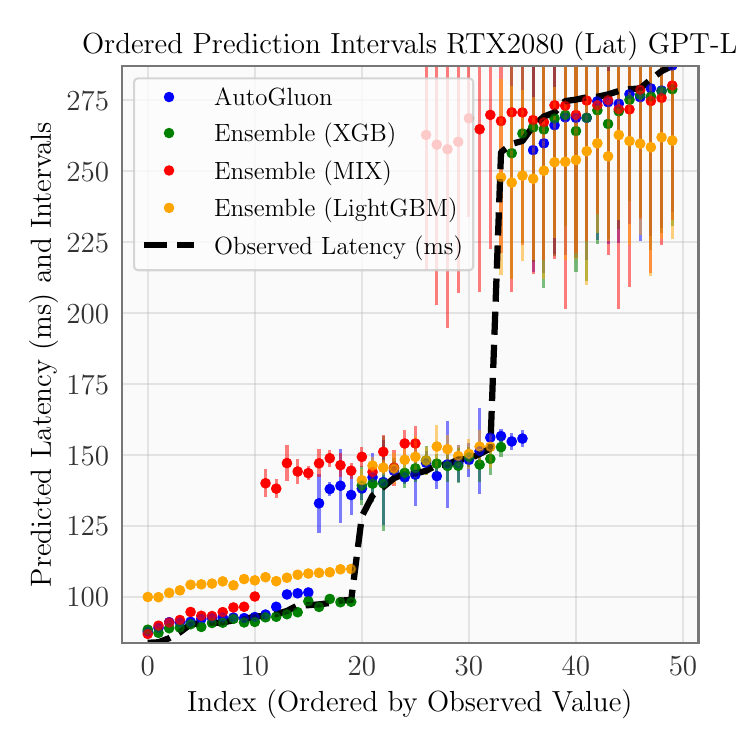}
\end{minipage}
\begin{minipage}{.16\linewidth}
  \centering
\includegraphics[width=\linewidth]{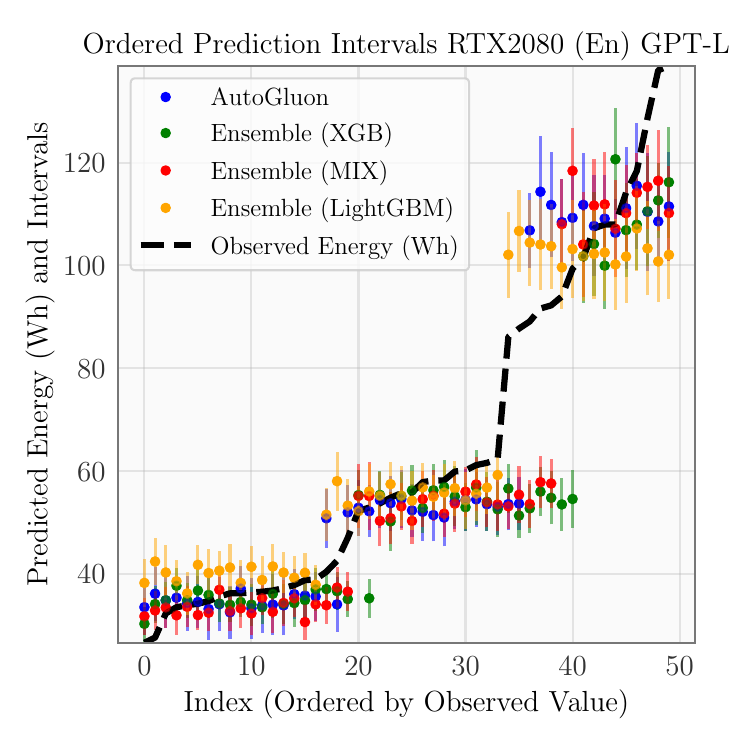}
\end{minipage}
\caption{Prediction Intervals RTX2080}
\label{fig:interval_bounds_rtx2080}

\end{figure}

\begin{figure}[H]

\begin{minipage}{.16\linewidth}
  \centering
  \includegraphics[width=\linewidth]{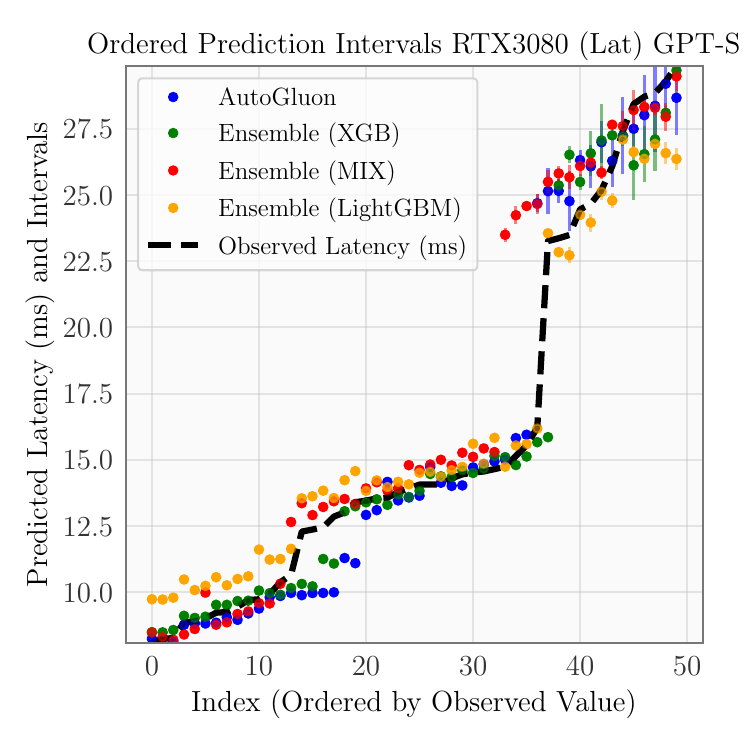}
\end{minipage}%
\begin{minipage}{.16\linewidth}
  \centering
\includegraphics[width=\linewidth]{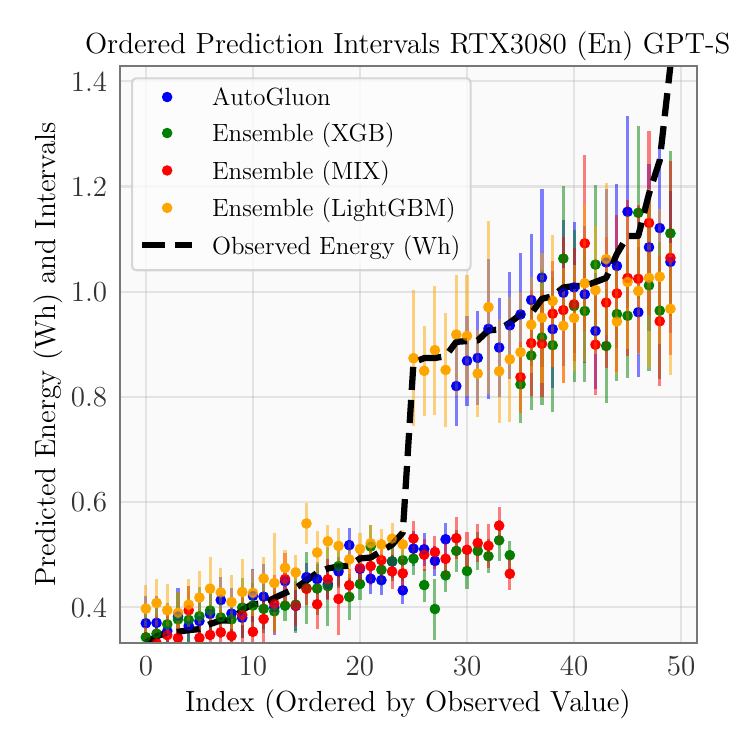}
\end{minipage}
\begin{minipage}{.16\linewidth}
  \centering
\includegraphics[width=\linewidth]{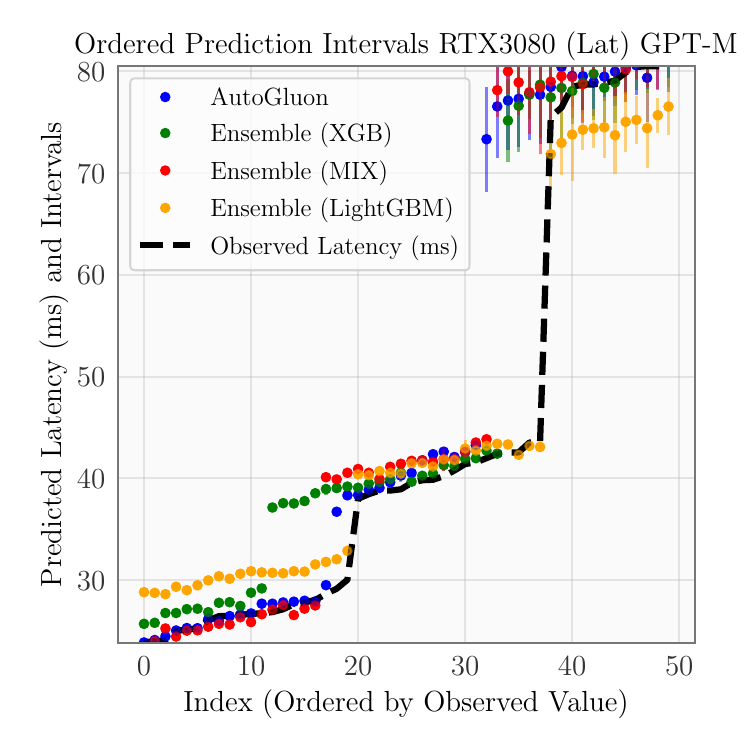}
\end{minipage}
\begin{minipage}{.16\linewidth}
  \centering
\includegraphics[width=\linewidth]{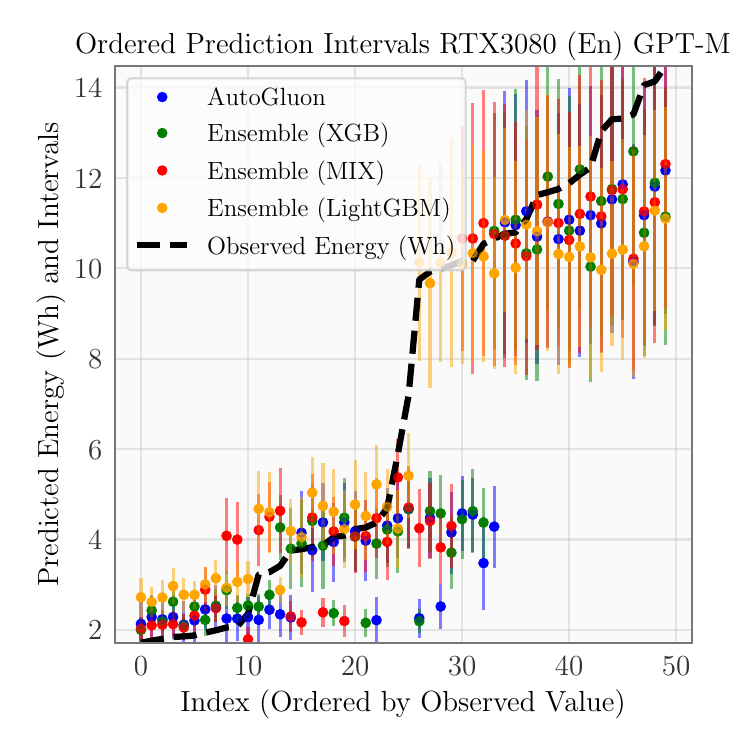}
\end{minipage}
\begin{minipage}{.16\linewidth}
  \centering
\includegraphics[width=\linewidth]{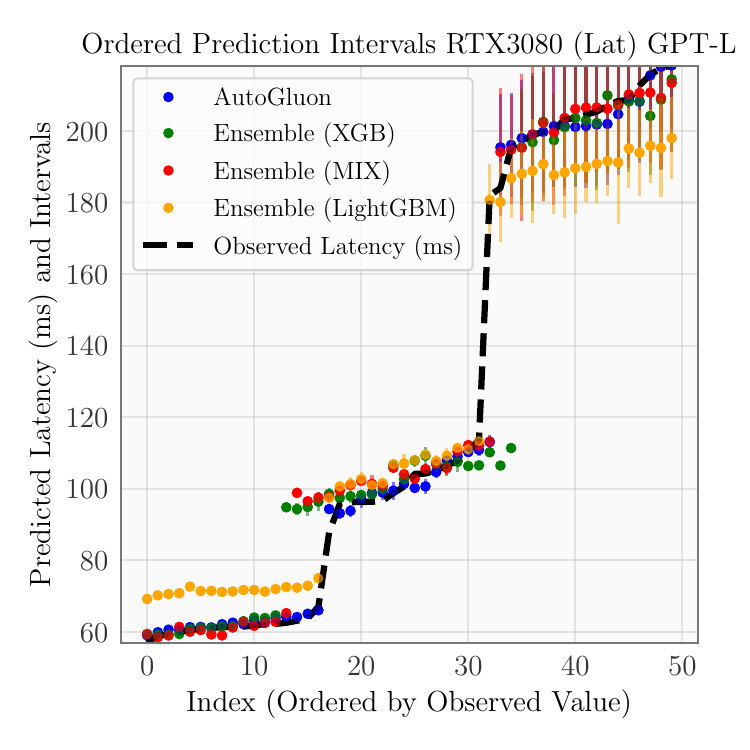}
\end{minipage}
\begin{minipage}{.16\linewidth}
  \centering
\includegraphics[width=\linewidth]{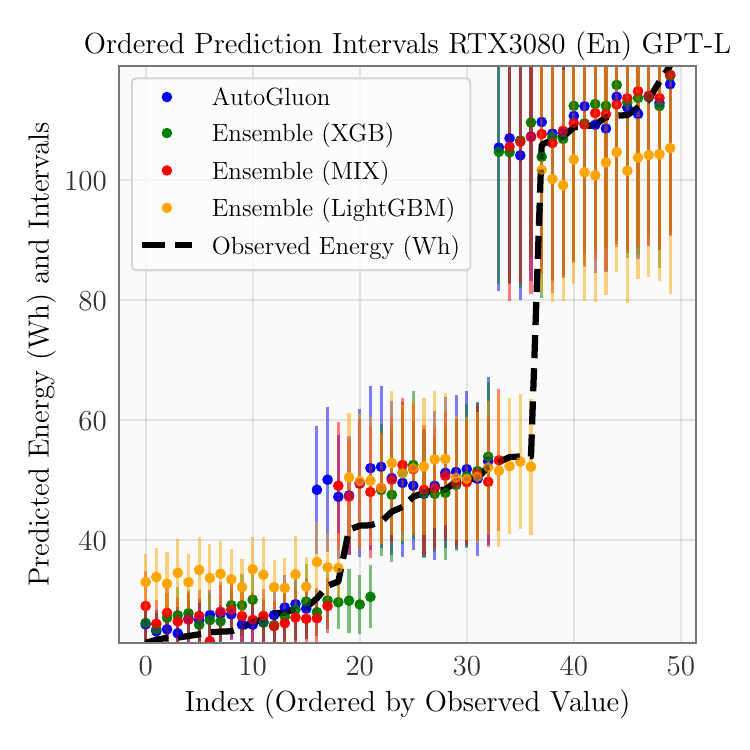}
\end{minipage}
\caption{Prediction Intervals RTX3080}
\label{fig:interval_bounds_rtx3080}

\end{figure}

\begin{figure}[H]

\begin{minipage}{.16\linewidth}
  \centering
  \includegraphics[width=\linewidth]{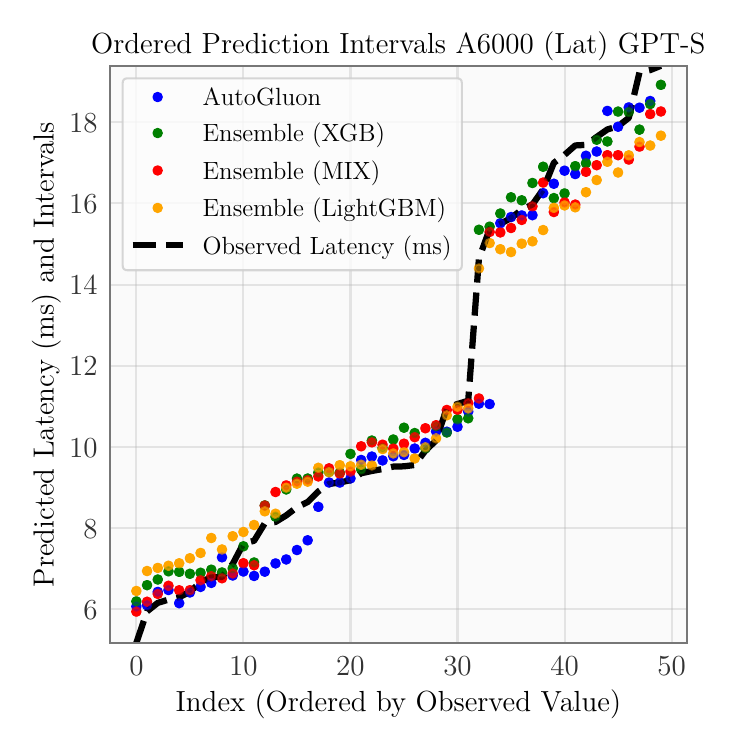}
\end{minipage}%
\begin{minipage}{.16\linewidth}
  \centering
\includegraphics[width=\linewidth]{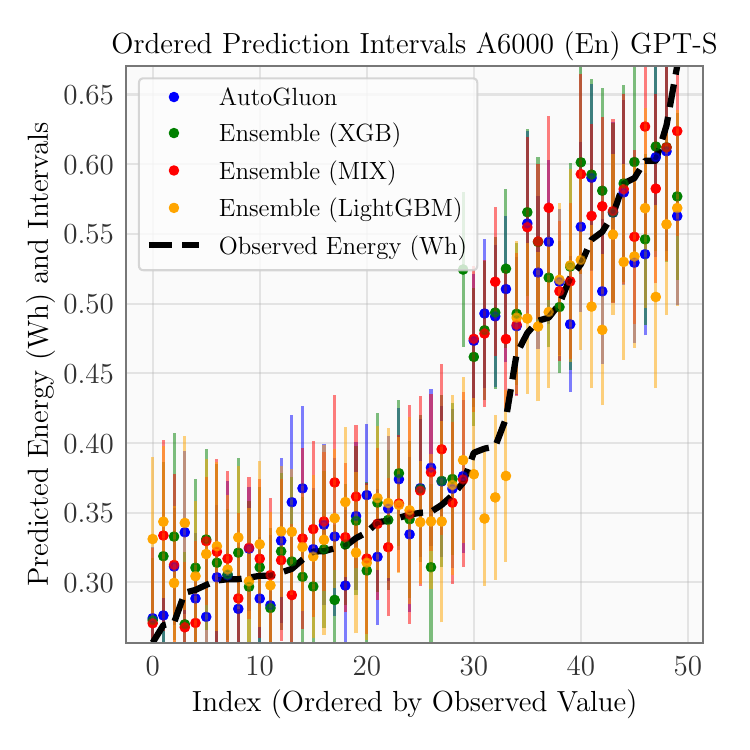}
\end{minipage}
\begin{minipage}{.16\linewidth}
  \centering
\includegraphics[width=\linewidth]{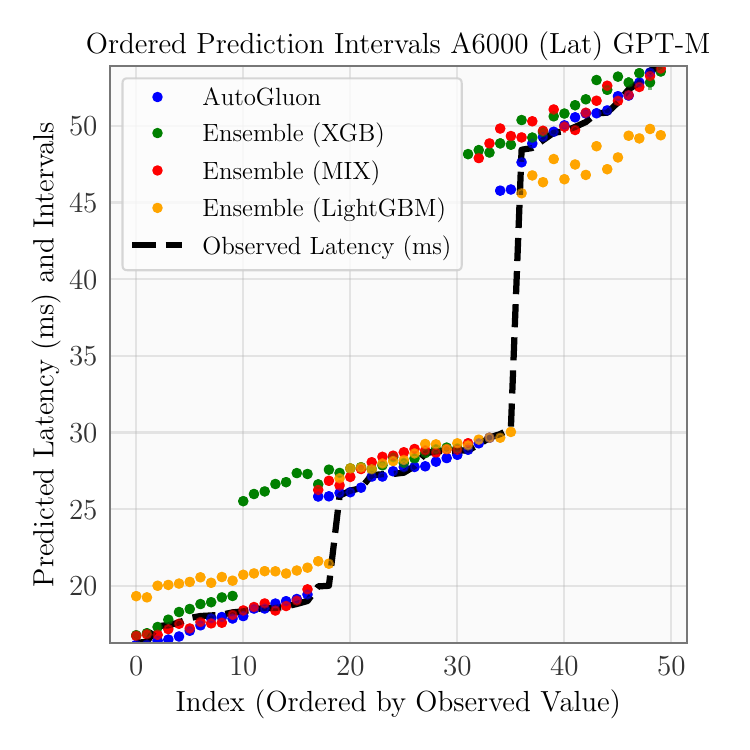}
\end{minipage}
\begin{minipage}{.16\linewidth}
  \centering
\includegraphics[width=\linewidth]{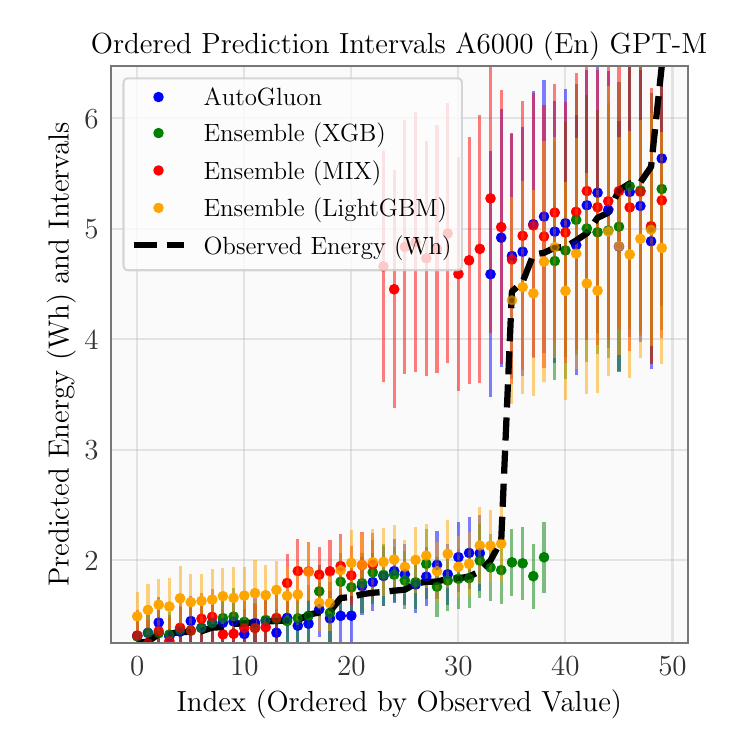}
\end{minipage}
\begin{minipage}{.16\linewidth}
  \centering
\includegraphics[width=\linewidth]{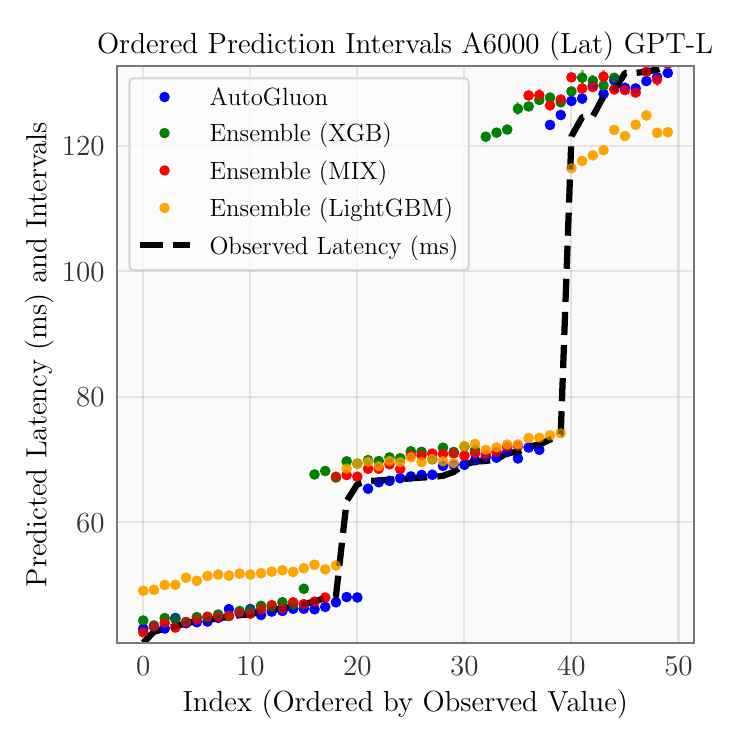}
\end{minipage}
\begin{minipage}{.16\linewidth}
  \centering
\includegraphics[width=\linewidth]{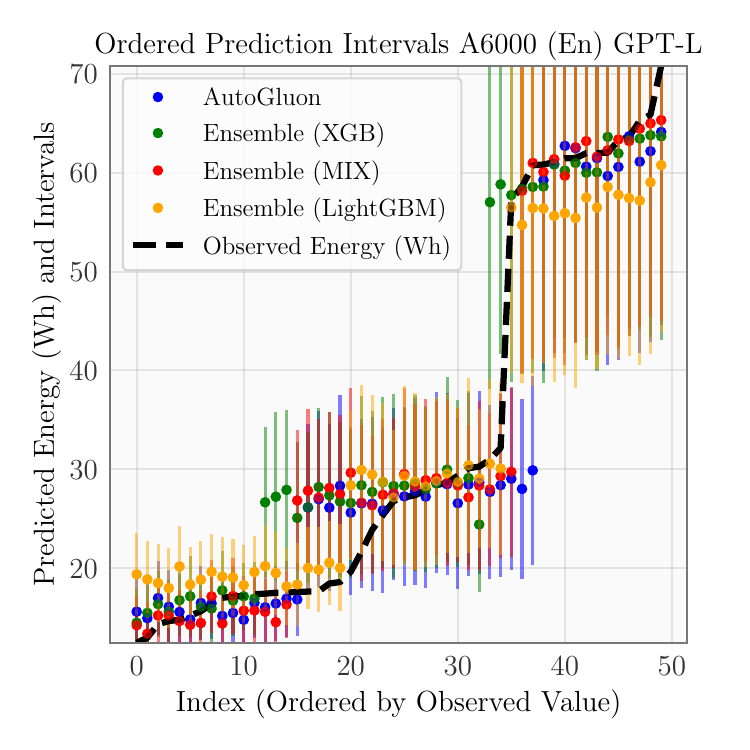}
\end{minipage}
\caption{Prediction Intervals A6000}
\label{fig:interval_bounds_a6000}

\end{figure}

\begin{figure}[H]

\begin{minipage}{.16\linewidth}
  \centering
  \includegraphics[width=\linewidth]{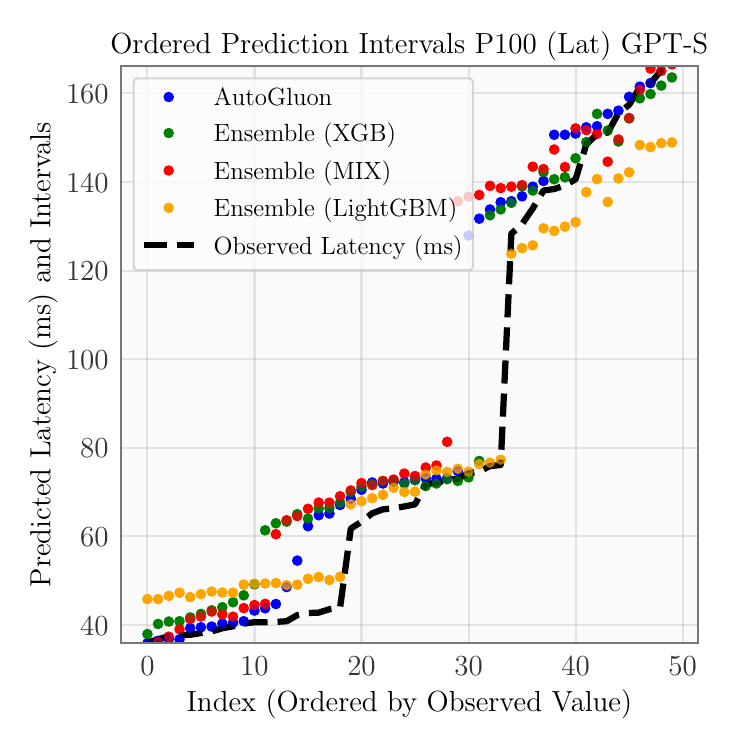}
\end{minipage}%
\begin{minipage}{.16\linewidth}
  \centering
\includegraphics[width=\linewidth]{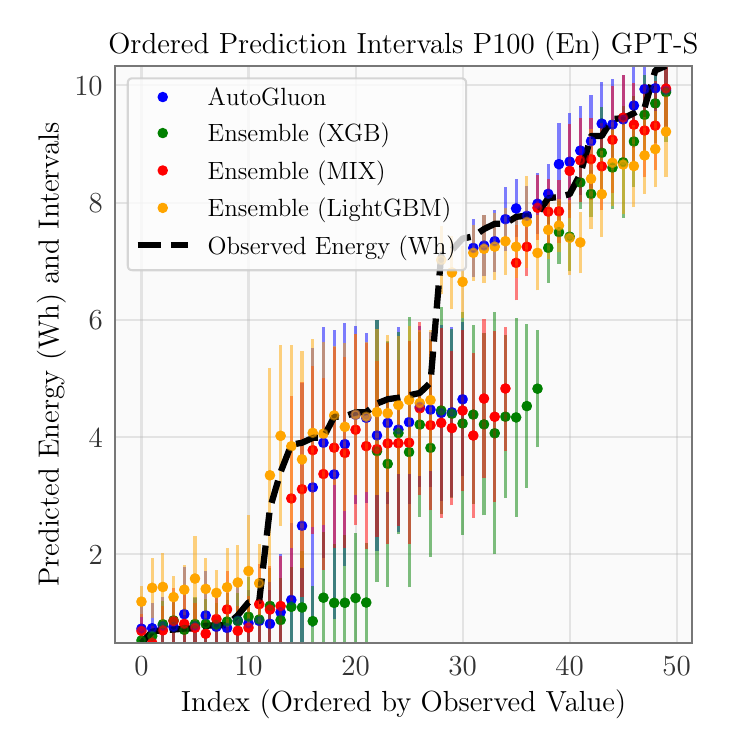}
\end{minipage}
\begin{minipage}{.16\linewidth}
  \centering
\includegraphics[width=\linewidth]{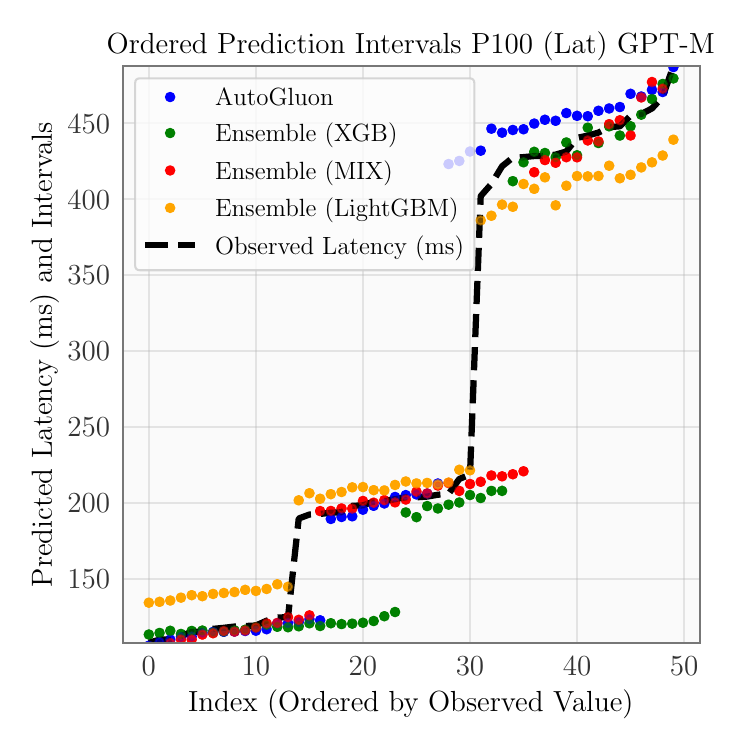}
\end{minipage}
\begin{minipage}{.16\linewidth}
  \centering
\includegraphics[width=\linewidth]{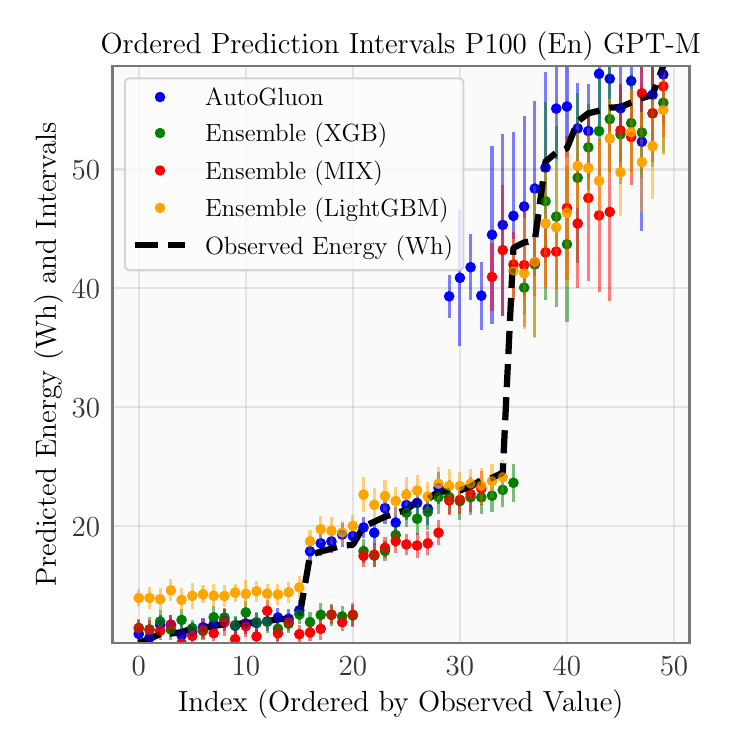}
\end{minipage}
\begin{minipage}{.16\linewidth}
  \centering
\includegraphics[width=\linewidth]{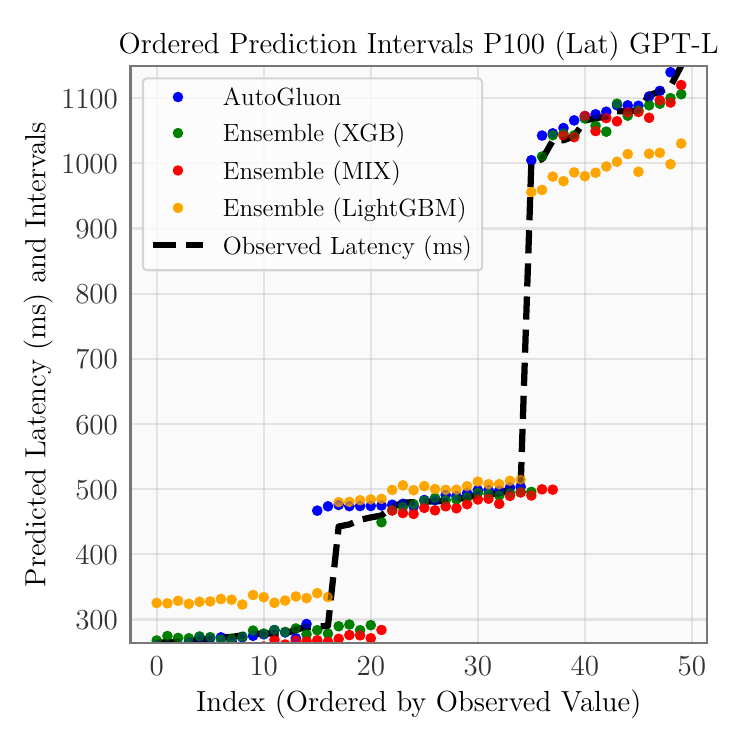}
\end{minipage}
\begin{minipage}{.16\linewidth}
  \centering
\includegraphics[width=\linewidth]{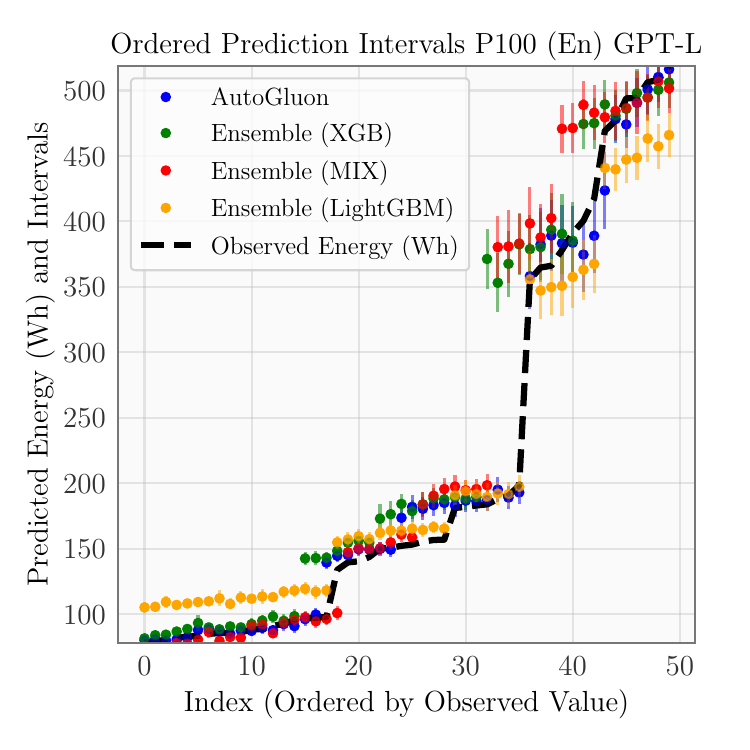}
\end{minipage}
\caption{Prediction Intervals P100}
\label{fig:interval_bounds_p100}

\end{figure}

\begin{figure}[H]

\begin{minipage}{.16\linewidth}
  \centering
  \includegraphics[width=\linewidth]{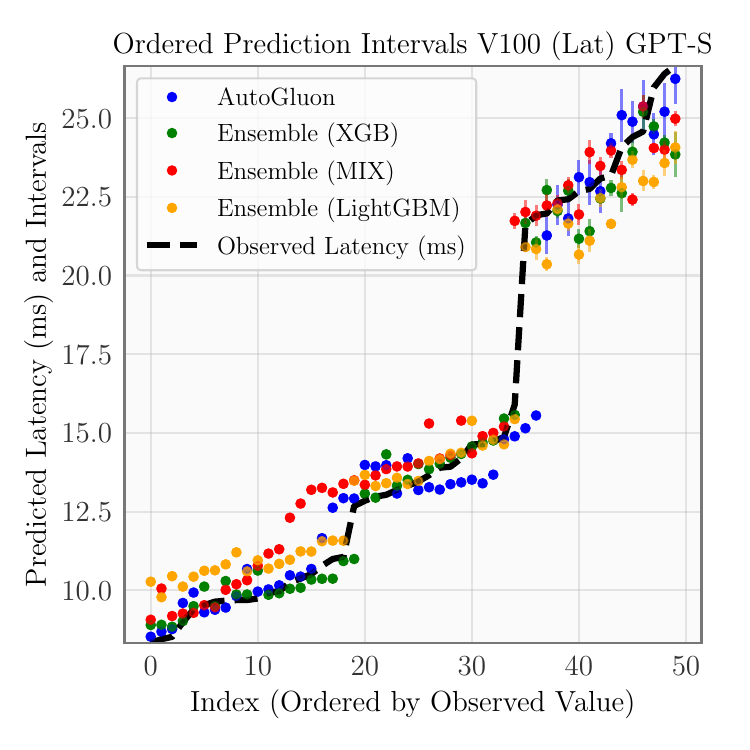}
\end{minipage}%
\begin{minipage}{.16\linewidth}
  \centering
\includegraphics[width=\linewidth]{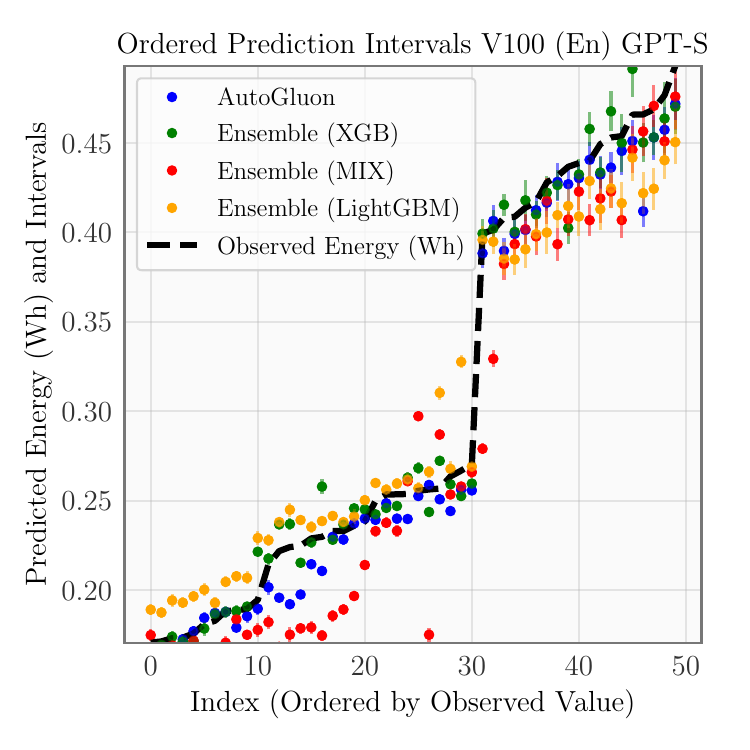}
\end{minipage}
\begin{minipage}{.16\linewidth}
  \centering
\includegraphics[width=\linewidth]{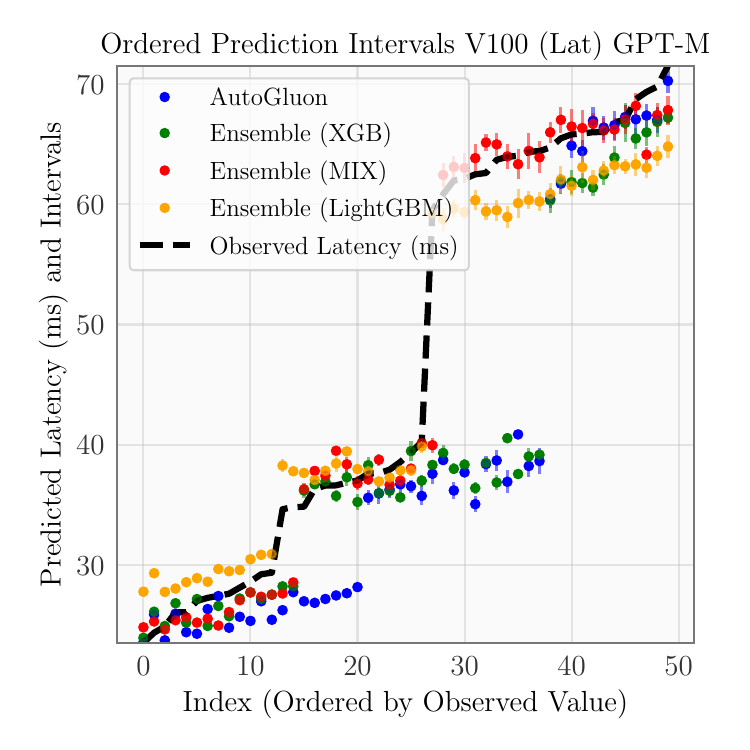}
\end{minipage}
\begin{minipage}{.16\linewidth}
  \centering
\includegraphics[width=\linewidth]{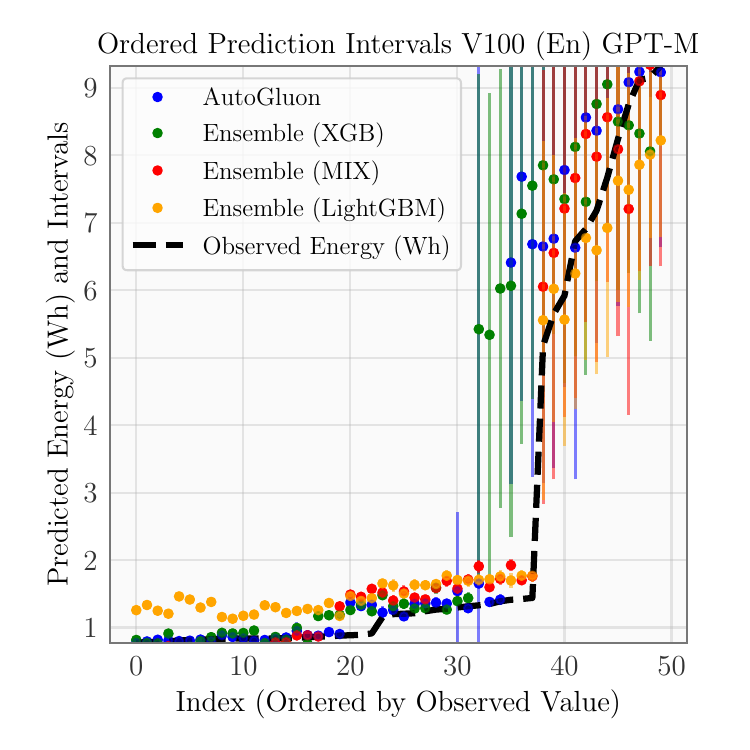}
\end{minipage}
\begin{minipage}{.16\linewidth}
  \centering
\includegraphics[width=\linewidth]{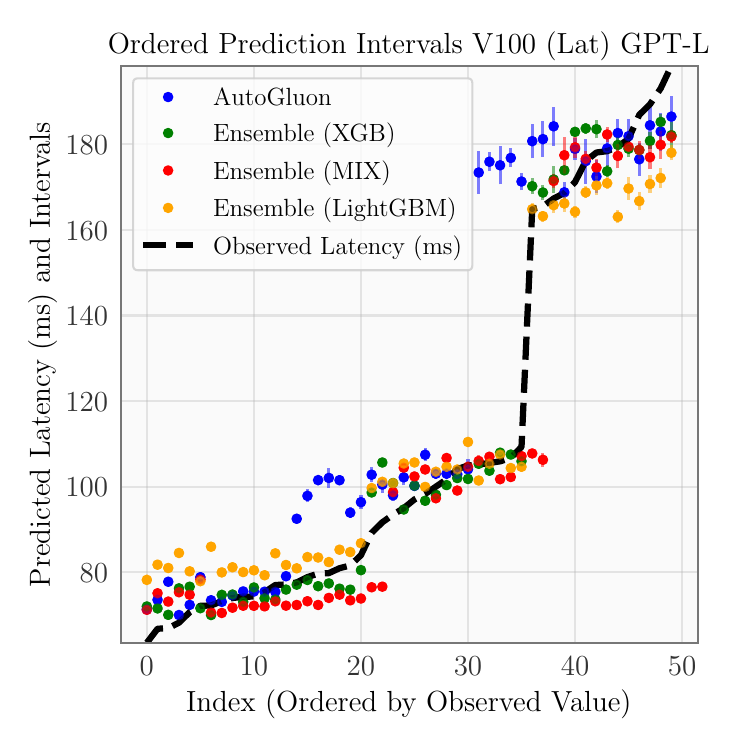}
\end{minipage}
\begin{minipage}{.16\linewidth}
  \centering
\includegraphics[width=\linewidth]{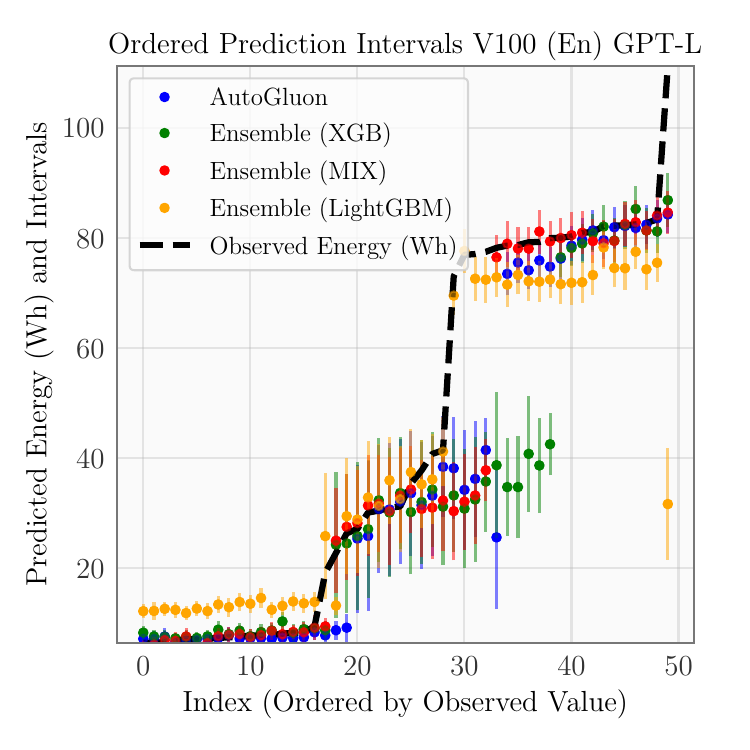}
\end{minipage}
\caption{Prediction Intervals V100}
\label{fig:interval_bounds_v100}

\end{figure}

\begin{figure}[H]
\begin{minipage}{.16\linewidth}
  \centering
  \includegraphics[width=\linewidth]{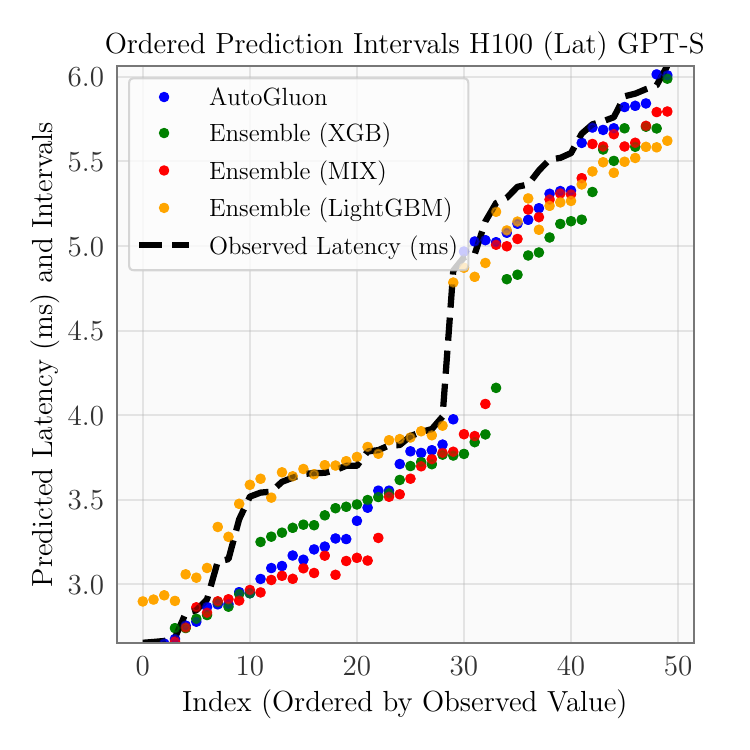}
\end{minipage}%
\begin{minipage}{.16\linewidth}
  \centering
\includegraphics[width=\linewidth]{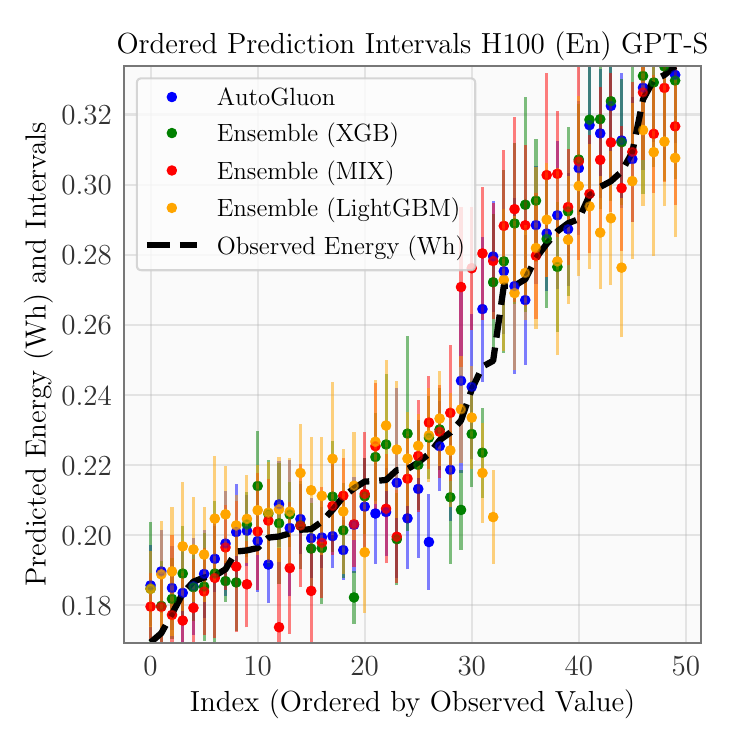}
\end{minipage}
\begin{minipage}{.16\linewidth}
  \centering
\includegraphics[width=\linewidth]{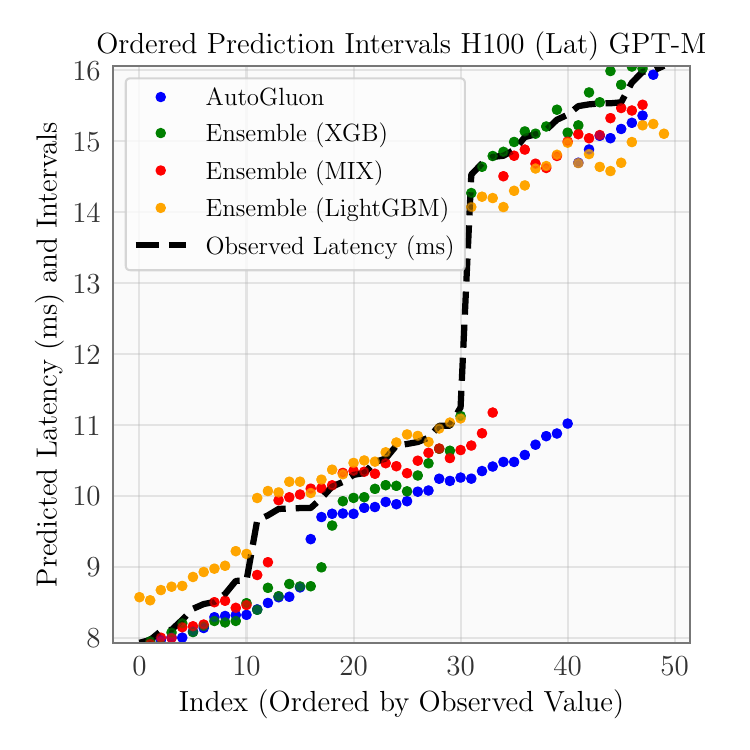}
\end{minipage}
\begin{minipage}{.16\linewidth}
  \centering
\includegraphics[width=\linewidth]{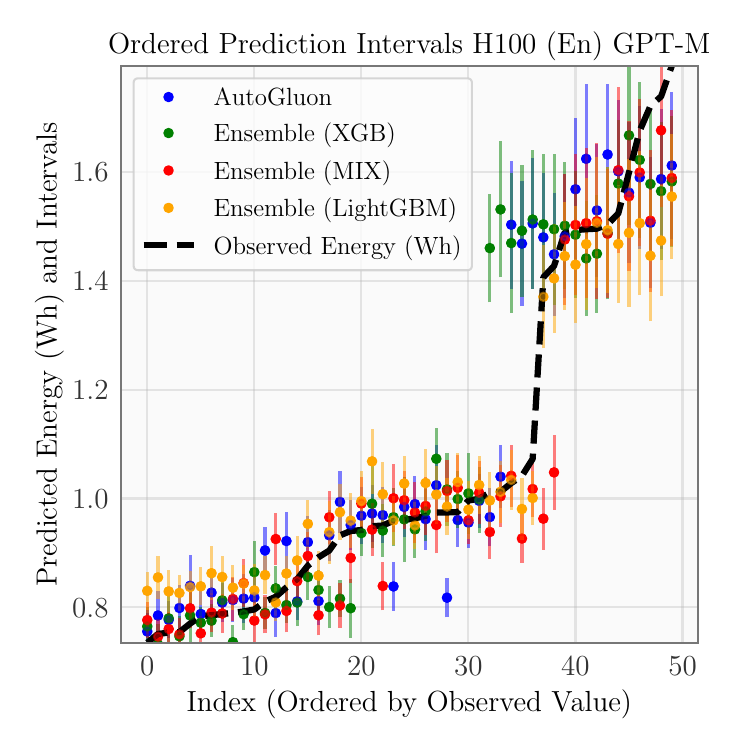}
\end{minipage}
\begin{minipage}{.16\linewidth}
  \centering
\includegraphics[width=\linewidth]{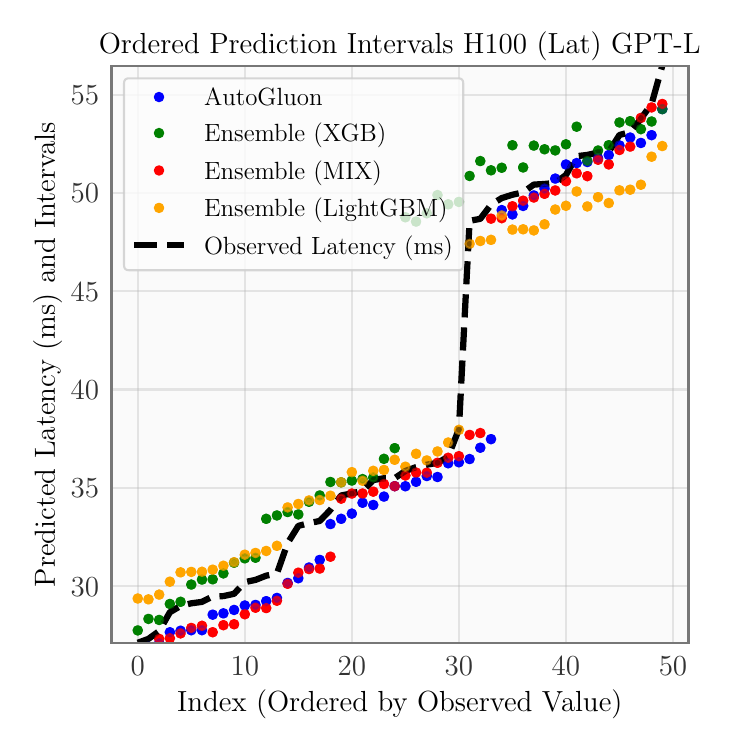}
\end{minipage}
\begin{minipage}{.16\linewidth}
  \centering
\includegraphics[width=\linewidth]{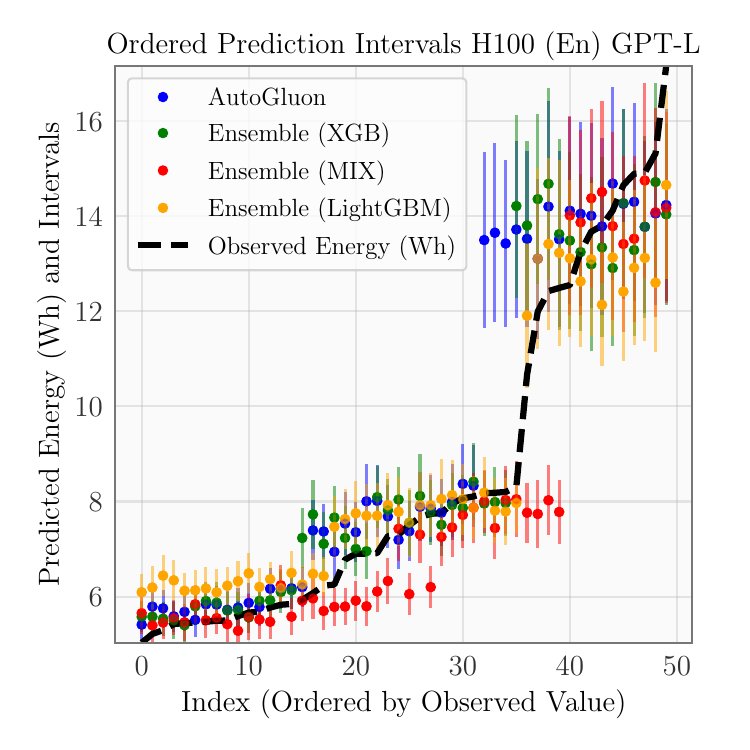}
\end{minipage}
\caption{Prediction Intervals H100}
\label{fig:interval_bounds_h100}

\end{figure}

\begin{figure}[H]
\begin{minipage}{.16\linewidth}
  \centering
  \includegraphics[width=\linewidth]{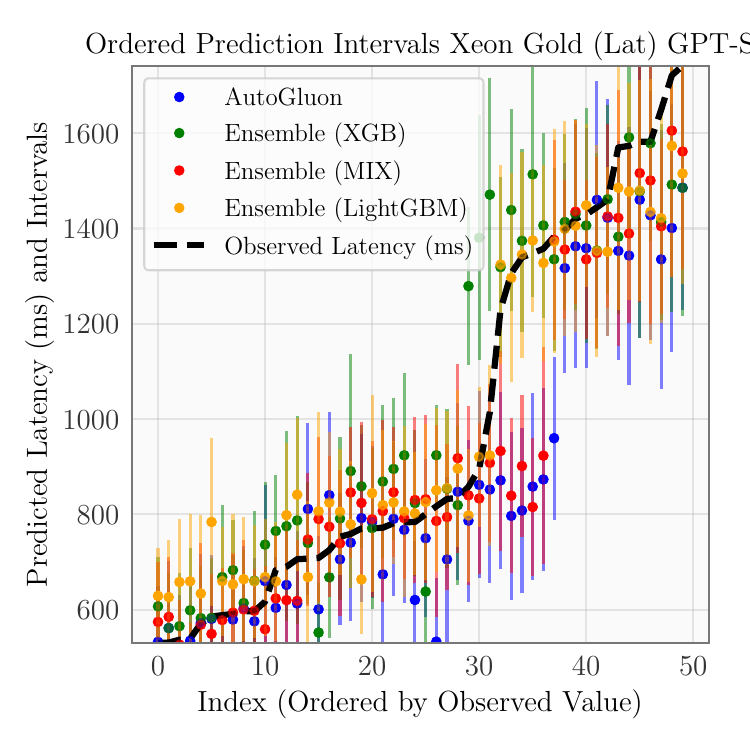}
\end{minipage}%
\begin{minipage}{.16\linewidth}
  \centering
\includegraphics[width=\linewidth]{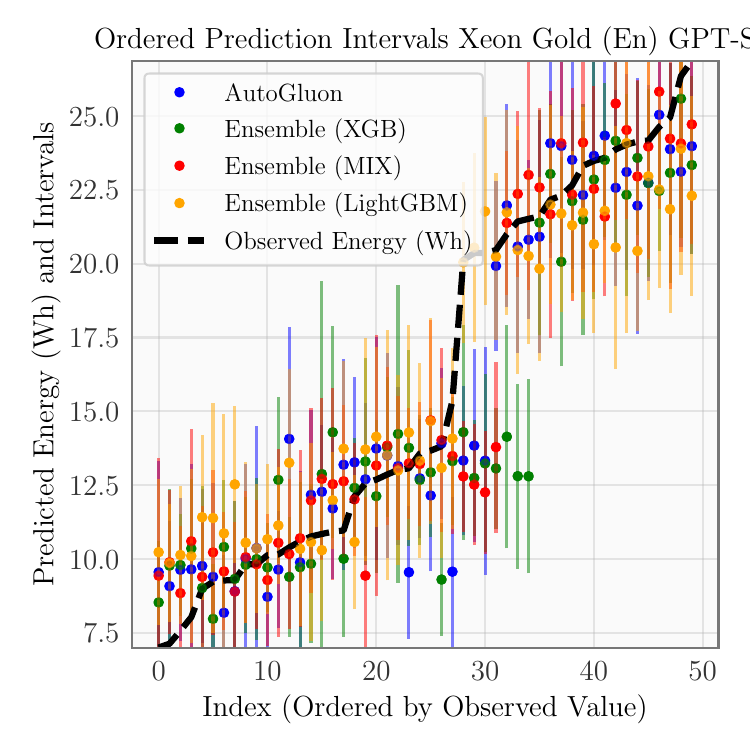}
\end{minipage}
\begin{minipage}{.16\linewidth}
  \centering
\includegraphics[width=\linewidth]{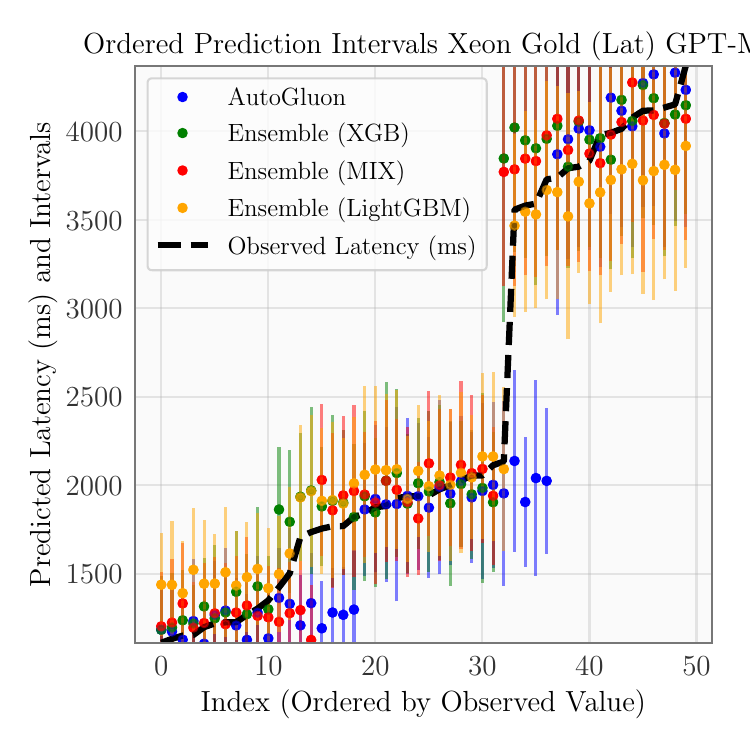}
\end{minipage}
\begin{minipage}{.16\linewidth}
  \centering
\includegraphics[width=\linewidth]{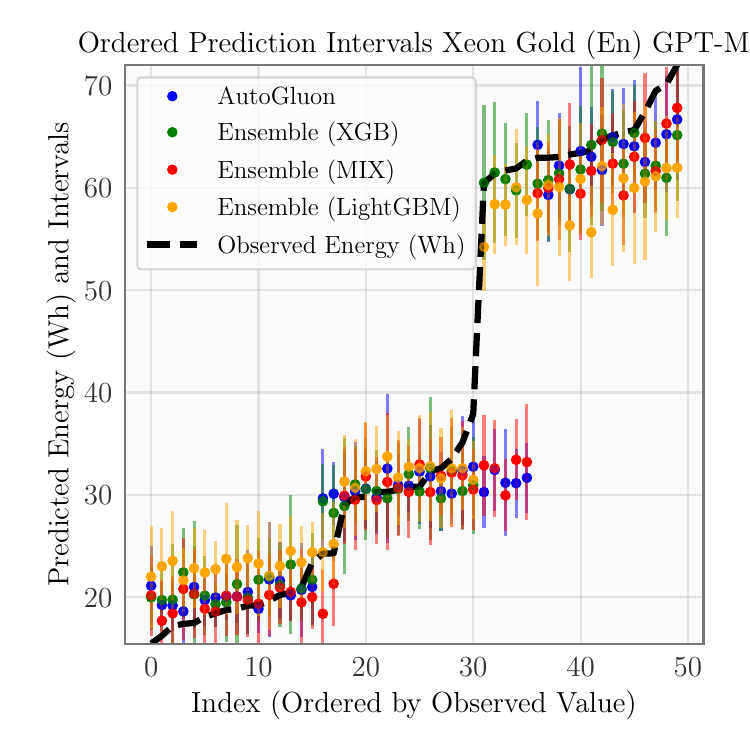}
\end{minipage}
\begin{minipage}{.16\linewidth}
  \centering
\includegraphics[width=\linewidth]{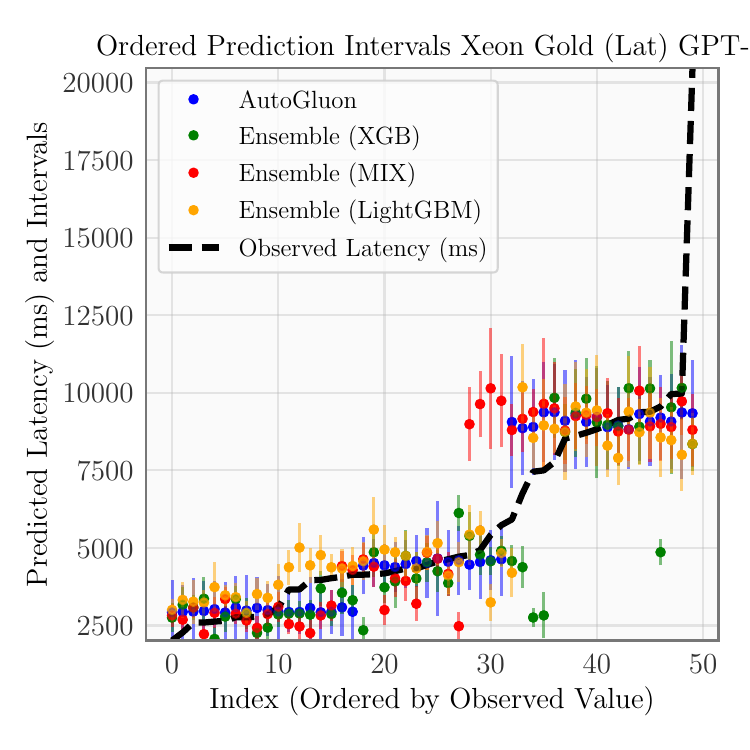}
\end{minipage}
\begin{minipage}{.16\linewidth}
  \centering
\includegraphics[width=\linewidth]{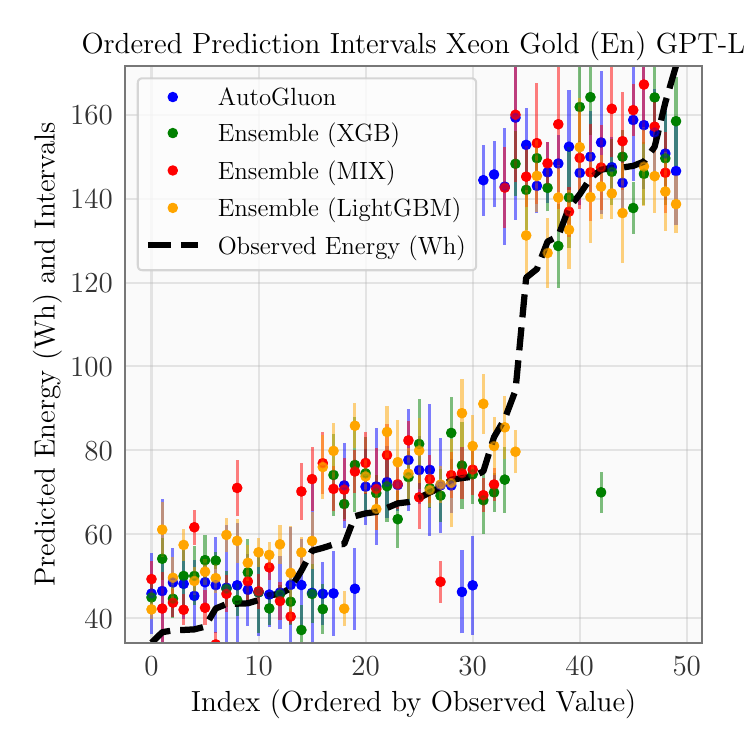}
\end{minipage}
\caption{Prediction Intervals Xeon Gold}
\label{fig:interval_bounds_xeon_gold}

\end{figure}

\begin{figure}[H]
\begin{minipage}{.16\linewidth}
  \centering
  \includegraphics[width=\linewidth]{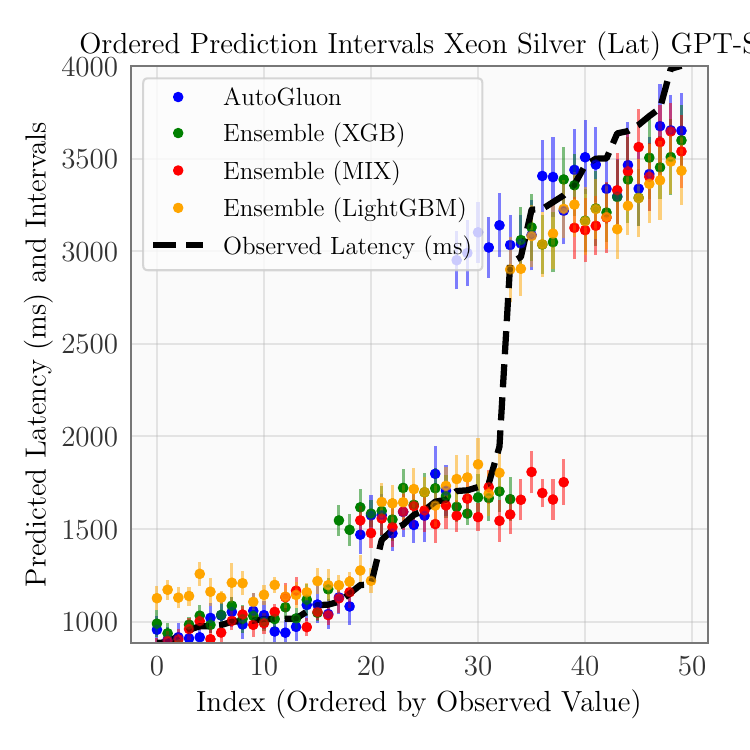}
\end{minipage}%
\begin{minipage}{.16\linewidth}
  \centering
\includegraphics[width=\linewidth]{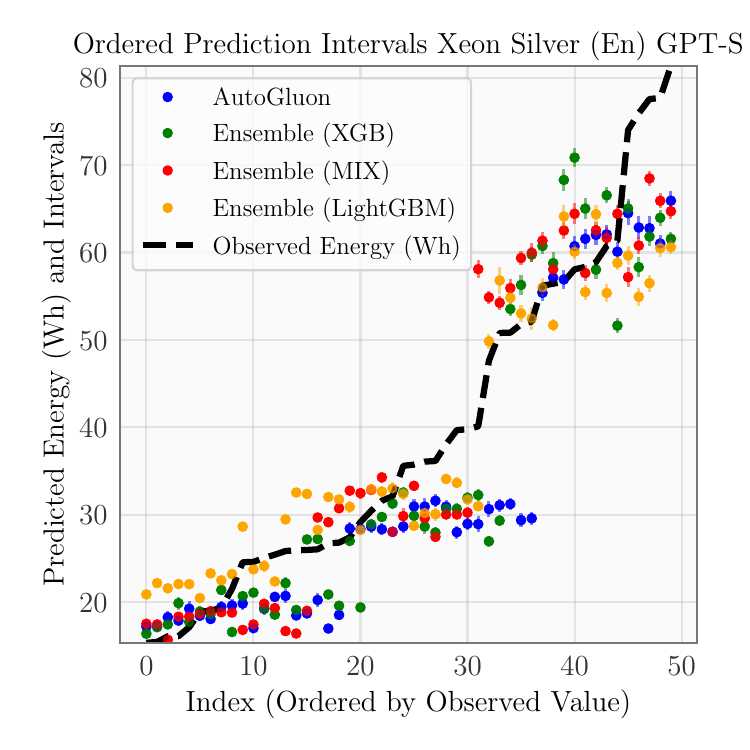}
\end{minipage}
\begin{minipage}{.16\linewidth}
  \centering
\includegraphics[width=\linewidth]{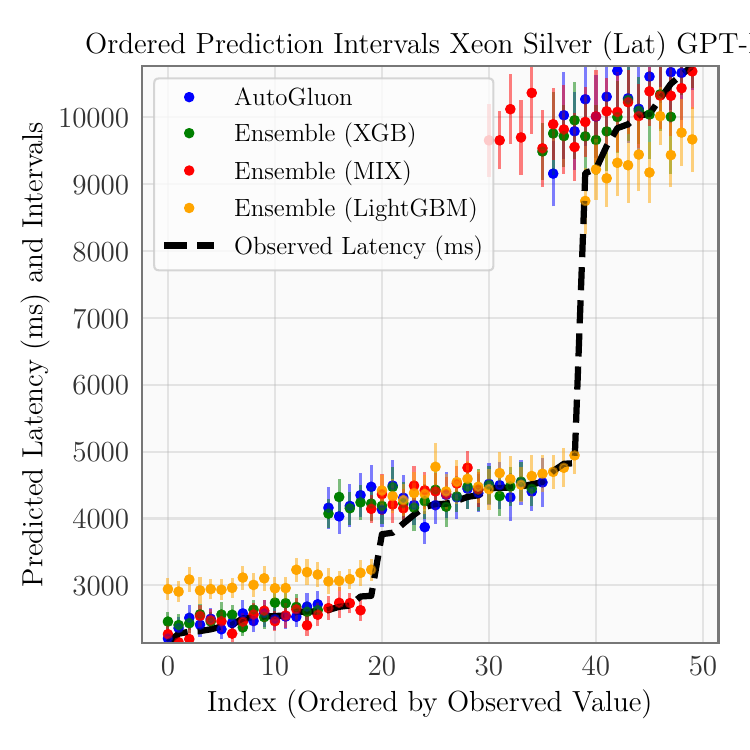}
\end{minipage}
\begin{minipage}{.16\linewidth}
  \centering
\includegraphics[width=\linewidth]{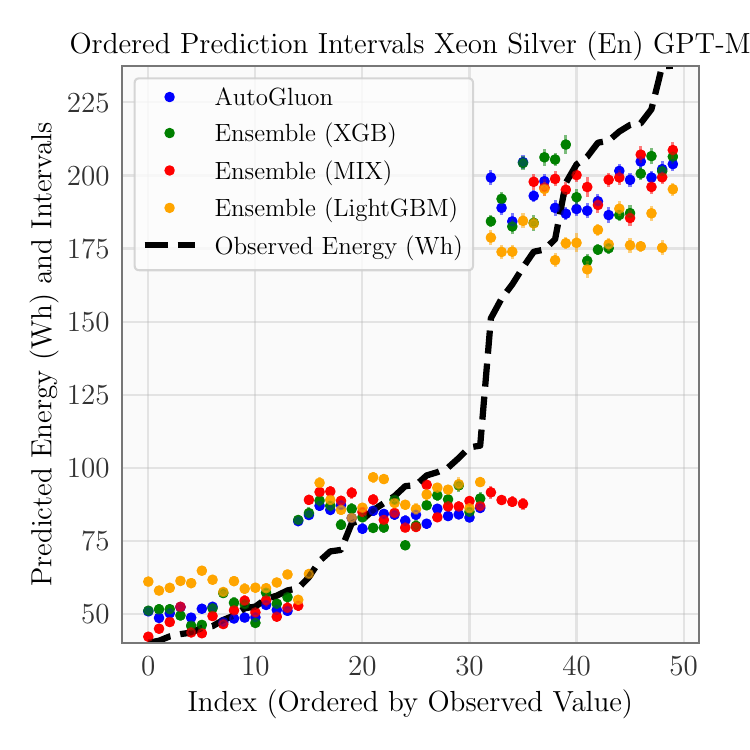}
\end{minipage}
\begin{minipage}{.16\linewidth}
  \centering
\includegraphics[width=\linewidth]{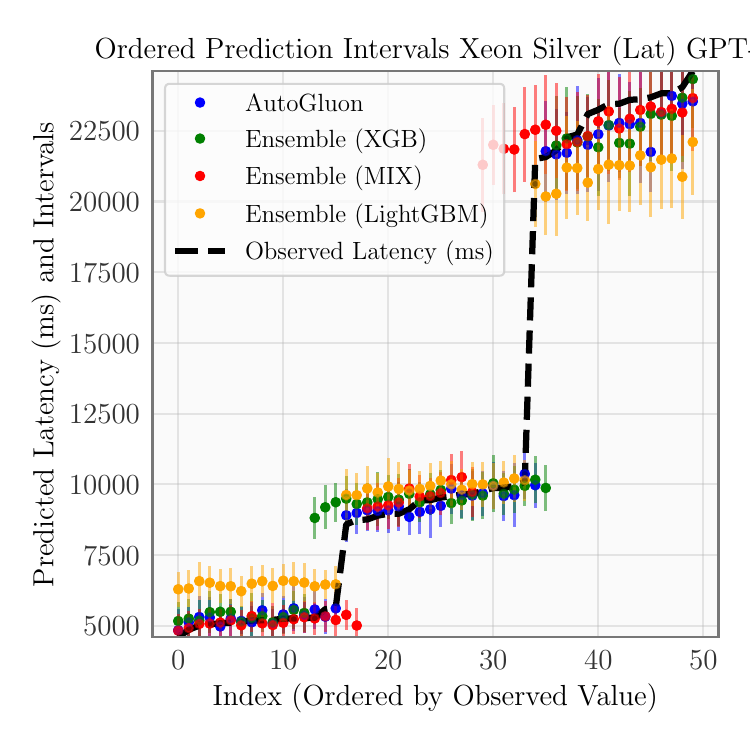}
\end{minipage}
\begin{minipage}{.16\linewidth}
  \centering
\includegraphics[width=\linewidth]{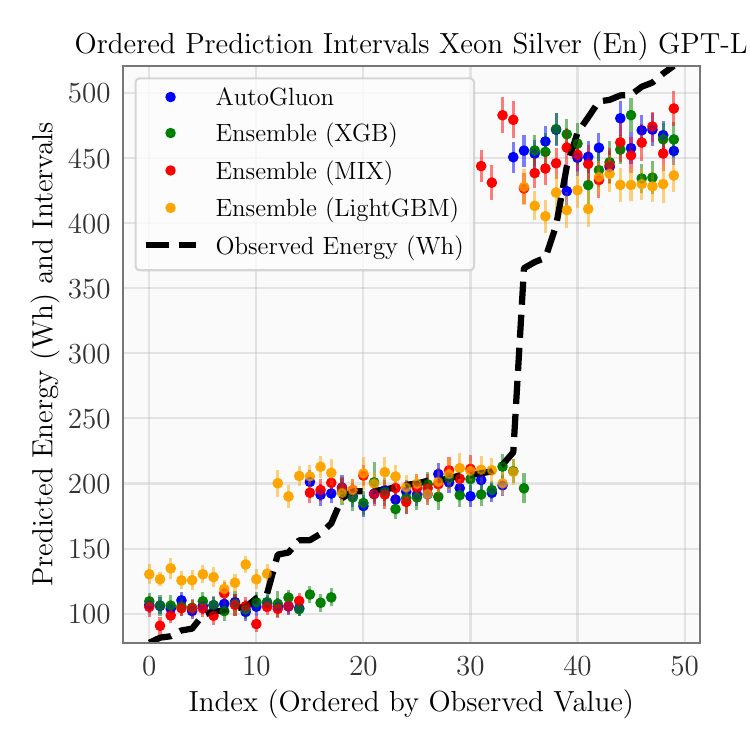}
\end{minipage}
\caption{Prediction Intervals Xeon Silver}
\label{fig:interval_bounds_xeon_silver}

\end{figure}

\begin{figure}[H]
\begin{minipage}{.16\linewidth}
  \centering
  \includegraphics[width=\linewidth]{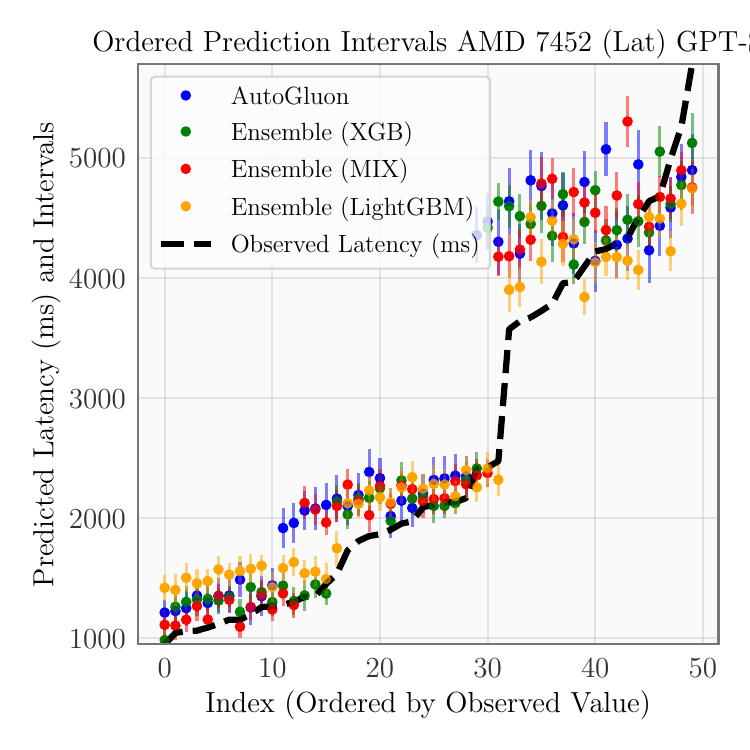}
\end{minipage}%
\begin{minipage}{.16\linewidth}
  \centering
\includegraphics[width=\linewidth]{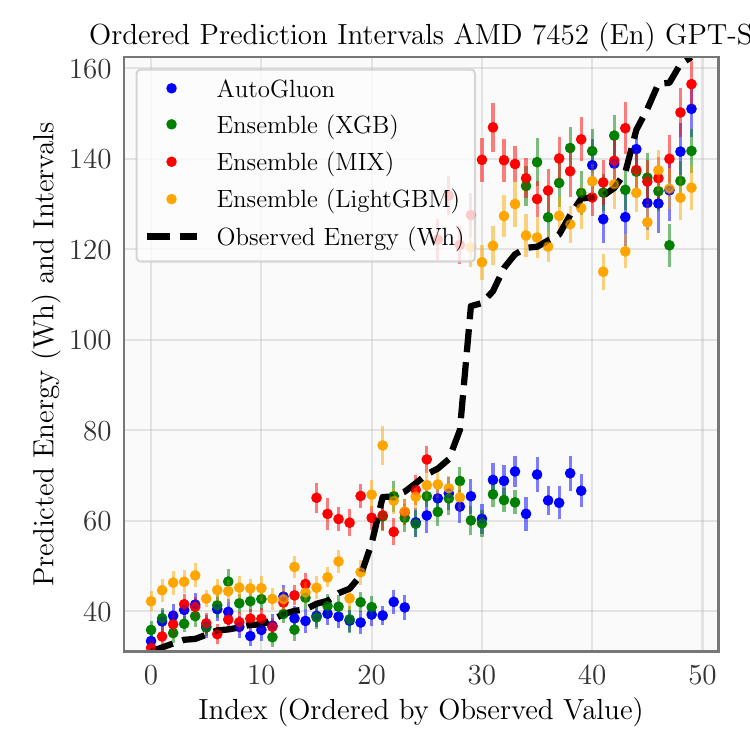}
\end{minipage}
\begin{minipage}{.16\linewidth}
  \centering
\includegraphics[width=\linewidth]{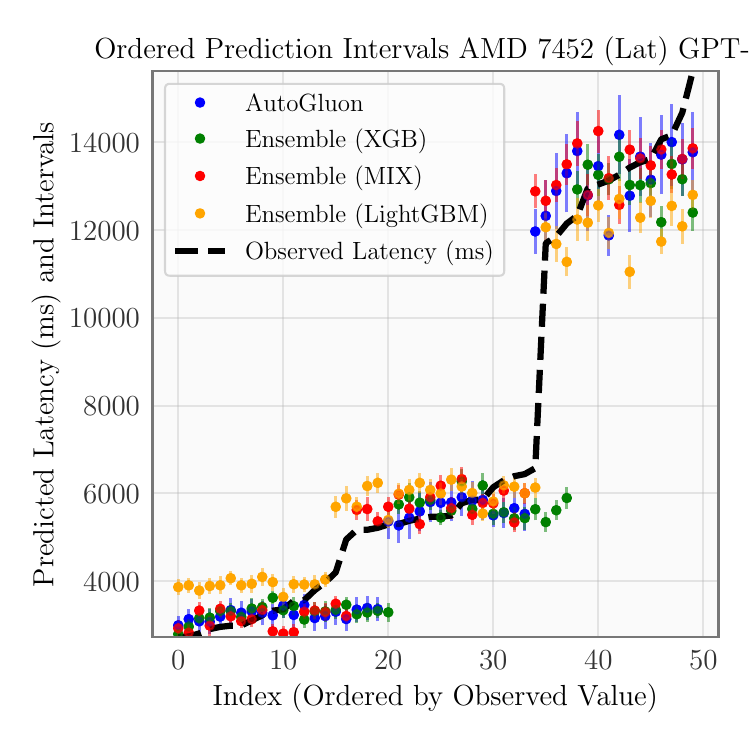}
\end{minipage}
\begin{minipage}{.16\linewidth}
  \centering
\includegraphics[width=\linewidth]{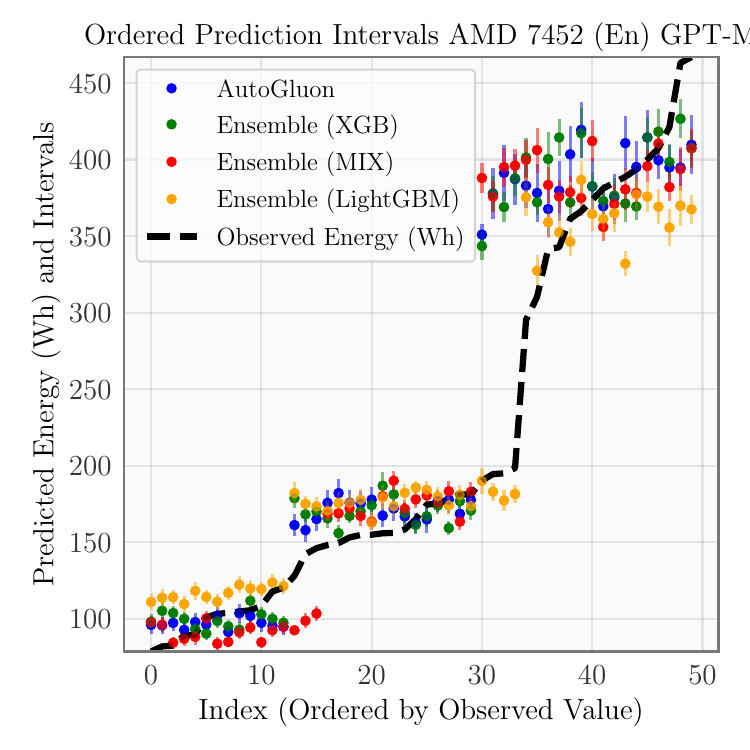}
\end{minipage}
\begin{minipage}{.16\linewidth}
  \centering
\includegraphics[width=\linewidth]{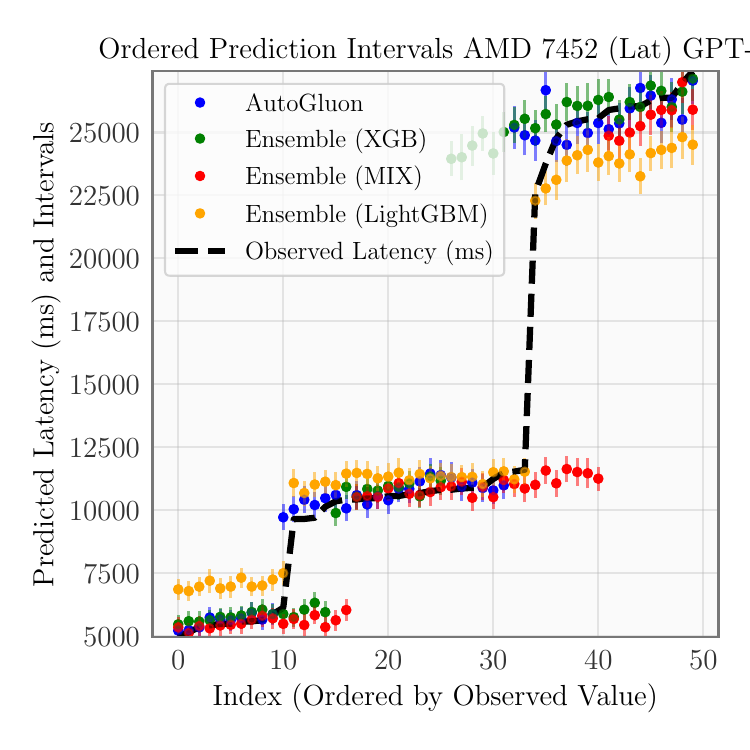}
\end{minipage}
\begin{minipage}{.16\linewidth}
  \centering
\includegraphics[width=\linewidth]{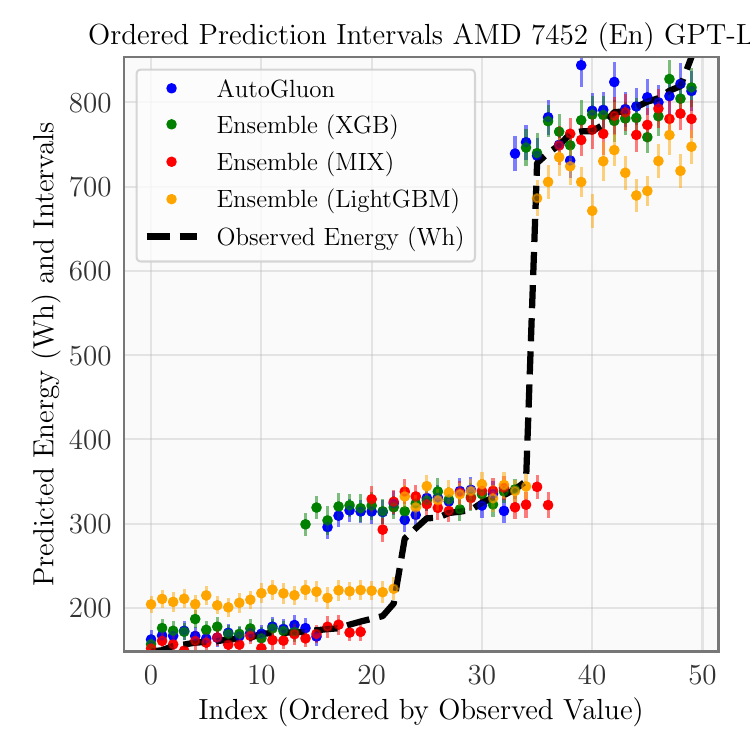}
\end{minipage}
\caption{Prediction Intervals CPU AMD 7452}
\label{fig:interval_amd_7452}

\end{figure}

\begin{figure}[H]
\begin{minipage}{.16\linewidth}
  \centering
  \includegraphics[width=\linewidth]{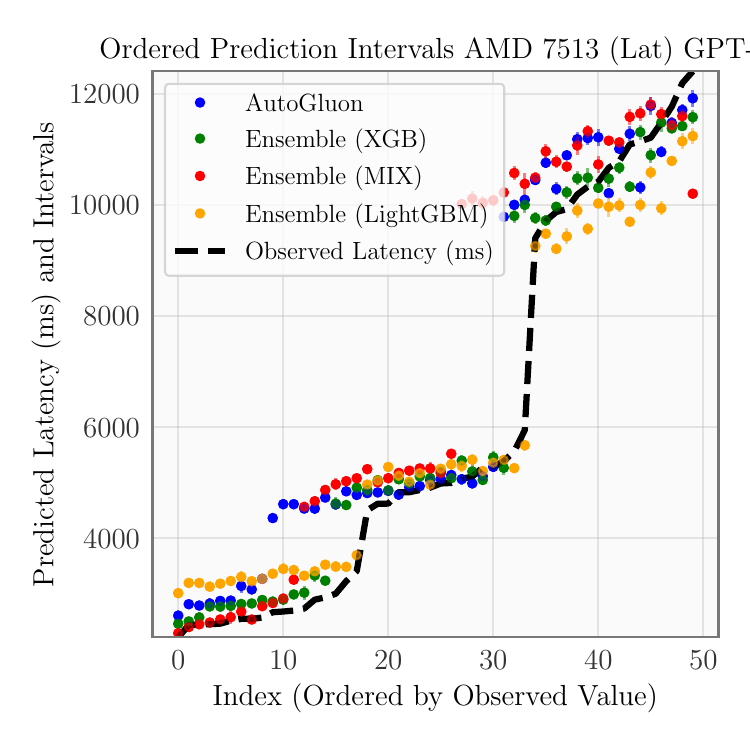}
\end{minipage}%
\begin{minipage}{.16\linewidth}
  \centering
\includegraphics[width=\linewidth]{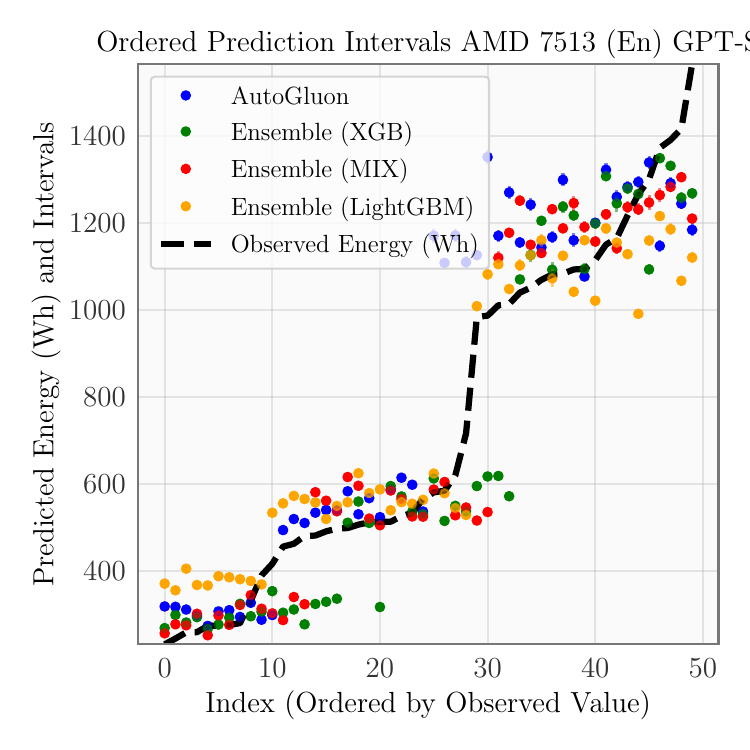}
\end{minipage}
\begin{minipage}{.16\linewidth}
  \centering
\includegraphics[width=\linewidth]{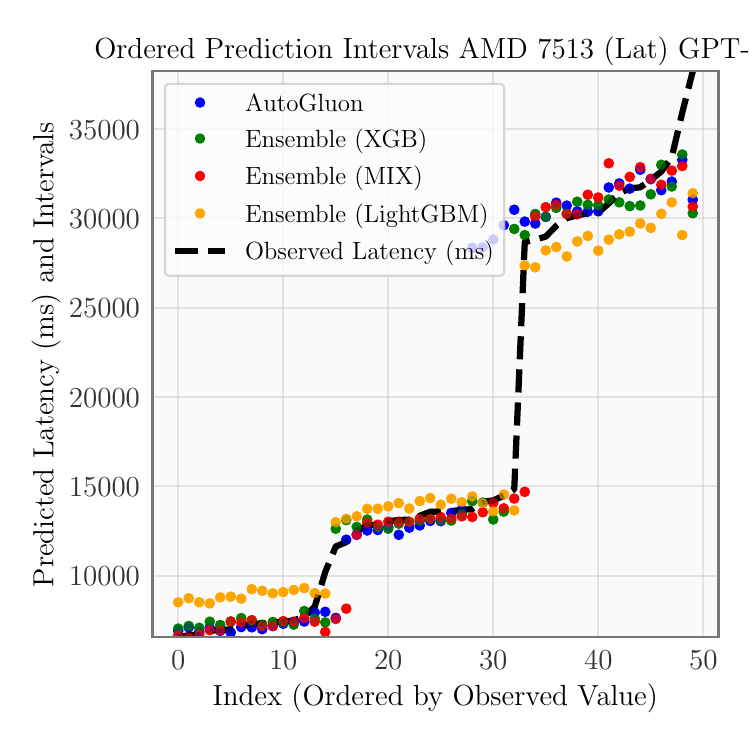}
\end{minipage}
\begin{minipage}{.16\linewidth}
  \centering
\includegraphics[width=\linewidth]{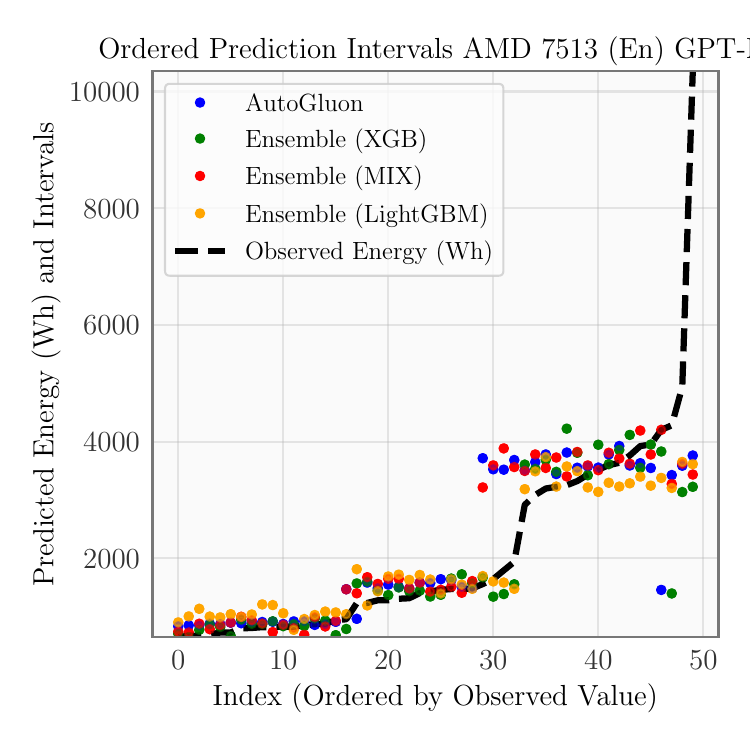}
\end{minipage}
\begin{minipage}{.16\linewidth}
  \centering
\includegraphics[width=\linewidth]{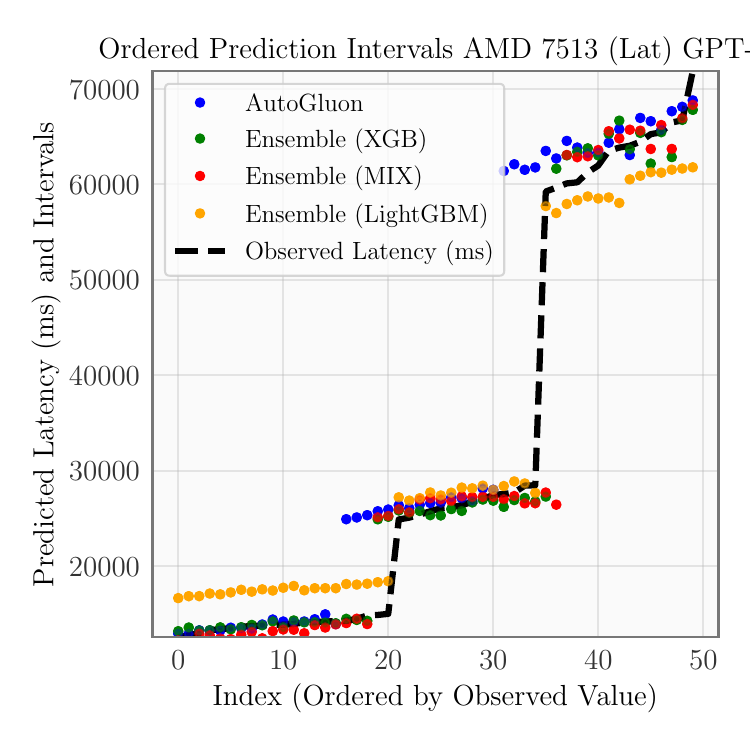}
\end{minipage}
\begin{minipage}{.16\linewidth}
  \centering
\includegraphics[width=\linewidth]{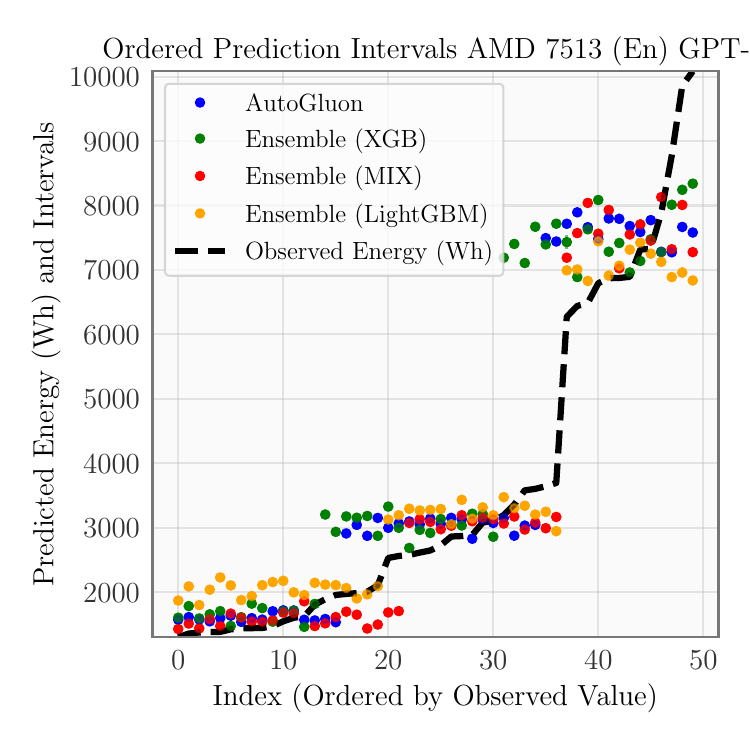}
\end{minipage}
\caption{Prediction Intervals CPU AMD 7513}
\label{fig:interval_amd_7513}

\end{figure}

\begin{figure}[H]
\begin{minipage}{.16\linewidth}
  \centering
  \includegraphics[width=\linewidth]{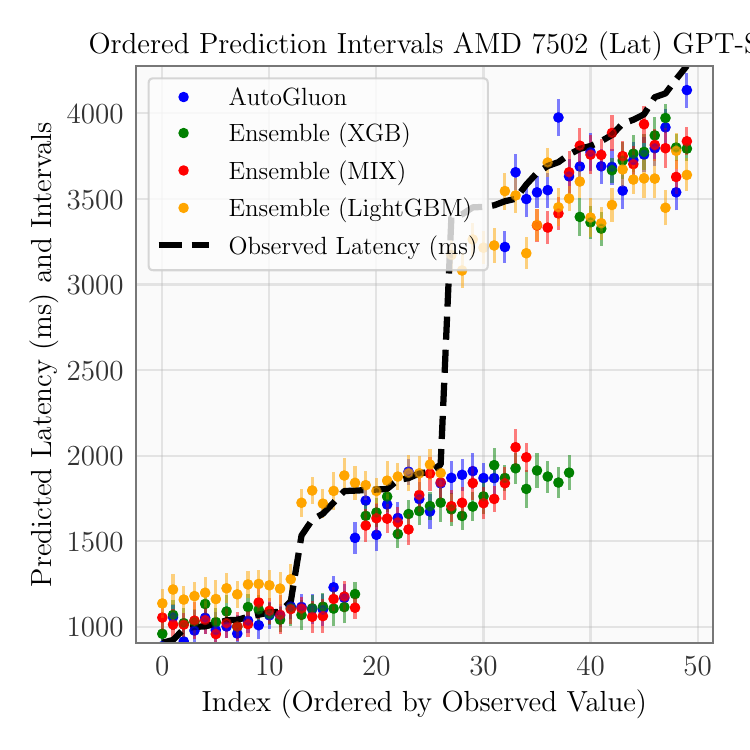}
\end{minipage}%
\begin{minipage}{.16\linewidth}
  \centering
\includegraphics[width=\linewidth]{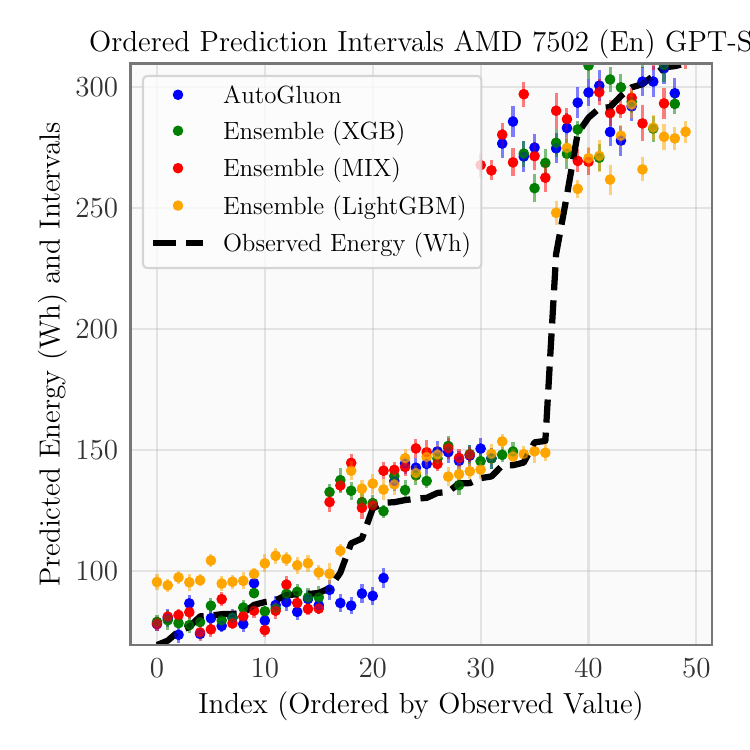}
\end{minipage}
\begin{minipage}{.16\linewidth}
  \centering
\includegraphics[width=\linewidth]{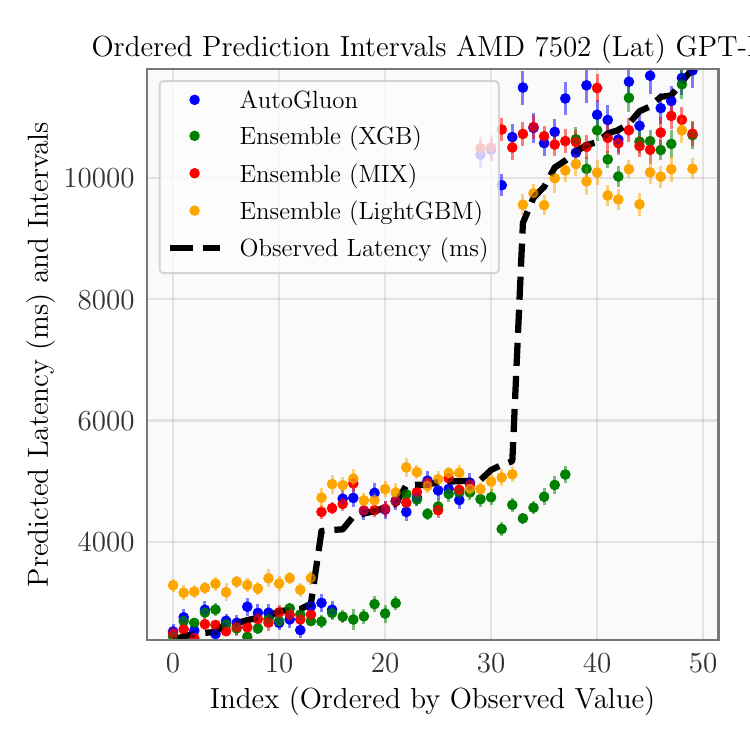}
\end{minipage}
\begin{minipage}{.16\linewidth}
  \centering
\includegraphics[width=\linewidth]{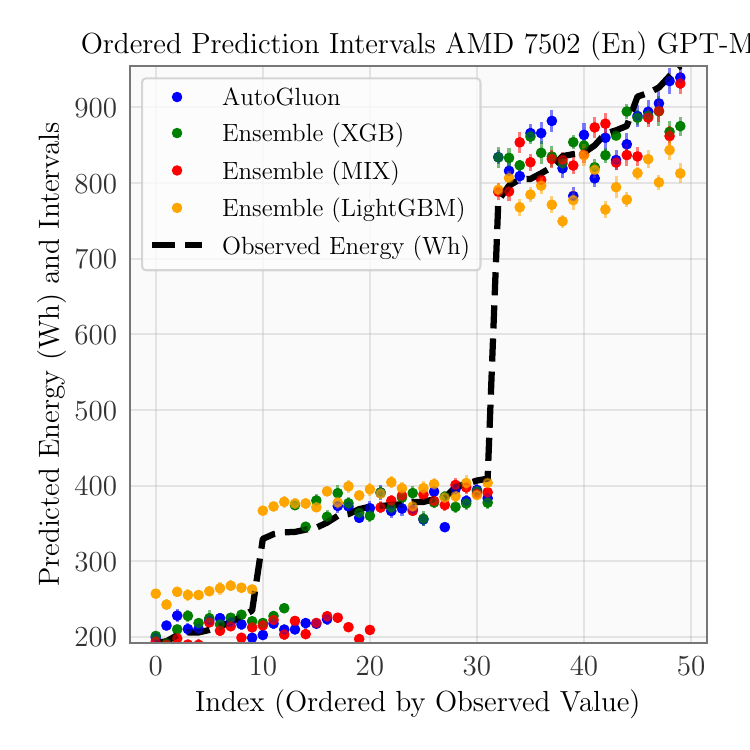}
\end{minipage}
\begin{minipage}{.16\linewidth}
  \centering
\includegraphics[width=\linewidth]{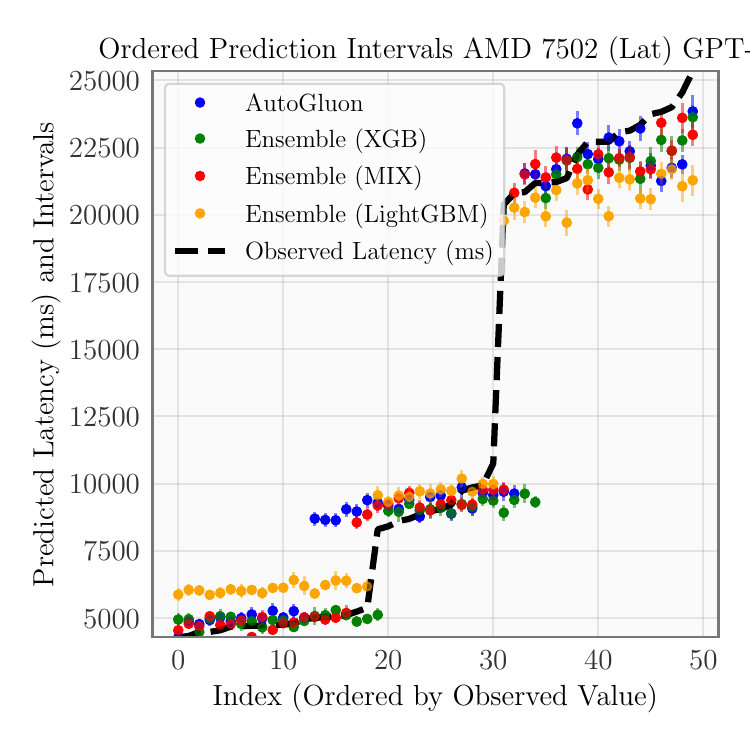}
\end{minipage}
\begin{minipage}{.16\linewidth}
  \centering
\includegraphics[width=\linewidth]{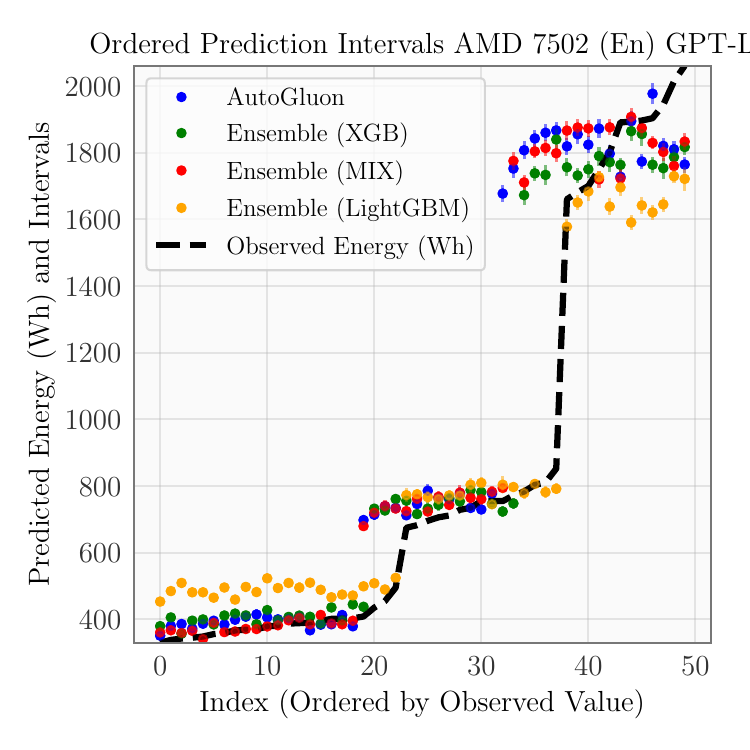}
\end{minipage}
\caption{Prediction Intervals CPU AMD 7502}
\label{fig:interval_amd_7502}

\end{figure}

In Tables \ref{tab:surrogatemetrics_combined_gpu_s}-\ref{tab:surrogatemetrics_combined_cpu_l} and Figures \ref{fig:calib_area_a100}-\ref{fig:interval_amd_7502} we show additional results that complement results shown in Table~\ref{tab:surrogatemetrics} and Figure~\ref{fig:uct_plots}.

\begin{landscape}
    \begin{table}[ht]\centering
    \caption{Various performance metrics of surrogates evaluated to predict latencies and energies on GPT-S for different GPUs. The arrow on the side of the metric name indicates if lower or higher is better.}
    \resizebox{\linewidth}{!}{%
    \begin{tabular}{llclclclclclclclclclclcl} 
    \toprule
    \multirow{3}{*}{\textbf{GPU}} & \multirow{3}{*}{\textbf{Surrogate}} & \multicolumn{12}{c}{\textbf{Accuracy}}                                                                                                                                                                                                                                                                                                   & \multicolumn{6}{c}{\textbf{Calibration}}                                                                                                                      \\ 
    \cmidrule(lr){3-14}\cmidrule(lr){15-20}
    & & \multicolumn{2}{c}{\textbf{MAE} $\downarrow$} & \multicolumn{2}{c}{\textbf{RMSE} $\downarrow$} & \multicolumn{2}{c}{\textbf{MDAE} $\downarrow$} & \multicolumn{2}{c}{\textbf{MARPD} $\downarrow$} & \multicolumn{2}{c}{\textbf{R\textsuperscript{2}} $\uparrow$} & \multicolumn{2}{c}{\textbf{Corr.} $\uparrow$} & \multicolumn{2}{c}{\textbf{RMS Cal.} $\downarrow$} & \multicolumn{2}{c}{\textbf{MA Cal.} $\downarrow$} & \multicolumn{2}{c}{\textbf{Miscal. Area} $\downarrow$} \\ 
    \midrule
    \multicolumn{1}{c}{} & & \textbf{latencies} & \multicolumn{1}{c}{\textbf{energies}} & \textbf{latencies} & \multicolumn{1}{c}{\textbf{energies}} & \textbf{latencies} & \multicolumn{1}{c}{\textbf{energies}} & \textbf{latencies} & \multicolumn{1}{c}{\textbf{energies}} & \textbf{latencies} & \multicolumn{1}{c}{\textbf{energies}} & \textbf{latencies} & \multicolumn{1}{c}{\textbf{energies}} & \textbf{latencies} & \multicolumn{1}{c}{\textbf{energies}} & \textbf{latencies} & \multicolumn{1}{c}{\textbf{energies}} & \textbf{latencies} & \multicolumn{1}{c}{\textbf{energies}} \\ 
    \cmidrule(lr){3-4} \cmidrule(lr){5-6} \cmidrule(lr){7-8} \cmidrule(lr){9-10} \cmidrule(lr){11-12} \cmidrule(lr){13-14} \cmidrule(lr){15-16} \cmidrule(lr){17-18} \cmidrule(lr){19-20}
    \multirow{4}{*}{A100} & \textbf{AutoGluon} & \textbf{0.39} & \textbf{0.03} & \textbf{0.50} & \textbf{0.28} & \textbf{0.33} & \textbf{0.02} & \textbf{0.23} & \textbf{5.68} & \textbf{1.00} & \textbf{0.17} & \textbf{1.00} & \textbf{0.42} & \textbf{0.46} & 0.09 & \textbf{0.40} & 0.08 & \textbf{0.41} & 0.08 \\ 
    & \textbf{Ensemble (XGB)} & 1.81 & 0.05 & 2.37 & 0.29 & 1.48 & 0.02 & 1.03 & 8.32 & 1.00 & 0.10 & 1.00 & 0.34 & 0.54 & \textbf{0.08} & 0.47 & 0.07 & 0.47 & 0.07 \\ 
    & \textbf{Ensemble (LGB)} & 8.74 & 0.05 & 10.14 & 0.29 & 8.37 & 0.03 & 5.06 & 9.70 & 0.97 & 0.14 & 1.00 & 0.37 & 0.57 & 0.08 & 0.49 & \textbf{0.06} & 0.49 & \textbf{0.06} \\ 
    & \textbf{Ensemble (Mix)} & 1.93 & 0.04 & 2.51 & 0.29 & 1.57 & 0.02 & 1.09 & 7.42 & 1.00 & 0.14 & 1.00 & 0.38 & 0.55 & 0.08 & 0.47 & 0.07 & 0.47 & 0.07 \\ 
    \cmidrule{1-20}
    \multirow{4}{*}{A40} & \textbf{AutoGluon} & \textbf{0.52} & \textbf{0.09} & \textbf{0.76} & \textbf{0.15} & \textbf{0.32} & \textbf{0.03} & \textbf{0.18} & \textbf{6.16} & \textbf{1.00} & \textbf{0.93} & \textbf{1.00} & \textbf{0.96} & \textbf{0.35} & 0.21 & \textbf{0.31} & 0.18 & \textbf{0.31} & 0.18 \\ 
    & \textbf{Ensemble (XGB)} & 3.16 & 0.09 & 4.19 & 0.16 & 2.43 & 0.03 & 1.26 & 6.56 & 1.00 & 0.92 & 1.00 & 0.96 & 0.54 & 0.16 & 0.46 & 0.14 & 0.47 & 0.15 \\ 
    & \textbf{Ensemble (LGB)} & 14.90 & 0.12 & 17.40 & 0.18 & 14.80 & 0.08 & 6.21 & 11.26 & 0.97 & 0.90 & 1.00 & 0.96 & 0.57 & \textbf{0.09} & 0.49 & \textbf{0.07} & 0.49 & \textbf{0.08} \\ 
    & \textbf{Ensemble (Mix)} & 3.16 & 0.09 & 4.32 & 0.15 & 2.30 & 0.03 & 1.20 & 6.49 & 1.00 & 0.93 & 1.00 & 0.96 & 0.54 & 0.17 & 0.46 & 0.15 & 0.47 & 0.15 \\ 
    \cmidrule{1-20}
    \multirow{4}{*}{H100} & \textbf{AutoGluon} & \textbf{0.18} & \textbf{0.02} & \textbf{0.25} & \textbf{0.03} & \textbf{0.14} & \textbf{0.01} & \textbf{0.23} & \textbf{3.34} & \textbf{1.00} & \textbf{0.90} & \textbf{1.00} & \textbf{0.95} & \textbf{0.19} & \textbf{0.05} & \textbf{0.17} & \textbf{0.04} & \textbf{0.17} & \textbf{0.04} \\ 
    & \textbf{Ensemble (XGB)} & 0.76 & 0.02 & 0.99 & 0.03 & 0.61 & 0.01 & 0.98 & 3.57 & 1.00 & 0.90 & 1.00 & 0.95 & 0.48 & 0.05 & 0.41 & 0.05 & 0.42 & 0.05 \\ 
    & \textbf{Ensemble (LGB)} & 3.45 & 0.02 & 4.02 & 0.04 & 3.34 & 0.02 & 4.44 & 4.64 & 0.97 & 0.87 & 1.00 & 0.95 & 0.55 & 0.18 & 0.48 & 0.16 & 0.48 & 0.16 \\ 
    & \textbf{Ensemble (Mix)} & 0.77 & 0.02 & 1.00 & 0.03 & 0.64 & 0.01 & 0.99 & 3.50 & 1.00 & 0.90 & 1.00 & 0.95 & 0.47 & 0.05 & 0.41 & 0.05 & 0.42 & 0.05 \\ 
    \cmidrule{1-20}
    \multirow{4}{*}{RTX2080} & \textbf{AutoGluon} & \textbf{20.01} & \textbf{0.09} & \textbf{28.40} & \textbf{0.14} & \textbf{11.91} & \textbf{0.07} & \textbf{4.54} & \textbf{6.45} & \textbf{0.98} & \textbf{0.99} & \textbf{0.99} & \textbf{0.99} & \textbf{0.35} & 0.26 & \textbf{0.29} & 0.23 & \textbf{0.29} & 0.23 \\ 
    & \textbf{Ensemble (XGB)} & 20.59 & 0.10 & 29.58 & 0.15 & 12.17 & 0.07 & 4.60 & 6.95 & 0.97 & 0.98 & 0.99 & 0.99 & 0.42 & 0.24 & 0.35 & 0.21 & 0.36 & 0.21 \\
    & \textbf{Ensemble (LGB)} & 32.25 & 0.20 & 42.97 & 0.25 & 22.52 & 0.18 & 7.14 & 14.77 & 0.94 & 0.96 & 0.99 & 0.99 & 0.53 & \textbf{0.05} & 0.46 & \textbf{0.04} & 0.46 & \textbf{0.04} \\
    & \textbf{Ensemble (Mix)} & 20.79 & 0.10 & 29.25 & 0.15 & 13.27 & 0.08 & 4.71 & 7.52 & 0.97 & 0.98 & 0.99 & 0.99 & 0.42 & 0.21 & 0.36 & 0.19 & 0.37 & 0.20 \\ 
    \cmidrule{1-20}
    \multirow{4}{*}{RTX3080} & \textbf{AutoGluon} & \textbf{6.21} & \textbf{0.06} & \textbf{8.83} & \textbf{0.12} & \textbf{3.69} & \textbf{0.03} & \textbf{1.97} & \textbf{4.10} & \textbf{1.00} & \textbf{0.96} & \textbf{1.00} & \textbf{0.98} & \textbf{0.43} & 0.03 & \textbf{0.36} & 0.03 & \textbf{0.37} & 0.03 \\
    & \textbf{Ensemble (XGB)} & 7.18 & 0.06 & 10.25 & 0.12 & 4.26 & 0.04 & 2.21 & 4.67 & 1.00 & 0.95 & 1.00 & 0.98 & 0.47 & \textbf{0.02} & 0.41 & \textbf{0.02} & 0.41 & \textbf{0.02} \\
    & \textbf{Ensemble (LGB)} & 22.76 & 0.10 & 27.36 & 0.16 & 21.79 & 0.08 & 7.45 & 8.40 & 0.97 & 0.92 & 1.00 & 0.98 & 0.56 & 0.25 & 0.48 & 0.22 & 0.49 & 0.23 \\ 
    & \textbf{Ensemble (Mix)} & 7.49 & 0.07 & 10.43 & 0.13 & 5.05 & 0.04 & 2.37 & 4.74 & 1.00 & 0.95 & 1.00 & 0.98 & 0.48 & 0.02 & 0.42 & 0.02 & 0.42 & 0.02 \\
    \cmidrule{1-20}
    \multirow{4}{*}{A6000} & \textbf{AutoGluon} & \textbf{1.74} & \textbf{0.05} & \textbf{2.68} & \textbf{0.06} & \textbf{1.03} & \textbf{0.04} & \textbf{0.85} & \textbf{5.95} & \textbf{1.00} & \textbf{0.93} & \textbf{1.00} & \textbf{0.96} & \textbf{0.47} & 0.04 & \textbf{0.40} & 0.03 & \textbf{0.41} & 0.03 \\
    & \textbf{Ensemble (XGB)} & 3.34 & 0.05 & 4.45 & 0.06 & 2.42 & 0.04 & 1.54 & 6.17 & 1.00 & 0.92 & 1.00 & 0.96 & 0.53 & \textbf{0.02} & 0.45 & \textbf{0.02} & 0.46 & \textbf{0.02} \\
    & \textbf{Ensemble (LGB)} & 13.46 & 0.05 & 15.83 & 0.07 & 13.16 & 0.04 & 6.48 & 6.85 & 0.97 & 0.90 & 1.00 & 0.96 & 0.57 & 0.04 & 0.49 & 0.04 & 0.49 & 0.04 \\ 
    & \textbf{Ensemble (Mix)} & 3.39 & 0.05 & 4.54 & 0.06 & 2.50 & 0.04 & 1.54 & 6.22 & 1.00 & 0.92 & 1.00 & 0.96 & 0.54 & 0.03 & 0.46 & 0.03 & 0.47 & 0.03 \\
    \cmidrule{1-20}
    \multirow{4}{*}{V100} & \textbf{AutoGluon} & \textbf{7.49} & \textbf{0.02} & \textbf{10.51} & \textbf{0.14} & \textbf{3.94} & \textbf{0.01} & \textbf{2.21} & \textbf{2.64} & \textbf{0.99} & \textbf{0.70} & \textbf{1.00} & \textbf{0.84} & \textbf{0.44} & \textbf{0.36} & \textbf{0.38} & \textbf{0.32} & \textbf{0.39} & \textbf{0.33} \\ 
    & \textbf{Ensemble (XGB)} & 8.03 & 0.02 & 11.42 & 0.15 & 4.60 & 0.01 & 2.41 & 3.68 & 0.99 & 0.64 & 1.00 & 0.80 & 0.47 & 0.37 & 0.41 & 0.33 & 0.41 & 0.33 \\ 
    & \textbf{Ensemble (LGB)} & 18.87 & 0.04 & 23.54 & 0.15 & 16.73 & 0.03 & 6.32 & 7.22 & 0.96 & 0.66 & 1.00 & 0.83 & 0.55 & 0.52 & 0.47 & 0.45 & 0.48 & 0.45 \\ 
    & \textbf{Ensemble (Mix)} & 8.10 & 0.02 & 11.40 & 0.14 & 4.82 & 0.01 & 2.43 & 3.40 & 0.99 & 0.69 & 1.00 & 0.83 & 0.48 & 0.37 & 0.42 & 0.32 & 0.42 & 0.33 \\
    \cmidrule{1-20}
    \multirow{4}{*}{P100} & \textbf{AutoGluon} & \textbf{3.62} & \textbf{0.30} & \textbf{5.31} & \textbf{0.40} & \textbf{2.36} & \textbf{0.23} & \textbf{0.21} & \textbf{6.13} & \textbf{1.00} & \textbf{1.00} & \textbf{1.00} & \textbf{1.00} & \textbf{0.24} & 0.26 & \textbf{0.22} & 0.23 & \textbf{0.22} & 0.23 \\
    & \textbf{Ensemble (XGB)} & 22.53 & 0.34 & 31.55 & 0.45 & 16.57 & 0.27 & 1.35 & 7.06 & 1.00 & 1.00 & 1.00 & 1.00 & 0.52 & 0.21 & 0.45 & 0.19 & 0.45 & 0.19 \\ 
    & \textbf{Ensemble (LGB)} & 140.75 & 0.99 & 160.96 & 1.18 & 144.00 & 1.04 & 9.42 & 22.64 & 0.97 & 0.97 & 1.00 & 1.00 & 0.57 & 0.21 & 0.49 & \textbf{0.16} & 0.49 & \textbf{0.16} \\ 
    & \textbf{Ensemble (Mix)} & 26.29 & 0.39 & 36.45 & 0.49 & 19.65 & 0.33 & 1.55 & 7.87 & 1.00 & 0.99 & 1.00 & 1.00 & 0.52 & \textbf{0.18} & 0.45 & 0.16 & 0.46 & 0.16 \\  
    \bottomrule
    \end{tabular}
    }
    \vspace{-3ex}
    \label{tab:surrogatemetrics_combined_gpu_s}
    \end{table}
\end{landscape}

\begin{landscape}
    \begin{table}[ht]\centering
    \caption{Various performance metrics of surrogates evaluated to predict latencies and energies on GPT-S for different CPUs. The arrow on the side of the metric name indicates if lower or higher is better.}
    \resizebox{\linewidth}{!}{%
    \begin{tabular}{llclclclclclclclclclclcl} 
    \toprule
    \multirow{3}{*}{\textbf{CPU}} & \multirow{3}{*}{\textbf{Surrogate}} & \multicolumn{12}{c}{\textbf{Accuracy}}                                                                                                                                                                                                                                                                                                   & \multicolumn{6}{c}{\textbf{Calibration}}                                                                                                                      \\ 
    \cmidrule(lr){3-14}\cmidrule(lr){15-20}
    & & \multicolumn{2}{c}{\textbf{MAE} $\downarrow$} & \multicolumn{2}{c}{\textbf{RMSE} $\downarrow$} & \multicolumn{2}{c}{\textbf{MDAE} $\downarrow$} & \multicolumn{2}{c}{\textbf{MARPD} $\downarrow$} & \multicolumn{2}{c}{\textbf{R\textsuperscript{2}} $\uparrow$} & \multicolumn{2}{c}{\textbf{Corr.} $\uparrow$} & \multicolumn{2}{c}{\textbf{RMS Cal.} $\downarrow$} & \multicolumn{2}{c}{\textbf{MA Cal.} $\downarrow$} & \multicolumn{2}{c}{\textbf{Miscal. Area} $\downarrow$} \\ 
    \midrule
    \multicolumn{1}{c}{} & & \textbf{latencies} & \multicolumn{1}{c}{\textbf{energies}} & \textbf{latencies} & \multicolumn{1}{c}{\textbf{energies}} & \textbf{latencies} & \multicolumn{1}{c}{\textbf{energies}} & \textbf{latencies} & \multicolumn{1}{c}{\textbf{energies}} & \textbf{latencies} & \multicolumn{1}{c}{\textbf{energies}} & \textbf{latencies} & \multicolumn{1}{c}{\textbf{energies}} & \textbf{latencies} & \multicolumn{1}{c}{\textbf{energies}} & \textbf{latencies} & \multicolumn{1}{c}{\textbf{energies}} & \textbf{latencies} & \multicolumn{1}{c}{\textbf{energies}} \\ 
    \cmidrule(lr){3-4} \cmidrule(lr){5-6} \cmidrule(lr){7-8} \cmidrule(lr){9-10} \cmidrule(lr){11-12} \cmidrule(lr){13-14} \cmidrule(lr){15-16} \cmidrule(lr){17-18} \cmidrule(lr){19-20}
    \multirow{4}{*}{Xeon Silver} & \textbf{AutoGluon} & \textbf{1534.66} & \textbf{8.94} & \textbf{1982.43} & \textbf{11.88} & \textbf{1242.68} & \textbf{6.52} & \textbf{3.89} & \textbf{11.69} & \textbf{0.99} & \textbf{0.91} & \textbf{1.00} & \textbf{0.95} & \textbf{0.14} & \textbf{0.50} & \textbf{0.13} & \textbf{0.43} & \textbf{0.13} & \textbf{0.43} \\
    & \textbf{Ensemble (XGB)} & 1606.20 & 9.14 & 2107.51 & 12.21 & 1297.02 & 6.92 & 4.11 & 12.08 & 0.99 & 0.90 & 0.99 & 0.95 & 0.21 & 0.53 & 0.19 & 0.45 & 0.20 & 0.46 \\ 
    & \textbf{Ensemble (LGB)} & 3243.75 & 10.09 & 4020.23 & 13.66 & 2833.39 & 8.42 & 9.15 & 14.42 & 0.96 & 0.88 & 0.99 & 0.95 & 0.39 & 0.54 & 0.34 & 0.46 & 0.35 & 0.47 \\ 
    & \textbf{Ensemble (Mix)} & 1633.05 & 9.05 & 2146.05 & 12.09 & 1285.48 & 6.95 & 4.14 & 11.81 & 0.99 & 0.90 & 0.99 & 0.95 & 0.22 & 0.52 & 0.20 & 0.45 & 0.20 & 0.46 \\ 
    \cmidrule{1-20}
    \multirow{4}{*}{Xeon Gold} & \textbf{AutoGluon} & \textbf{1161.51} & \textbf{2.62} & \textbf{1634.77} & \textbf{3.37} & \textbf{833.97} & \textbf{2.08} & \textbf{6.49} & \textbf{9.40} & \textbf{0.96} & \textbf{0.92} & \textbf{0.98} & \textbf{0.96} & 0.12 & \textbf{0.02} & 0.10 & \textbf{0.02} & 0.10 & \textbf{0.02} \\ 
    & \textbf{Ensemble (XGB)} & 1224.12 & 2.69 & 1695.22 & 3.47 & 896.84 & 2.19 & 6.88 & 9.66 & 0.95 & 0.92 & 0.98 & 0.96 & \textbf{0.06} & 0.04 & \textbf{0.05} & 0.04 & \textbf{0.05} & 0.04 \\ 
    & \textbf{Ensemble (LGB)} & 1690.01 & 3.20 & 2144.80 & 3.98 & 1456.95 & 2.78 & 9.58 & 11.52 & 0.93 & 0.89 & 0.98 & 0.96 & 0.09 & 0.12 & 0.09 & 0.11 & 0.09 & 0.11 \\ 
    & \textbf{Ensemble (Mix)} & 1200.12 & 2.65 & 1677.90 & 3.42 & 874.13 & 2.16 & 6.68 & 9.47 & 0.95 & 0.92 & 0.98 & 0.96 & 0.07 & 0.04 & 0.06 & 0.04 & 0.06 & 0.04 \\ 
    \cmidrule{1-20}
    \multirow{4}{*}{AMD 7502} & \textbf{AutoGluon} & \textbf{1570.47} & \textbf{15.84} & \textbf{2297.45} & \textbf{22.15} & \textbf{1021.43} & \textbf{11.24} & \textbf{3.84} & \textbf{4.89} & \textbf{0.99} & \textbf{0.98} & \textbf{0.99} & \textbf{0.99} & \textbf{0.17} & \textbf{0.35} & \textbf{0.15} & \textbf{0.31} & \textbf{0.15} & \textbf{0.31} \\ 
    & \textbf{Ensemble (XGB)} & 1697.98 & 16.83 & 2420.66 & 23.22 & 1107.53 & 12.16 & 4.12 & 5.16 & 0.99 & 0.98 & 0.99 & 0.99 & 0.20 & 0.39 & 0.17 & 0.34 & 0.17 & 0.35 \\ 
    & \textbf{Ensemble (LGB)} & 3764.57 & 31.28 & 4531.32 & 38.53 & 3489.26 & 26.99 & 9.94 & 10.26 & 0.96 & 0.95 & 0.99 & 0.99 & 0.45 & 0.51 & 0.39 & 0.44 & 0.40 & 0.44 \\ 
    & \textbf{Ensemble (Mix)} & 1693.33 & 16.82 & 2425.90 & 23.07 & 1110.10 & 12.23 & 4.13 & 5.21 & 0.99 & 0.98 & 0.99 & 0.99 & 0.20 & 0.41 & 0.17 & 0.36 & 0.17 & 0.36 \\ 
    \cmidrule{1-20}
    \multirow{4}{*}{AMD 7513} & \textbf{AutoGluon} & \textbf{3321.83} & \textbf{150.24} & \textbf{5518.96} & \textbf{221.10} & \textbf{2174.08} & \textbf{103.02} & \textbf{2.76} & \textbf{10.55} & \textbf{0.99} & \textbf{0.93} & \textbf{1.00} & \textbf{0.96} & \textbf{0.38} & \textbf{0.53} & \textbf{0.34} & \textbf{0.46} & \textbf{0.34} & \textbf{0.46} \\ 
    & \textbf{Ensemble (XGB)} & 3761.37 & 157.08 & 6054.88 & 230.02 & 2633.74 & 108.79 & 3.23 & 11.14 & 0.99 & 0.92 & 1.00 & 0.96 & 0.41 & 0.54 & 0.36 & 0.46 & 0.36 & 0.47 \\ 
    & \textbf{Ensemble (LGB)} & 11204.40 & 187.74 & 13456.74 & 268.02 & 10465.21 & 150.15 & 10.68 & 14.91 & 0.96 & 0.89 & 1.00 & 0.96 & 0.55 & 0.54 & 0.47 & 0.47 & 0.48 & 0.47 \\ 
    & \textbf{Ensemble (Mix)} & 3852.39 & 154.29 & 6098.78 & 226.36 & 2635.35 & 104.72 & 3.22 & 10.85 & 0.99 & 0.92 & 1.00 & 0.96 & 0.41 & 0.54 & 0.36 & 0.47 & 0.36 & 0.47 \\
    \cmidrule{1-20}
    \multirow{4}{*}{AMD 7452} & \textbf{AutoGluon} & \textbf{4879.20} & \textbf{14.75} & \textbf{6100.78} & \textbf{18.95} & 3820.30 & 11.39 & \textbf{9.02} & \textbf{9.21} & \textbf{0.95} & \textbf{0.95} & \textbf{0.98} & \textbf{0.97} & \textbf{0.41} & \textbf{0.47} & \textbf{0.36} & \textbf{0.41} & \textbf{0.37} & \textbf{0.41} \\ 
    & \textbf{Ensemble (XGB)} & 4930.64 & 15.06 & 6253.26 & 19.54 & 3817.53 & 11.60 & 9.16 & 9.46 & 0.95 & 0.95 & 0.98 & 0.97 & 0.44 & 0.47 & 0.39 & 0.41 & 0.39 & 0.42 \\ 
    & \textbf{Ensemble (LGB)} & 5907.72 & 18.06 & 7569.56 & 23.38 & 4669.03 & 14.40 & 11.97 & 12.29 & 0.93 & 0.92 & 0.98 & 0.97 & 0.45 & 0.48 & 0.40 & 0.41 & 0.40 & 0.42 \\ 
    & \textbf{Ensemble (Mix)} & 4903.38 & 14.94 & 6172.90 & 19.31 & \textbf{3738.50} & \textbf{11.25} & 9.12 & 9.36 & 0.95 & 0.95 & 0.98 & 0.97 & 0.46 & 0.48 & 0.40 & 0.42 & 0.40 & 0.42 \\
    
    \bottomrule
    \end{tabular}
    }
    \vspace{-3ex}
    \label{tab:surrogatemetrics_combined_cpu_s}
    \end{table}
\end{landscape}
\begin{landscape}
    \begin{table}[ht]\centering
    \caption{Various performance metrics of surrogates evaluated to predict latencies and energies on GPT-M for different GPUs. The arrow on the side of the metric name indicates if lower or higher is better.}
    \resizebox{\linewidth}{!}{%
    \begin{tabular}{llclclclclclclclclclclcl} 
    \toprule
    \multirow{3}{*}{\textbf{GPU}} & \multirow{3}{*}{\textbf{Surrogate}} & \multicolumn{12}{c}{\textbf{Accuracy}}                                                                                                                                                                                                                                                                                                   & \multicolumn{6}{c}{\textbf{Calibration}}                                                                                                                      \\ 
    \cmidrule(lr){3-14}\cmidrule(lr){15-20}
    & & \multicolumn{2}{c}{\textbf{MAE} $\downarrow$} & \multicolumn{2}{c}{\textbf{RMSE} $\downarrow$} & \multicolumn{2}{c}{\textbf{MDAE} $\downarrow$} & \multicolumn{2}{c}{\textbf{MARPD} $\downarrow$} & \multicolumn{2}{c}{\textbf{R\textsuperscript{2}} $\uparrow$} & \multicolumn{2}{c}{\textbf{Corr.} $\uparrow$} & \multicolumn{2}{c}{\textbf{RMS Cal.} $\downarrow$} & \multicolumn{2}{c}{\textbf{MA Cal.} $\downarrow$} & \multicolumn{2}{c}{\textbf{Miscal. Area} $\downarrow$} \\ 
    \midrule
    \multicolumn{1}{c}{} & & \textbf{latencies} & \multicolumn{1}{c}{\textbf{energies}} & \textbf{latencies} & \multicolumn{1}{c}{\textbf{energies}} & \textbf{latencies} & \multicolumn{1}{c}{\textbf{energies}} & \textbf{latencies} & \multicolumn{1}{c}{\textbf{energies}} & \textbf{latencies} & \multicolumn{1}{c}{\textbf{energies}} & \textbf{latencies} & \multicolumn{1}{c}{\textbf{energies}} & \textbf{latencies} & \multicolumn{1}{c}{\textbf{energies}} & \textbf{latencies} & \multicolumn{1}{c}{\textbf{energies}} & \textbf{latencies} & \multicolumn{1}{c}{\textbf{energies}} \\ 
    \cmidrule(lr){3-4} \cmidrule(lr){5-6} \cmidrule(lr){7-8} \cmidrule(lr){9-10} \cmidrule(lr){11-12} \cmidrule(lr){13-14} \cmidrule(lr){15-16} \cmidrule(lr){17-18} \cmidrule(lr){19-20}
    \multirow{4}{*}{A100} & \textbf{AutoGluon} & \textbf{2.31} & \textbf{0.10} & \textbf{3.59} & \textbf{0.36} & \textbf{1.17} & \textbf{0.06} & \textbf{0.74} & \textbf{6.50} & \textbf{1.00} & \textbf{0.84} & \textbf{1.00} & \textbf{0.92} & \textbf{0.45} & 0.40 & \textbf{0.39} & 0.35 & \textbf{0.39} & 0.36 \\ &
\textbf{Ensemble (XGB)} & 3.35 & 0.13 & 4.66 & 0.39 & 2.30 & 0.07 & 1.17 & 7.94 & 1.00 & 0.81 & 1.00 & 0.90 & 0.50 & 0.36 & 0.43 & 0.32 & 0.44 & 0.32 \\ &
\textbf{Ensemble (LGB)} & 13.88 & 0.16 & 16.12 & 0.39 & 13.82 & 0.13 & 5.37 & 12.00 & 0.97 & 0.81 & 1.00 & 0.91 & 0.57 & \textbf{0.24} & 0.49 & \textbf{0.21} & 0.49 & \textbf{0.22} \\ &
\textbf{Ensemble (Mix)} & 3.21 & 0.12 & 4.50 & 0.37 & 2.21 & 0.07 & 1.11 & 7.32 & 1.00 & 0.82 & 1.00 & 0.91 & 0.48 & 0.37 & 0.42 & 0.33 & 0.43 & 0.33 \\ 
    \cmidrule{1-20}
    \multirow{4}{*}{A40} & \textbf{AutoGluon} & \textbf{1.12} & \textbf{0.26} & \textbf{1.68} & \textbf{0.35} & \textbf{0.67} & \textbf{0.18} & \textbf{0.28} & \textbf{6.38} & \textbf{1.00} & \textbf{0.98} & \textbf{1.00} & \textbf{0.99} & \textbf{0.45} & 0.20 & \textbf{0.39} & 0.18 & \textbf{0.40} & 0.18 \\ &
\textbf{Ensemble (XGB)} & 4.16 & 0.27 & 5.58 & 0.36 & 3.19 & 0.20 & 1.17 & 6.82 & 1.00 & 0.98 & 1.00 & 0.99 & 0.54 & 0.18 & 0.47 & 0.16 & 0.47 & 0.16 \\ &
\textbf{Ensemble (LGB)} & 23.39 & 0.48 & 26.62 & 0.56 & 24.93 & 0.45 & 7.11 & 13.17 & 0.97 & 0.95 & 1.00 & 0.99 & 0.57 & \textbf{0.08} & 0.49 & \textbf{0.08} & 0.49 & \textbf{0.08} \\ &
\textbf{Ensemble (Mix)} & 4.01 & 0.27 & 5.47 & 0.36 & 2.94 & 0.19 & 1.09 & 6.63 & 1.00 & 0.98 & 1.00 & 0.99 & 0.54 & 0.19 & 0.46 & 0.17 & 0.47 & 0.1 \\ 
    \cmidrule{1-20}
    \multirow{4}{*}{H100} & \textbf{AutoGluon} & \textbf{0.27} & \textbf{0.05} & \textbf{0.37} & \textbf{0.07} & \textbf{0.21} & \textbf{0.04} & \textbf{0.24} & \textbf{4.57} & \textbf{1.00} & \textbf{0.95} & \textbf{1.00} & \textbf{0.98} & \textbf{0.22} & \textbf{0.24} & \textbf{0.19} & \textbf{0.21} & \textbf{0.20} & \textbf{0.21} \\ &
\textbf{Ensemble (XGB)} & 0.91 & 0.05 & 1.19 & 0.07 & 0.74 & 0.04 & 0.81 & 4.72 & 1.00 & 0.95 & 1.00 & 0.98 & 0.45 & 0.24 & 0.39 & 0.22 & 0.40 & 0.22 \\ &
\textbf{Ensemble (LGB)} & 4.49 & 0.07 & 5.14 & 0.09 & 4.54 & 0.06 & 4.04 & 6.25 & 0.97 & 0.92 & 1.00 & 0.98 & 0.56 & 0.31 & 0.48 & 0.27 & 0.48 & 0.27 \\ &
\textbf{Ensemble (Mix)} & 0.87 & 0.05 & 1.15 & 0.07 & 0.68 & 0.04 & 0.76 & 4.68 & 1.00 & 0.95 & 1.00 & 0.98 & 0.45 & 0.24 & 0.39 & 0.21 & 0.40 & 0.22 \\ 
    \cmidrule{1-20}
    \multirow{4}{*}{RTX2080} & \textbf{AutoGluon} & \textbf{9.13} & \textbf{0.17} & \textbf{12.76} & \textbf{0.23} & \textbf{6.70} & \textbf{0.12} & \textbf{1.36} & \textbf{3.04} & \textbf{1.00} & \textbf{1.00} & \textbf{1.00} & \textbf{1.00} & \textbf{0.11} & 0.35 & \textbf{0.07} & 0.31 & \textbf{0.07} & 0.31 \\ &
\textbf{Ensemble (XGB)} & 11.91 & 0.19 & 16.15 & 0.27 & 8.86 & 0.14 & 1.83 & 3.35 & 0.99 & 1.00 & 1.00 & 1.00 & 0.16 & 0.32 & 0.12 & 0.28 & 0.12 & 0.29 \\ &
\textbf{Ensemble (LGB)} & 34.99 & 0.60 & 41.31 & 0.70 & 34.08 & 0.57 & 5.73 & 9.74 & 0.97 & 0.97 & 1.00 & 1.00 & 0.37 & \textbf{0.09} & 0.33 & \textbf{0.07} & 0.33 & \textbf{0.07} \\ &
\textbf{Ensemble (Mix)} & 11.58 & 0.20 & 15.82 & 0.27 & 8.78 & 0.15 & 1.76 & 3.43 & 0.99 & 1.00 & 1.00 & 1.00 & 0.16 & 0.32 & 0.12 & 0.28 & 0.13 & 0.28 \\ 
    \cmidrule{1-20}
    \multirow{4}{*}{RTX3080} & \textbf{AutoGluon} & \textbf{2.90} & \textbf{0.82} & \textbf{4.62} & \textbf{1.06} & \textbf{1.70} & \textbf{0.59} & \textbf{0.54} & 16.97 & \textbf{1.00} & \textbf{0.93} & \textbf{1.00} & \textbf{0.97} & \textbf{0.14} & \textbf{0.20} & \textbf{0.11} & \textbf{0.18} & \textbf{0.11} & \textbf{0.19} \\ &
\textbf{Ensemble (XGB)} & 5.82 & 0.83 & 8.10 & 1.08 & 4.52 & 0.63 & 1.28 & 17.18 & 1.00 & 0.93 & 1.00 & 0.96 & 0.26 & 0.24 & 0.21 & 0.21 & 0.21 & 0.21 \\ &
\textbf{Ensemble (LGB)} & 34.62 & 1.00 & 38.99 & 1.28 & 36.60 & 0.87 & 7.87 & 22.33 & 0.97 & 0.90 & 1.00 & 0.96 & 0.55 & 0.32 & 0.47 & 0.27 & 0.48 & 0.27 \\ &
\textbf{Ensemble (Mix)} & 5.95 & 0.83 & 8.17 & 1.07 & 4.63 & 0.63 & 1.30 & \textbf{16.84} & 1.00 & 0.93 & 1.00 & 0.96 & 0.27 & 0.24 & 0.22 & 0.21 & 0.22 & 0.21 \\
    \cmidrule{1-20}
    \multirow{4}{*}{A6000} & \textbf{AutoGluon} & \textbf{0.70} & \textbf{0.11} & \textbf{1.06} & \textbf{0.18} & \textbf{0.46} & \textbf{0.07} & \textbf{0.21} & \textbf{3.97} & \textbf{1.00} & \textbf{0.99} & \textbf{1.00} & \textbf{0.99} & \textbf{0.32} & 0.20 & \textbf{0.28} & 0.18 & \textbf{0.28} & 0.18 \\ &
\textbf{Ensemble (XGB)} & 3.66 & 0.12 & 4.93 & 0.19 & 2.82 & 0.07 & 1.16 & 4.25 & 1.00 & 0.99 & 1.00 & 0.99 & 0.52 & 0.18 & 0.45 & 0.16 & 0.46 & 0.16 \\ &
\textbf{Ensemble (LGB)} & 20.92 & 0.27 & 23.81 & 0.34 & 22.15 & 0.23 & 7.19 & 10.84 & 0.97 & 0.96 & 1.00 & 0.99 & 0.57 & 0.21 & 0.49 & 0.19 & 0.49 & 0.19 \\ &
\textbf{Ensemble (Mix)} & 3.52 & 0.14 & 4.86 & 0.20 & 2.57 & 0.10 & 1.08 & 5.25 & 1.00 & 0.99 & 1.00 & 0.99 & 0.52 & \textbf{0.10} & 0.45 & \textbf{0.09} & 0.46 & \textbf{0.09} \\
    \cmidrule{1-20}
    \multirow{4}{*}{V100} & \textbf{AutoGluon} & \textbf{9.43} & \textbf{0.16} & \textbf{11.64} & \textbf{0.35} & \textbf{7.92} & \textbf{0.05} & \textbf{2.60} & \textbf{4.46} & \textbf{0.99} & \textbf{0.99} & \textbf{1.00} & \textbf{0.99} & \textbf{0.37} & \textbf{0.08} & \textbf{0.33} & \textbf{0.07} & \textbf{0.33} & \textbf{0.07} \\ &
\textbf{Ensemble (XGB)} & 10.24 & 0.24 & 12.67 & 0.45 & 8.31 & 0.10 & 2.71 & 7.80 & 0.99 & 0.98 & 1.00 & 0.99 & 0.42 & 0.20 & 0.37 & 0.15 & 0.37 & 0.15 \\ &
\textbf{Ensemble (LGB)} & 26.38 & 0.54 & 30.14 & 0.67 & 25.11 & 0.42 & 6.63 & 25.46 & 0.97 & 0.96 & 1.00 & 0.99 & 0.54 & 0.40 & 0.46 & 0.34 & 0.47 & 0.35 \\ &
\textbf{Ensemble (Mix)} & 10.46 & 0.32 & 12.69 & 0.46 & 9.43 & 0.23 & 2.81 & 15.49 & 0.99 & 0.98 & 1.00 & 0.99 & 0.42 & 0.31 & 0.37 & 0.25 & 0.37 & 0.25 \\
    \cmidrule{1-20}
    \multirow{4}{*}{P100} & \textbf{AutoGluon} & \textbf{6.87} & \textbf{0.66} & \textbf{10.69} & \textbf{0.96} & \textbf{4.52} & \textbf{0.42} & \textbf{0.28} & \textbf{2.38} & \textbf{1.00} & \textbf{1.00} & \textbf{1.00} & \textbf{1.00} & \textbf{0.32} & 0.06 & \textbf{0.28} & 0.04 & \textbf{0.28} & 0.04 \\ &
\textbf{Ensemble (XGB)} & 33.86 & 0.73 & 48.02 & 1.06 & 24.22 & 0.46 & 1.39 & 2.63 & 1.00 & 1.00 & 1.00 & 1.00 & 0.52 & \textbf{0.03} & 0.45 & \textbf{0.03} & 0.46 & \textbf{0.03} \\ &
\textbf{Ensemble (LGB)} & 213.74 & 2.55 & 241.02 & 3.03 & 224.82 & 2.33 & 10.04 & 11.36 & 0.97 & 0.97 & 1.00 & 1.00 & 0.57 & 0.43 & 0.49 & 0.38 & 0.49 & 0.38 \\ &
\textbf{Ensemble (Mix)} & 38.65 & 0.89 & 52.59 & 1.21 & 28.81 & 0.69 & 1.60 & 3.61 & 1.00 & 0.99 & 1.00 & 1.00 & 0.53 & 0.08 & 0.46 & 0.06 & 0.47 & 0.07 \\  
    \bottomrule
    \end{tabular}
    }
    \vspace{-3ex}
    \label{tab:surrogatemetrics_combined_gpu_m}
    \end{table}
\end{landscape}


\begin{landscape}
    \begin{table}[ht]\centering
    \caption{Various performance metrics of surrogates evaluated to predict latencies and energies on GPT-M for different CPUs. The arrow on the side of the metric name indicates if lower or higher is better.}
    \resizebox{\linewidth}{!}{%
    \begin{tabular}{llclclclclclclclclclclcl} 
    \toprule
    \multirow{3}{*}{\textbf{CPU}} & \multirow{3}{*}{\textbf{Surrogate}} & \multicolumn{12}{c}{\textbf{Accuracy}}                                                                                                                                                                                                                                                                                                   & \multicolumn{6}{c}{\textbf{Calibration}}                                                                                                                      \\ 
    \cmidrule(lr){3-14}\cmidrule(lr){15-20}
    & & \multicolumn{2}{c}{\textbf{MAE} $\downarrow$} & \multicolumn{2}{c}{\textbf{RMSE} $\downarrow$} & \multicolumn{2}{c}{\textbf{MDAE} $\downarrow$} & \multicolumn{2}{c}{\textbf{MARPD} $\downarrow$} & \multicolumn{2}{c}{\textbf{R\textsuperscript{2}} $\uparrow$} & \multicolumn{2}{c}{\textbf{Corr.} $\uparrow$} & \multicolumn{2}{c}{\textbf{RMS Cal.} $\downarrow$} & \multicolumn{2}{c}{\textbf{MA Cal.} $\downarrow$} & \multicolumn{2}{c}{\textbf{Miscal. Area} $\downarrow$} \\ 
    \midrule
    \multicolumn{1}{c}{} & & \textbf{latencies} & \multicolumn{1}{c}{\textbf{energies}} & \textbf{latencies} & \multicolumn{1}{c}{\textbf{energies}} & \textbf{latencies} & \multicolumn{1}{c}{\textbf{energies}} & \textbf{latencies} & \multicolumn{1}{c}{\textbf{energies}} & \textbf{latencies} & \multicolumn{1}{c}{\textbf{energies}} & \textbf{latencies} & \multicolumn{1}{c}{\textbf{energies}} & \textbf{latencies} & \multicolumn{1}{c}{\textbf{energies}} & \textbf{latencies} & \multicolumn{1}{c}{\textbf{energies}} & \textbf{latencies} & \multicolumn{1}{c}{\textbf{energies}} \\ 
    \cmidrule(lr){3-4} \cmidrule(lr){5-6} \cmidrule(lr){7-8} \cmidrule(lr){9-10} \cmidrule(lr){11-12} \cmidrule(lr){13-14} \cmidrule(lr){15-16} \cmidrule(lr){17-18} \cmidrule(lr){19-20}
    \multirow{4}{*}{Xeon Silver} & \textbf{AutoGluon} & \textbf{1713.09} & \textbf{13.02} & \textbf{2355.46} & \textbf{16.40} & \textbf{1261.81} & 10.12 & \textbf{3.50} & \textbf{11.85} & \textbf{0.99} & \textbf{0.94} & \textbf{1.00} & \textbf{0.97} & \textbf{0.08} & 0.54 & \textbf{0.07} & 0.46 & \textbf{0.07} & 0.47 \\ &
\textbf{Ensemble (XGB)} & 1867.00 & 13.19 & 2544.82 & 16.74 & 1379.22 & 10.25 & 3.78 & 12.04 & 0.99 & 0.93 & 1.00 & 0.97 & 0.14 & \textbf{0.54} & 0.12 & \textbf{0.46} & 0.12 & \textbf{0.47} \\ &
\textbf{Ensemble (LGB)} & 5077.16 & 15.70 & 5981.26 & 20.00 & 4830.55 & 13.47 & 10.78 & 15.19 & 0.97 & 0.91 & 1.00 & 0.97 & 0.44 & 0.54 & 0.38 & 0.46 & 0.39 & 0.47 \\ &
\textbf{Ensemble (Mix)} & 1913.78 & 13.05 & 2597.53 & 16.48 & 1464.40 & \textbf{9.92} & 3.87 & 11.93 & 0.99 & 0.94 & 1.00 & 0.97 & 0.14 & 0.54 & 0.12 & 0.46 & 0.12 & 0.47 \\ 
    \cmidrule{1-20}
    \multirow{4}{*}{Xeon Gold} & \textbf{AutoGluon} & \textbf{997.52} & \textbf{1.78} & \textbf{1342.44} & \textbf{2.36} & \textbf{749.29} & \textbf{1.37} & \textbf{4.66} & \textbf{5.51} & \textbf{0.99} & \textbf{0.98} & \textbf{0.99} & \textbf{0.99} & 0.21 & \textbf{0.02} & 0.19 & \textbf{0.02} & 0.19 & \textbf{0.02} \\ &
\textbf{Ensemble (XGB)} & 1059.66 & 1.88 & 1399.18 & 2.46 & 835.65 & 1.49 & 5.00 & 5.82 & 0.99 & 0.98 & 0.99 & 0.99 & 0.19 & 0.02 & 0.17 & 0.02 & 0.17 & 0.02 \\ &
\textbf{Ensemble (LGB)} & 2038.55 & 3.23 & 2439.66 & 3.88 & 1886.11 & 2.99 & 9.75 & 9.79 & 0.96 & 0.96 & 0.99 & 0.99 & \textbf{0.08} & 0.22 & \textbf{0.07} & 0.20 & \textbf{0.07} & 0.20 \\ &
\textbf{Ensemble (Mix)} & 1042.62 & 1.88 & 1395.96 & 2.47 & 795.68 & 1.47 & 4.81 & 5.74 & 0.99 & 0.98 & 0.99 & 0.99 & 0.20 & 0.02 & 0.18 & 0.02 & 0.18 & 0.02 \\ 
    \cmidrule{1-20}
    \multirow{4}{*}{AMD 7502} & \textbf{AutoGluon} & \textbf{2793.38} & \textbf{22.27} & \textbf{3832.68} & \textbf{29.64} & \textbf{2039.52} & \textbf{16.81} & \textbf{4.75} & \textbf{4.81} & \textbf{0.99} & \textbf{0.99} & \textbf{0.99} & \textbf{0.99} & \textbf{0.36} & \textbf{0.45} & \textbf{0.32} & \textbf{0.40} & \textbf{0.32} & \textbf{0.40} \\ &
\textbf{Ensemble (XGB)} & 2889.71 & 22.93 & 3927.64 & 30.61 & 2083.31 & 17.06 & 4.95 & 4.98 & 0.99 & 0.99 & 0.99 & 0.99 & 0.41 & 0.47 & 0.36 & 0.41 & 0.36 & 0.41 \\ &
\textbf{Ensemble (LGB)} & 5664.76 & 44.12 & 6951.41 & 54.57 & 5092.86 & 39.69 & 11.17 & 11.02 & 0.96 & 0.96 & 0.99 & 0.99 & 0.51 & 0.51 & 0.44 & 0.44 & 0.44 & 0.45 \\ &
\textbf{Ensemble (Mix)} & 2888.22 & 22.81 & 3930.69 & 30.58 & 2101.56 & 16.93 & 4.95 & 4.94 & 0.99 & 0.99 & 0.99 & 0.99 & 0.41 & 0.46 & 0.36 & 0.40 & 0.36 & 0.40 \\ 
    \cmidrule{1-20}
    \multirow{4}{*}{AMD 7513} & \textbf{AutoGluon} & \textbf{3297.57} & \textbf{315.32} & \textbf{6512.33} & \textbf{708.51} & \textbf{1709.93} & \textbf{174.70} & \textbf{1.92} & \textbf{14.36} & \textbf{1.00} & \textbf{0.74} & \textbf{1.00} & \textbf{0.86} & \textbf{0.46} & 0.55 & \textbf{0.40} & 0.48 & \textbf{0.40} & 0.48 \\ &
\textbf{Ensemble (XGB)} & 4020.05 & 344.26 & 7071.85 & 724.33 & 2639.66 & 200.98 & 2.55 & 15.96 & 1.00 & 0.72 & 1.00 & 0.85 & 0.49 & 0.55 & 0.43 & 0.48 & 0.43 & 0.48 \\ &
\textbf{Ensemble (LGB)} & 15744.34 & 367.84 & 18417.64 & 744.75 & 15733.28 & 256.65 & 11.25 & 19.37 & 0.97 & 0.71 & 1.00 & 0.85 & 0.57 & 0.56 & 0.49 & 0.48 & 0.49 & 0.49 \\ &
\textbf{Ensemble (Mix)} & 4390.21 & 329.13 & 7277.61 & 717.62 & 3043.23 & 183.58 & 2.72 & 14.82 & 0.99 & 0.73 & 1.00 & 0.85 & 0.50 & \textbf{0.55} & 0.44 & \textbf{0.48} & 0.44 & \textbf{0.48} \\
    \cmidrule{1-20}
    \multirow{4}{*}{AMD 7452} & \textbf{AutoGluon} & \textbf{6422.40} & \textbf{19.16} & \textbf{8400.06} & \textbf{25.91} & \textbf{4603.69} & \textbf{13.53} & \textbf{8.76} & \textbf{8.80} & \textbf{0.96} & \textbf{0.96} & \textbf{0.98} & \textbf{0.98} & \textbf{0.40} & \textbf{0.43} & \textbf{0.34} & \textbf{0.37} & \textbf{0.35} & \textbf{0.38} \\ &
\textbf{Ensemble (XGB)} & 6583.99 & 19.77 & 8650.39 & 26.64 & 4800.54 & 14.12 & 9.01 & 9.17 & 0.96 & 0.96 & 0.98 & 0.98 & 0.46 & 0.47 & 0.40 & 0.41 & 0.41 & 0.42 \\ &
\textbf{Ensemble (LGB)} & 8414.41 & 25.28 & 11082.78 & 33.29 & 6746.91 & 20.86 & 13.18 & 13.17 & 0.93 & 0.93 & 0.98 & 0.98 & 0.49 & 0.51 & 0.43 & 0.44 & 0.43 & 0.44 \\ &
\textbf{Ensemble (Mix)} & 6535.60 & 19.66 & 8556.66 & 26.48 & 5013.65 & 14.41 & 8.93 & 9.06 & 0.96 & 0.96 & 0.98 & 0.98 & 0.46 & 0.47 & 0.40 & 0.41 & 0.40 & 0.41 \\
    
    \bottomrule
    \end{tabular}
    }
    \vspace{-3ex}
    \label{tab:surrogatemetrics_combined_cpu_m}
    \end{table}
\end{landscape}
\begin{landscape}
    \begin{table}[ht]\centering
    \caption{Various performance metrics of surrogates evaluated to predict latencies and energies on GPT-L for different GPUs. The arrow on the side of the metric name indicates if lower or higher is better.}
    \resizebox{\linewidth}{!}{%
    \begin{tabular}{llclclclclclclclclclclcl} 
    \toprule
    \multirow{3}{*}{\textbf{GPU}} & \multirow{3}{*}{\textbf{Surrogate}} & \multicolumn{12}{c}{\textbf{Accuracy}}                                                                                                                                                                                                                                                                                                   & \multicolumn{6}{c}{\textbf{Calibration}}                                                                                                                      \\ 
    \cmidrule(lr){3-14}\cmidrule(lr){15-20}
    & & \multicolumn{2}{c}{\textbf{MAE} $\downarrow$} & \multicolumn{2}{c}{\textbf{RMSE} $\downarrow$} & \multicolumn{2}{c}{\textbf{MDAE} $\downarrow$} & \multicolumn{2}{c}{\textbf{MARPD} $\downarrow$} & \multicolumn{2}{c}{\textbf{R\textsuperscript{2}} $\uparrow$} & \multicolumn{2}{c}{\textbf{Corr.} $\uparrow$} & \multicolumn{2}{c}{\textbf{RMS Cal.} $\downarrow$} & \multicolumn{2}{c}{\textbf{MA Cal.} $\downarrow$} & \multicolumn{2}{c}{\textbf{Miscal. Area} $\downarrow$} \\ 
    \midrule
    \multicolumn{1}{c}{} & & \textbf{latencies} & \multicolumn{1}{c}{\textbf{energies}} & \textbf{latencies} & \multicolumn{1}{c}{\textbf{energies}} & \textbf{latencies} & \multicolumn{1}{c}{\textbf{energies}} & \textbf{latencies} & \multicolumn{1}{c}{\textbf{energies}} & \textbf{latencies} & \multicolumn{1}{c}{\textbf{energies}} & \textbf{latencies} & \multicolumn{1}{c}{\textbf{energies}} & \textbf{latencies} & \multicolumn{1}{c}{\textbf{energies}} & \textbf{latencies} & \multicolumn{1}{c}{\textbf{energies}} & \textbf{latencies} & \multicolumn{1}{c}{\textbf{energies}} \\ 
    \cmidrule(lr){3-4} \cmidrule(lr){5-6} \cmidrule(lr){7-8} \cmidrule(lr){9-10} \cmidrule(lr){11-12} \cmidrule(lr){13-14} \cmidrule(lr){15-16} \cmidrule(lr){17-18} \cmidrule(lr){19-20}
    \multirow{4}{*}{A100} & \textbf{AutoGluon} & \textbf{1.43} & \textbf{0.34} & \textbf{1.97} & \textbf{0.56} & \textbf{1.14} & \textbf{0.16} & \textbf{0.70} & \textbf{6.63} & \textbf{1.00} & \textbf{0.98} & \textbf{1.00} & \textbf{0.99} & 0.31 & 0.39 & 0.24 & 0.34 & 0.24 & 0.34 \\ &
\textbf{Ensemble (XGB)} & 2.07 & 0.36 & 2.85 & 0.58 & 1.55 & 0.18 & 1.00 & 7.55 & 1.00 & 0.97 & 1.00 & 0.99 & 0.30 & 0.35 & 0.22 & 0.31 & 0.23 & 0.32 \\ &
\textbf{Ensemble (LGB)} & 9.28 & 0.67 & 10.63 & 0.84 & 9.24 & 0.54 & 4.75 & 18.27 & 0.97 & 0.95 & 1.00 & 0.99 & 0.46 & \textbf{0.13} & 0.41 & \textbf{0.10} & 0.41 & \textbf{0.10} \\ &
\textbf{Ensemble (Mix)} & 2.16 & 0.38 & 2.94 & 0.58 & 1.68 & 0.21 & 1.04 & 8.26 & 1.00 & 0.97 & 1.00 & 0.99 & \textbf{0.29} & 0.34 & \textbf{0.22} & 0.30 & \textbf{0.22} & 0.31 \\ 
    \cmidrule{1-20}
    \multirow{4}{*}{A40} & \textbf{AutoGluon} & \textbf{0.61} & \textbf{0.18} & \textbf{0.90} & \textbf{0.23} & \textbf{0.41} & \textbf{0.15} & \textbf{0.27} & \textbf{2.43} & \textbf{1.00} & \textbf{1.00} & \textbf{1.00} & \textbf{1.00} & \textbf{0.44} & 0.38 & \textbf{0.39} & 0.34 & \textbf{0.39} & 0.35 \\ &
\textbf{Ensemble (XGB)} & 2.44 & 0.23 & 3.37 & 0.28 & 1.81 & 0.19 & 1.11 & 2.88 & 1.00 & 1.00 & 1.00 & 1.00 & 0.54 & 0.36 & 0.47 & 0.32 & 0.47 & 0.32 \\ &
\textbf{Ensemble (LGB)} & 15.23 & 0.83 & 17.17 & 0.95 & 16.50 & 0.87 & 7.69 & 10.53 & 0.97 & 0.97 & 1.00 & 1.00 & 0.57 & \textbf{0.06} & 0.49 & \textbf{0.05} & 0.49 & \textbf{0.05} \\ &
\textbf{Ensemble (Mix)} & 2.41 & 0.22 & 3.34 & 0.28 & 1.74 & 0.19 & 1.07 & 2.79 & 1.00 & 1.00 & 1.00 & 1.00 & 0.54 & 0.36 & 0.47 & 0.32 & 0.47 & 0.33 \\ 
    \cmidrule{1-20}
    \multirow{4}{*}{H100} & \textbf{AutoGluon} & \textbf{0.15} & \textbf{0.13} & \textbf{0.21} & \textbf{0.17} & \textbf{0.11} & 0.09 & \textbf{0.15} & \textbf{5.37} & \textbf{1.00} & \textbf{0.96} & \textbf{1.00} & \textbf{0.98} & \textbf{0.22} & 0.16 & \textbf{0.20} & 0.14 & \textbf{0.20} & 0.14 \\ &
\textbf{Ensemble (XGB)} & 0.62 & 0.13 & 0.83 & 0.18 & 0.47 & 0.09 & 0.63 & 5.50 & 1.00 & 0.96 & 1.00 & 0.98 & 0.48 & 0.16 & 0.42 & 0.14 & 0.42 & 0.14 \\ &
\textbf{Ensemble (LGB)} & 3.61 & 0.18 & 4.11 & 0.23 & 3.79 & 0.15 & 3.84 & 8.09 & 0.97 & 0.93 & 1.00 & 0.98 & 0.56 & 0.27 & 0.48 & 0.25 & 0.49 & 0.25 \\ &
\textbf{Ensemble (Mix)} & 0.57 & 0.13 & 0.78 & 0.17 & 0.41 & \textbf{0.08} & 0.57 & 5.41 & 1.00 & 0.96 & 1.00 & 0.98 & 0.47 & \textbf{0.16} & 0.41 & \textbf{0.14} & 0.41 & \textbf{0.14} \\ 
    \cmidrule{1-20}
    \multirow{4}{*}{RTX2080} & \textbf{AutoGluon} & \textbf{2.40} & \textbf{1.80} & \textbf{3.43} & \textbf{2.58} & \textbf{1.53} & 1.12 & \textbf{0.59} & \textbf{10.68} & \textbf{1.00} & \textbf{0.90} & \textbf{1.00} & \textbf{0.95} & 0.16 & \textbf{0.24} & 0.14 & \textbf{0.22} & 0.14 & \textbf{0.22} \\ &
\textbf{Ensemble (XGB)} & 5.21 & 1.83 & 6.98 & 2.63 & 4.06 & 1.15 & 1.30 & 10.91 & 1.00 & 0.90 & 1.00 & 0.95 & \textbf{0.16} & 0.29 & \textbf{0.13} & 0.25 & \textbf{0.13} & 0.25 \\ &
\textbf{Ensemble (LGB)} & 28.00 & 2.09 & 31.84 & 2.92 & 30.68 & 1.42 & 7.61 & 13.14 & 0.97 & 0.88 & 1.00 & 0.95 & 0.35 & 0.35 & 0.32 & 0.30 & 0.32 & 0.31 \\ &
\textbf{Ensemble (Mix)} & 4.79 & 1.83 & 6.60 & 2.62 & 3.46 & \textbf{1.09} & 1.14 & 10.92 & 1.00 & 0.90 & 1.00 & 0.95 & 0.16 & 0.30 & 0.13 & 0.26 & 0.13 & 0.27 \\ 
    \cmidrule{1-20}
    \multirow{4}{*}{RTX3080} & \textbf{AutoGluon} & \textbf{1.49} & \textbf{0.68} & \textbf{2.13} & \textbf{0.77} & \textbf{0.99} & 0.66 & \textbf{0.50} & \textbf{6.50} & \textbf{1.00} & \textbf{0.99} & \textbf{1.00} & \textbf{1.00} & \textbf{0.13} & 0.07 & \textbf{0.12} & 0.06 & \textbf{0.12} & 0.06 \\ &
\textbf{Ensemble (XGB)} & 3.95 & 0.70 & 5.40 & 0.81 & 2.89 & 0.65 & 1.36 & 6.59 & 1.00 & 0.99 & 1.00 & 1.00 & 0.21 & 0.07 & 0.15 & 0.06 & 0.16 & 0.06 \\ &
\textbf{Ensemble (LGB)} & 22.68 & 1.46 & 25.65 & 1.71 & 24.92 & 1.42 & 8.86 & 12.08 & 0.97 & 0.96 & 1.00 & 1.00 & 0.52 & 0.12 & 0.45 & 0.11 & 0.46 & 0.11 \\ &
\textbf{Ensemble (Mix)} & 3.86 & 0.69 & 5.32 & 0.81 & 2.78 & \textbf{0.65} & 1.32 & 6.56 & 1.00 & 0.99 & 1.00 & 1.00 & 0.23 & \textbf{0.07} & 0.17 & \textbf{0.06} & 0.17 & \textbf{0.06} \\
    \cmidrule{1-20}
    \multirow{4}{*}{A6000} & \textbf{AutoGluon} & \textbf{1.56} & \textbf{0.45} & \textbf{2.24} & \textbf{0.61} & \textbf{1.08} & \textbf{0.33} & \textbf{0.74} & \textbf{7.13} & \textbf{1.00} & \textbf{0.98} & \textbf{1.00} & \textbf{0.99} & \textbf{0.48} & 0.20 & \textbf{0.42} & 0.17 & \textbf{0.42} & 0.18 \\ &
\textbf{Ensemble (XGB)} & 2.22 & 0.47 & 3.06 & 0.63 & 1.57 & 0.34 & 1.11 & 7.21 & 1.00 & 0.98 & 1.00 & 0.99 & 0.51 & 0.19 & 0.44 & 0.17 & 0.45 & 0.17 \\ &
\textbf{Ensemble (LGB)} & 13.40 & 0.84 & 15.18 & 1.04 & 14.58 & 0.67 & 7.47 & 12.16 & 0.97 & 0.95 & 1.00 & 0.99 & 0.57 & \textbf{0.03} & 0.49 & \textbf{0.02} & 0.49 & \textbf{0.02} \\ &
\textbf{Ensemble (Mix)} & 2.08 & 0.46 & 2.95 & 0.63 & 1.45 & 0.34 & 0.99 & 7.36 & 1.00 & 0.98 & 1.00 & 0.99 & 0.50 & 0.17 & 0.43 & 0.15 & 0.44 & 0.15 \\
    \cmidrule{1-20}
    \multirow{4}{*}{V100} & \textbf{AutoGluon} & 11.61 & \textbf{0.30} & \textbf{13.29} & \textbf{2.99} & 11.30 & \textbf{0.15} & 4.23 & \textbf{3.74} & \textbf{0.98} & \textbf{0.86} & \textbf{0.99} & \textbf{0.93} & \textbf{0.52} & 0.21 & \textbf{0.45} & 0.19 & \textbf{0.45} & 0.19 \\ &
\textbf{Ensemble (XGB)} & \textbf{11.50} & 0.44 & 13.66 & 3.02 & \textbf{10.46} & 0.25 & \textbf{4.16} & 6.43 & 0.98 & 0.86 & 0.99 & 0.93 & 0.54 & \textbf{0.05} & 0.47 & \textbf{0.05} & 0.47 & \textbf{0.05} \\ &
\textbf{Ensemble (LGB)} & 17.77 & 1.32 & 22.53 & 3.31 & 13.23 & 1.34 & 6.55 & 25.45 & 0.96 & 0.83 & 0.99 & 0.93 & 0.54 & 0.34 & 0.46 & 0.30 & 0.47 & 0.30 \\ &
\textbf{Ensemble (Mix)} & 11.84 & 0.43 & 13.63 & 3.03 & 10.69 & 0.24 & 4.35 & 6.30 & 0.98 & 0.86 & 0.99 & 0.93 & 0.55 & 0.07 & 0.47 & 0.06 & 0.48 & 0.06 \\
    \cmidrule{1-20}
    \multirow{4}{*}{P100} & \textbf{AutoGluon} & \textbf{5.59} & \textbf{0.78} & \textbf{7.81} & \textbf{1.27} & \textbf{4.28} & \textbf{0.40} & \textbf{0.52} & \textbf{1.31} & \textbf{1.00} & \textbf{1.00} & \textbf{1.00} & \textbf{1.00} & \textbf{0.29} & 0.13 & \textbf{0.26} & 0.11 & \textbf{0.26} & 0.12 \\ &
\textbf{Ensemble (XGB)} & 21.84 & 1.17 & 29.68 & 1.77 & 16.52 & 0.73 & 1.61 & 2.18 & 1.00 & 1.00 & 1.00 & 1.00 & 0.49 & \textbf{0.06} & 0.43 & \textbf{0.06} & 0.43 & \textbf{0.06} \\ &
\textbf{Ensemble (LGB)} & 131.33 & 5.74 & 146.45 & 6.60 & 139.03 & 5.69 & 10.74 & 13.53 & 0.97 & 0.97 & 1.00 & 1.00 & 0.57 & 0.50 & 0.49 & 0.44 & 0.49 & 0.44 \\ &
\textbf{Ensemble (Mix)} & 23.97 & 1.64 & 31.74 & 2.20 & 18.94 & 1.30 & 1.76 & 3.66 & 1.00 & 1.00 & 1.00 & 1.00 & 0.49 & 0.20 & 0.42 & 0.18 & 0.43 & 0.18 \\  
    \bottomrule
    \end{tabular}
    }
    \vspace{-3ex}
    \label{tab:surrogatemetrics_combined_gpu_l}
    \end{table}
\end{landscape}


\begin{landscape}
    \begin{table}[ht]\centering
    \caption{Various performance metrics of surrogates evaluated to predict latencies and energies on GPT-L for different CPUs. The arrow on the side of the metric name indicates if lower or higher is better.}
    \resizebox{\linewidth}{!}{%
    \begin{tabular}{llclclclclclclclclclclcl} 
    \toprule
    \multirow{3}{*}{\textbf{CPU}} & \multirow{3}{*}{\textbf{Surrogate}} & \multicolumn{12}{c}{\textbf{Accuracy}}                                                                                                                                                                                                                                                                                                   & \multicolumn{6}{c}{\textbf{Calibration}}                                                                                                                      \\ 
    \cmidrule(lr){3-14}\cmidrule(lr){15-20}
    & & \multicolumn{2}{c}{\textbf{MAE} $\downarrow$} & \multicolumn{2}{c}{\textbf{RMSE} $\downarrow$} & \multicolumn{2}{c}{\textbf{MDAE} $\downarrow$} & \multicolumn{2}{c}{\textbf{MARPD} $\downarrow$} & \multicolumn{2}{c}{\textbf{R\textsuperscript{2}} $\uparrow$} & \multicolumn{2}{c}{\textbf{Corr.} $\uparrow$} & \multicolumn{2}{c}{\textbf{RMS Cal.} $\downarrow$} & \multicolumn{2}{c}{\textbf{MA Cal.} $\downarrow$} & \multicolumn{2}{c}{\textbf{Miscal. Area} $\downarrow$} \\ 
    \midrule
    \multicolumn{1}{c}{} & & \textbf{latencies} & \multicolumn{1}{c}{\textbf{energies}} & \textbf{latencies} & \multicolumn{1}{c}{\textbf{energies}} & \textbf{latencies} & \multicolumn{1}{c}{\textbf{energies}} & \textbf{latencies} & \multicolumn{1}{c}{\textbf{energies}} & \textbf{latencies} & \multicolumn{1}{c}{\textbf{energies}} & \textbf{latencies} & \multicolumn{1}{c}{\textbf{energies}} & \textbf{latencies} & \multicolumn{1}{c}{\textbf{energies}} & \textbf{latencies} & \multicolumn{1}{c}{\textbf{energies}} & \textbf{latencies} & \multicolumn{1}{c}{\textbf{energies}} \\ 
    \cmidrule(lr){3-4} \cmidrule(lr){5-6} \cmidrule(lr){7-8} \cmidrule(lr){9-10} \cmidrule(lr){11-12} \cmidrule(lr){13-14} \cmidrule(lr){15-16} \cmidrule(lr){17-18} \cmidrule(lr){19-20}
    \multirow{4}{*}{Xeon Silver} & \textbf{AutoGluon} & \textbf{636.15} & 5.61 & \textbf{864.06} & \textbf{7.70} & \textbf{494.13} & 3.93 & \textbf{2.54} & \textbf{9.84} & \textbf{1.00} & \textbf{0.96} & \textbf{1.00} & \textbf{0.98} & 0.17 & 0.44 & 0.15 & 0.38 & 0.15 & 0.39 \\ &
\textbf{Ensemble (XGB)} & 753.47 & 5.72 & 1025.05 & 7.81 & 578.99 & 4.01 & 2.96 & 10.02 & 1.00 & 0.96 & 1.00 & 0.98 & 0.08 & \textbf{0.42} & 0.07 & \textbf{0.36} & 0.07 & \textbf{0.37} \\ &
\textbf{Ensemble (LGB)} & 2824.18 & 8.02 & 3236.41 & 9.97 & 2733.85 & 6.99 & 11.64 & 14.87 & 0.97 & 0.93 & 1.00 & 0.98 & 0.41 & 0.48 & 0.37 & 0.41 & 0.37 & 0.42 \\ &
\textbf{Ensemble (Mix)} & 794.59 & \textbf{5.60} & 1061.38 & 7.73 & 619.52 & \textbf{3.82} & 3.12 & 10.00 & 1.00 & 0.96 & 1.00 & 0.98 & \textbf{0.06} & 0.42 & \textbf{0.05} & 0.36 & \textbf{0.06} & 0.37 \\ 
    \cmidrule{1-20}
    \multirow{4}{*}{Xeon Gold} & \textbf{AutoGluon} & \textbf{1808.83} & \textbf{2.94} & \textbf{4993.72} & \textbf{9.06} & \textbf{961.39} & \textbf{1.64} & \textbf{12.44} & \textbf{12.52} & \textbf{0.65} & \textbf{0.60} & \textbf{0.81} & \textbf{0.77} & \textbf{0.04} & \textbf{0.22} & \textbf{0.03} & \textbf{0.20} & \textbf{0.03} & \textbf{0.20} \\ &
\textbf{Ensemble (XGB)} & 2071.91 & 3.33 & 5172.30 & 9.21 & 1192.99 & 2.01 & 14.28 & 14.39 & 0.62 & 0.58 & 0.79 & 0.76 & 0.23 & 0.36 & 0.21 & 0.32 & 0.21 & 0.32 \\ &
\textbf{Ensemble (LGB)} & 2276.83 & 3.72 & 5208.21 & 9.42 & 1546.31 & 2.52 & 16.82 & 17.10 & 0.62 & 0.56 & 0.80 & 0.76 & 0.32 & 0.42 & 0.28 & 0.37 & 0.28 & 0.37 \\ &
\textbf{Ensemble (Mix)} & 1956.75 & 3.15 & 5056.95 & 9.17 & 1058.07 & 1.74 & 13.15 & 13.29 & 0.64 & 0.59 & 0.80 & 0.77 & 0.21 & 0.35 & 0.18 & 0.31 & 0.18 & 0.31 \\ 
    \cmidrule{1-20}
    \multirow{4}{*}{AMD 7502} & \textbf{AutoGluon} & \textbf{979.59} & \textbf{10.85} & \textbf{1401.41} & \textbf{15.13} & \textbf{637.77} & \textbf{7.47} & \textbf{3.47} & \textbf{4.78} & \textbf{0.99} & \textbf{0.99} & \textbf{1.00} & \textbf{0.99} & \textbf{0.32} & 0.46 & \textbf{0.29} & 0.40 & \textbf{0.29} & 0.41 \\ &
\textbf{Ensemble (XGB)} & 1061.60 & 11.30 & 1488.69 & 15.60 & 747.04 & 8.16 & 3.90 & 5.16 & 0.99 & 0.99 & 1.00 & 0.99 & 0.36 & 0.47 & 0.32 & 0.41 & 0.32 & 0.41 \\ &
\textbf{Ensemble (LGB)} & 2963.13 & 24.72 & 3487.50 & 29.94 & 2788.52 & 22.54 & 12.75 & 13.26 & 0.96 & 0.96 & 1.00 & 0.99 & 0.53 & 0.54 & 0.46 & 0.47 & 0.46 & 0.47 \\ &
\textbf{Ensemble (Mix)} & 1077.05 & 11.22 & 1504.92 & 15.53 & 777.37 & 8.19 & 3.88 & 4.97 & 0.99 & 0.99 & 1.00 & 0.99 & 0.34 & \textbf{0.44} & 0.30 & \textbf{0.38} & 0.31 & \textbf{0.39} \\ 
    \cmidrule{1-20}
    \multirow{4}{*}{AMD 7513} & \textbf{AutoGluon} & \textbf{1341.23} & \textbf{138.82} & \textbf{2336.42} & \textbf{193.98} & \textbf{747.38} & \textbf{92.12} & \textbf{1.55} & \textbf{13.84} & \textbf{1.00} & \textbf{0.91} & \textbf{1.00} & \textbf{0.96} & \textbf{0.48} & \textbf{0.55} & \textbf{0.42} & \textbf{0.48} & \textbf{0.42} & \textbf{0.48} \\ &
\textbf{Ensemble (XGB)} & 1821.68 & 141.56 & 2867.87 & 198.08 & 1203.03 & 97.95 & 2.32 & 14.26 & 1.00 & 0.91 & 1.00 & 0.95 & 0.53 & 0.56 & 0.45 & 0.48 & 0.46 & 0.48 \\ &
\textbf{Ensemble (LGB)} & 8438.54 & 158.86 & 9602.56 & 224.77 & 8622.26 & 129.98 & 12.73 & 18.19 & 0.97 & 0.88 & 1.00 & 0.95 & 0.57 & 0.56 & 0.49 & 0.48 & 0.49 & 0.49 \\ &
\textbf{Ensemble (Mix)} & 1990.32 & 141.12 & 2967.07 & 197.03 & 1402.25 & 103.07 & 2.56 & 14.02 & 1.00 & 0.91 & 1.00 & 0.95 & 0.53 & 0.56 & 0.46 & 0.48 & 0.46 & 0.49 \\
    \cmidrule{1-20}
    \multirow{4}{*}{AMD 7452} & \textbf{AutoGluon} & \textbf{766.81} & \textbf{4.16} & \textbf{1101.54} & \textbf{6.15} & \textbf{535.75} & \textbf{2.72} & \textbf{2.52} & \textbf{4.20} & \textbf{1.00} & \textbf{0.99} & \textbf{1.00} & \textbf{1.00} & \textbf{0.05} & \textbf{0.26} & \textbf{0.04} & \textbf{0.23} & \textbf{0.05} & \textbf{0.23} \\ &
\textbf{Ensemble (XGB)} & 895.53 & 4.40 & 1262.25 & 6.44 & 636.53 & 3.02 & 2.98 & 4.55 & 1.00 & 0.99 & 1.00 & 1.00 & 0.14 & 0.29 & 0.12 & 0.26 & 0.13 & 0.26 \\ &
\textbf{Ensemble (LGB)} & 3265.62 & 10.37 & 3747.71 & 12.55 & 3294.40 & 9.69 & 12.28 & 12.92 & 0.97 & 0.96 & 1.00 & 0.99 & 0.50 & 0.50 & 0.43 & 0.43 & 0.44 & 0.44 \\ &
\textbf{Ensemble (Mix)} & 920.19 & 4.45 & 1277.03 & 6.47 & 681.07 & 2.98 & 3.07 & 4.57 & 1.00 & 0.99 & 1.00 & 0.99 & 0.14 & 0.29 & 0.13 & 0.26 & 0.13 & 0.26 \\
    \bottomrule
    \end{tabular}
    }
    \vspace{-3ex}
    \label{tab:surrogatemetrics_combined_cpu_l}
    \end{table}
\end{landscape}
\clearpage
\newpage

\clearpage


\clearpage
\section{Comparison with NAS-Bench-301}
Given the immense size of the search spaces (approximately $10^{36}$), training architectures from scratch is impractical. Our work is inspired by surrogate benchmarks such as those proposed by \citep{zela-iclr22a}. However, HW-GPT-Bench presents several key differences compared to NB301. First, unlike NB301, which trains architectures from scratch—impractical for larger models and dataset sizes—we utilize an efficient weight-sharing-based supernet. The performance of the inherited subnetwork directly serves as a reliable performance estimation proxy, and these architectures can be fine-tuned if necessary. We also employ a sandwich scheme to train the supernet by sampling the largest, smallest, and a set of random architectures.
Second, while NB301 uses the DARTS pre-training pipeline, we introduce three novel search spaces and design a training pipeline specifically for supernet training. Additionally, we focus on the application domain of language modeling, in contrast to NB301’s primary focus on image classification.
Lastly, unlike NB301, our work supports a range of hardware devices and provides well-calibrated latency predictions. To our knowledge, we are the first to study the calibration of surrogate models for latency predictions.
\section{Importance Analysis}
\label{sec:importance_app}
In this section, we provide details on the methods we used throughout the paper to analyze and interpret the data collected from the HW-GPT-Bench search space.

\subsection{OLS Covariate Analysis.} 
Ordinary Least Squares (OLS) is a statistical method used for estimating the parameters of a linear regression model. The primary goal of OLS is to minimize the sum of the squared residuals, which are the differences between the observed values and the values predicted by the model.
In the context of OLS, covariate analysis involves examining the relationships between independent variables (covariates) and the dependent variable. This analysis helps to understand how each covariate contributes to the prediction of the dependent variable and the overall model performance. To fit the linear regression model and conduct the analysis we used $\mathtt{statsmodel.regression.linear\_model.OLS}$~\footnote{\url{https://www.statsmodels.org/dev/examples/notebooks/generated/ols.html}}.

\subsubsection{Key Concepts in OLS}

\textbf{Linear Regression Model}: The model assumes a linear relationship between the dependent variable \( Y \) (perplexity) and one or more independent variables (covariates). The general form of the model is:
    \[
    Y = \beta_0 + \beta_1 e + \beta_2 l + \sum_{i=1}^{l} \beta_{i+2} h^i + \sum_{i=1}^{l} \beta_{i+2+l} m^i + \beta_{2l+3} b + \epsilon ,
    \]
    where \( \beta_0 \) is the intercept, \( \beta_i \) (for \( i = 1, \ldots, 2l+3 \)) are the coefficients of the covariates, and \( \epsilon \) represents the error term. $l$ is the number of layers, $e$ is the embedding dimension, $m^i$ and $h^i$ are the MLP ratio and number of heads on layer $i$, respectively, and b is the bias.

\textbf{Objective of OLS}: The goal is to estimate the coefficients \( \beta \) such that we minimize the sum of the squared residuals (SSR):
    \[
    SSR = \sum_{i=1}^{N} (Y_i - \hat{Y}_i)^2,
    \]
    where \( Y_i \) is the observed ground truth perplexity and \( \hat{Y}_i \) is the predicted perplexity by the Linear Regression model.

\subsubsection{Covariate Analysis}

Covariate analysis in the context of OLS involves investigating the effect of each independent variable (embedding dimension, number of layers, etc.) on the dependent variable, i.e. perplexity. This includes:

\begin{itemize}[leftmargin=*]
    \item \textbf{Estimating Coefficients}: Determining the values of \( \beta_1, \beta_2, \ldots, \beta_{2l+3} \) which represent the change in perplexity for a one-unit change in the respective architectural dimension, holding other variables constant.

    \item \textbf{Statistical Significance}: Assessing the significance of each coefficient using t-tests. The null hypothesis \( H_0 \) states that the coefficient is zero (no effect). The p-value indicates whether the null hypothesis can be rejected.

    \item \textbf{Standard Errors}: Providing a measure of the variability of the coefficient estimates. Smaller standard errors suggest more precise estimates.

    \item \textbf{Goodness-of-Fit}: Evaluating how well the model explains the variability of perplexity using metrics such as R-squared and adjusted R-squared. R-squared indicates the proportion of the variance in perplexity that is predictable from the independent variables.
\end{itemize}

The results of the OLS analysis for perplexity on the collected samples are presented below. These results include:
\begin{enumerate}
    \item \textbf{Coefficients (\( \beta \))}: Estimates for each covariate.
    \item \textbf{Standard Errors}: Indicate the precision of the coefficient estimates.
    \item \textbf{t-Values}: Used to test the hypothesis that a coefficient is significantly different from zero.
    \item \textbf{P-Values}: Indicate the significance level of each coefficient.
    \item \textbf{R}$^2$: The proportion of the variance in the dependent variable explained by the model.
    \item \textbf{Adjusted R-squared}: Adjusted for the number of predictors in the model, providing a more accurate measure of goodness-of-fit for models with multiple covariates.
\end{enumerate}

\begin{Verbatim}[commandchars=\\\{\}]
==============================================================================
                        OLS Regression Results GPT-L
==============================================================================
Dep. Variable:             perplexity   R-squared:                       0.901
Model:                            OLS   Adj. R-squared:                  0.901
Method:                 Least Squares   F-statistic:                 1.817e+04
No. Observations:               10000   AIC:                         3.395e+04
Df Residuals:                    9994   BIC:                         3.399e+04
Df Model:                           5
Covariance Type:            nonrobust
==============================================================================
                 coef    std err          t      P>|t|      [0.025      0.975]
------------------------------------------------------------------------------
const         23.6263      0.013   1788.886      0.000      23.600      23.652
num_layers    -0.0426      0.013     -3.227      0.001      -0.069      -0.017
embed_dim     -3.9787      0.013   -301.209      0.000      -4.005      -3.953
mean_mlp_ratio -0.1164     0.013     -8.812      0.000      -0.142      -0.091
mean_heads    -0.0612      0.013     -4.634      0.000      -0.087      -0.035
bias          -0.0630      0.013     -4.769      0.000      -0.089      -0.037
==============================================================================
Omnibus:                    75511.319   Durbin-Watson:                   2.026
Prob(Omnibus):                  0.000   Jarque-Bera (JB):             1537.046
Skew:                          -0.635   Prob(JB):                         0.00
Kurtosis:                       1.559   Cond. No.                         1.03
==============================================================================
\end{Verbatim}

\begin{verbatim}
==============================================================================
                        OLS Regression Results GPT-M
==============================================================================
Dep. Variable:             perplexity   R-squared:                       0.914
Model:                            OLS   Adj. R-squared:                  0.914
Method:                 Least Squares   F-statistic:                 2.117e+04
No. Observations:               10000   AIC:                         3.791e+04
Df Residuals:                    9994   BIC:                         3.796e+04
Df Model:                           5
Covariance Type:            nonrobust
==============================================================================
                 coef    std err          t      P>|t|      [0.025      0.975]
------------------------------------------------------------------------------
const         28.1305      0.016   1747.009      0.000      28.099      28.162
num_layers    -0.0983      0.016     -6.104      0.000      -0.130      -0.067
embed_dim     -5.2315      0.016   -324.795      0.000      -5.263      -5.200
mean_mlp_ratio -0.1445     0.016     -8.975      0.000      -0.176      -0.113
mean_heads    -0.0744      0.016     -4.623      0.000      -0.106      -0.043
bias          -0.0650      0.016     -4.035      0.000      -0.097      -0.033
==============================================================================
Omnibus:                    66013.096   Durbin-Watson:                   2.021
Prob(Omnibus):                  0.000   Jarque-Bera (JB):             1509.031
Skew:                          -0.600   Prob(JB):                         0.00
Kurtosis:                       1.524   Cond. No.                         1.03
==============================================================================
\end{verbatim}
\begin{Verbatim}[commandchars=\\\{\}]
==============================================================================
                       OLS Regression Results GPT-S
==============================================================================
Dep. Variable:             perplexity   R-squared:                       0.949
Model:                            OLS   Adj. R-squared:                  0.949
Method:                 Least Squares   F-statistic:                 3.754e+04
No. Observations:               10000   AIC:                         3.667e+04
Df Residuals:                    9994   BIC:                         3.671e+04
Df Model:                           5
Covariance Type:            nonrobust
==============================================================================
                 coef    std err          t      P>|t|      [0.025      0.975]
------------------------------------------------------------------------------
const          32.3973    0.015   2141.130      0.000      32.368      32.427
num_layers     -0.4912    0.015    -32.459      0.000      -0.521      -0.462
embed_dim      -6.5321    0.015   -431.656      0.000      -6.562      -6.502
mean_mlp_ratio -0.2199    0.015    -14.530      0.000      -0.250      -0.190
mean_heads     -0.2739    0.015    -18.096      0.000      -0.304      -0.244
bias           -0.0251    0.015     -1.661      0.097      -0.055       0.005
==============================================================================
Omnibus:                    88675.889   Durbin-Watson:                   2.044
Prob(Omnibus):                  0.000   Jarque-Bera (JB):             1321.623
Skew:                          -0.544   Prob(JB):                    1.03e-287
Kurtosis:                       1.589   Cond. No.                         1.03
==============================================================================
\end{Verbatim}

\subsection{Power-laws for GPT-wide spaces}
We define the power-law fits for GPT-wide spaces below. Similar to the observation in \ref{sec:analysis}, we see that the embedding size has the most effect on perplexity, followed by number of layers, mlp expansion ratio and number of heads. The bias plays a minimal role in determining perplexity of an architecture. 
\begin{align*}
\text{GPT-S-wide:} & \quad y = 1116.453\cdot l^{-0.212} \cdot e^{-0.3770} \cdot m^{-0.0514} \cdot h^{-0.0647} \cdot b^{-0.000190}, \\
\text{GPT-M-wide:} & \quad y = 618.0753 \cdot l^{-0.1795} \cdot e^{-0.3401} \cdot m^{-0.0711} \cdot h^{-0.0556} \cdot b^{-0.0050}, \\
\text{GPT-L-wide:} & \quad y = 498.9920  \cdot l^{-0.1659} \cdot e^{-0.3204} \cdot m^{-0.0692} \cdot h^{-0.053} \cdot b^{-0.0081}.
\end{align*}
\subsection{Recursive Feature Elimination} 
Recursive Feature Elimination (RFE) is a feature selection method used in machine learning to identify the most relevant features in a dataset. It works by recursively fitting a model and removing the least important feature(s) based on model coefficients or importance scores until the maximum number of features is reached. Features in our case are the architectural choices, i.e. embedding dimension size, number of layers, etc. The process involves the following steps:

\begin{enumerate}
    \item \textbf{Model Fitting}: The model is initially trained on the entire set of features.
    \item \textbf{Feature Ranking}: Features are ranked based on their importance scores derived from the fitted model.
    \item \textbf{Feature Elimination}: The least important feature(s) are removed from the dataset.
    \item \textbf{Repetition}: Steps 1-3 are repeated recursively on the pruned feature set until the desired number of features is reached or only a single feature remains.
\end{enumerate}

\begin{figure}[ht]
  
        \centering
        \includegraphics[width=\textwidth]{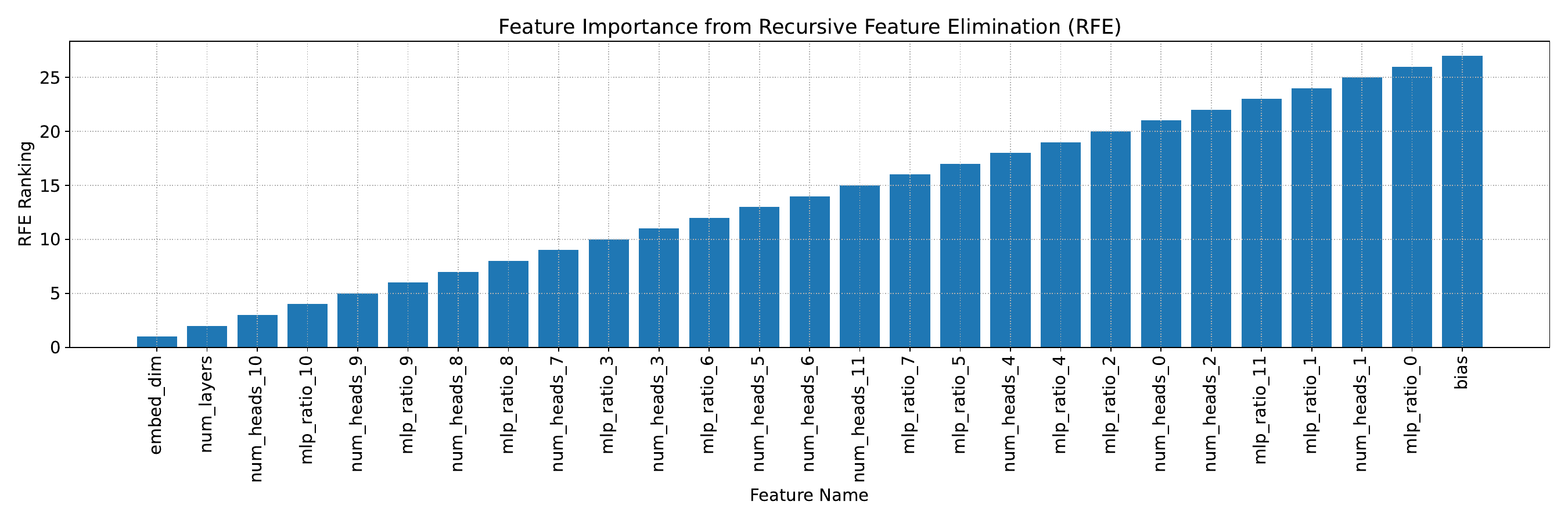}
        
    \caption{Detailed feature ranking from RFE for GPT-S.}
    \label{fig:rfegpts}
\end{figure}

\begin{figure}[ht]
  
        \centering
        \includegraphics[width=\textwidth]{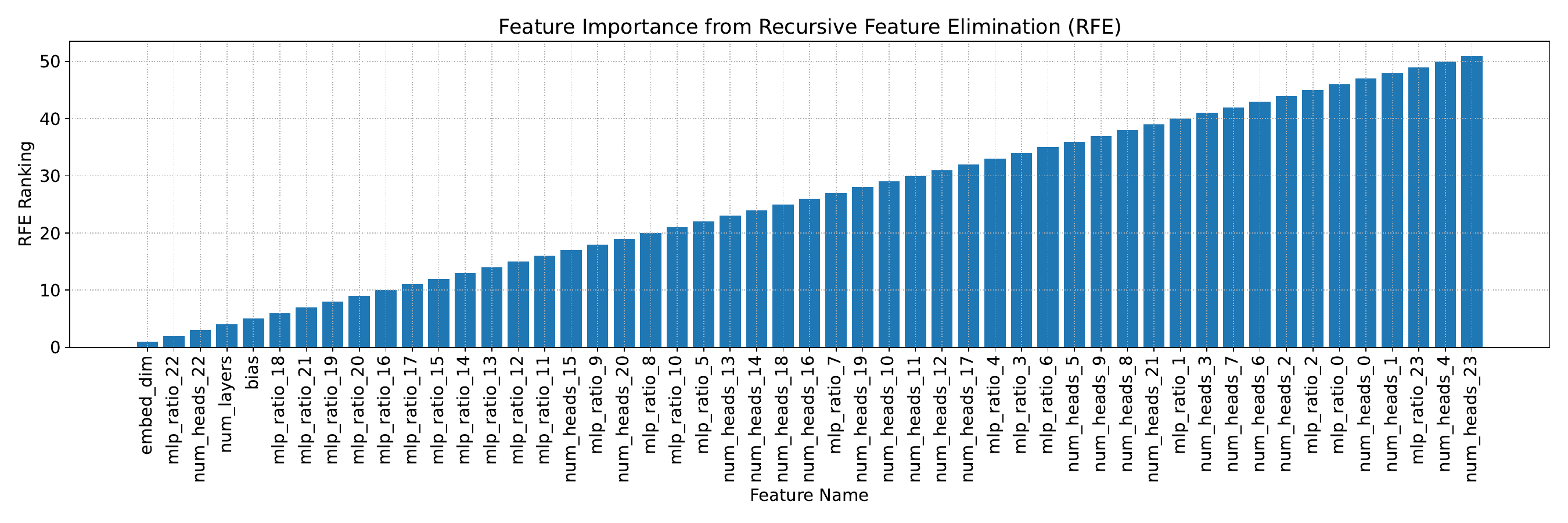}
        
    \caption{Detailed feature ranking from RFE for GPT-M.}
    \label{fig:rfegptm}
\end{figure}

\begin{figure}[ht]
  
        \centering
        \includegraphics[width=\textwidth]{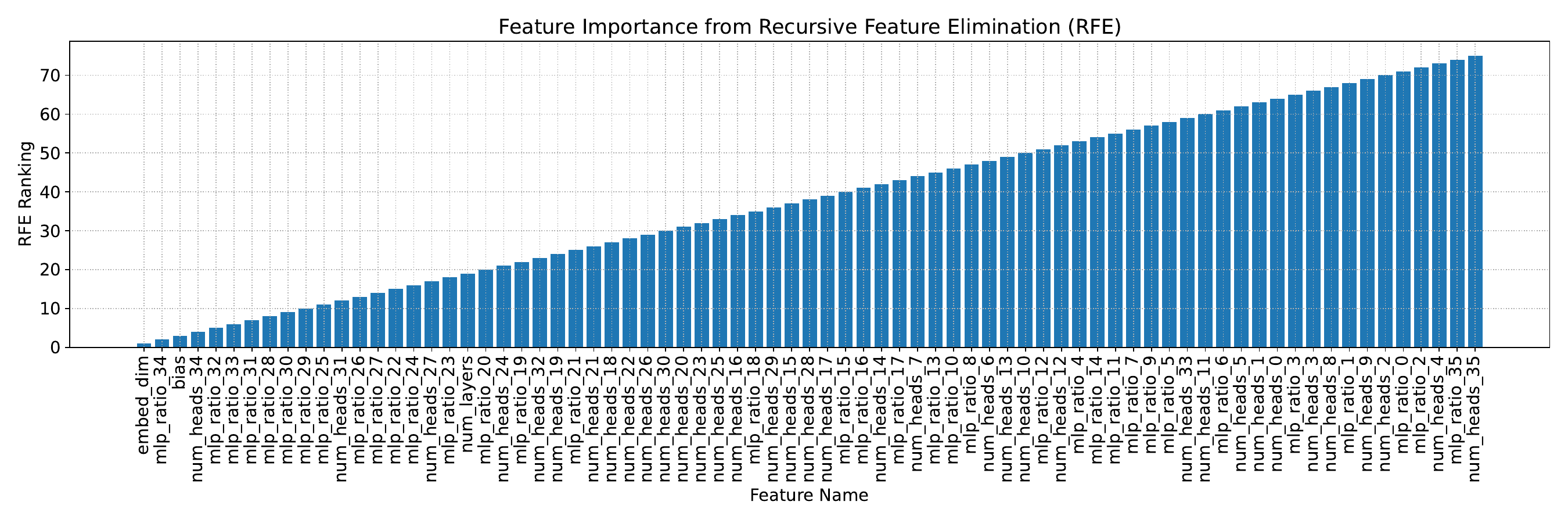}
        
    \caption{Detailed feature ranking from RFE for GPT-L.}
    \label{fig:rfegptl}
\end{figure}

We implement RFE using $\mathtt{sklearn}$~\footnote{\url{https://scikit-learn.org/stable/auto_examples/feature_selection/plot_rfe_digits.html}}, which provides an efficient and easy-to-use RFE function. We employ a \texttt{RandomForest} regressor as the base model due to its ability to handle high-dimensional data and capture non-linear relationships. Specifically, we use a \texttt{RandomForest} regressor with 50 estimators, which balances model complexity and computational efficiency.
In Figures \ref{fig:rfegpts}, \ref{fig:rfegptm} and \ref{fig:rfegptl}, we present the ranking of all search space dimensions for GPT-S, GPT-M and GPT-L spaces. 


\section{Details on Empirical Attainment Function (EAF)}
\label{sec:eaf_app}
Running multi-objective optimization algorithms multiple times can yield different Pareto fronts. Computing uncertainty estimates of Pareto fronts over multiple runs of an algorithm is important to ensure that we appropriately compare different algorithms in a statistically meaningful way.
Simply superimposing the Pareto fronts across multiple runs is insufficient in depicting how the Pareto fronts tend to vary. The Empirical Attainment Function (EAF)~\citep{fonseca1996, eaf} is a statistical measure used in multi-objective optimization to describe the distribution of outcomes achieved by an optimization algorithm over multiple runs. It provides a way to empirically estimate the probability that a given point in the objective space is attained (i.e., dominated or matched) by the solutions generated by the algorithm. 

For two solution vectors of the multi-objective function, $\mathbf{y}$ and $\mathbf{z}$, $\mathbf{y}$ weakly dominates $\mathbf{z}$ ($\mathbf{y} \preceq \mathbf{z}$) if the following conditions hold:
\begin{enumerate}[leftmargin=*]
    \item \textbf{Not worse on any objective:} $\mathbf{y}$ is at least as good as $\mathbf{z}$ on all objectives. This means for each objective function, the value in $\mathbf{y}$ is either equal to or better than the corresponding value in $\mathbf{z}$.
    \item \textbf{Strictly better on at least one objective (or indifferent on all):} $\mathbf{y}$ must be strictly better than $\mathbf{z}$ on at least one objective function. Alternatively, it can be equal on all objectives.
\end{enumerate}
Given a set of Pareto fronts $\{F^1,\dots ,F^n \}$ obtained from $n$ independent runs with different random seeds, the EAF is defined by the following empirical attainment function:
$$
\varepsilon(\mathbf{z}) = \frac{1}{n} \sum_{i=1}^{n} \mathbb{I}[F^i \preceq \mathbf{z}],
$$
where $\varepsilon(\mathbf{z})$ represents the EAF value for a specific objective vector $\mathbf{z}$ in the objective space. $\mathbb{I}[F^i \preceq \mathbf{z}]$ is an indicator function that is 1 if the objective vector $\mathbf{z}$ is weakly dominated by the $F^i$, i.e. the Pareto front from the i-th run, and 0 otherwise. $F^i \preceq \mathbf{z}$ means that there exists at least an objective vector $\mathbf{y}$ in $F^i$ at least as good as $\mathbf{z}$. This doesn't necessarily mean every single run achieved a better outcome than $\mathbf{z}$, but at least one did.
In simpler terms, the EAF value at a given objective vector $\mathbf{z}$ represents the portion of independent runs where a Pareto front achieved at least that good of an objective vector (weakly dominated $\mathbf{z}$).

The EAF can be used to visualize uncertainty in the Pareto front by plotting different EAS levels. For example, $S^{1/2}$  represents the set of objective vectors that are weakly dominated by at least half of the independent runs (50\% EAF level).
To compute the Empirical Attainment Surfaces (EAS) in this paper we used the implementation from \citet{watanabe2023pareto} \footnote{\url{https://github.com/nabenabe0928/empirical-attainment-func}}.
\vspace{-3mm}
\section{Inheriting v/s Finetuning Subnetworks}
\vspace{-3mm}
We validate the effectiveness of out perplexity surrogate by inheriting randomly sampled 100 subnetworks and comparing the correlation between the perplexity on simply inheriting the subnetworks v/s finetuning the subnetworks further upon inheritance for 5000 update steps. We observe that the inherited subnetwork performance strongly correlates with fine-tuned subnetworks as indicated by Table \ref{tab:supernet-kendall-tau}.
\begin{table}[h]
    \centering
    \begin{tabular}{l c}
        \toprule
        Supernet & Kendall-Tau \\
        \midrule
        GPT-S & 0.9626 \\
        GPT-M & 0.9580 \\
        GPT-L & 0.9286 \\
        \bottomrule
    \end{tabular}
    \caption{Kendall-Tau values between perplexities after inheriting and perplexities after finetuning (for 5000 steps) for different Supernets for a set of 100 random architectures}
    \label{tab:supernet-kendall-tau}
\end{table}
\vspace{-3mm}
\section{Distribution of Architecture Latencies}
\begin{figure}[t]
    \centering
    \begin{minipage}{0.45\textwidth}
        \centering
        \includegraphics[width=\linewidth]{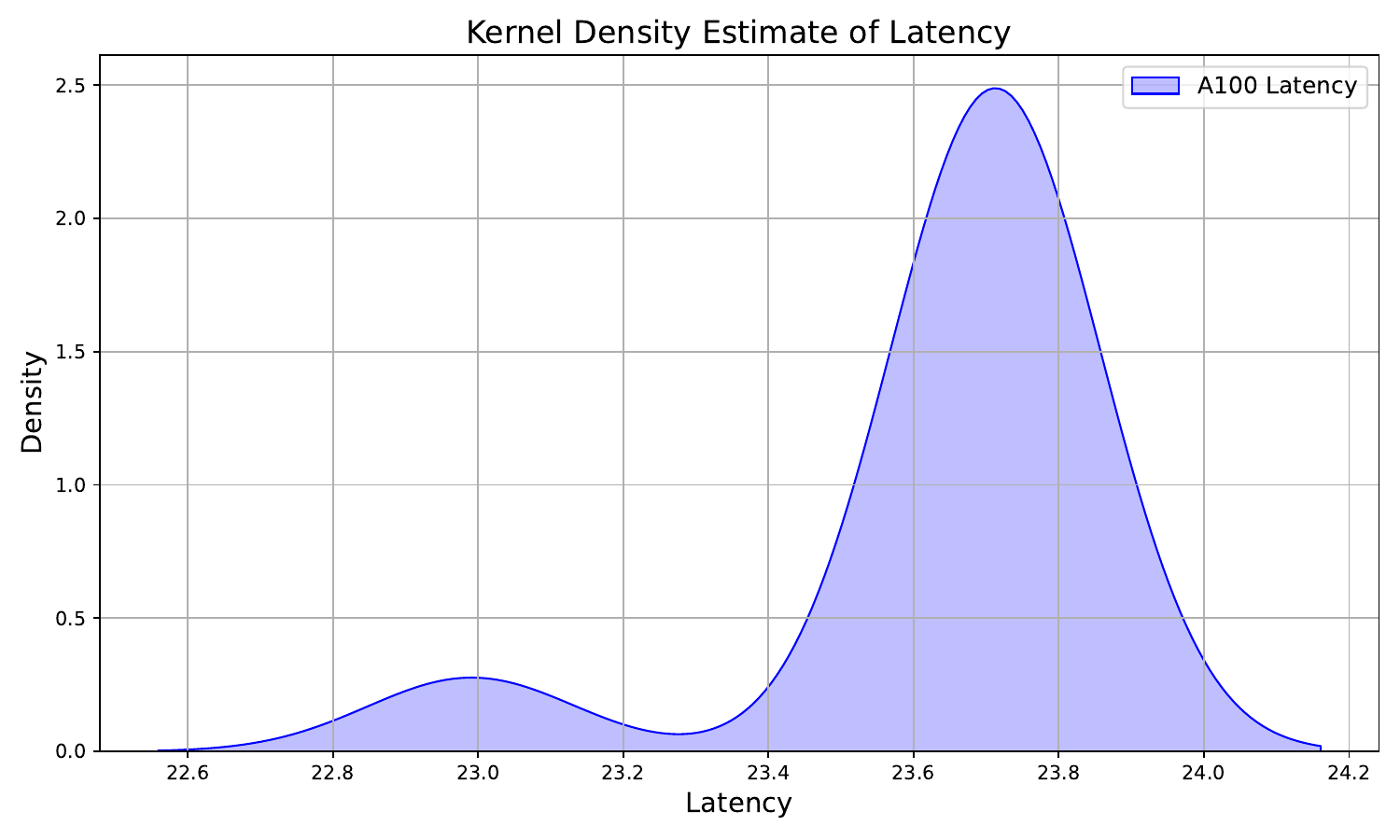}
        \caption{GPU-A100}
        \label{fig:gpulat}
    \end{minipage}\hfill
    \begin{minipage}{0.45\textwidth}
        \centering
        \includegraphics[width=\linewidth]{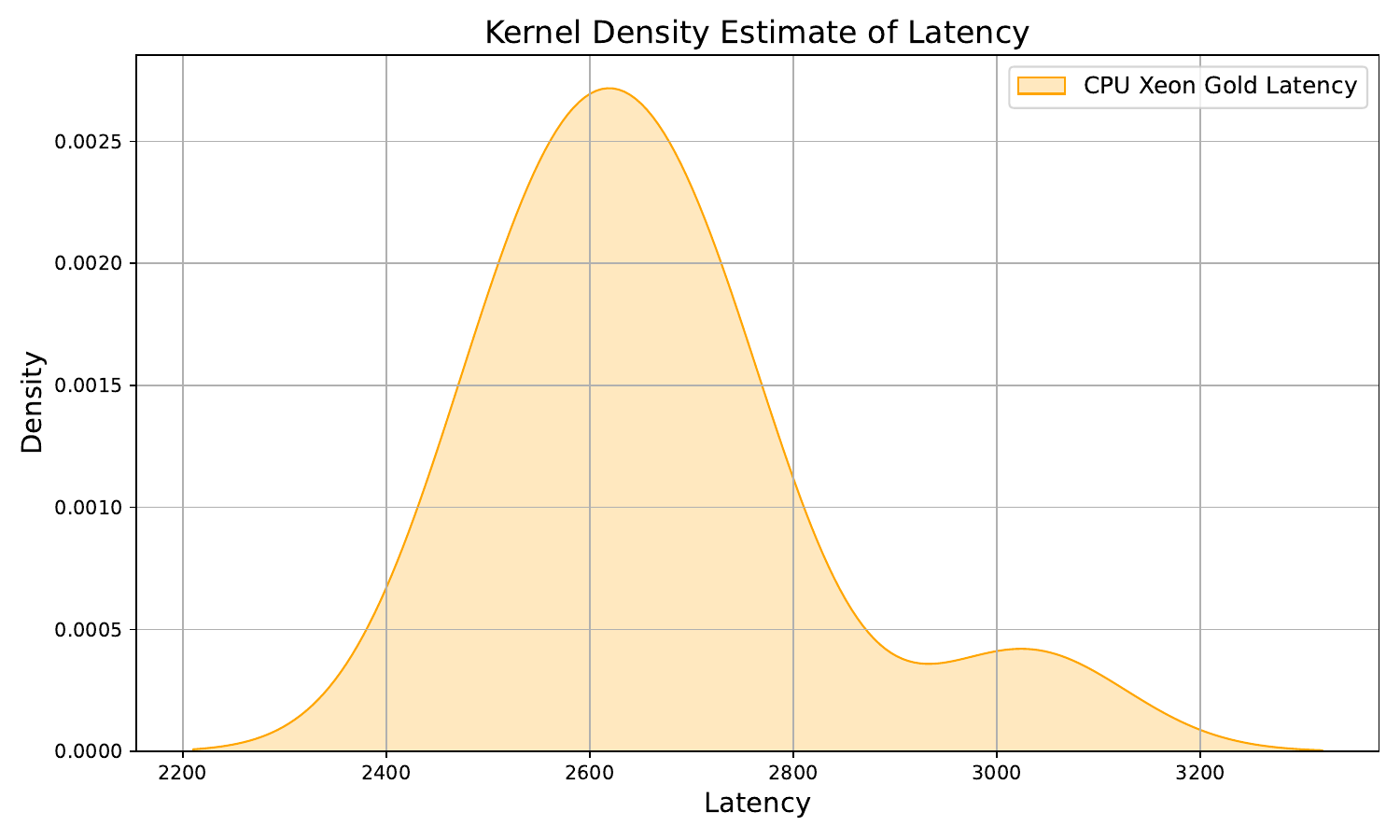}
        \caption{CPU-Xeon Gold}
        \label{fig:cpulat}
    \end{minipage}
    \caption{Distribution of Latencies on different devices}
    \vspace{-6mm}
\end{figure}

\vspace{-3mm}
We fit a kernel-density-estimator to the collected subnetwork latencies. As observed in Figure \ref{fig:gpulat} and \ref{fig:cpulat}, the distribution of CPU latency is more noisier than GPU latency. 
\vspace{-6mm}
\vspace{-2mm}
\section{Scatter Plots for GPT-wide}
\begin{figure}[ht]
\vspace{-4mm}
    \begin{minipage}{\textwidth}
        \centering
        \includegraphics[width=\linewidth]{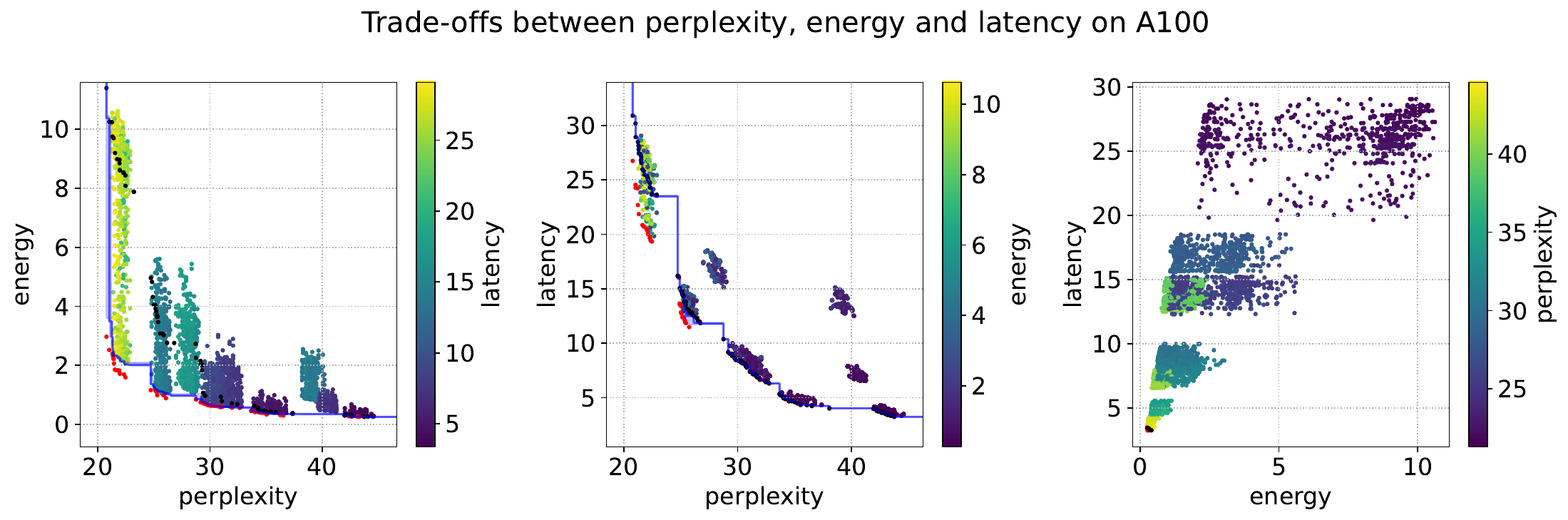}
        \label{fig:plot3}
    \end{minipage}\hfill
        \begin{minipage}{\textwidth}
        \centering
        \includegraphics[width=\linewidth]{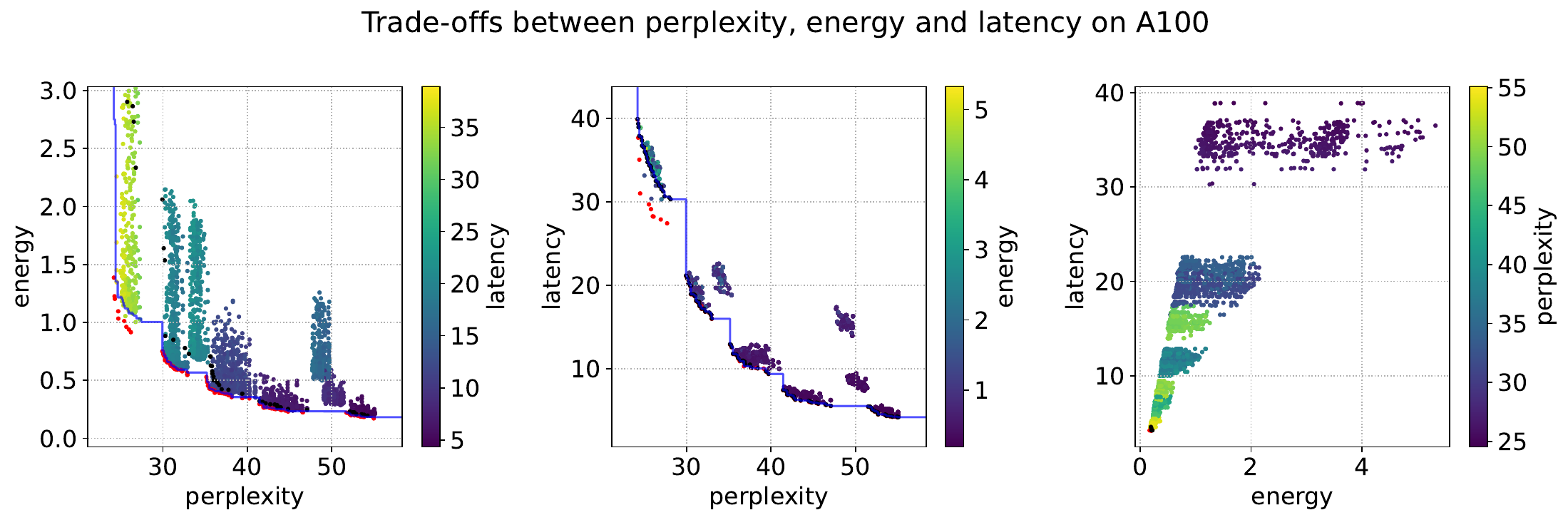}
        
    \end{minipage}
\hfill
        \begin{minipage}{\textwidth}
        \centering
        \includegraphics[width=\linewidth]{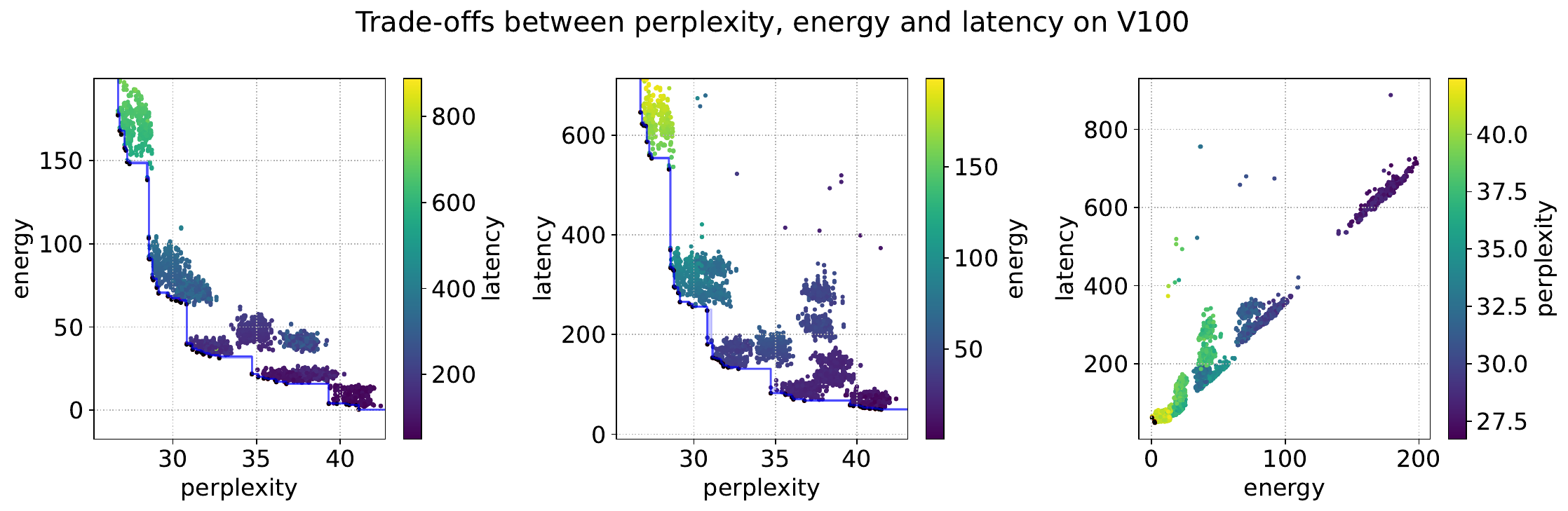}
        
    \end{minipage}
    \caption{Trade-off scatter plots for GPT-M-wide, GPT-L-wide and GPT-XL-wide}
    \label{fig:scatter_plota_gpt_wide}
    \vspace{-3mm}
\end{figure}

\vspace{-3mm}
We introduced 4 new search spaces (with more widely spaced choices), mainly GPT-S-wide, GPT-M-wide, GPT-L-wide and GPT-XL-wide and present the scatter plots on the collected ground truth subnetworks for these spaces in Figure \ref{fig:scatter_plota_gpt_wide}. 

\vspace{-3mm}
\section{Multi-objective NAS algorithms}
\label{sec:algos_full}

In this section, we provide more details on the multi-objective NAS methods we run on HW-GPT-Bench in Section~\ref{sec:experiments}. We use their implementation in SyneTune~\citep{salinas2022syne}\footnote{\url{https://syne-tune.readthedocs.io/en/latest/getting_started.html\#supported-multi-objective-optimization-methods}}.

\begin{itemize}[leftmargin=*]
    \item \textbf{Random Search (RS)}. RS has been shown to be a strong baseline for single \citep{li2020random} and multi-objective \citep{Cai2020Once-for-All:,chen-iccv21b,sukthanker2024multi} architecture search. For this baseline we sample architectures uniformly at random from the search space and then compute the Pareto-Front from these architecture samples. In larger search space, random search, while being embarrasingly parallel and often performant, is not guaranteed to yield the optimal architectures. 
    \item \textbf{Multi-Objective Regularized Evolution Algorithm (MOREA).} Regularized Evolution (RE) or Aging Evolution \citep{real2019regularized} has been quite successfully applied for neural architecture search. Regularized Evolution works by evolving a population of candidates using mutation and periodically discarding the oldest members of the population, inducing a \emph{regularization} effect. In SyneTune RE is extended to Multi-Objective Regularized Evolution (MOREA) by scoring the population via a multi-objective priority based on non-dominated sorting. Parents are then be sampled from the population based on this priority score. 
    \item \textbf{Non-dominated Sorting Genetic Algorithm II (NSGA-II)}. NSGA-II~\citep{deb2002fast}, is a multi-objective evolutionary algorithm to obtain a Pareto-Set of architectures. It employs non-dominated sorting to rank  architecures based on their dominance relationships and \textbf{crowding distance} to maintain a diverse population. Through selection, crossover, and mutation, NSGA-II iteratively evolves populations toward the Pareto front, offering a range of trade-off solutions. 
    \item \textbf{Local Seach (LS).} SyneTune adapts LS to explore the vicinity of Pareto-optimal points in multi-objective optimization problems, aiming to iteratively refine Pareto-optimal solutions solutions within defined neighborhoods. The method is described in more detail in \citet{klein2023structural}.
    \item \textbf{Bayesian Optimization with Random Scalarizations (RSBO). } RSBO~\citep{paria2020flexible} uses an acquisition function that takes as input multiple random scalarizations of the objectives being optimized, to obtain the Pareto-optimal set which minimizes the Bayesian regret. 
    \item \textbf{Bayesian Optimization with Linear Scalarizations (LSBO).} Similar to RSBO, this method optimizes a single objective corresponding to a fixed linear combination of two objectives instead of randomizing the scalarizations at each BO iteration.
    \item \textbf{Expected Hypervolume Improvement (EHVI).} EHVI~\citep{daulton2020differentiable} is a Bayesian optimization method with an acquisition function designed to efficiently explore the Pareto front in multi-objective optimization problems. It quantifies the expected improvement in hypervolume, which measures the volume of the objective space dominated by Pareto-optimal solutions, that a candidate solution can offer. EHVI guides the search towards regions of the objective space likely to contain better trade-offs, aiding in the discovery of diverse Pareto-optimal solutions. 
    \item \textbf{Multi-objective Asynchronous Successive Halving (MOASHA).} MOASHA~\citep{schmucker-arxiv21a} is a multi-fidelity approach that leverages an asynchronous successive halving scheduler~\citep{li-mlsys20a} along with non-dominating sorting for budget allocation. It employs the NSGA-II selection mechanism and the $\epsilon$-net~\citep{Salinas2021AMP} exploration strategy, which ranks candidates within the same Pareto set by iteratively choosing the one with the greatest Euclidean distance from the previously selected candidates.
\end{itemize}
\vspace{-3mm}
\section{Additional experiments with MOO methods}
\vspace{-3mm}
In addition to the results presented in the paper we also run MOO methods on our benchmark for latencies and perplexity across different devices and search spaces in Figures \ref{fig:baselines_latencies_a100}-\ref{fig:baselines_latencies_amd_7513}. We present the EAFs resulting from running the baselines for multiple seeds and the hypervolume of the baselines over the number of surrogate evaluations.  Furthermore, we also present the results of running different MOO methods on the energy-perplexity objectives for different devices on GPT-L in Figures \ref{fig:baselines_energies_a100_a40_h100}-\ref{fig:baselines_energies_amd7502}. Interestingly for these two objectives local search is often very performant at higher budgets, outperforming other baselines like NSGA-2 and EHVI.

 \subsection{Experiments with 2 Objectives}
 \label{sec:2obj_app}
\begin{figure}[ht]
\centering
\begin{minipage}{.16\linewidth}
  \centering
  \includegraphics[width=\linewidth]{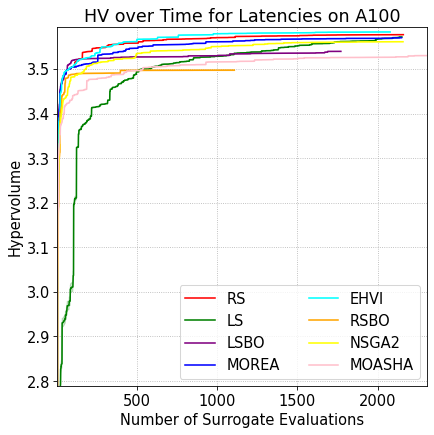}
\end{minipage}%
\begin{minipage}{.16\linewidth}
  \centering
\includegraphics[width=\linewidth]{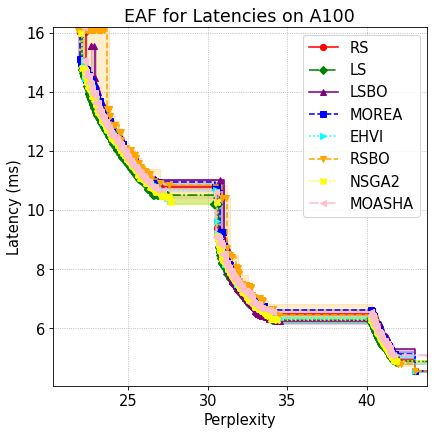}
\end{minipage}
\begin{minipage}{.16\linewidth}
  \centering
  \includegraphics[width=\linewidth]{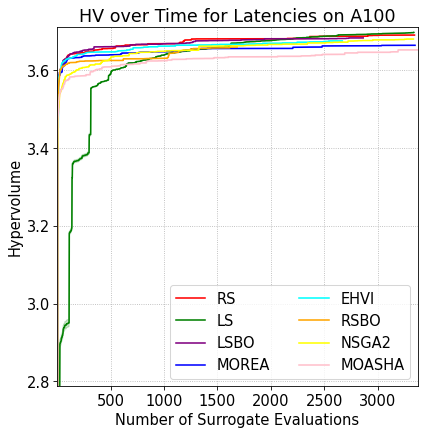}
\end{minipage}%
\begin{minipage}{.16\linewidth}
  \centering
\includegraphics[width=\linewidth]{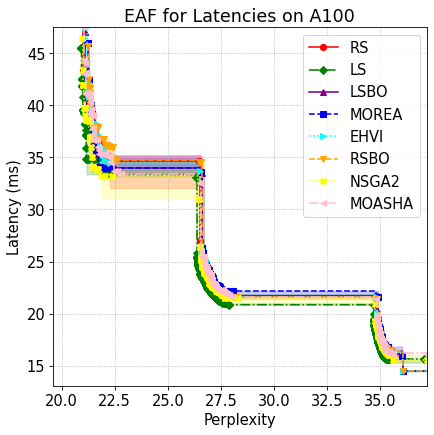}
\end{minipage}
\begin{minipage}{.16\linewidth}
  \centering
  \includegraphics[width=\linewidth]{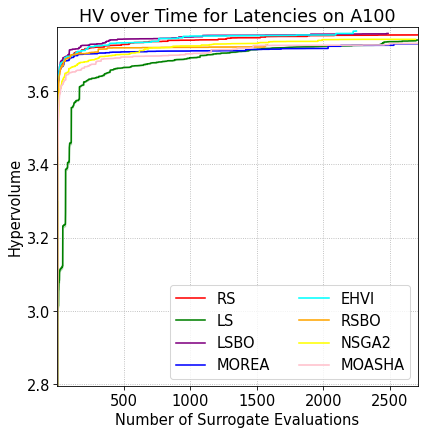}
\end{minipage}%
\begin{minipage}{.16\linewidth}
  \centering
\includegraphics[width=\linewidth]{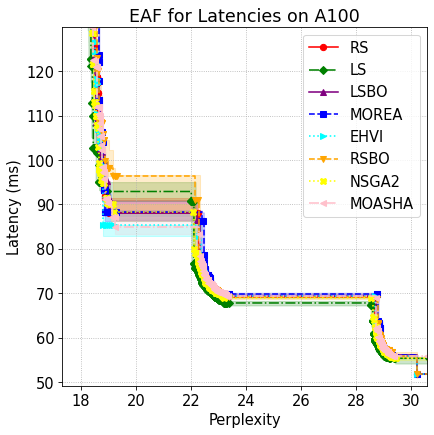}
\end{minipage}
\caption{Pareto fronts and HV over time on A100 for GPT-S (first two), GPT-M (second two) and GPT-L (last two).}
\label{fig:baselines_latencies_a100}
\vspace{-5mm}
\end{figure}

\begin{figure}[ht]
\centering
\begin{minipage}{.16\linewidth}
  \centering
  \includegraphics[width=\linewidth]{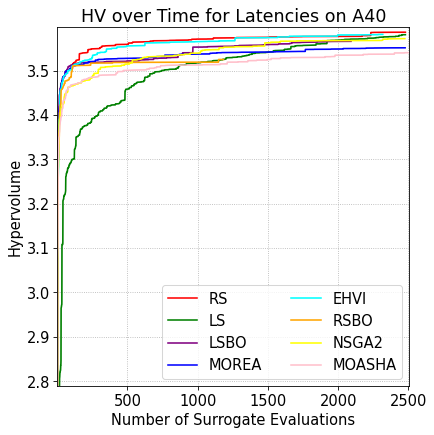}
\end{minipage}%
\begin{minipage}{.16\linewidth}
  \centering
\includegraphics[width=\linewidth]{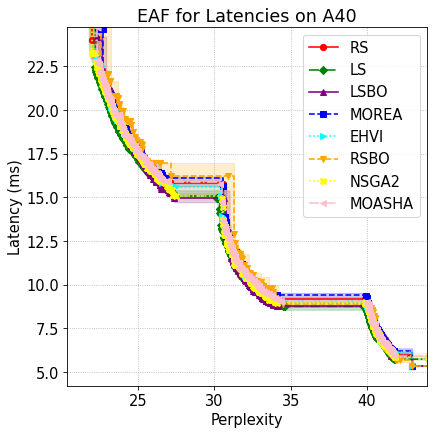}

\end{minipage}
\begin{minipage}{.16\linewidth}
  \centering
  \includegraphics[width=\linewidth]{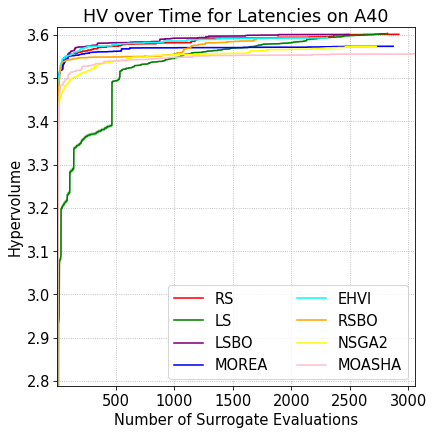}

\end{minipage}%
\begin{minipage}{.16\linewidth}
  \centering
\includegraphics[width=\linewidth]{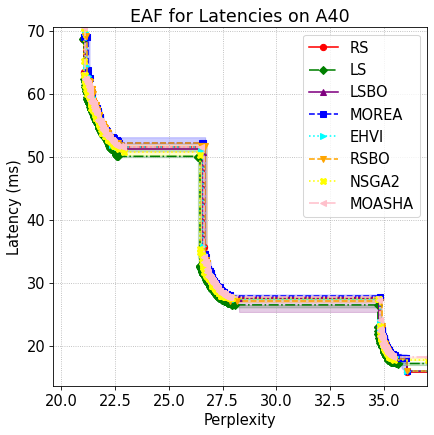}

\end{minipage}
\begin{minipage}{.16\linewidth}
  \centering
  \includegraphics[width=\linewidth]{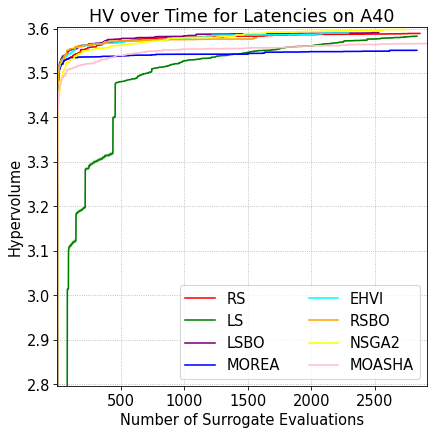}

\end{minipage}%
\begin{minipage}{.16\linewidth}
  \centering
\includegraphics[width=\linewidth]{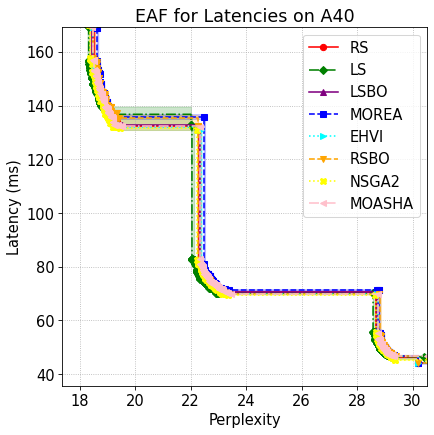}

\end{minipage}
\caption{Pareto fronts and HV over time on A40 for GPT-S (first two), GPT-M (second two) and GPT-L (last two).}
\label{fig:baselines_latencies_a40}
\vspace{-5mm}
\end{figure}

\begin{figure}[ht]
\centering
\begin{minipage}{.16\linewidth}
  \centering
  \includegraphics[width=\linewidth]{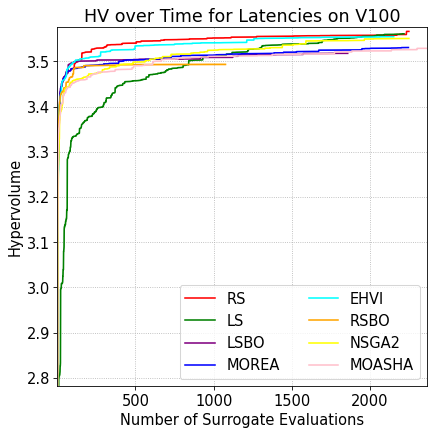}
\end{minipage}%
\begin{minipage}{.16\linewidth}
  \centering
\includegraphics[width=\linewidth]{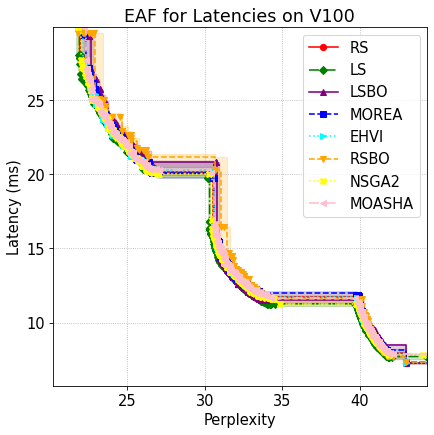}
\end{minipage}
\begin{minipage}{.16\linewidth}
  \centering
  \includegraphics[width=\linewidth]{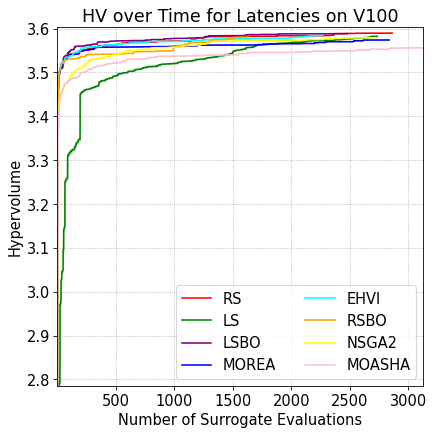}

\end{minipage}%
\begin{minipage}{.16\linewidth}
  \centering
\includegraphics[width=\linewidth]{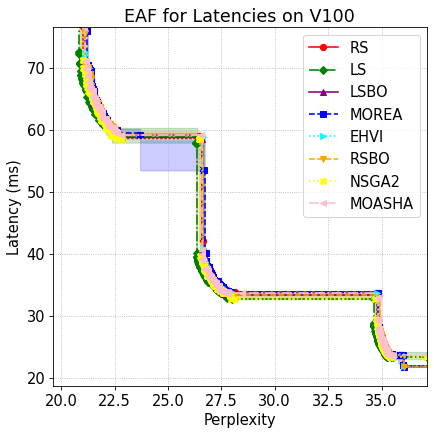}
\end{minipage}
\begin{minipage}{.16\linewidth}
  \centering
  \includegraphics[width=\linewidth]{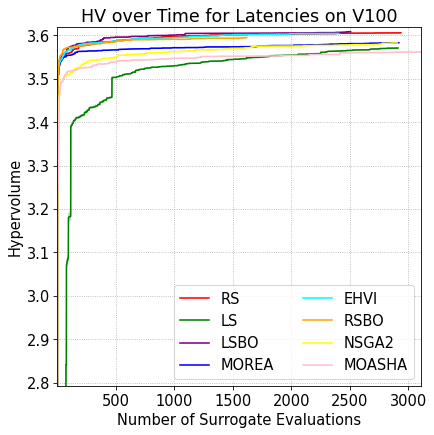}
\end{minipage}%
\begin{minipage}{.16\linewidth}
  \centering
\includegraphics[width=\linewidth]{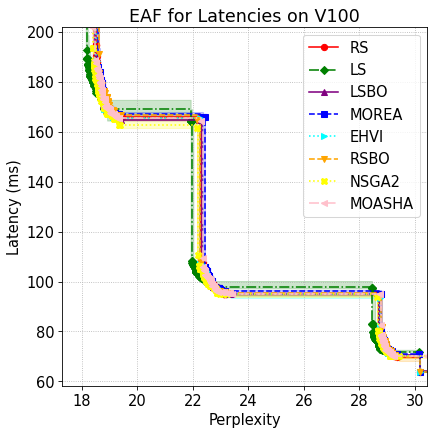}
\end{minipage}
\caption{Pareto fronts and HV over time on V100 for GPT-S (first two), GPT-M (second two) and GPT-L (last two).}
\label{fig:baselines_latencies_v100}
\vspace{-5mm}
\end{figure}

\begin{figure}[ht]
\centering
\begin{minipage}{.16\linewidth}
  \centering
  \includegraphics[width=\linewidth]{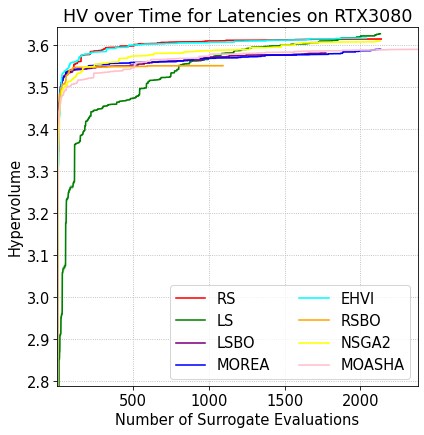}
\end{minipage}%
\begin{minipage}{.16\linewidth}
  \centering
\includegraphics[width=\linewidth]{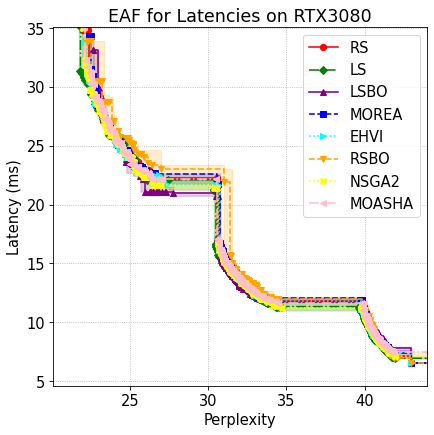}
\end{minipage}
\begin{minipage}{.16\linewidth}
  \centering
  \includegraphics[width=\linewidth]{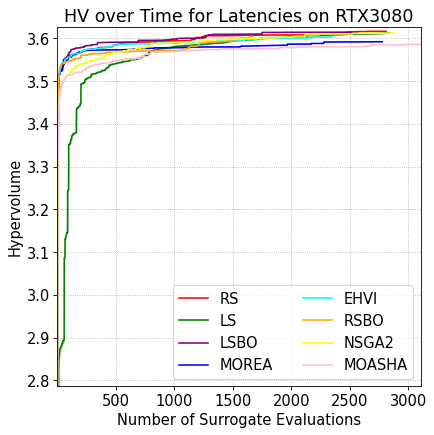}
\end{minipage}%
\begin{minipage}{.16\linewidth}
  \centering
\includegraphics[width=\linewidth]{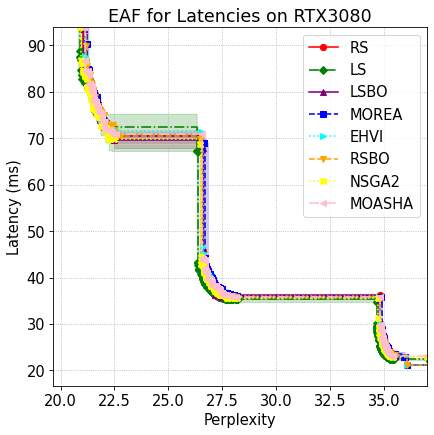}
\end{minipage}
\begin{minipage}{.16\linewidth}
  \centering
  \includegraphics[width=\linewidth]{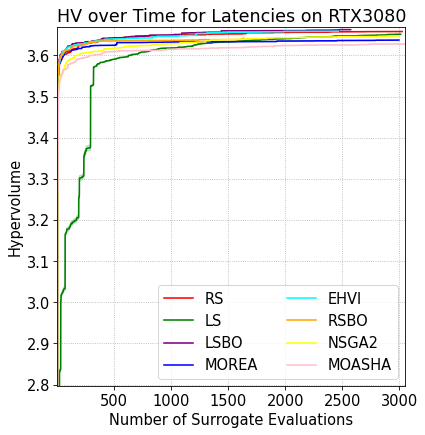}
\end{minipage}%
\begin{minipage}{.16\linewidth}
  \centering
\includegraphics[width=\linewidth]{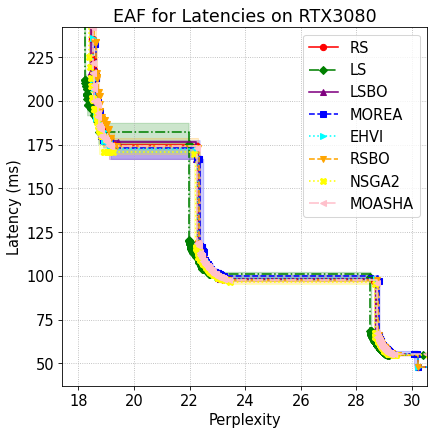}
\end{minipage}
\caption{Pareto fronts and HV over time on RTX3080 for GPT-S (first two), GPT-M (second two) and GPT-L (last two).}
\label{fig:baselines_latencies_rtx3080}
\vspace{-5mm}
\end{figure}

\begin{figure}[ht]
\centering
\begin{minipage}{.16\linewidth}
  \centering
  \includegraphics[width=\linewidth]{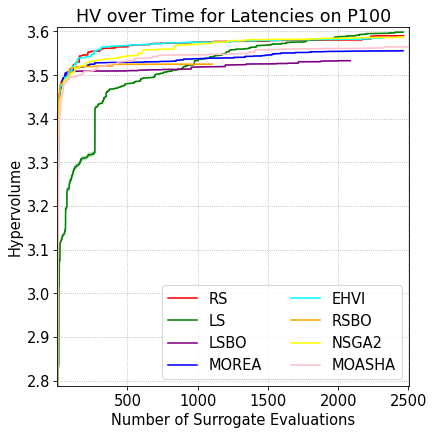}
\end{minipage}%
\begin{minipage}{.16\linewidth}
  \centering
\includegraphics[width=\linewidth]{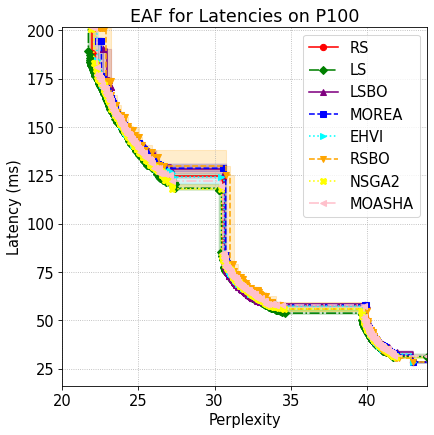}

\end{minipage}
\begin{minipage}{.16\linewidth}
  \centering
  \includegraphics[width=\linewidth]{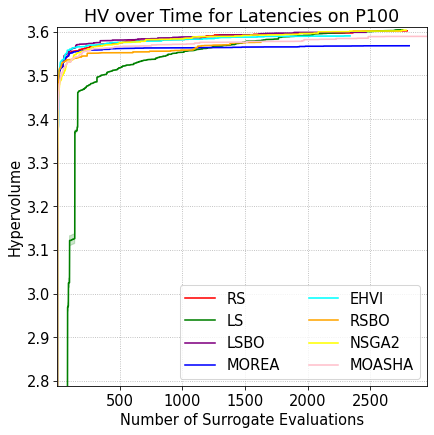}

\end{minipage}%
\begin{minipage}{.16\linewidth}
  \centering
\includegraphics[width=\linewidth]{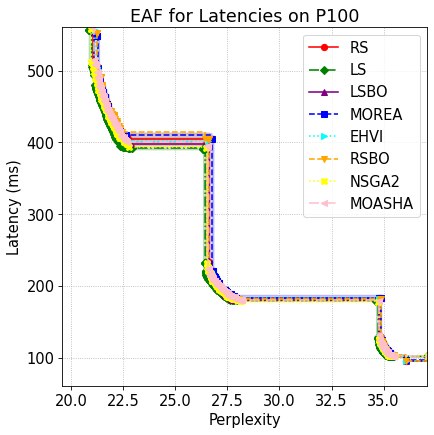}

\end{minipage}
\begin{minipage}{.16\linewidth}
  \centering
  \includegraphics[width=\linewidth]{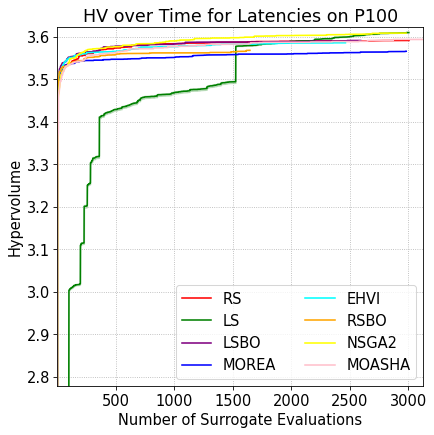}

\end{minipage}%
\begin{minipage}{.16\linewidth}
  \centering
\includegraphics[width=\linewidth]{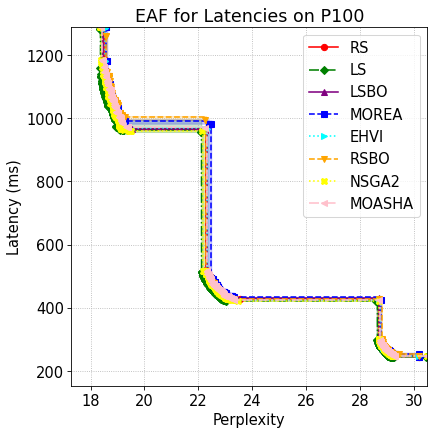}

\end{minipage}
\caption{Pareto fronts and HV over time on P100 for GPT-S (first two), GPT-M (second two) and GPT-L (last two).}
\label{fig:baselines_latencies_P100}
\vspace{-5mm}
\end{figure}

\begin{figure}[ht]
\centering
\begin{minipage}{.16\linewidth}
  \centering
  \includegraphics[width=\linewidth]{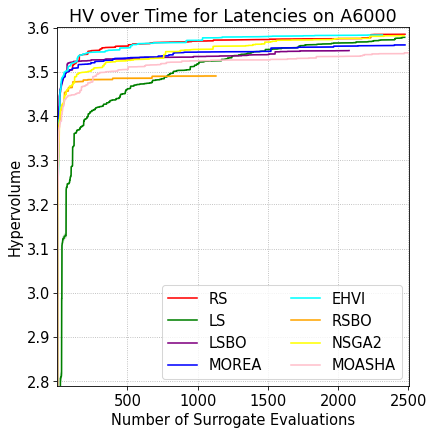}
\end{minipage}%
\begin{minipage}{.16\linewidth}
  \centering
\includegraphics[width=\linewidth]{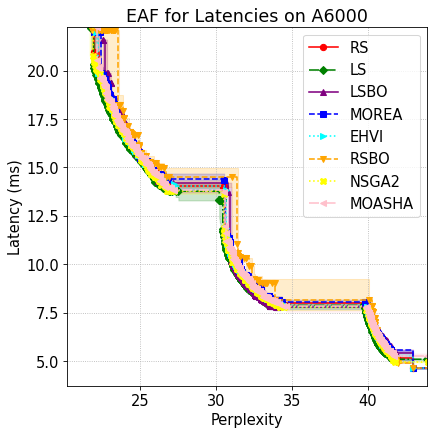}
\end{minipage}
\begin{minipage}{.16\linewidth}
  \centering
  \includegraphics[width=\linewidth]{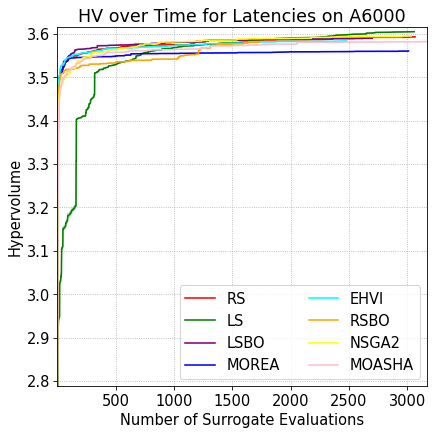}
\end{minipage}%
\begin{minipage}{.16\linewidth}
  \centering
\includegraphics[width=\linewidth]{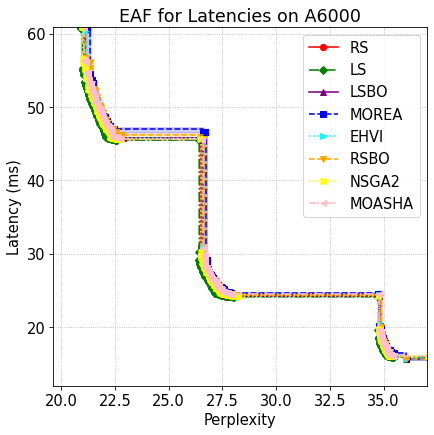}
\end{minipage}
\begin{minipage}{.16\linewidth}
  \centering
  \includegraphics[width=\linewidth]{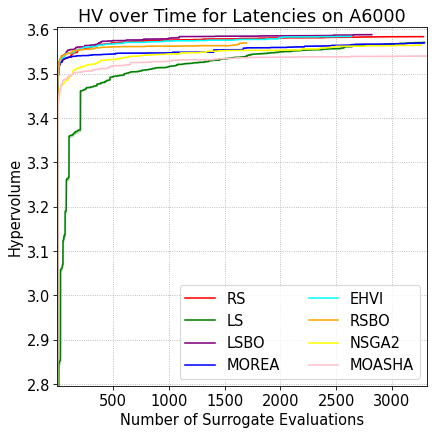}
\end{minipage}%
\begin{minipage}{.16\linewidth}
  \centering
\includegraphics[width=\linewidth]{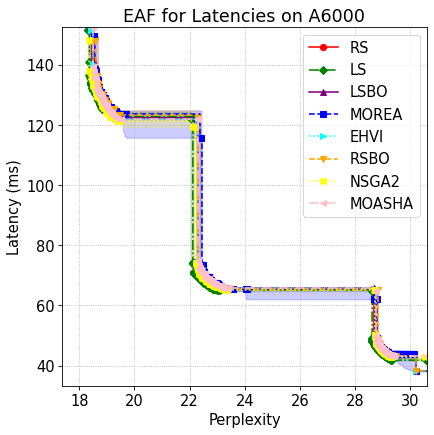}
\end{minipage}
\caption{Pareto fronts and HV over time on A6000 for GPT-S (first two), GPT-M (second two) and GPT-L (last two).}
\label{fig:baselines_latencies_a6000}
\vspace{-5mm}
\end{figure}

\begin{figure}[ht]
\centering
\begin{minipage}{.16\linewidth}
  \centering
  \includegraphics[width=\linewidth]{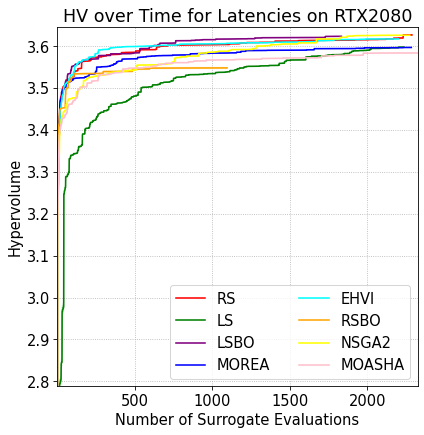}
\end{minipage}%
\begin{minipage}{.16\linewidth}
  \centering
\includegraphics[width=\linewidth]{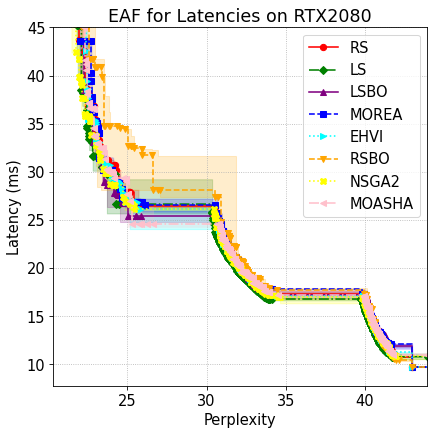}
\end{minipage}
\begin{minipage}{.16\linewidth}
  \centering
  \includegraphics[width=\linewidth]{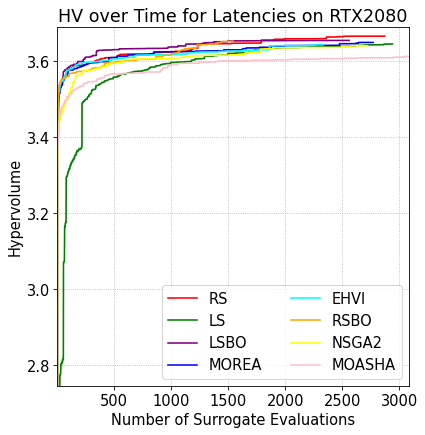}
\end{minipage}%
\begin{minipage}{.16\linewidth}
  \centering
\includegraphics[width=\linewidth]{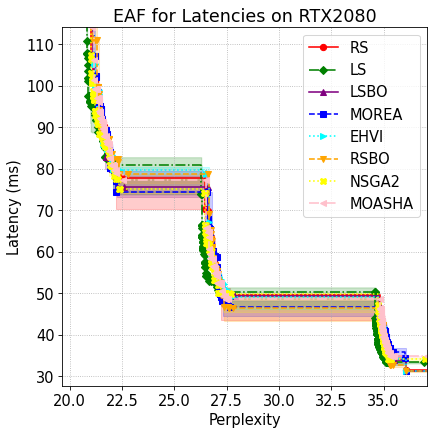}
\end{minipage}
\begin{minipage}{.16\linewidth}
  \centering
  \includegraphics[width=\linewidth]{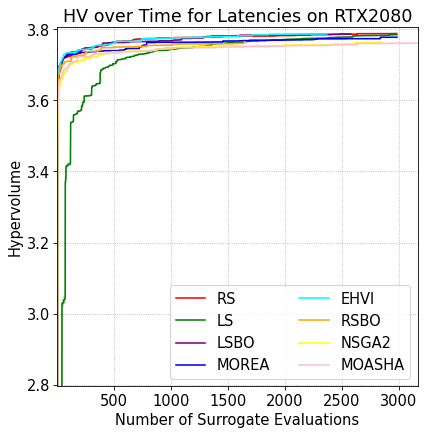}
\end{minipage}%
\begin{minipage}{.16\linewidth}
  \centering
\includegraphics[width=\linewidth]{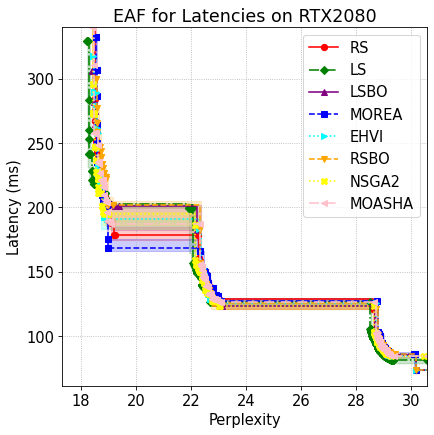}
\end{minipage}
\caption{Pareto fronts and HV over time on RTX2080 for GPT-S (first two), GPT-M (second two) and GPT-L (last two).}
\label{fig:baselines_latencies_rtx2080}
\vspace{-5mm}
\end{figure}

\begin{figure}[ht]
\centering
\begin{minipage}{.16\linewidth}
  \centering
  \includegraphics[width=\linewidth]{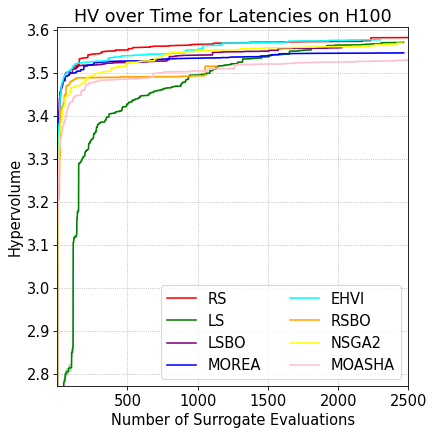}
\end{minipage}%
\begin{minipage}{.16\linewidth}
  \centering
\includegraphics[width=\linewidth]{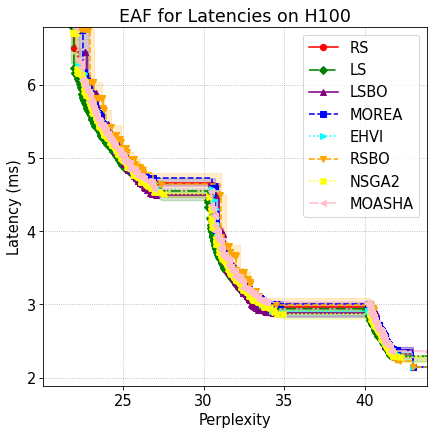}
\end{minipage}
\begin{minipage}{.16\linewidth}
  \centering
  \includegraphics[width=\linewidth]{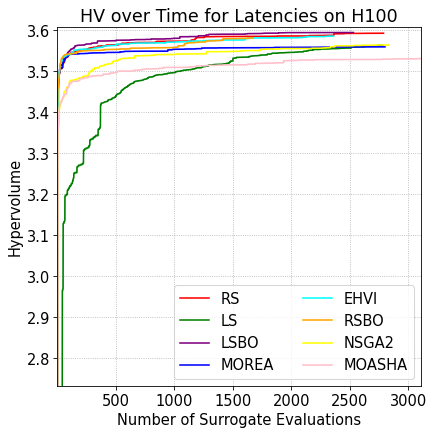}
\end{minipage}%
\begin{minipage}{.16\linewidth}
  \centering
\includegraphics[width=\linewidth]{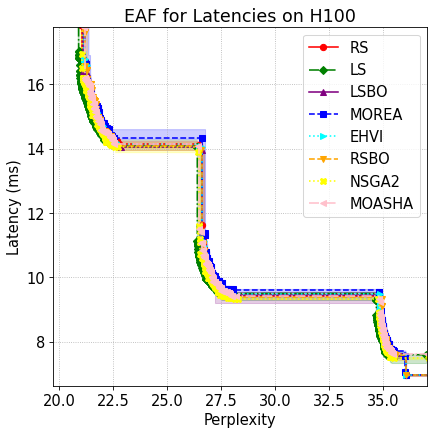}
\end{minipage}
\begin{minipage}{.16\linewidth}
  \centering
  \includegraphics[width=\linewidth]{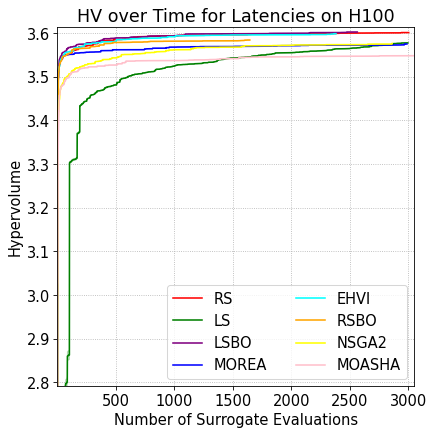}
\end{minipage}%
\begin{minipage}{.16\linewidth}
  \centering
\includegraphics[width=\linewidth]{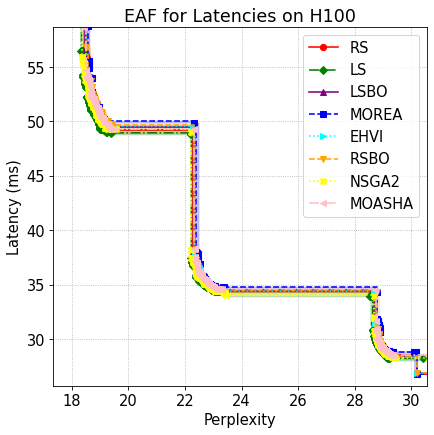}
\end{minipage}
\caption{Pareto fronts and HV over time on H100 for GPT-S (first two), GPT-M (second two) and GPT-L (last two).}
\label{fig:baselines_latencies_h100}
\vspace{-5mm}
\end{figure}

\begin{figure}[ht]
\centering
\begin{minipage}{.16\linewidth}
  \centering
  \includegraphics[width=\linewidth]{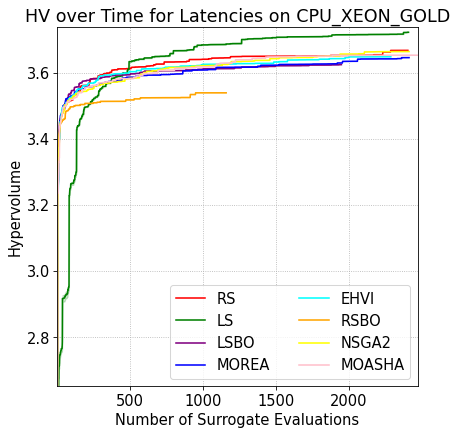}
\end{minipage}%
\begin{minipage}{.16\linewidth}
  \centering
\includegraphics[width=\linewidth]{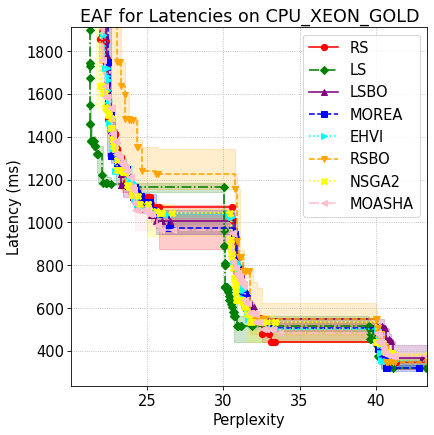}
\end{minipage}
\begin{minipage}{.16\linewidth}
  \centering
  \includegraphics[width=\linewidth]{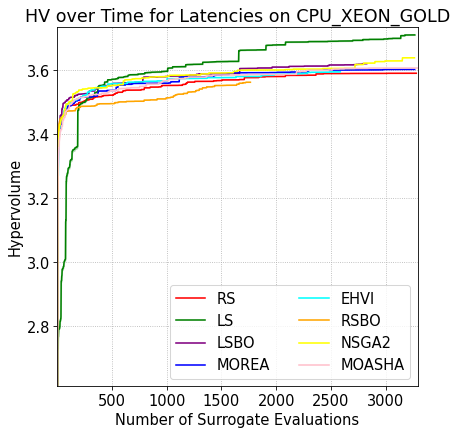}
\end{minipage}%
\begin{minipage}{.16\linewidth}
  \centering
\includegraphics[width=\linewidth]{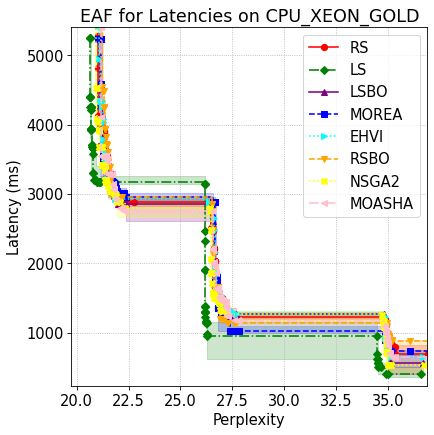}
\end{minipage}
\begin{minipage}{.16\linewidth}
  \centering
  \includegraphics[width=\linewidth]{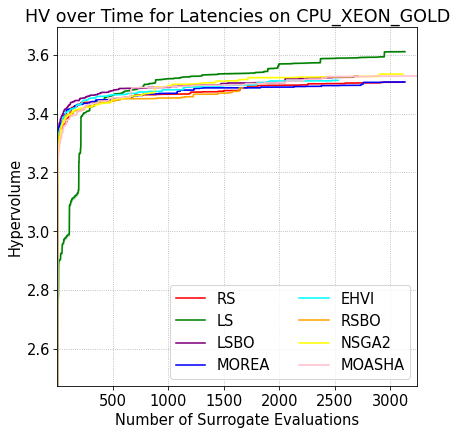}
\end{minipage}%
\begin{minipage}{.16\linewidth}
  \centering
\includegraphics[width=\linewidth]{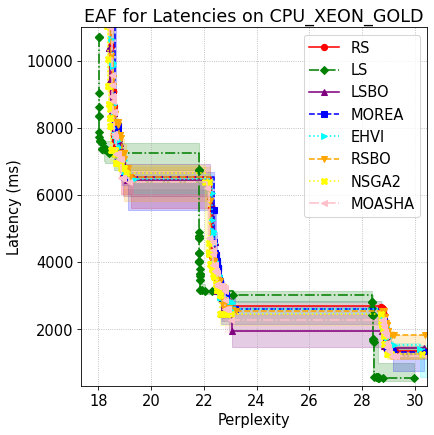}
\end{minipage}
\caption{
Pareto fronts and HV over time on CPU Xeon Gold for GPT-S (first two), GPT-M (second two) and GPT-L (last two).
}
\label{fig:baselines_latencies_gold}
\vspace{-5mm}
\end{figure}

\begin{figure}[ht]
\centering
\begin{minipage}{.16\linewidth}
  \centering
  \includegraphics[width=\linewidth]{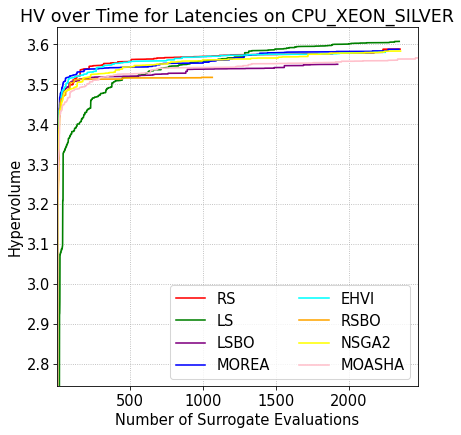}
\end{minipage}%
\begin{minipage}{.16\linewidth}
  \centering
\includegraphics[width=\linewidth]{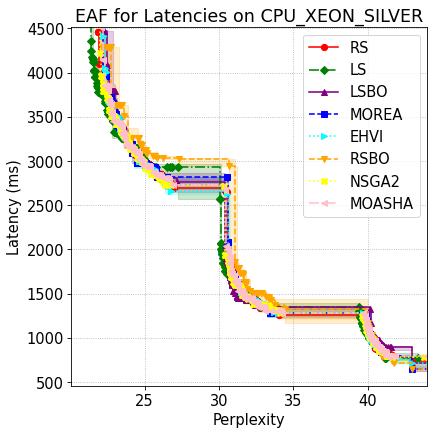}
\end{minipage}
\begin{minipage}{.16\linewidth}
  \centering
  \includegraphics[width=\linewidth]{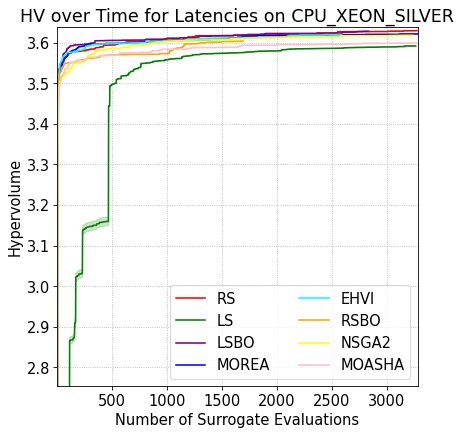}
\end{minipage}%
\begin{minipage}{.16\linewidth}
  \centering
\includegraphics[width=\linewidth]{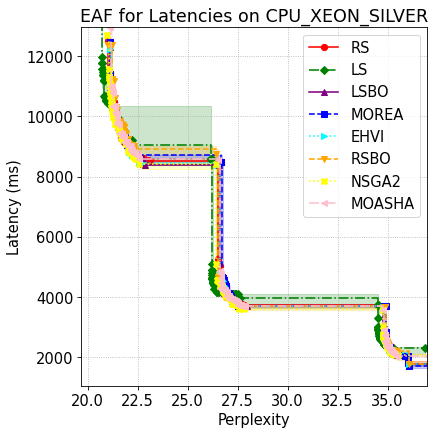}
\end{minipage}
\begin{minipage}{.16\linewidth}
  \centering
  \includegraphics[width=\linewidth]{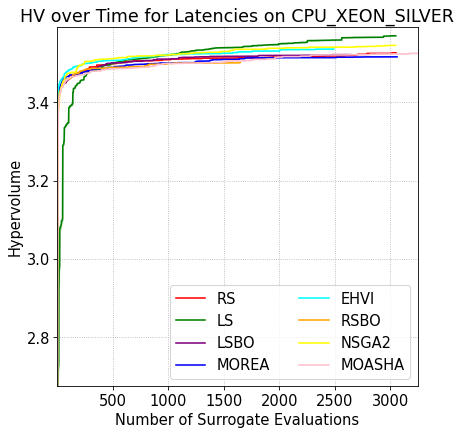}
\end{minipage}%
\begin{minipage}{.16\linewidth}
  \centering
\includegraphics[width=\linewidth]{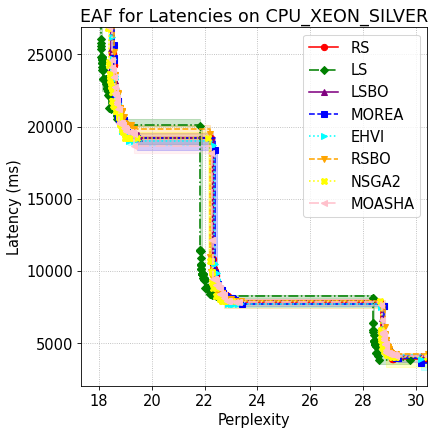}
\end{minipage}
\caption{Pareto fronts and HV over time on CPU Xeon Silver for GPT-S (first two), GPT-M (second two) and GPT-L (last two).}
\label{fig:baselines_latencies_silver}
\vspace{-5mm}
\end{figure}

\begin{figure}[ht]
\centering
\begin{minipage}{.16\linewidth}
  \centering
  \includegraphics[width=\linewidth]{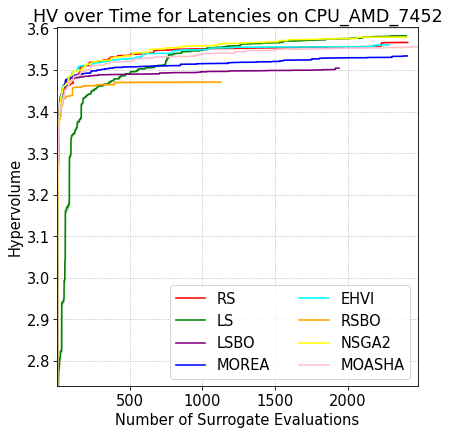}

\end{minipage}%
\begin{minipage}{.16\linewidth}
  \centering
\includegraphics[width=\linewidth]{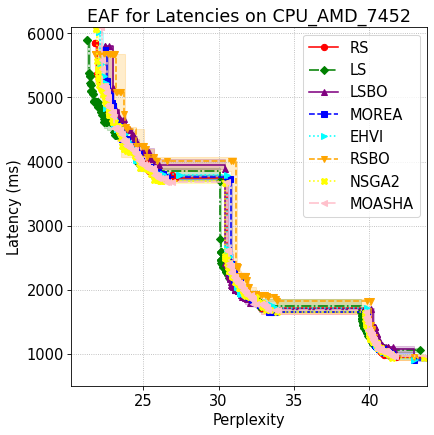}

\end{minipage}
\begin{minipage}{.16\linewidth}
  \centering
  \includegraphics[width=\linewidth]{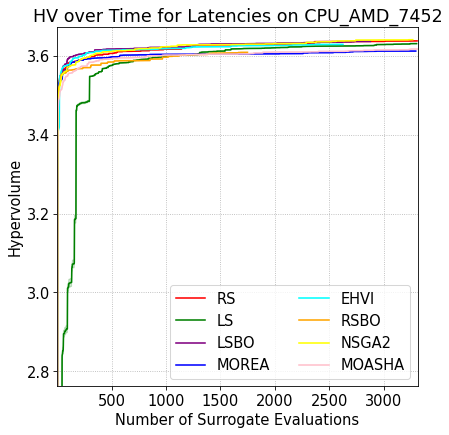}

\end{minipage}%
\begin{minipage}{.16\linewidth}
  \centering
\includegraphics[width=\linewidth]{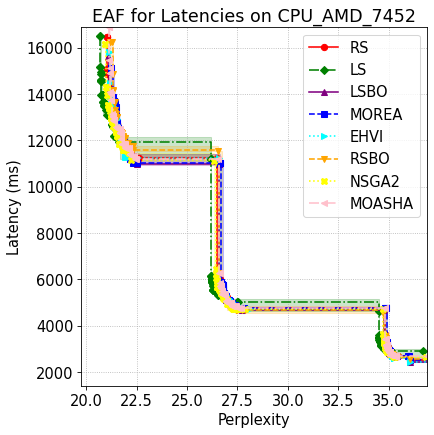}

\end{minipage}
\begin{minipage}{.16\linewidth}
  \centering
  \includegraphics[width=\linewidth]{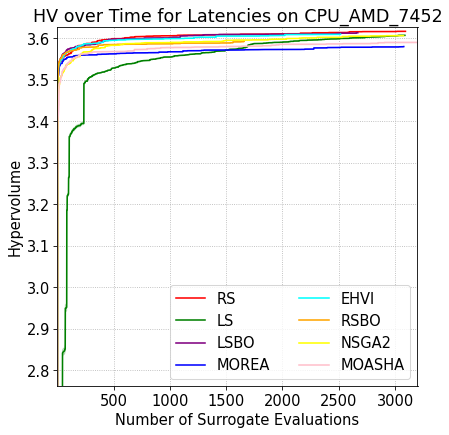}

\end{minipage}%
\begin{minipage}{.16\linewidth}
  \centering
\includegraphics[width=\linewidth]{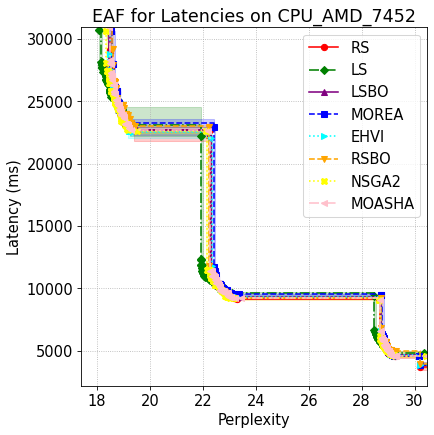}

\end{minipage}
\caption{Pareto fronts and HV over time on CPU AMD 7452 for GPT-S (first two), GPT-M (second two) and GPT-L (last two).}
\label{fig:baselines_latencies_amd_7452}
\vspace{-5mm}
\end{figure}

\begin{figure}[H]
\centering
\begin{minipage}{.16\linewidth}
  \centering
  \includegraphics[width=\linewidth]{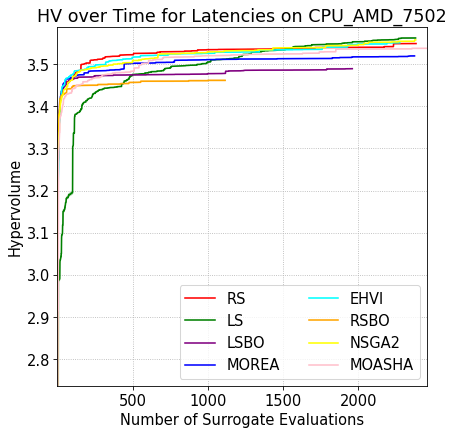}
\end{minipage}%
\begin{minipage}{.16\linewidth}
  \centering
\includegraphics[width=\linewidth]{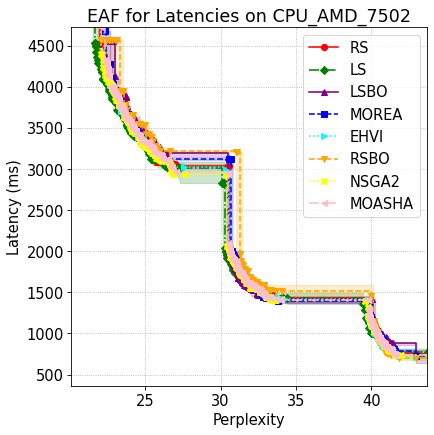}
\end{minipage}
\begin{minipage}{.16\linewidth}
  \centering
  \includegraphics[width=\linewidth]{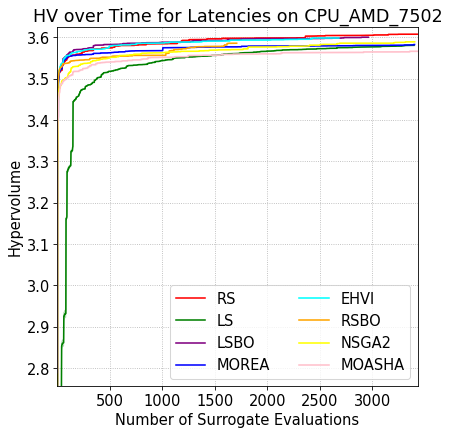}
\end{minipage}%
\begin{minipage}{.16\linewidth}
  \centering
\includegraphics[width=\linewidth]{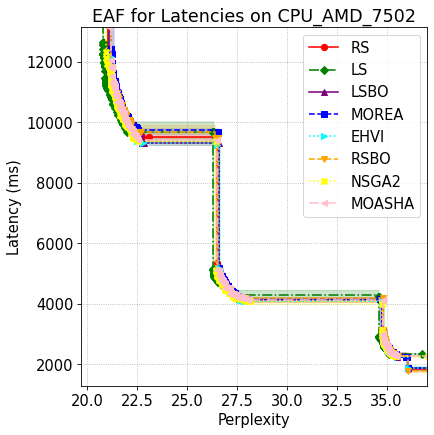}
\end{minipage}
\begin{minipage}{.16\linewidth}
  \centering
  \includegraphics[width=\linewidth]{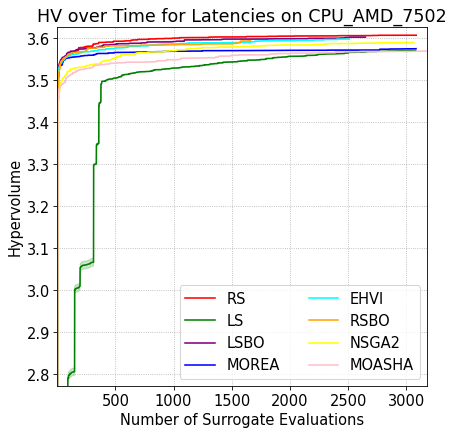}
\end{minipage}%
\begin{minipage}{.16\linewidth}
  \centering
\includegraphics[width=\linewidth]{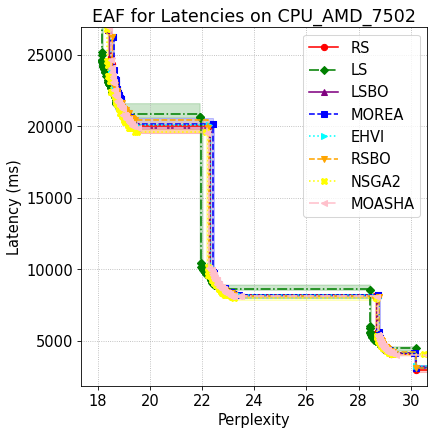}
\end{minipage}
\caption{Pareto fronts and HV over time on CPU AMD 7402 for GPT-S (first two), GPT-M (second two) and GPT-L (last two).}
\label{fig:baselines_latencies_amd_7502}
\vspace{-5mm}
\end{figure}

\begin{figure}[H]
\centering
\begin{minipage}{.16\linewidth}
  \centering
  \includegraphics[width=\linewidth]{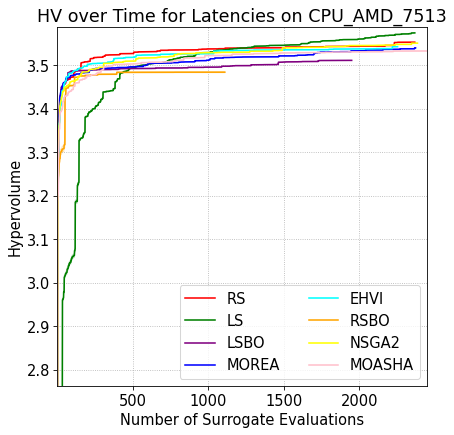}
\end{minipage}%
\begin{minipage}{.16\linewidth}
  \centering
\includegraphics[width=\linewidth]{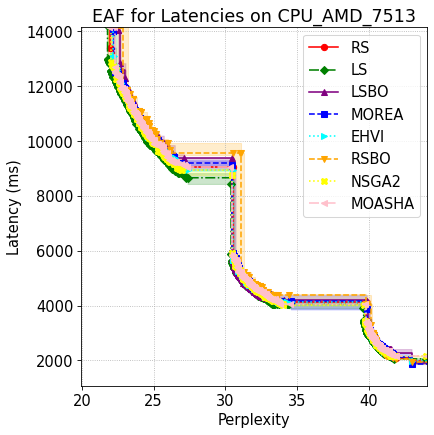}
\end{minipage}
\begin{minipage}{.16\linewidth}
  \centering
  \includegraphics[width=\linewidth]{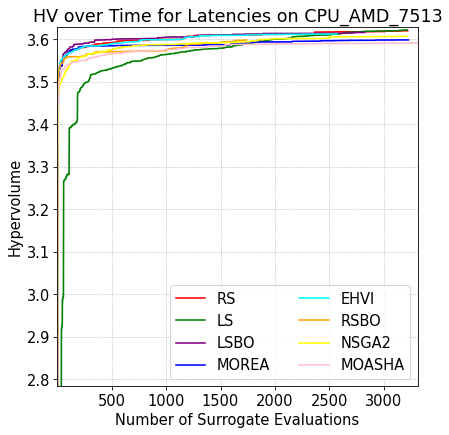}
\end{minipage}%
\begin{minipage}{.16\linewidth}
  \centering
\includegraphics[width=\linewidth]{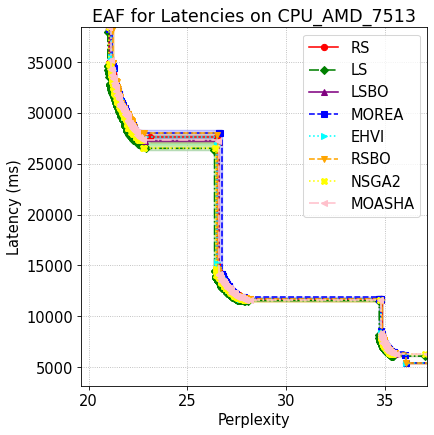}
\end{minipage}
\begin{minipage}{.16\linewidth}
  \centering
  \includegraphics[width=\linewidth]{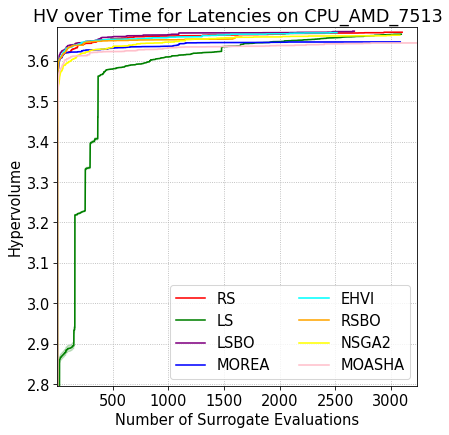}
\end{minipage}%
\begin{minipage}{.16\linewidth}
  \centering
\includegraphics[width=\linewidth]{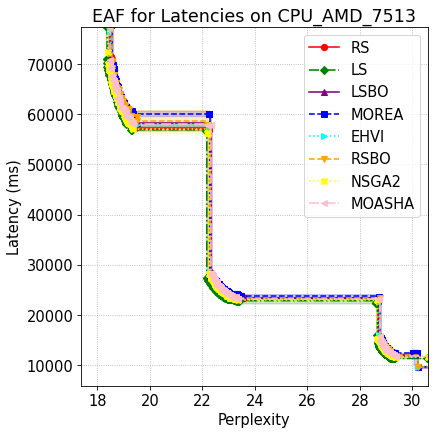}
\end{minipage}
\caption{Pareto fronts and HV over time on CPU AMD 7513 for GPT-S (first two), GPT-M (second two) and GPT-L (last two).}
\label{fig:baselines_latencies_amd_7513}
\vspace{-5mm}
\end{figure}

\begin{figure}[ht]
    \centering
    \begin{minipage}{.33\linewidth}
        \centering
        \includegraphics[width=.49\linewidth]{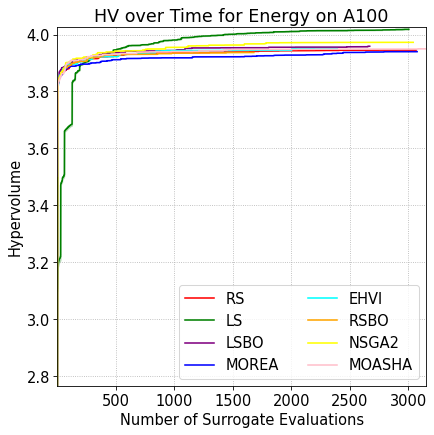}
        \includegraphics[width=.49\linewidth]{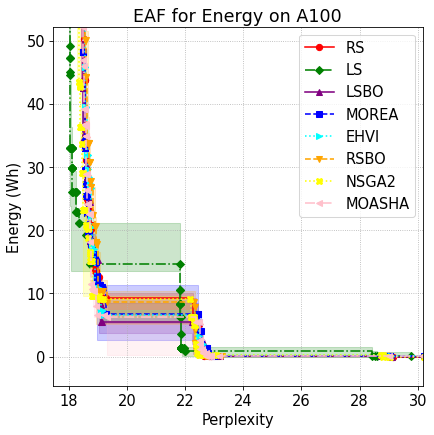}
    \end{minipage}%
    \hfill
    \begin{minipage}{.33\linewidth}
        \centering
        \includegraphics[width=.49\linewidth]{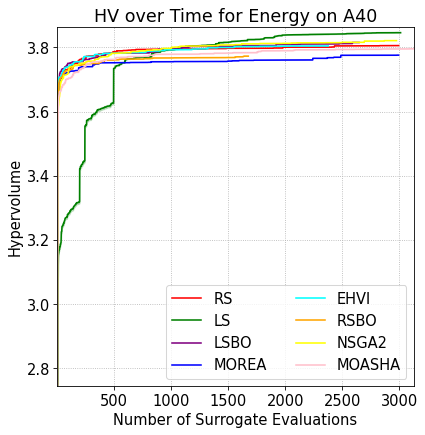}
        \includegraphics[width=.49\linewidth]{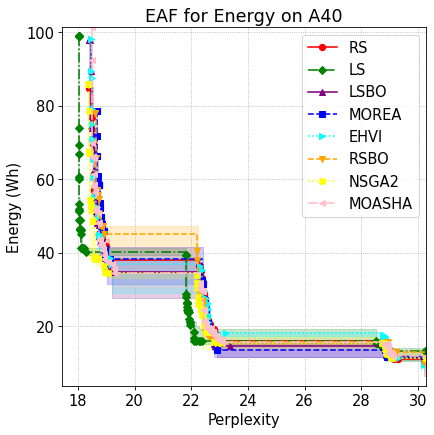}
    \end{minipage}%
    \hfill
    \begin{minipage}{.33\linewidth}
        \centering
        \includegraphics[width=.49\linewidth]{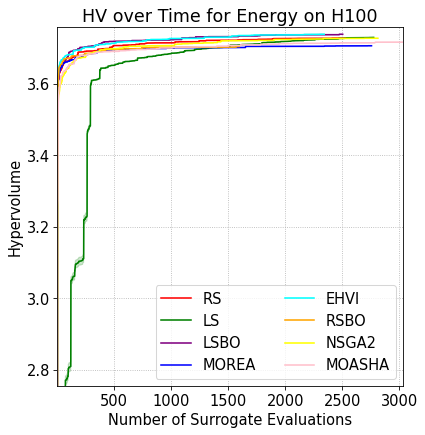}
        \includegraphics[width=.49\linewidth]{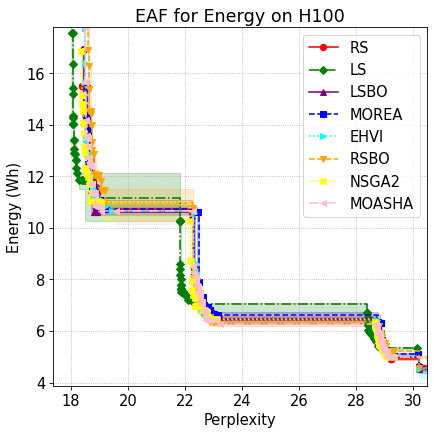}
    \end{minipage}
\caption{Pareto Fronts and HV on A100 (first 2), A40 (second 2) and H100 (last 2) for GPT-L}
\label{fig:baselines_energies_a100_a40_h100}
\end{figure}

\begin{figure}[ht]
    \centering
    \begin{minipage}{.33\linewidth}
        \centering
        \includegraphics[width=.49\linewidth]{hv_over_time_energies_rtx2080_l.png}
        \includegraphics[width=.49\linewidth]{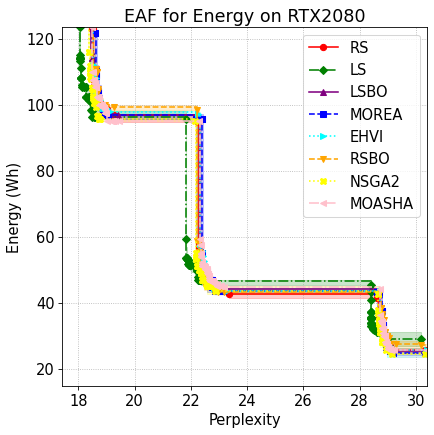}
    \end{minipage}%
    \hfill
    \begin{minipage}{.33\linewidth}
        \centering
        \includegraphics[width=.49\linewidth]{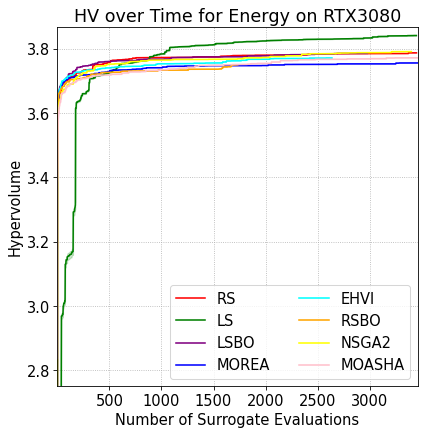}
        \includegraphics[width=.49\linewidth]{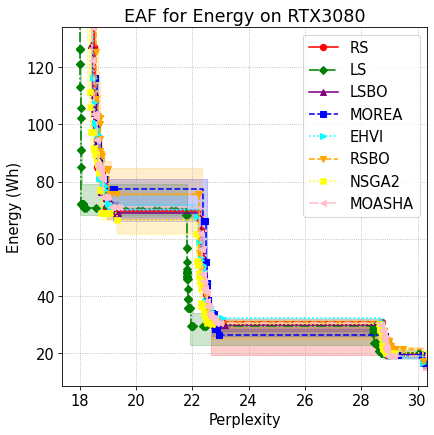}
    \end{minipage}%
    \hfill
    \begin{minipage}{.33\linewidth}
        \centering
        \includegraphics[width=.49\linewidth]{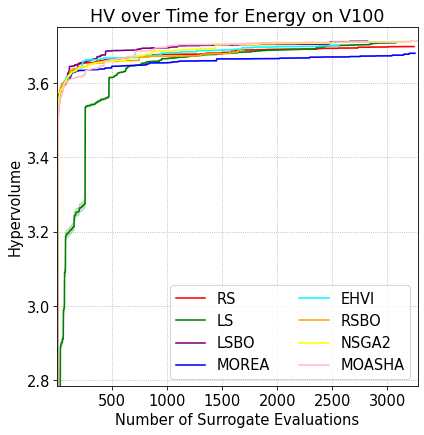}
        \includegraphics[width=.49\linewidth]{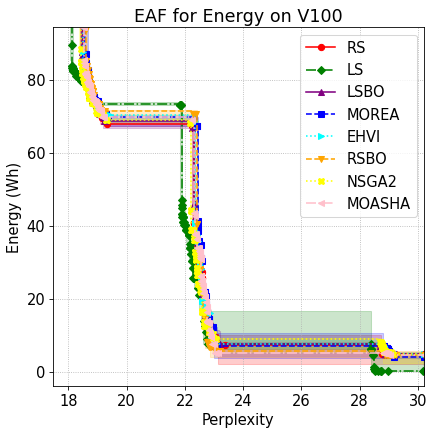}
    \end{minipage}
    \caption{Pareto Fronts and HV on RX2080 (first 2), RTX3080 (second 2) and V100 (last 2) for GPT-L}
    \label{fig:baselines_energies_rtx2080_rtx3080_v100}
\end{figure}

\begin{figure}[ht]
    \centering
    \begin{minipage}{.33\linewidth}
        \centering
        \includegraphics[width=.49\linewidth]{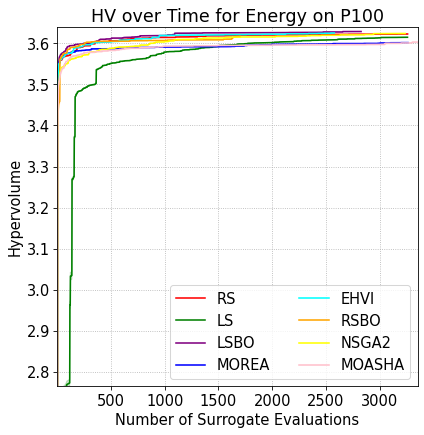}
        \includegraphics[width=.49\linewidth]{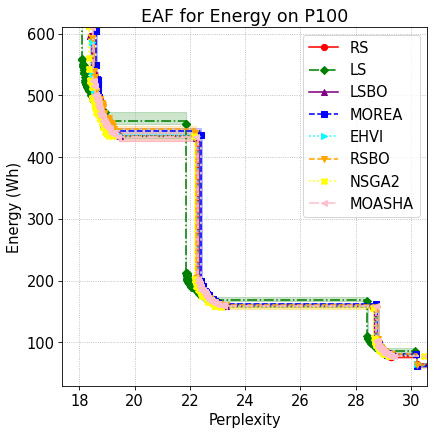}
    \end{minipage}%
    \hfill
    \begin{minipage}{.33\linewidth}
        \centering
        \includegraphics[width=.49\linewidth]{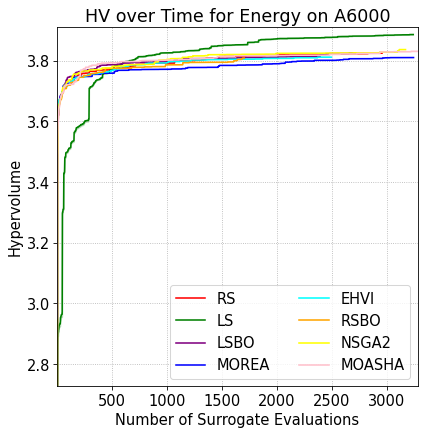}
        \includegraphics[width=.49\linewidth]{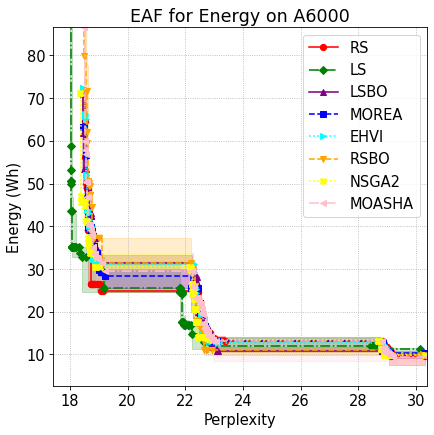}
    \end{minipage}%
    \hfill
    \begin{minipage}{.33\linewidth}
        \centering
        \includegraphics[width=.49\linewidth]{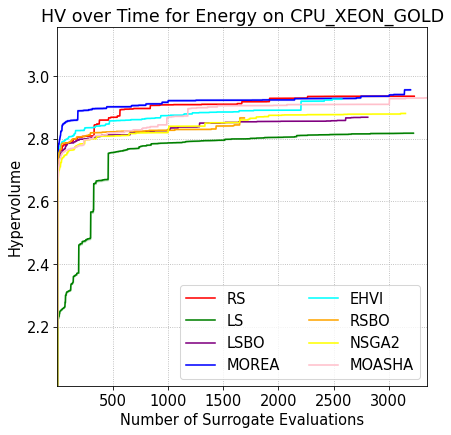}
        \includegraphics[width=.49\linewidth]{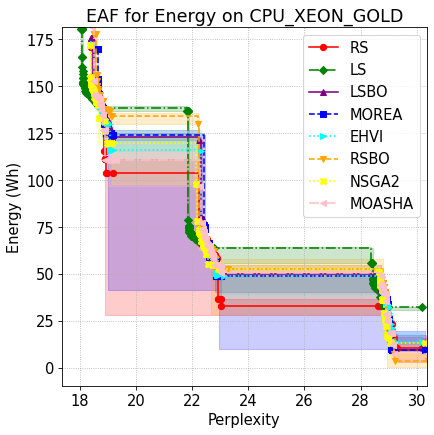}
    \end{minipage}
    \caption{Pareto Fronts and HV on P100 (first 2), A6000 (second 2) and Xeon Gold CPU (last 2) for GPT-L}
    \label{fig:baselines_energies_p100_a6000_gold}
\end{figure}

\begin{figure}[ht]
    \centering
    \begin{minipage}{.33\linewidth}
        \centering
        \includegraphics[width=.49\linewidth]{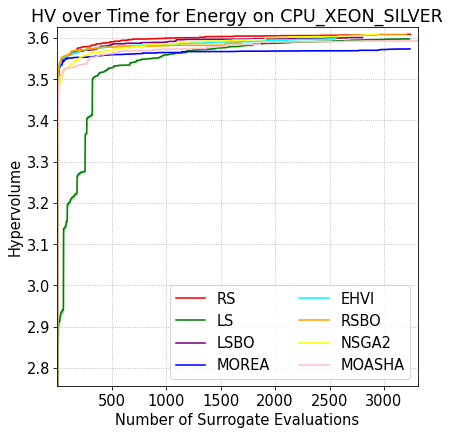}
        \includegraphics[width=.49\linewidth]{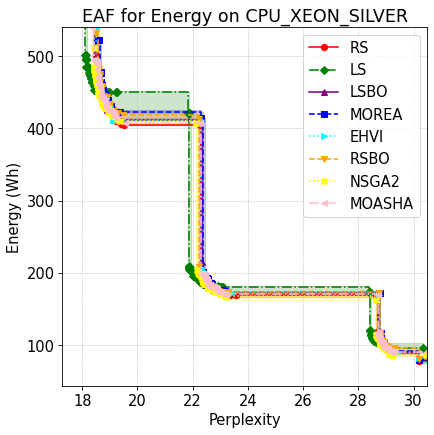}
    \end{minipage}%
    \hfill
    \begin{minipage}{.33\linewidth}
        \centering
        \includegraphics[width=.49\linewidth]{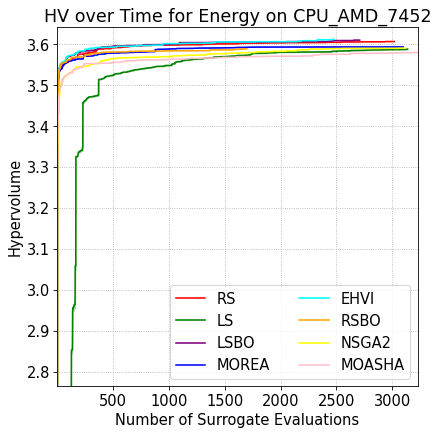}
        \includegraphics[width=.49\linewidth]{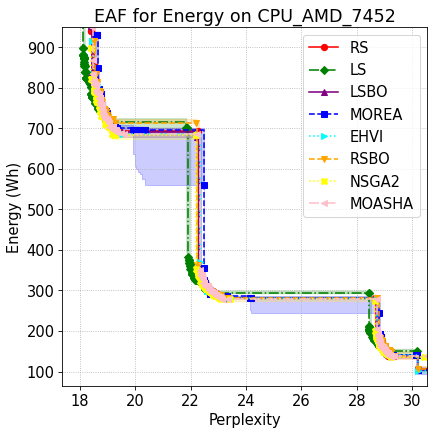}
    \end{minipage}%
    \hfill
    \begin{minipage}{.33\linewidth}
        \centering
        \includegraphics[width=.49\linewidth]{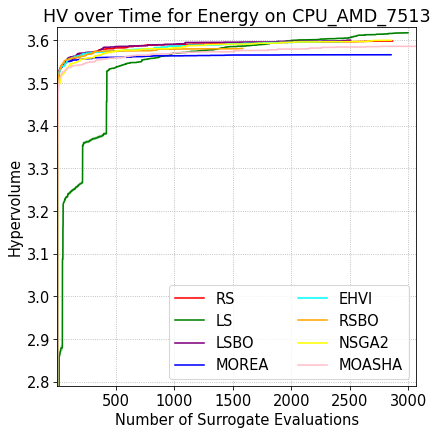}
        \includegraphics[width=.49\linewidth]{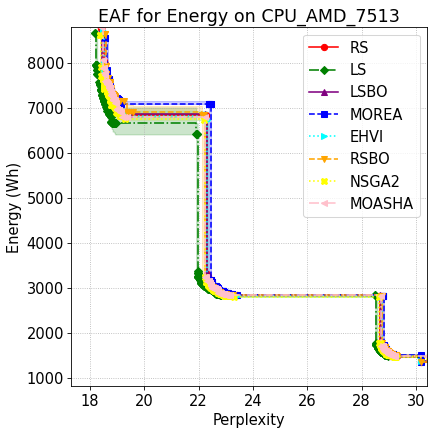}

    \end{minipage}
    \label{fig:baselines_energies_7452_7513_silver}
        \caption{Pareto Fronts and HV on Xeon Silver (first 2), AMD 7452 (second 2) and AMD 7513 (last 2) for GPT-L}
\end{figure}

\begin{figure}[ht]
\centering
\begin{minipage}{.33\textwidth}
  \centering
  \includegraphics[width=.49\linewidth]{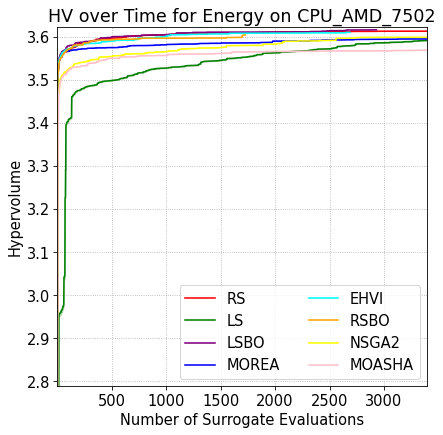}
\includegraphics[width=.49\linewidth]{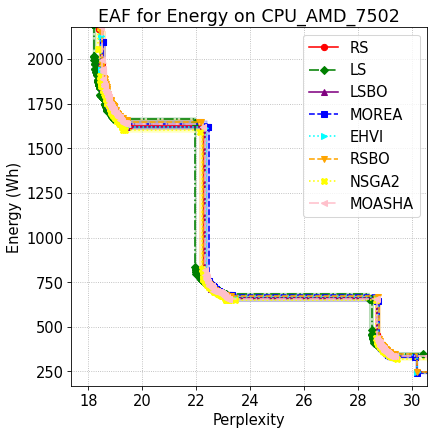}
\caption{Pareto Fronts and HV on AMD 7502 for GPT-L.}
\label{fig:baselines_energies_amd7502}
\end{minipage}
\end{figure}

We observe from Figures \ref{fig:baselines_latencies_a100}-\ref{fig:baselines_energies_amd7502} that NSGA-II and EHVI are amongst the top performing methods (even at lower budgets). LS typically has a low hypervolume in the beginning, however, often outperforms other methods with enough budget.





\subsection{Experiments with 3 Objectives}
\label{sec:3obj_app}
\begin{figure}[H]
\centering
\begin{minipage}{.24\linewidth}
  \centering
  \includegraphics[width=\linewidth]{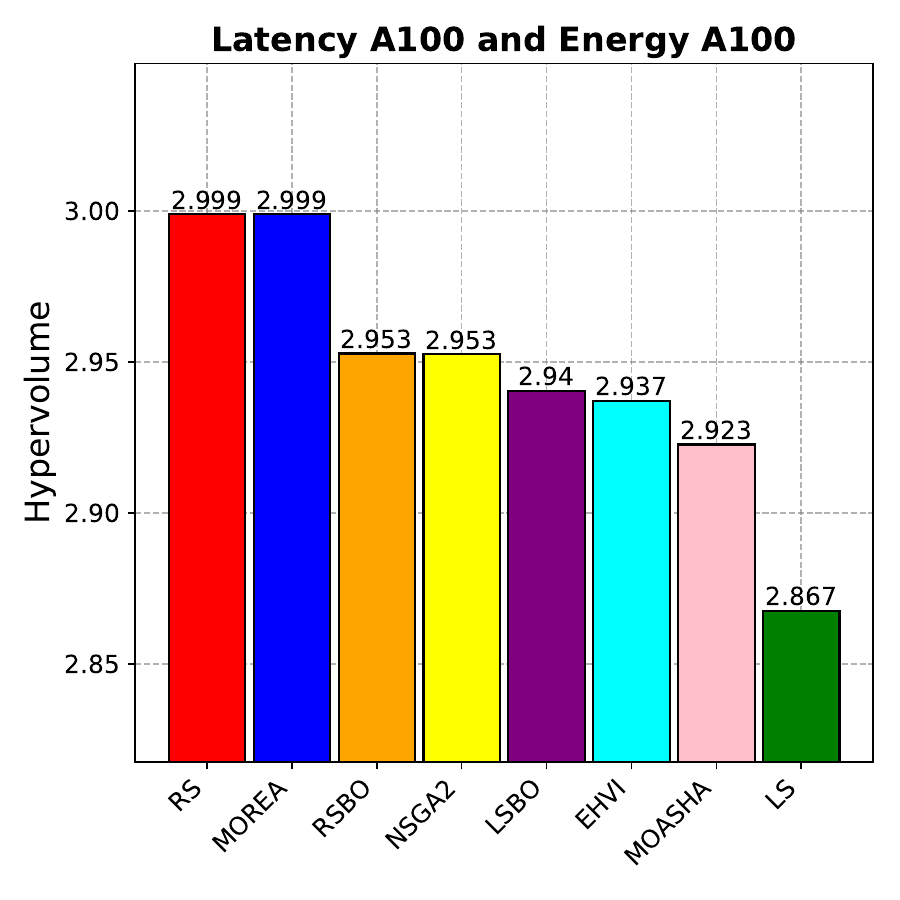}
\end{minipage}%
\begin{minipage}{.24\linewidth}
  \centering
\includegraphics[width=\linewidth]{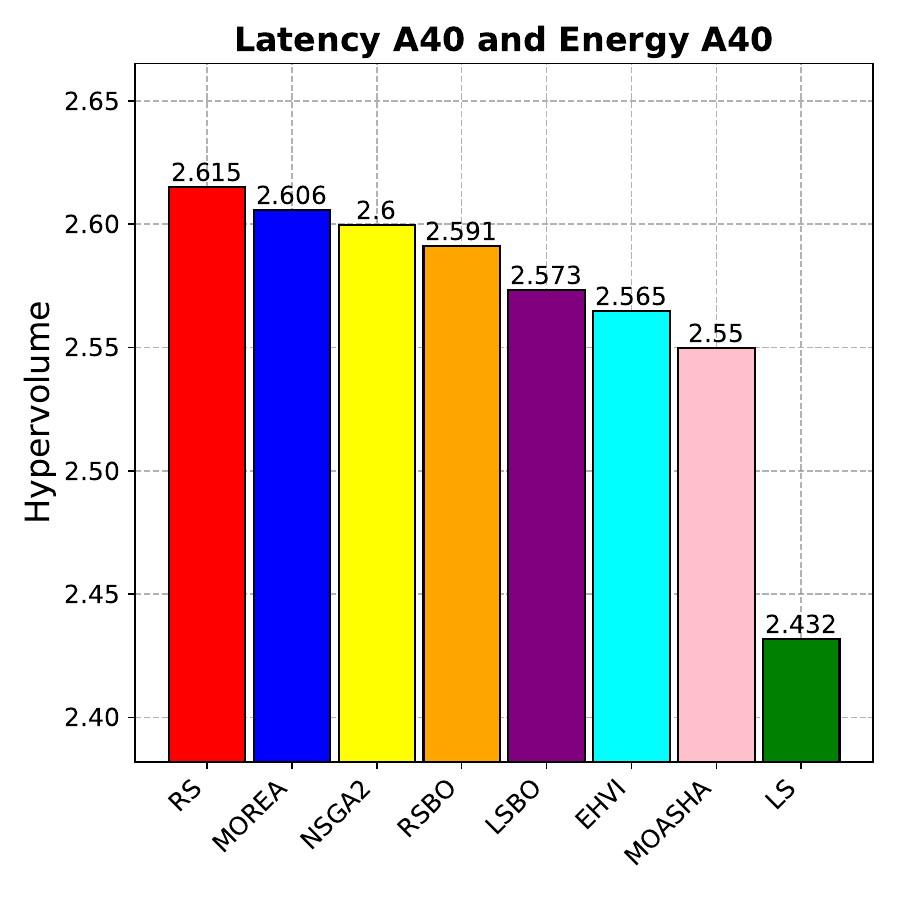}
\end{minipage}
\begin{minipage}{.24\linewidth}
  \centering
\includegraphics[width=\linewidth]{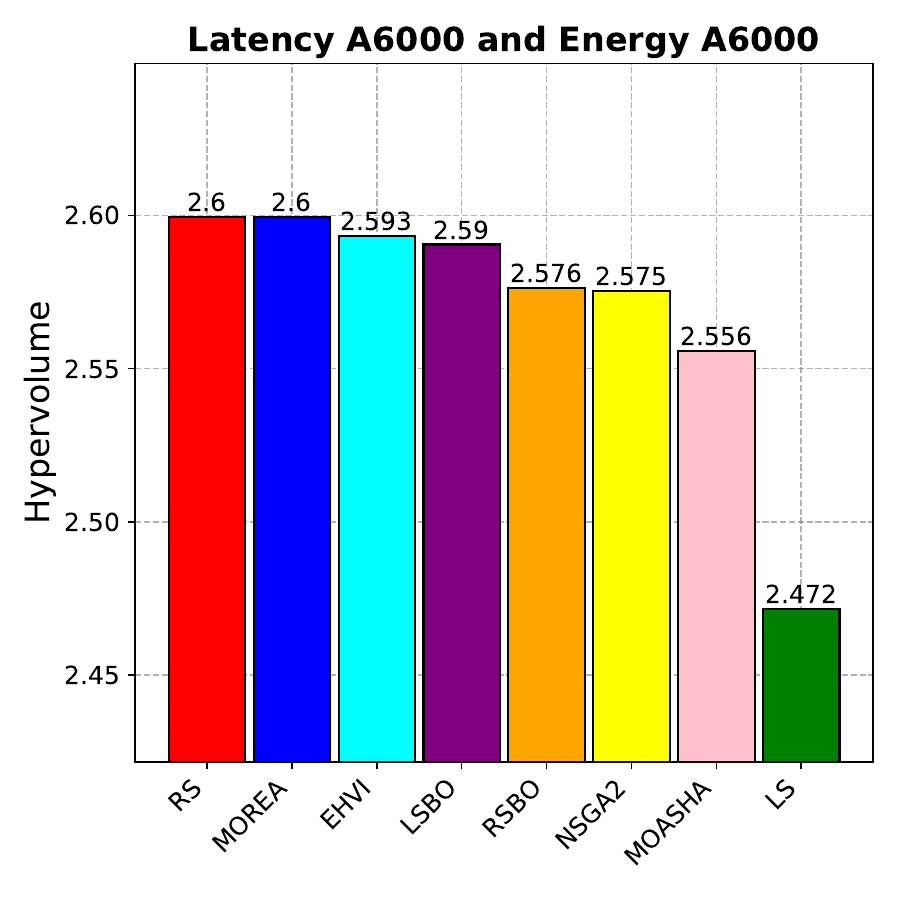}
\end{minipage}
\begin{minipage}{.24\linewidth}
  \centering
\includegraphics[width=\linewidth]{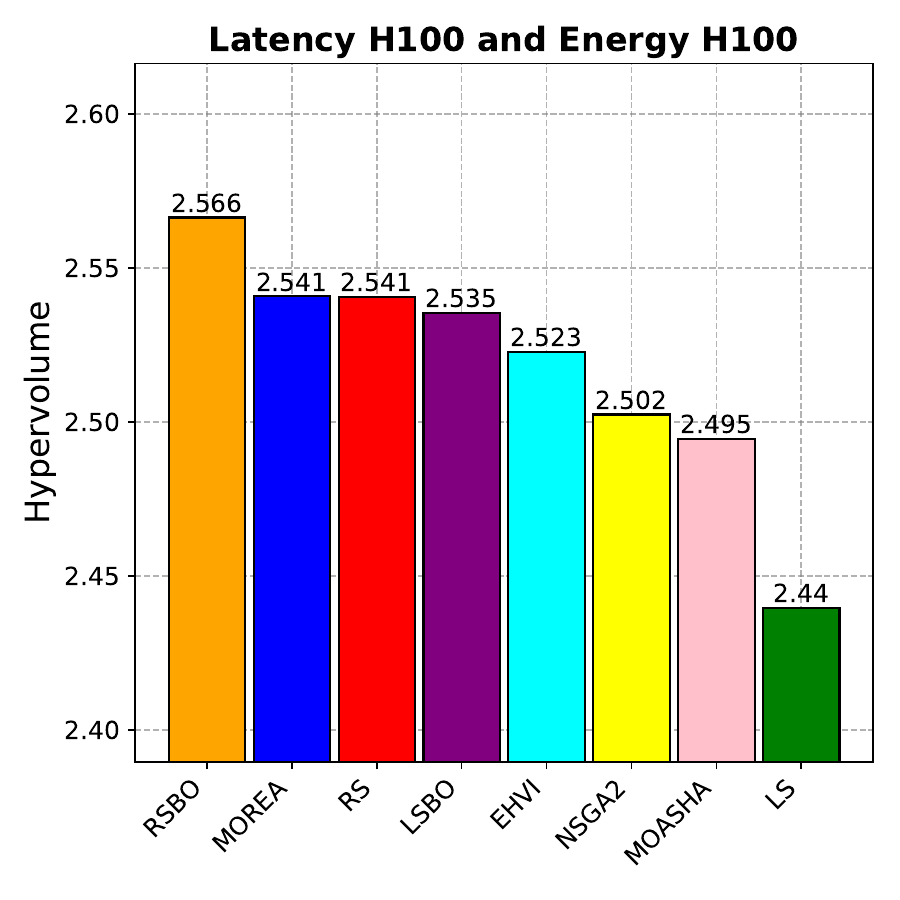}
\end{minipage}
\caption{Hypervolumes of baselines optimizing for perplexity, latency and energy usage on A100, A40, A6000 and H100.}
\label{fig:baselines_3d_app_1}
\vspace{-5mm}
\end{figure}

\begin{figure}[H]
\centering
\begin{minipage}{.24\linewidth}
  \centering
  \includegraphics[width=\linewidth]{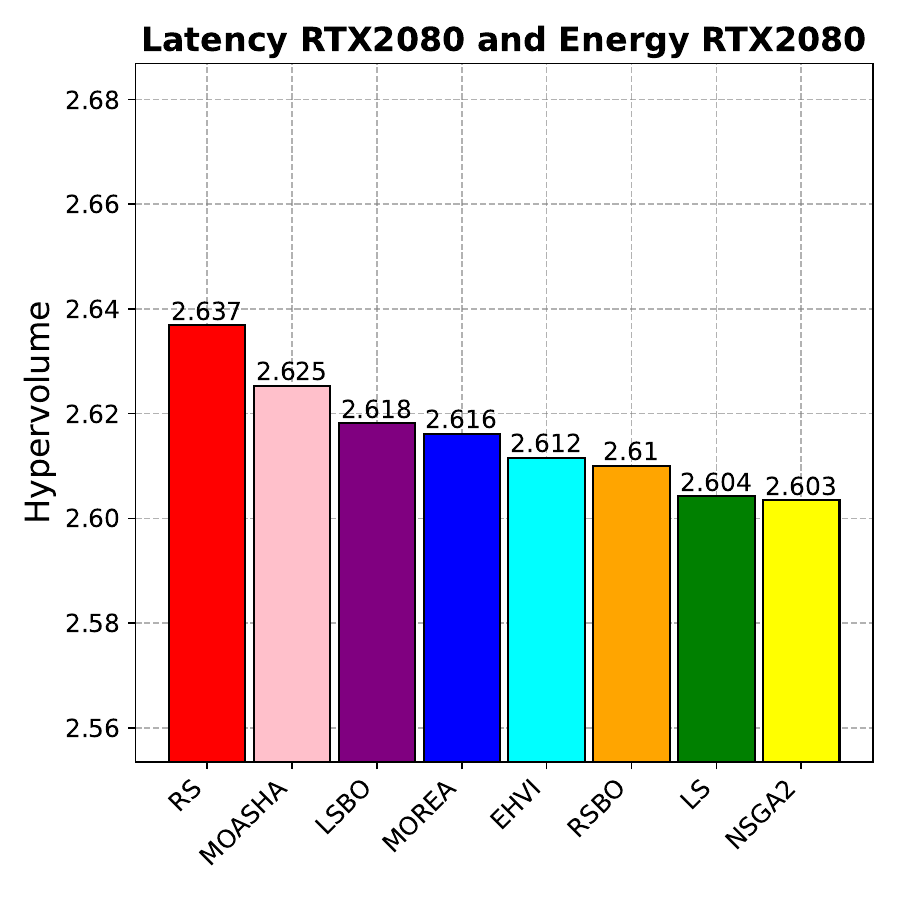}
\end{minipage}%
\begin{minipage}{.24\linewidth}
  \centering
\includegraphics[width=\linewidth]{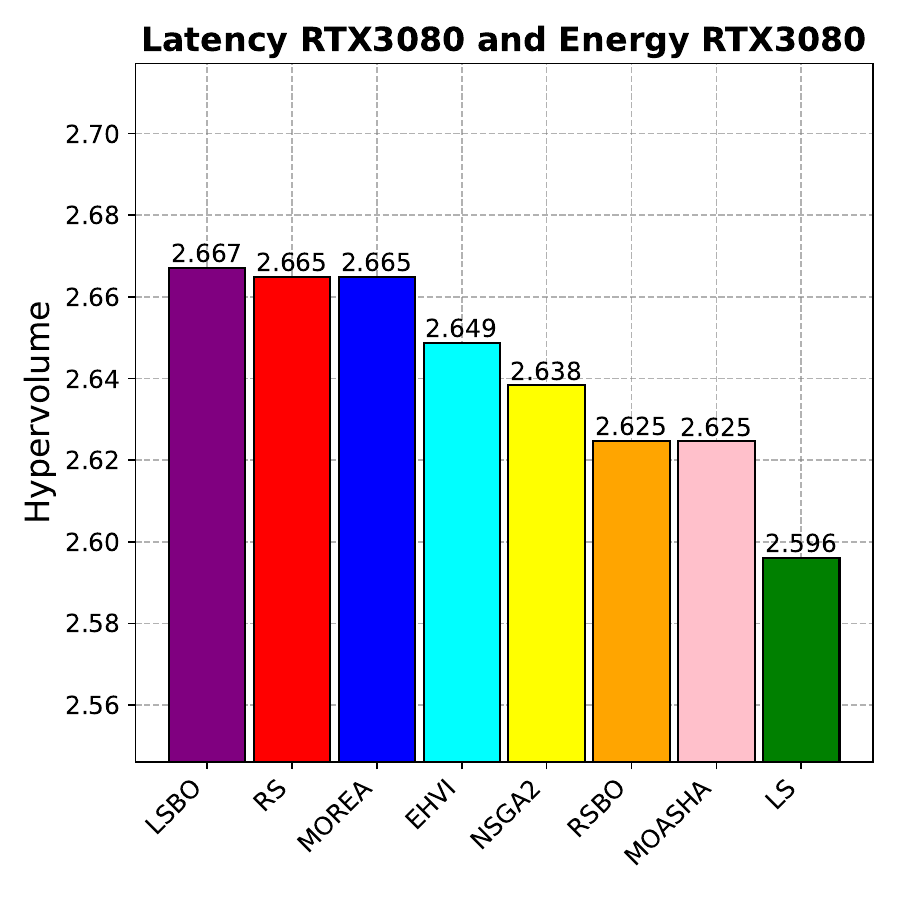}
\end{minipage}
\begin{minipage}{.24\linewidth}
  \centering
\includegraphics[width=\linewidth]{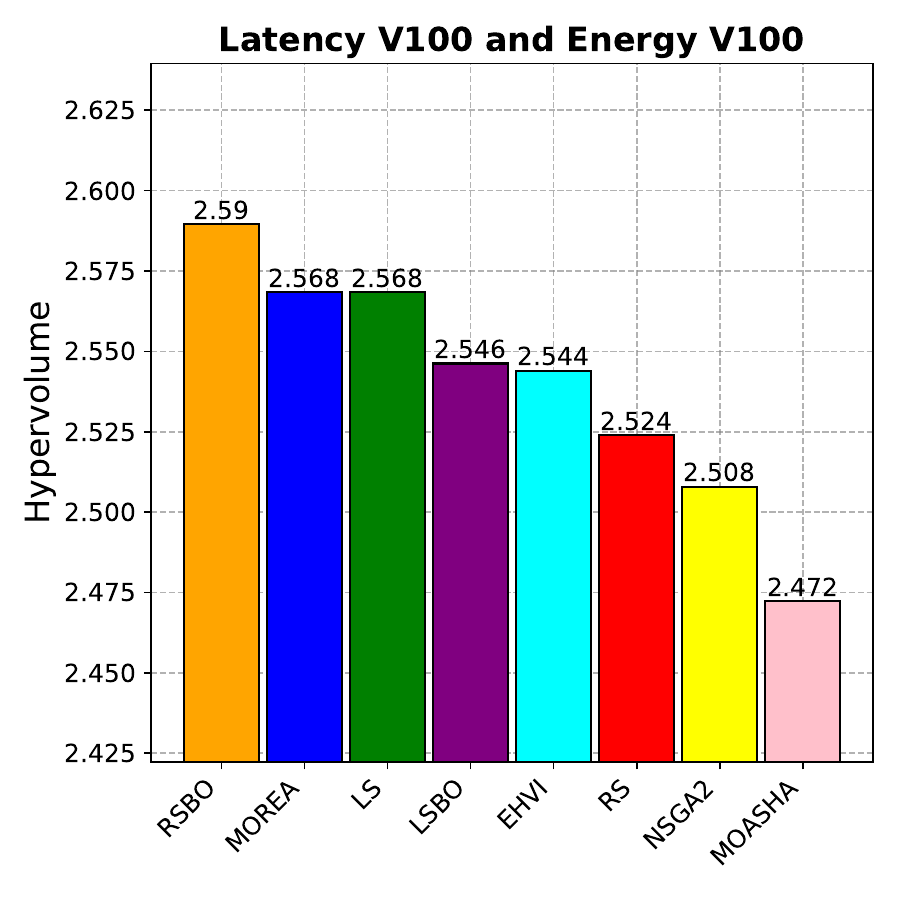}
\end{minipage}
\begin{minipage}{.24\linewidth}
  \centering
\includegraphics[width=\linewidth]{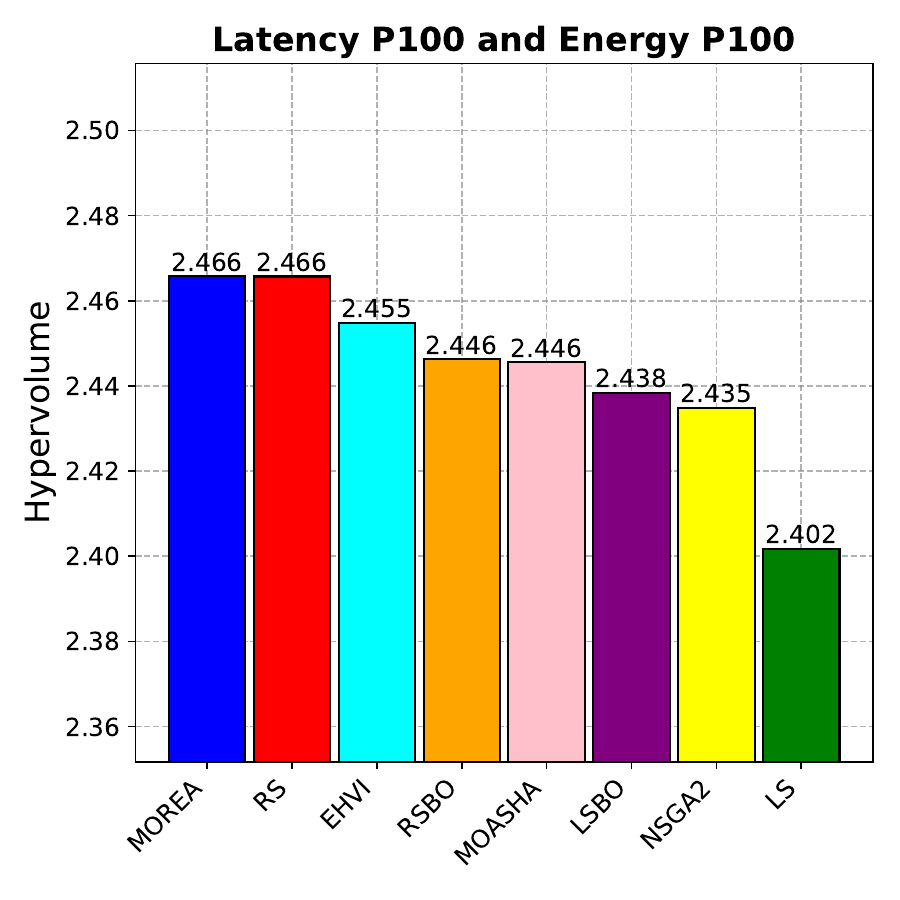}
\end{minipage}
\caption{Hypervolumes of baselines optimizing for perplexity, latency and energy usage on RTX2080, RTX3080, V100 and P100.}
\label{fig:baselines_3d_app_2}
\vspace{-5mm}
\end{figure}

In addition, we also use our benchmark to optimize energies and latencies, in conjunction with perplexity, on different devices for the GPT-L space, as presented in Figure \ref{fig:baselines_3d_app_1} and Figure \ref{fig:baselines_3d_app_2}. We run these experiments for a smaller time budget of 3 hours using SyneTune. We observe that at smaller time budgets random search and MOO methods based on Bayesian optimization are the top methods. 


\section{Correlations between different metrics}
\begin{figure}[ht]
\centering
\includegraphics[width=\textwidth, height=14cm]{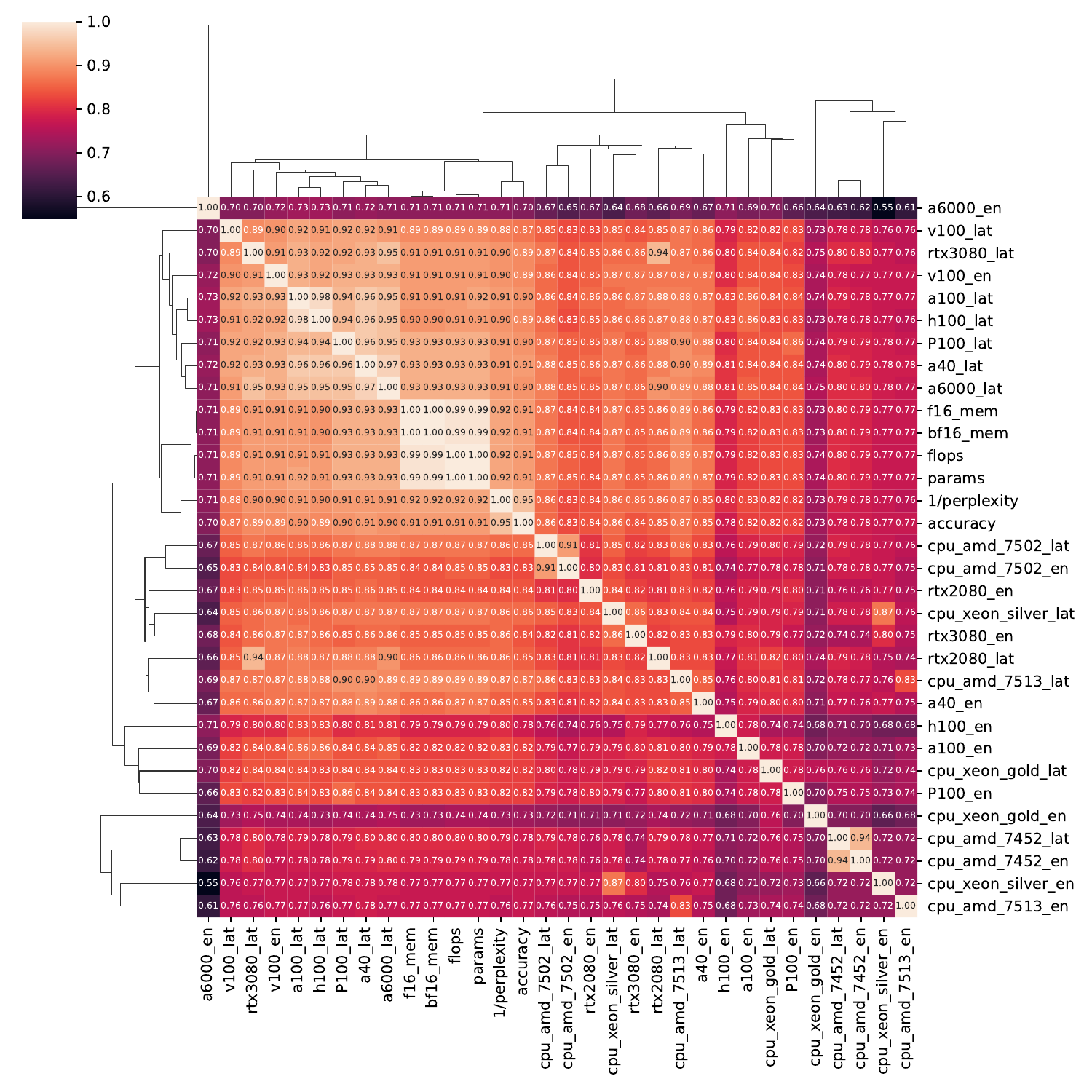}
\caption{Cross-Metric kendall-tau correlation plots for GPT-S}
\label{fig:correlation_scales_s}
\vspace{-4mm}
\end{figure}

\begin{figure}[ht]
\centering
\includegraphics[width=\textwidth, height=14cm]{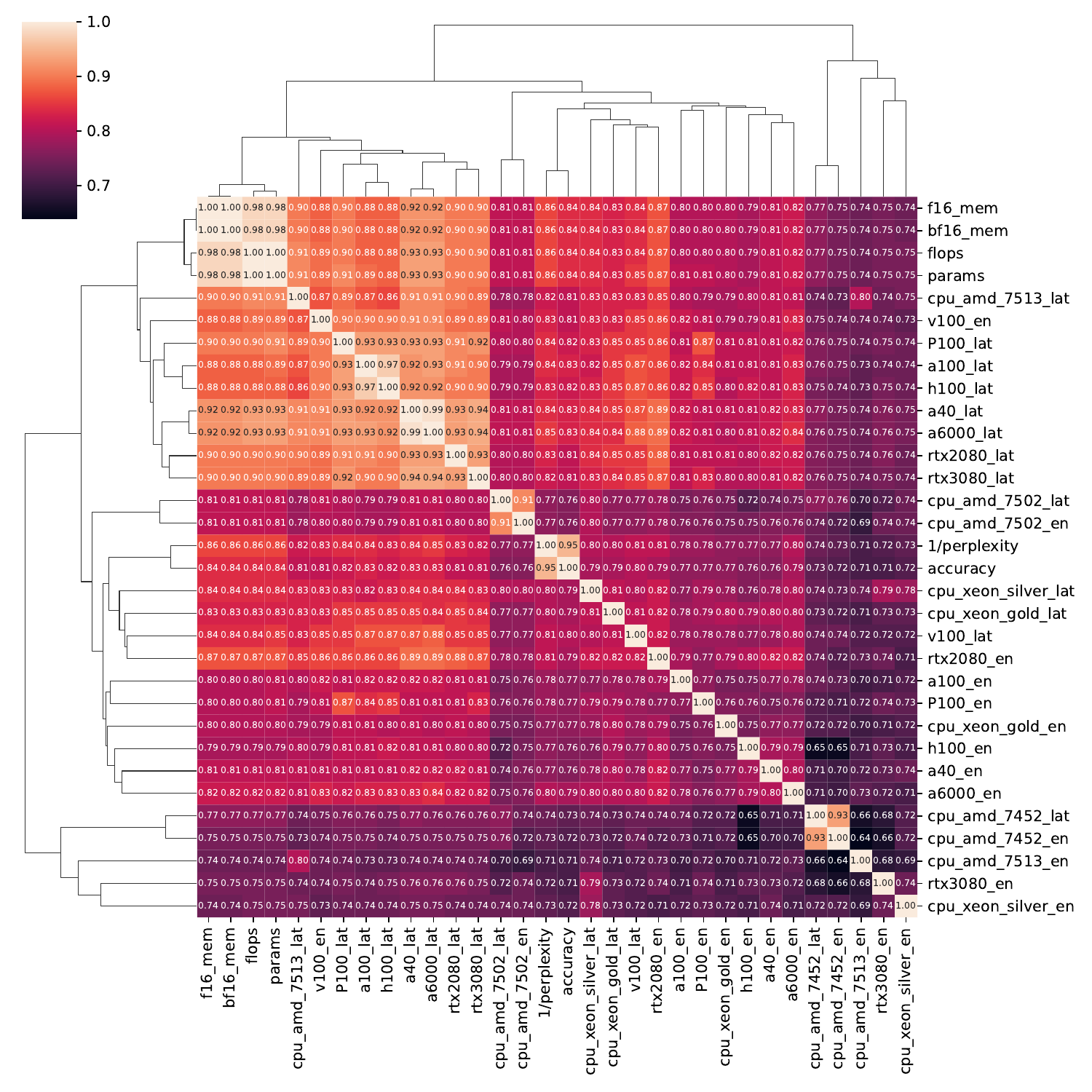}
\caption{Cross-metric Kendall-$\tau$ correlation plots for GPT-M.}
\label{fig:correlation_scales_m}
\vspace{-4mm}
\end{figure}

\begin{figure}[ht]
\centering
\includegraphics[width=\textwidth, height=14cm]{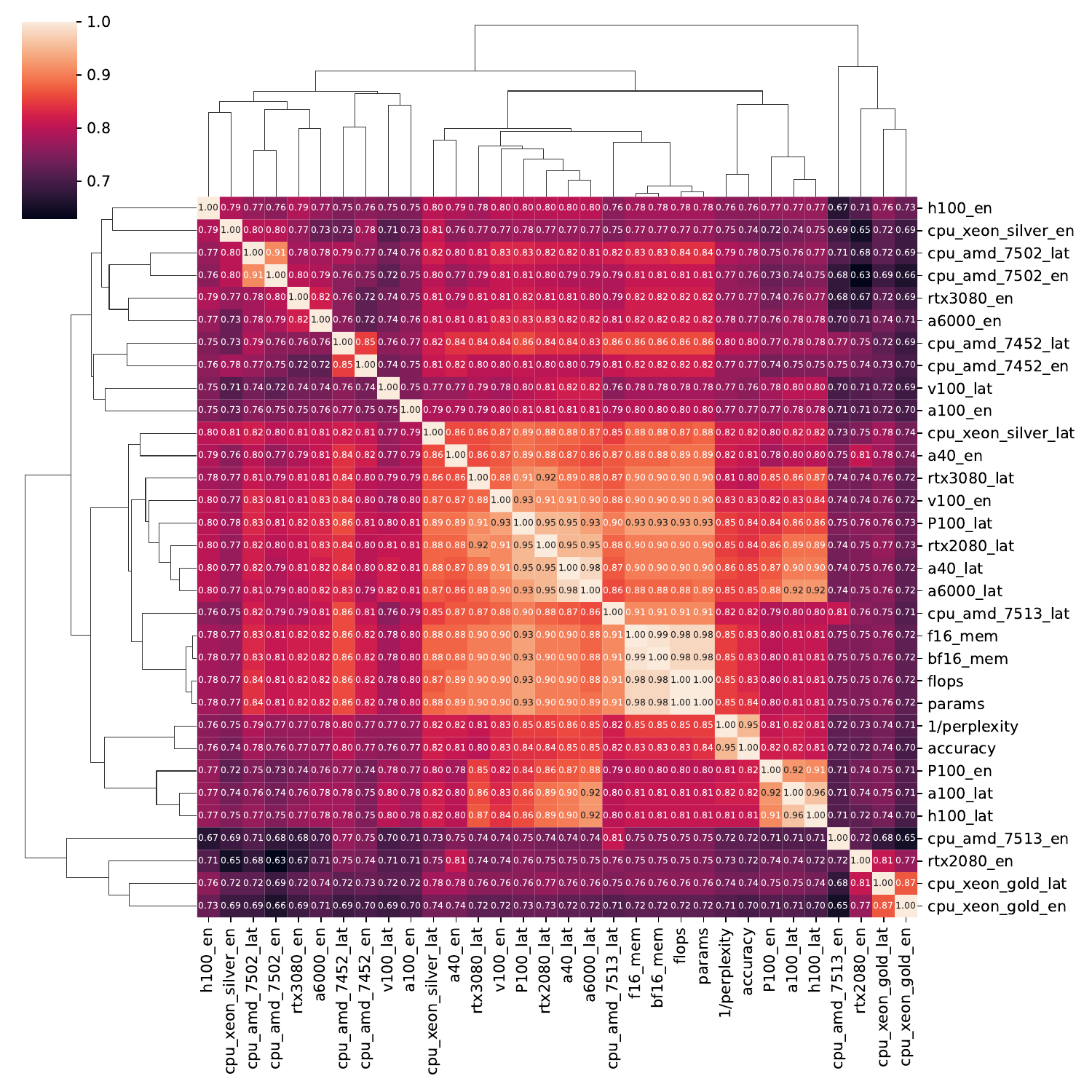}
\caption{Cross-metric Kendall-$\tau$ correlation plots for GPT-L.}
\label{fig:correlation_scales_l}
\vspace{-4mm}
\end{figure}

Figures \ref{fig:correlation_scales_s},\ref{fig:correlation_scales_m}, \ref{fig:correlation_scales_l} show the Kendall-$\tau$ rank correlation coefficient across all the metrics supported in HW-GPT-Bench. Given two sets of $n$ observations $\{y_i \}_{i=1}^n$ and $\{z_i \}_{i=1}^n$, the $\tau$ coefficient is computed as:
$$\tau = \frac{C - D}{\frac{1}{2} n (n-1)},$$
where $C$ and $D$ are the number of concordant and discordant pairs, respectively. For every pair \((i,j)\) where \(1 \leq i < j \leq n\):
\begin{enumerate}[leftmargin=*]
    \item A pair is \textbf{concordant} if the order of both elements in the pair is the same in both datasets: \((y_i < y_j \text{ and } z_i < z_j)\) or \((y_i > y_j \text{ and } z_i > z_j)\).
    \item A pair is \textbf{discordant} if the order of the elements in the pair is different in the two datasets: \((y_i < y_j \text{ and } z_i > z_j)\) or \((y_i > y_j \text{ and } z_i < z_j)\).
\end{enumerate}

\begin{figure}[ht]
\centering
\includegraphics[width=\textwidth, height=14cm]{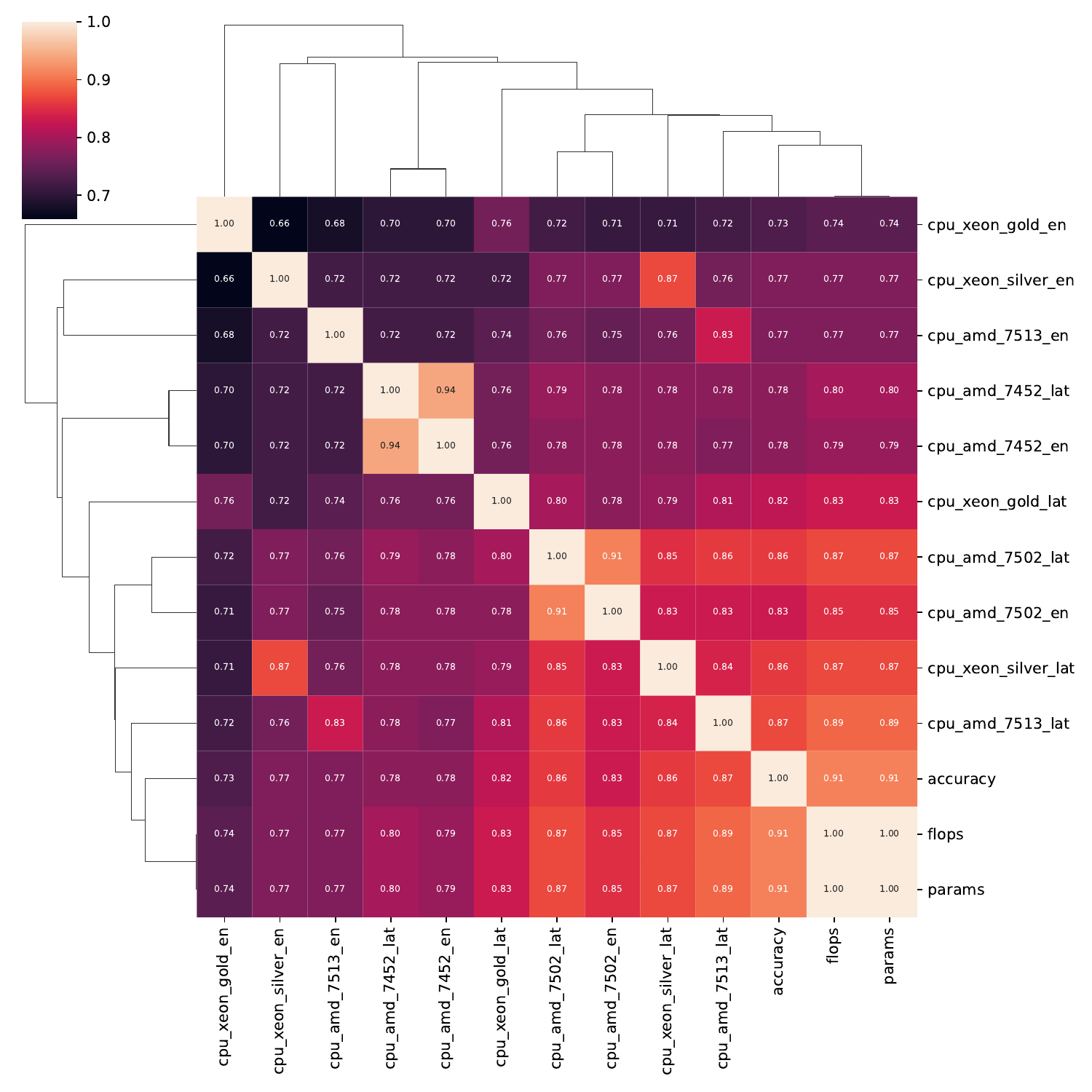}
\caption{Cross-metric Kendall-$\tau$ correlation plots for GPT-S (only CPU devices).}
\label{fig:correlation_scales_cpu_s}
\vspace{-4mm}
\end{figure}

\begin{figure}[ht]
\centering
\includegraphics[width=\textwidth, height=14cm]{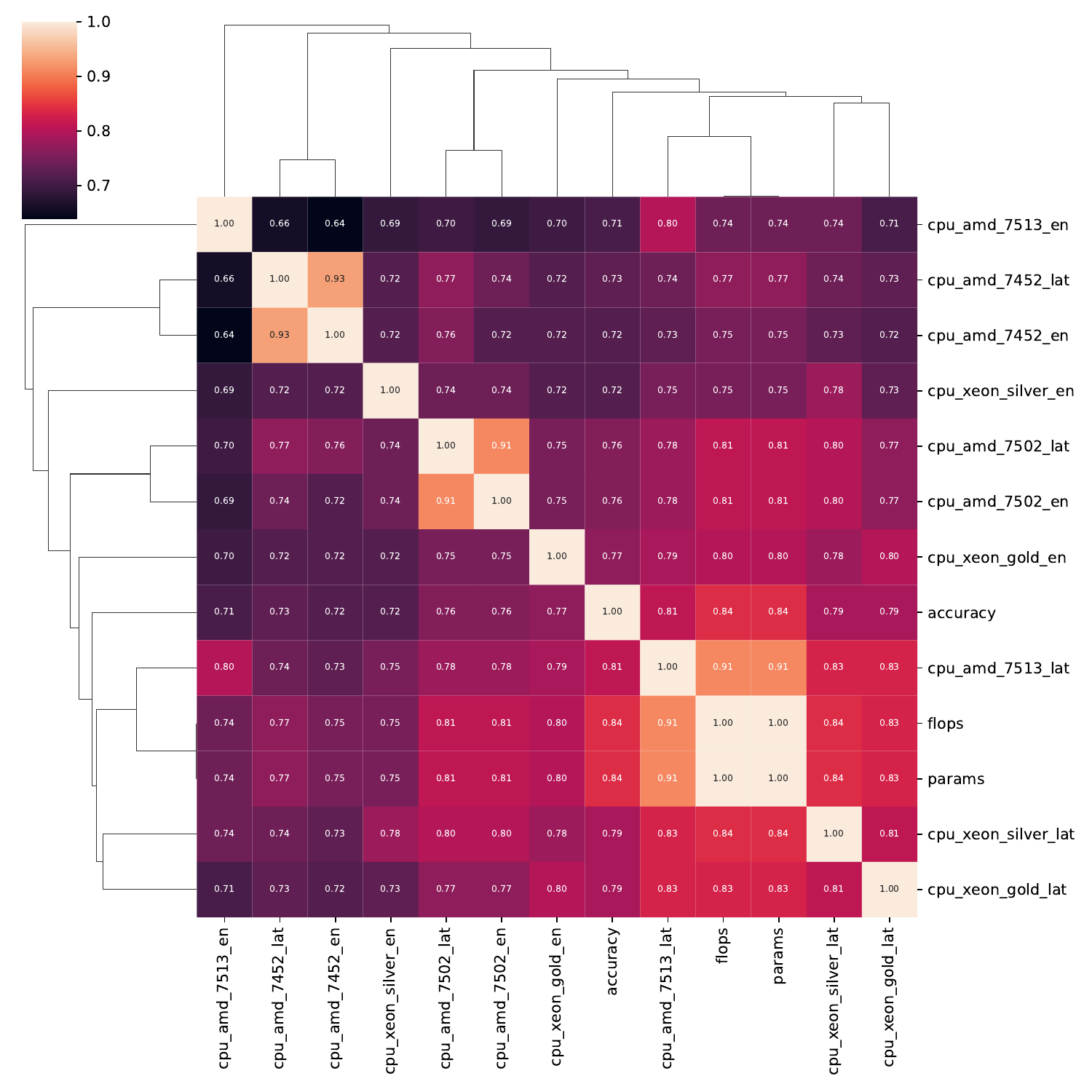}
\caption{Cross-metric Kendall-$\tau$ correlation plots for GPT-M (only CPU devices).}
\label{fig:correlation_scales_cpu_m}
\vspace{-4mm}
\end{figure}

\begin{figure}[ht]
\centering
\includegraphics[width=\textwidth, height=14cm]{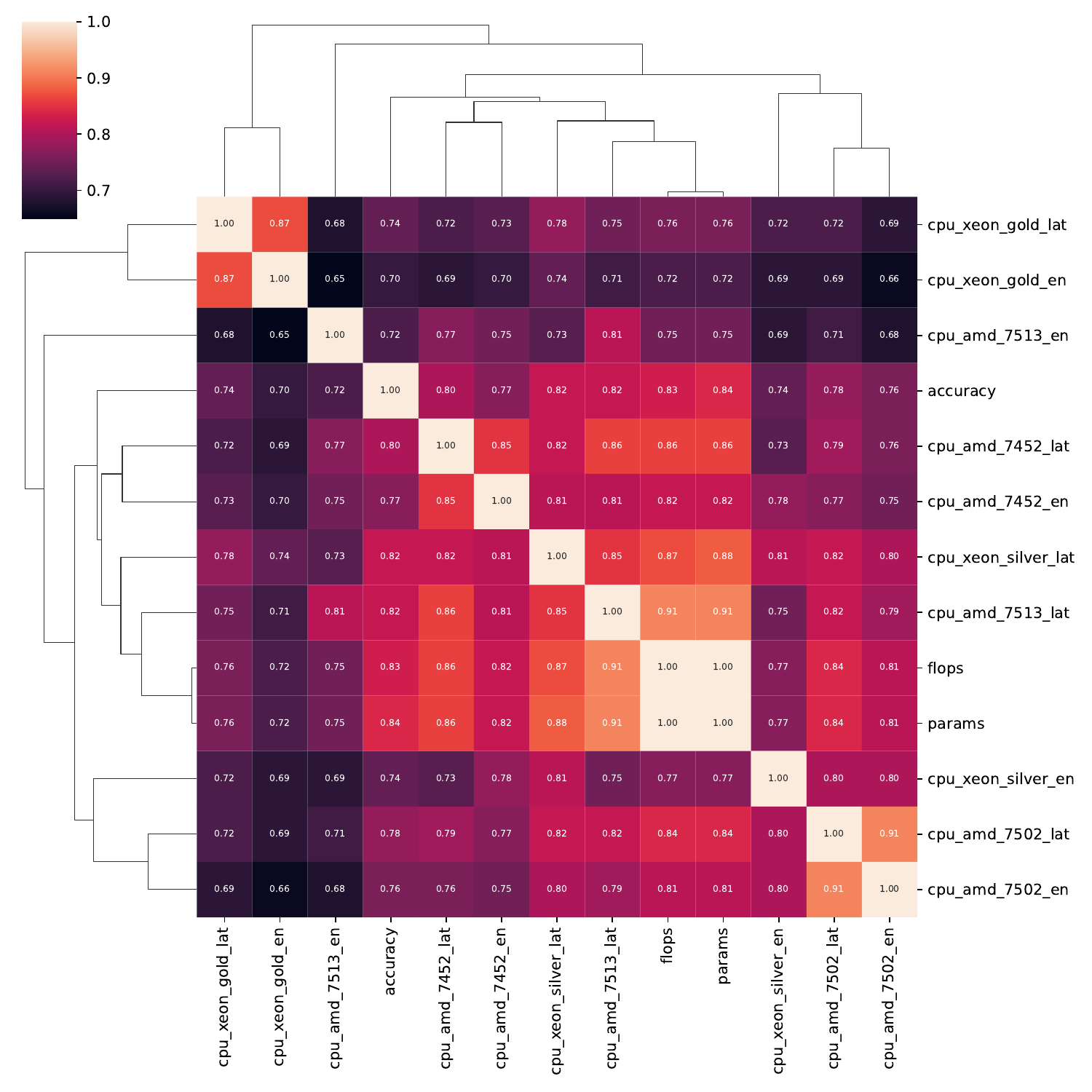}
\caption{Cross-metric Kendall-$\tau$ correlation plots for GPT-L (only CPU device).}
\label{fig:correlation_scales_cpu_l}
\vspace{-4mm}
\end{figure}

\begin{figure}[ht]
\centering
\includegraphics[width=\textwidth, height=14cm]{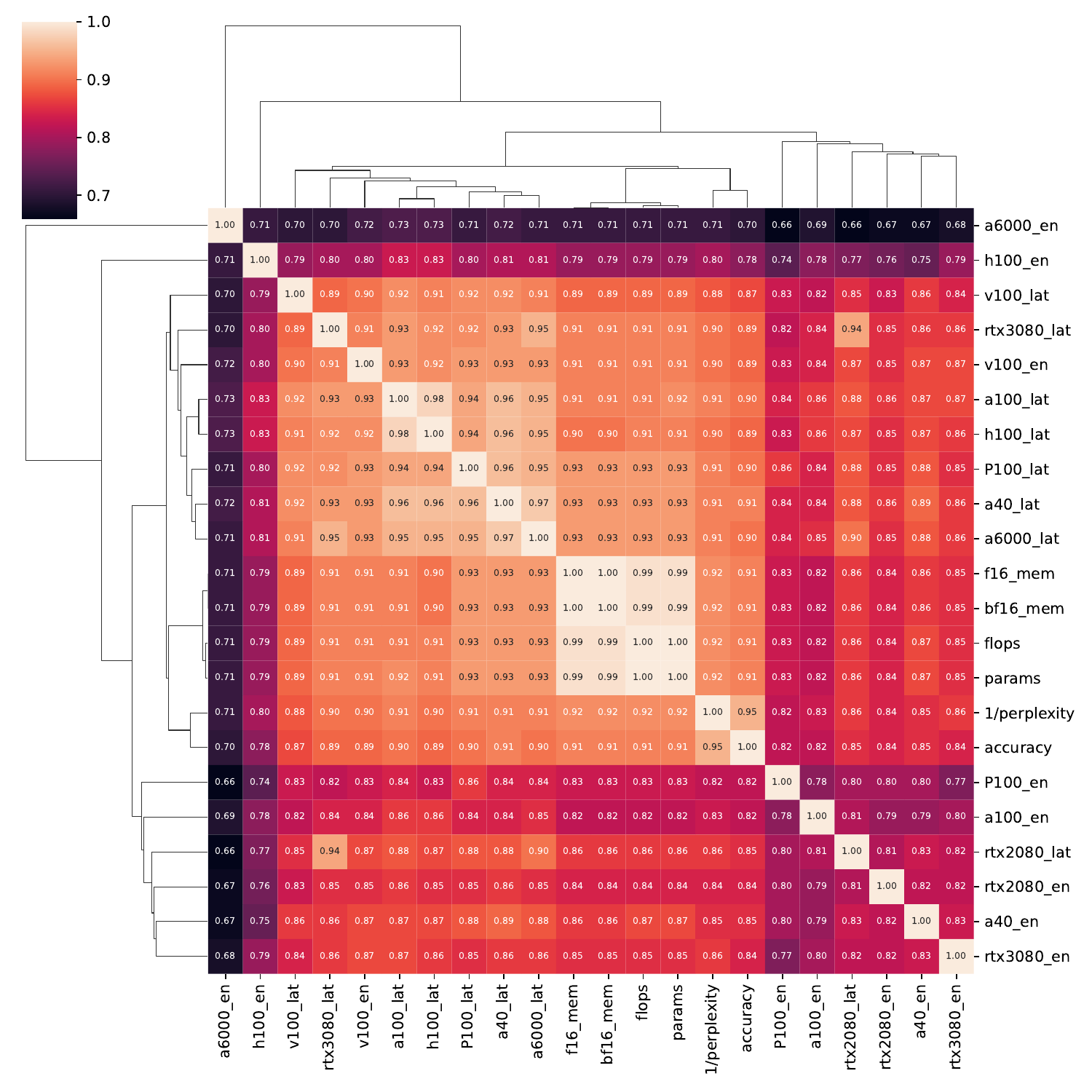}
\caption{Cross-metric Kendall-$\tau$ correlation plots for GPT-S (only GPU devices).}
\label{fig:correlation_scales_gpu_s}
\vspace{-4mm}
\end{figure}

\begin{figure}[ht]
\centering
\includegraphics[width=\textwidth, height=14cm]{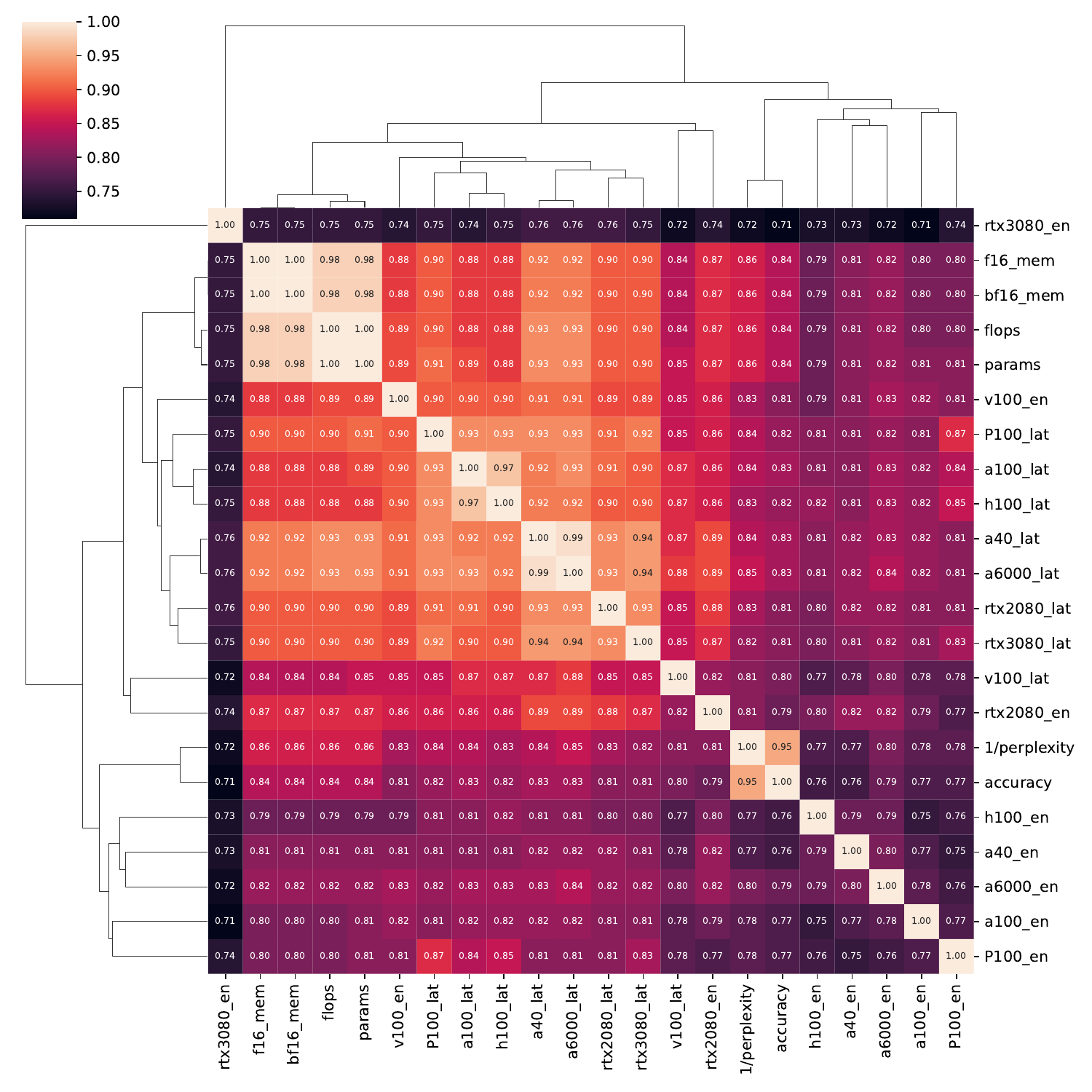}
\caption{Cross-metric Kendall-$\tau$ correlation plots for GPT-M (only GPU devices).}
\label{fig:correlation_scales_gpu_m}
\vspace{-4mm}
\end{figure}

\begin{figure}[ht]
\centering
\includegraphics[width=\textwidth, height=14cm]{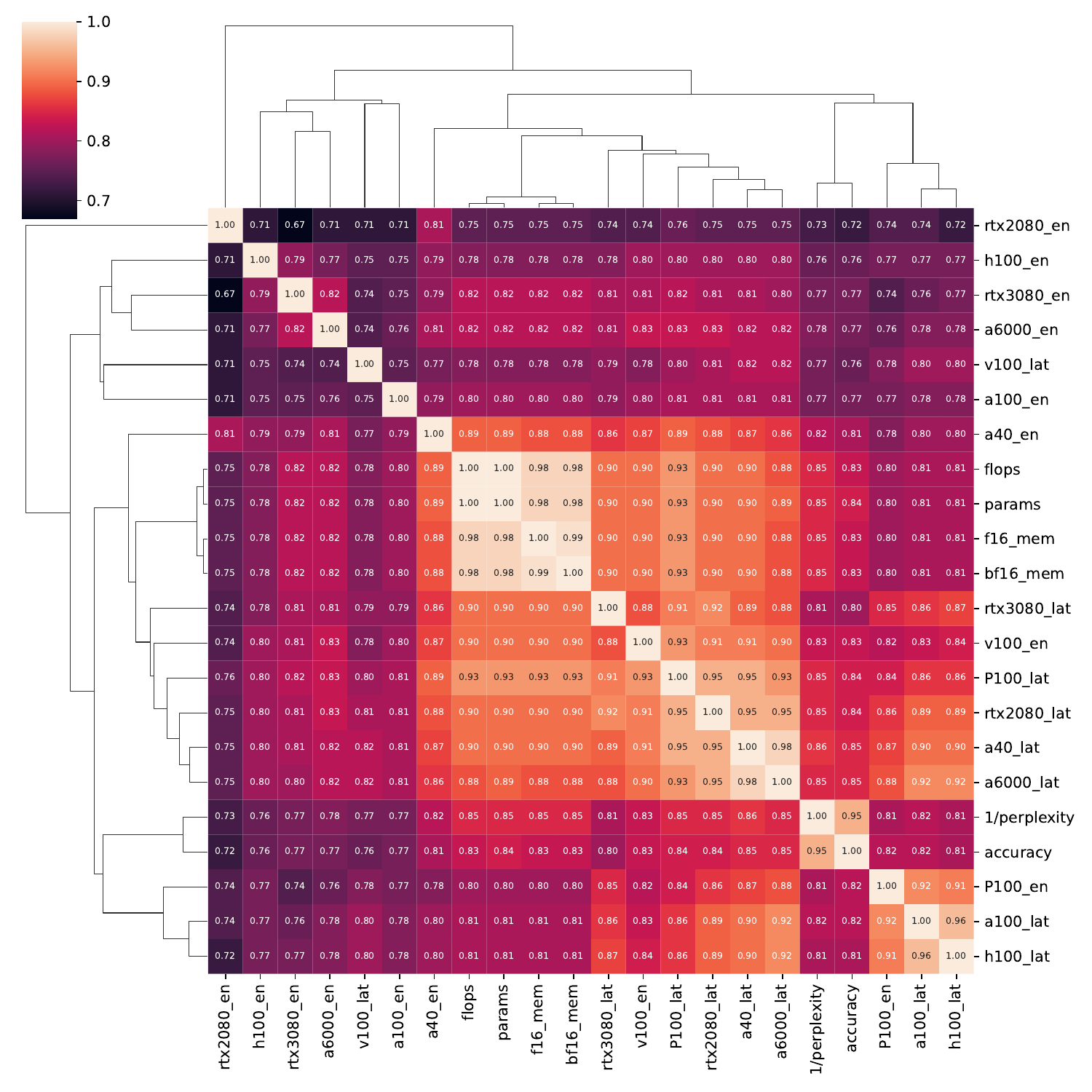}
\caption{Cross-metric Kendall-$\tau$ correlation plots for GPT-L (only GPU devices).}
\label{fig:correlation_scales_gpu_l}
\vspace{-4mm}
\end{figure}

To compute these values, we use the same 10k ($n=10000$) ground truth observations that we use to train the surrogate models. For metrics that contain multiple observations, such as latency and energy usage, we use the median value. For easier visualization, we stratify the aforementioned correlation plots by metrics relevant to GPUs (Figures \ref{fig:correlation_scales_gpu_s},\ref{fig:correlation_scales_gpu_m}, \ref{fig:correlation_scales_gpu_l}) and CPUs (Figures \ref{fig:correlation_scales_cpu_s},\ref{fig:correlation_scales_cpu_m}, \ref{fig:correlation_scales_cpu_l}).


\clearpage
\newpage
\section{Additional ECDF plots}
\begin{figure}[H]
    \centering
    \begin{subfigure}[b]{0.32\textwidth}
        \centering
        \includegraphics[width=\textwidth]{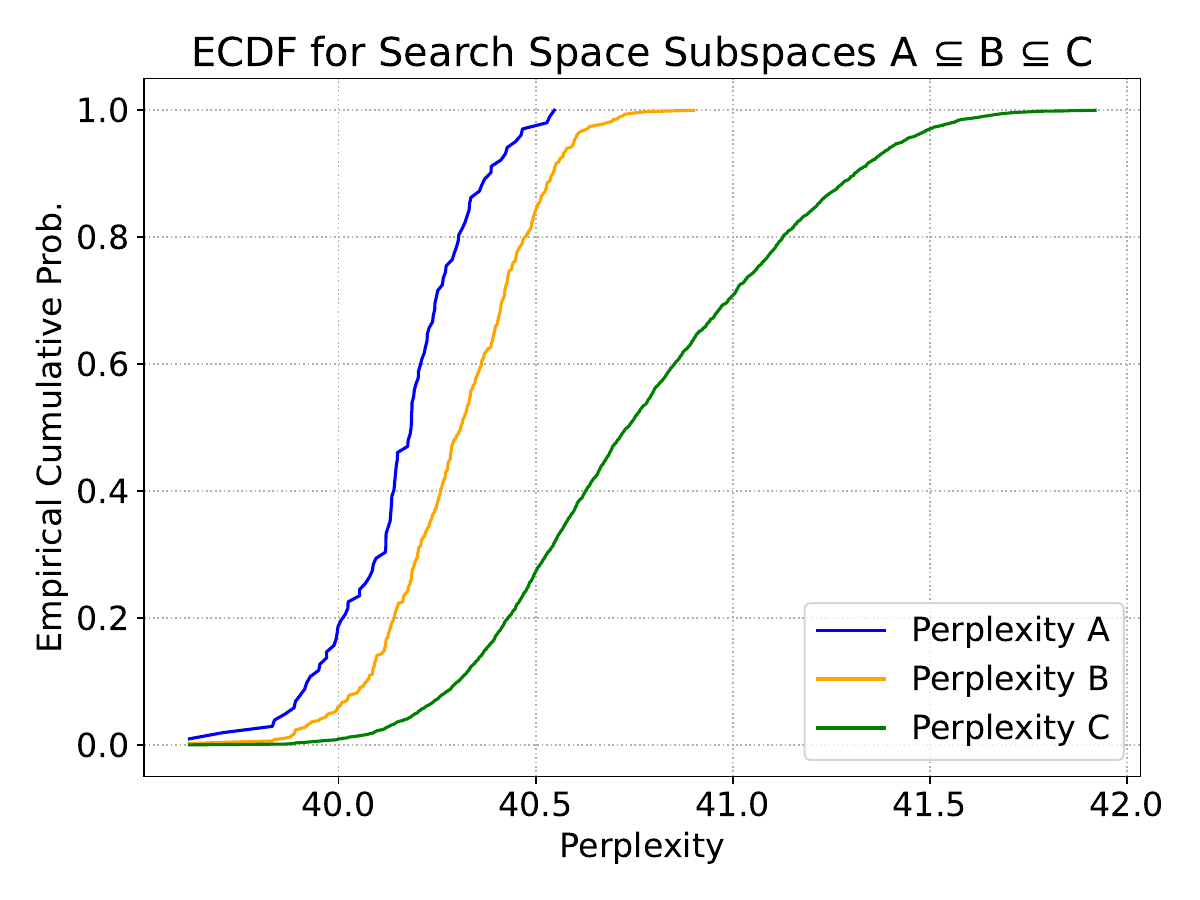}
        \caption{Embedding dimension 192}
        \label{fig:gpts192}
    \end{subfigure}
    \hfill
    \begin{subfigure}[b]{0.32\textwidth}
        \centering
        \includegraphics[width=\textwidth]{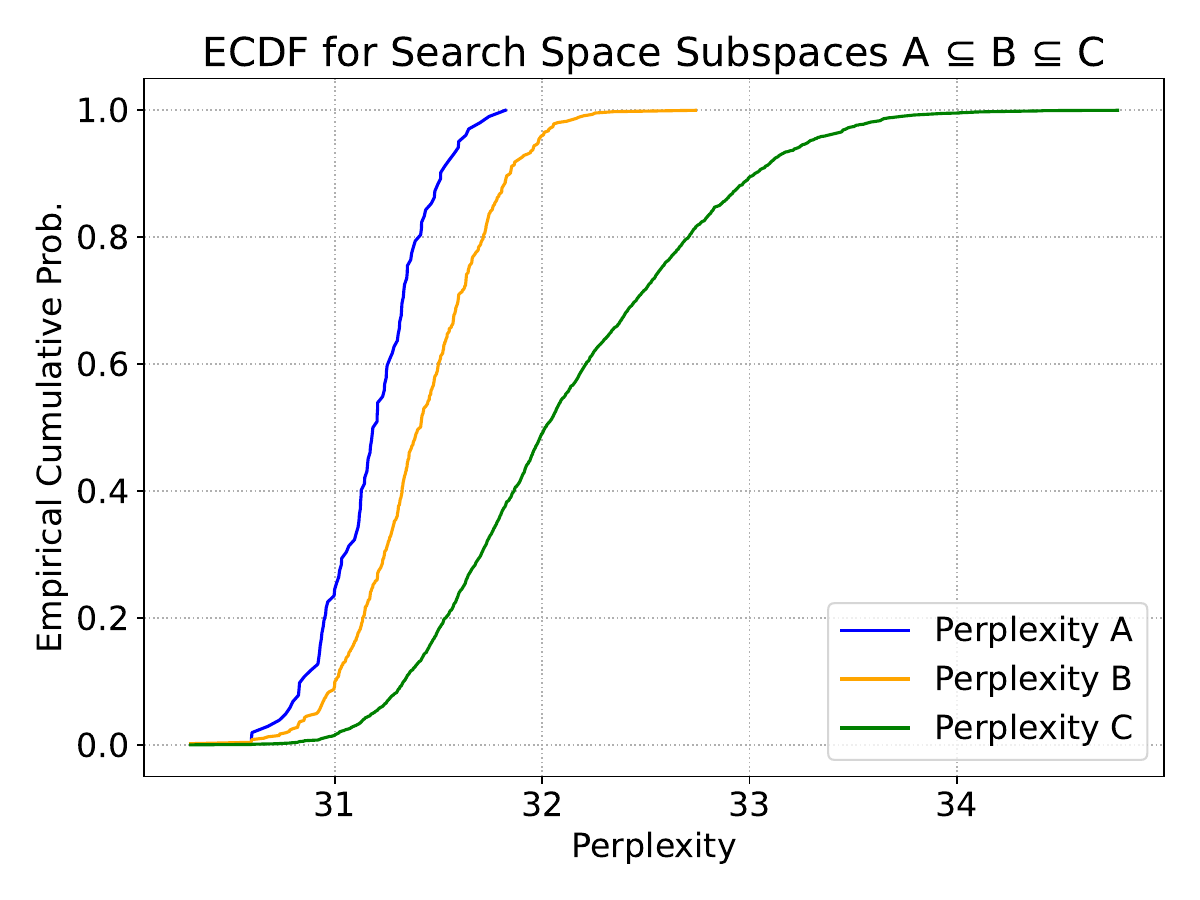}
        \caption{Embedding dimension 384}
        \label{fig:gpts384}
    \end{subfigure}
    \hfill
    \begin{subfigure}[b]{0.32\textwidth}
        \centering
        \includegraphics[width=\textwidth]{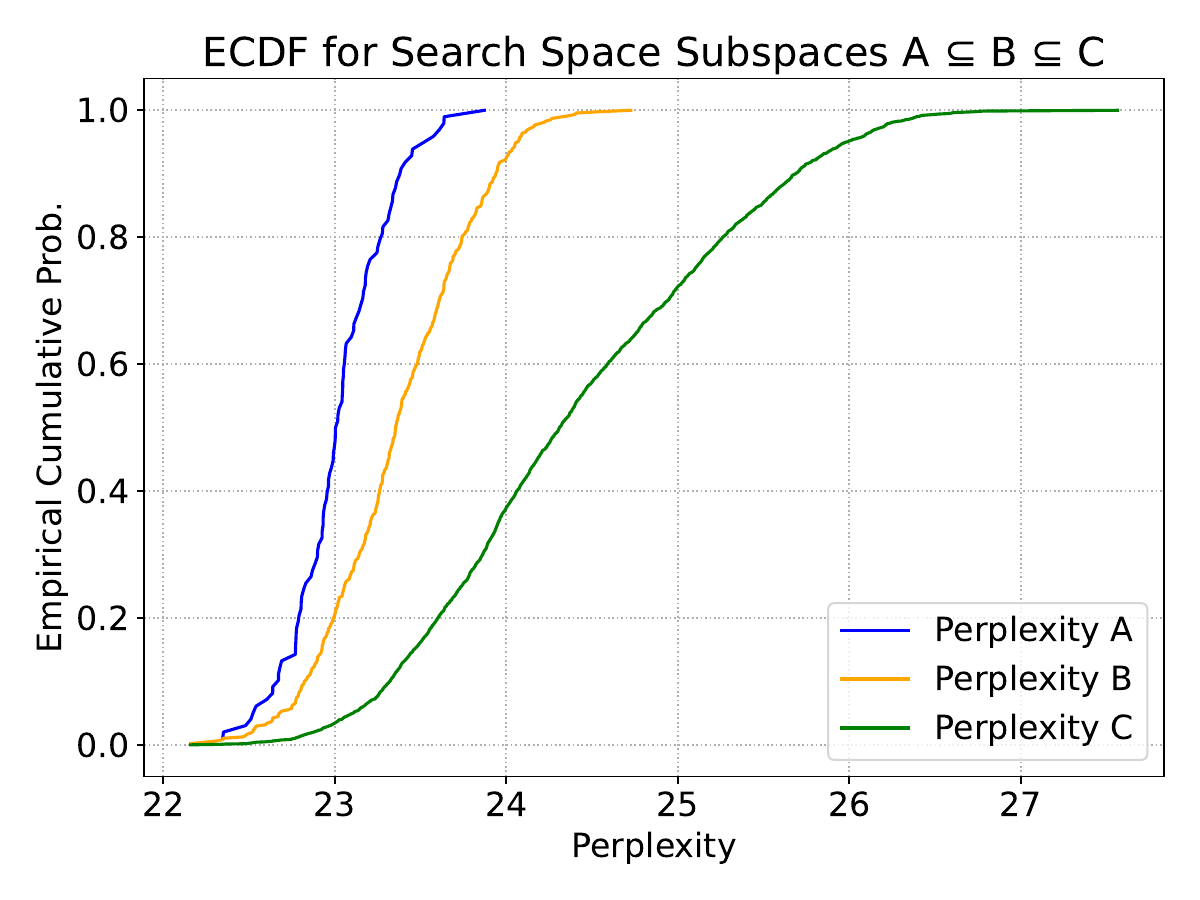}
        \caption{Embedding dimension 768}
        \label{fig:gpts768}
    \end{subfigure}
    \caption{ECDF plots for GPT-S.}
    \label{fig:gptsecdf}
\end{figure}

\begin{figure}[H]
    \centering
    \begin{subfigure}[b]{0.32\textwidth}
        \centering
        \includegraphics[width=\textwidth]{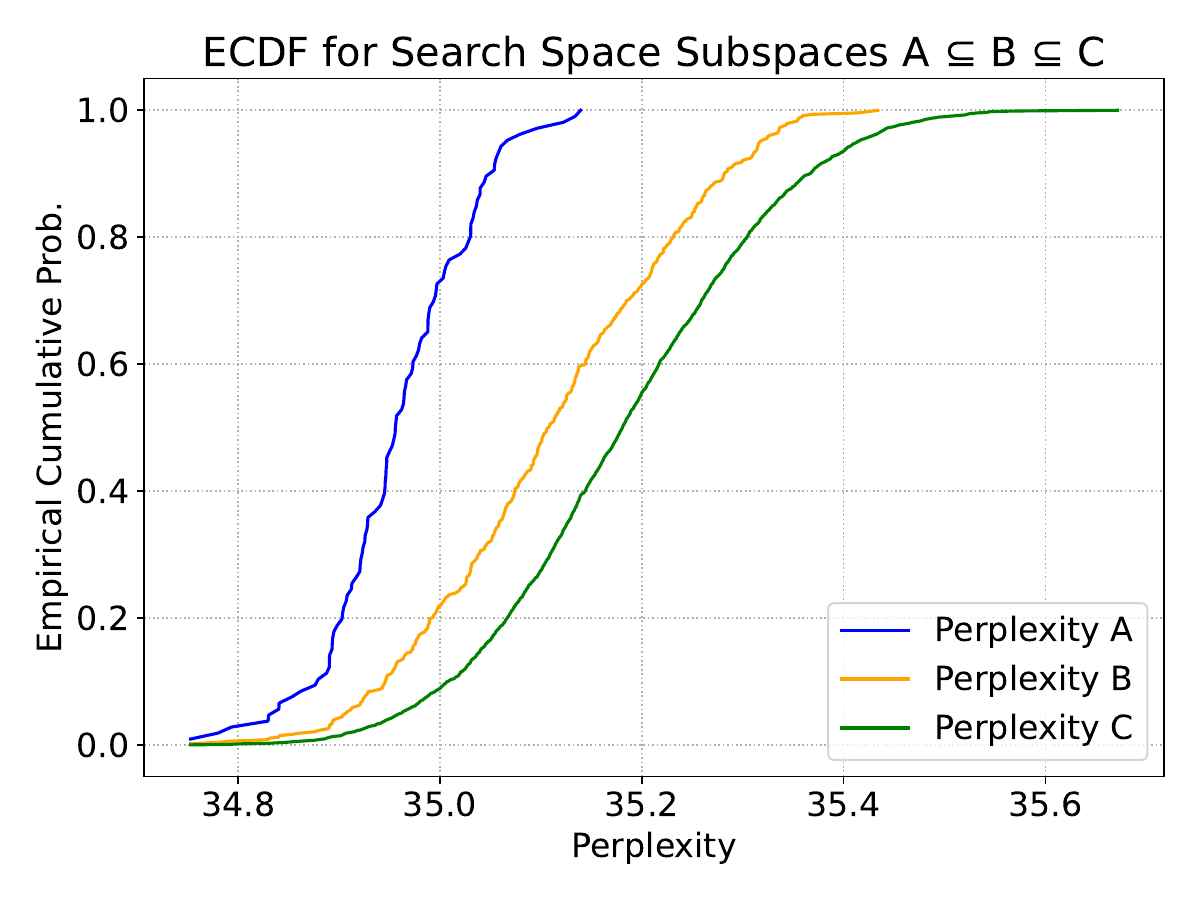}
        \caption{Embedding dimension 256}
        \label{fig:gptm256}
    \end{subfigure}
    \hfill
    \begin{subfigure}[b]{0.32\textwidth}
        \centering
        \includegraphics[width=\textwidth]{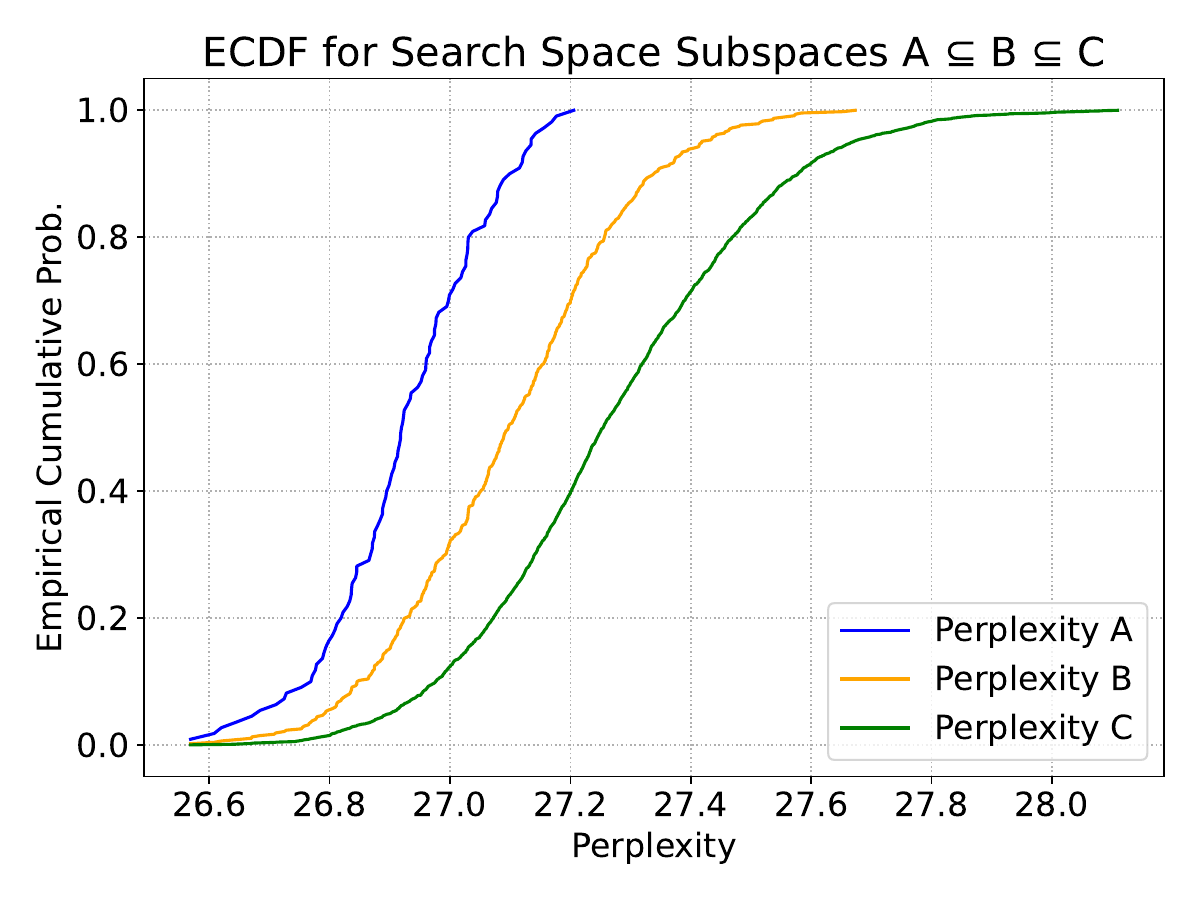}
        \caption{Embedding dimension 512}
        \label{fig:gptm512}
    \end{subfigure}
    \hfill
    \begin{subfigure}[b]{0.32\textwidth}
        \centering
        \includegraphics[width=\textwidth]{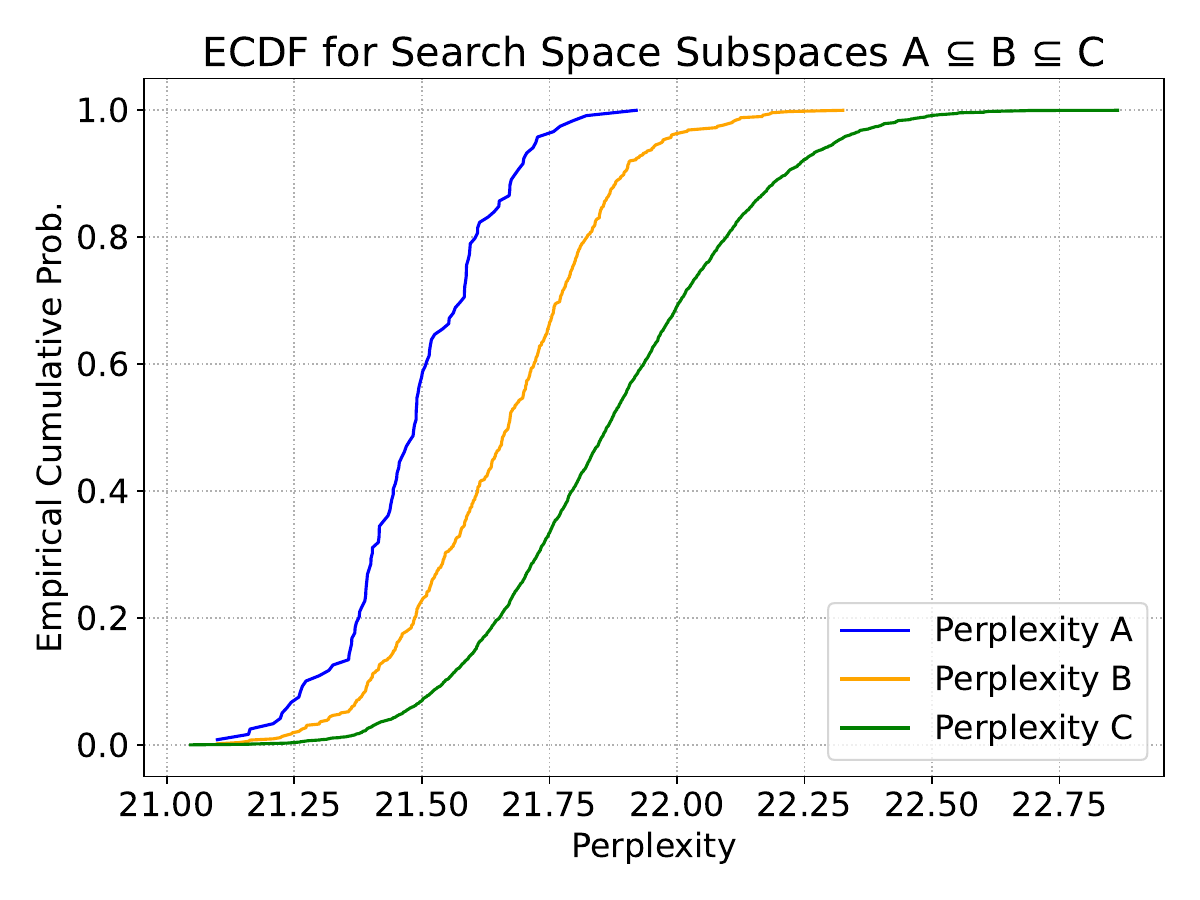}
        \caption{Embedding dimension 1024}
        \label{fig:gptm1024}
    \end{subfigure}
    \caption{ECDF plots for GPT-M.}
    \label{fig:gptmecdf}
\end{figure}

\begin{figure}[H]
    \centering
    \begin{subfigure}[b]{0.32\textwidth}
        \centering
        \includegraphics[width=\textwidth]{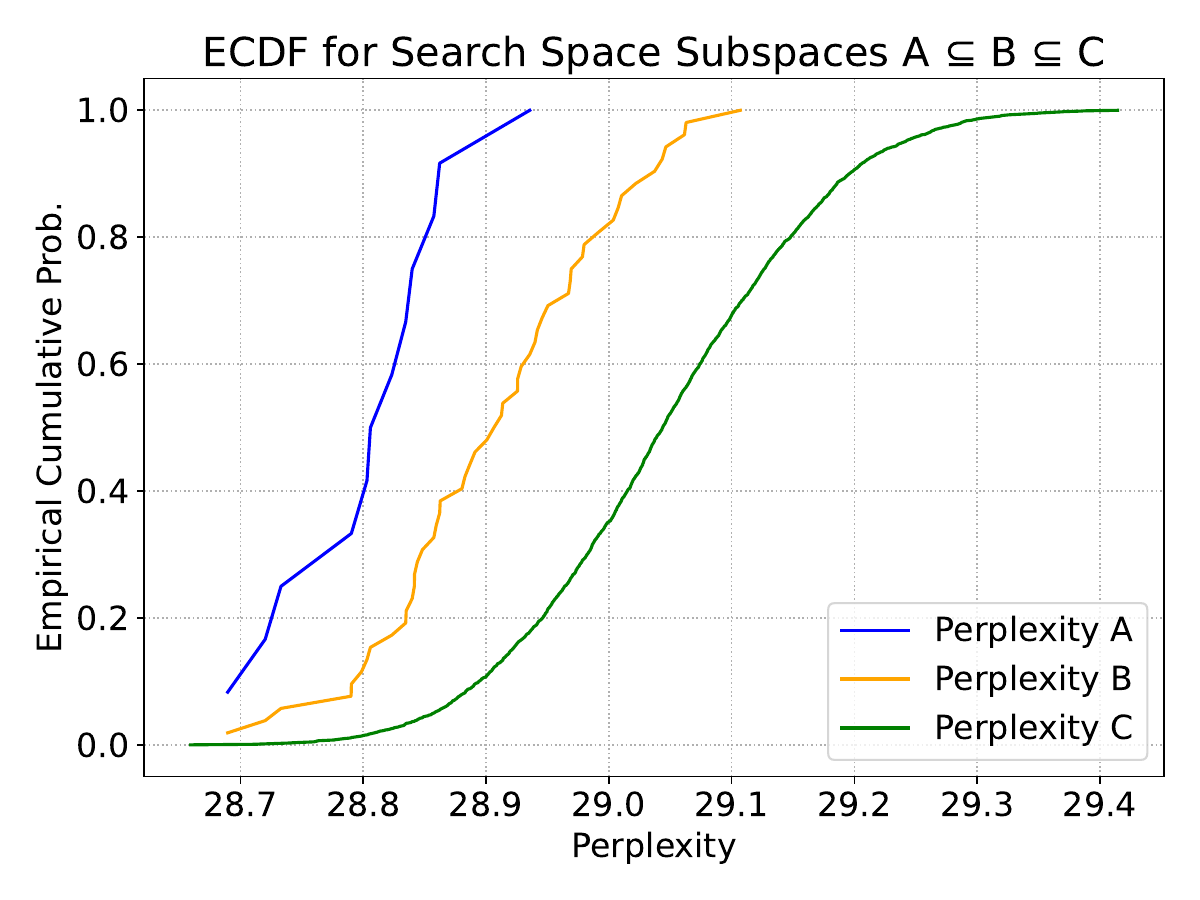}
        \caption{Embedding dimension 320}
        \label{fig:gptl320}
    \end{subfigure}
    \hfill
    \begin{subfigure}[b]{0.32\textwidth}
        \centering
        \includegraphics[width=\textwidth]{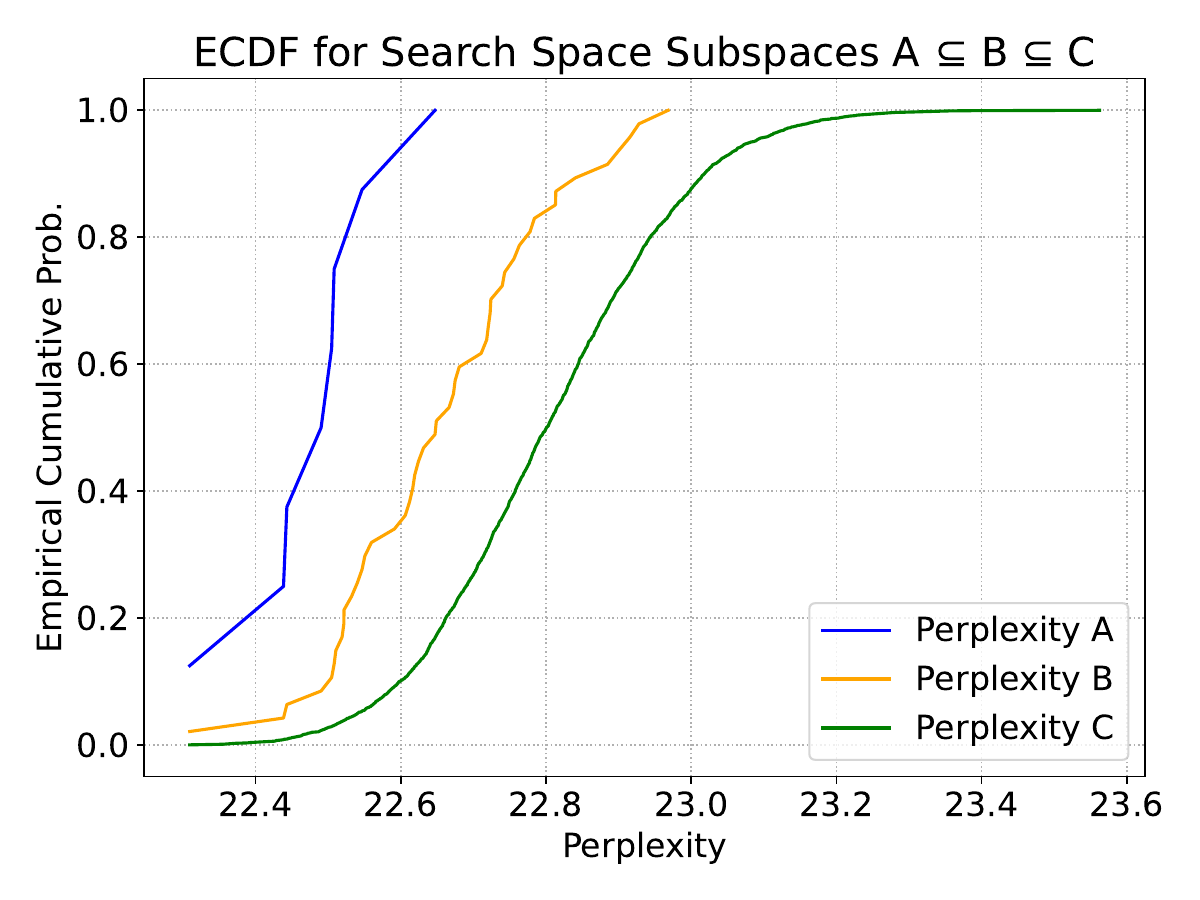}
        \caption{Embedding dimension 640}
        \label{fig:gptl640}
    \end{subfigure}
    \hfill
    \begin{subfigure}[b]{0.32\textwidth}
        \centering
        \includegraphics[width=\textwidth]{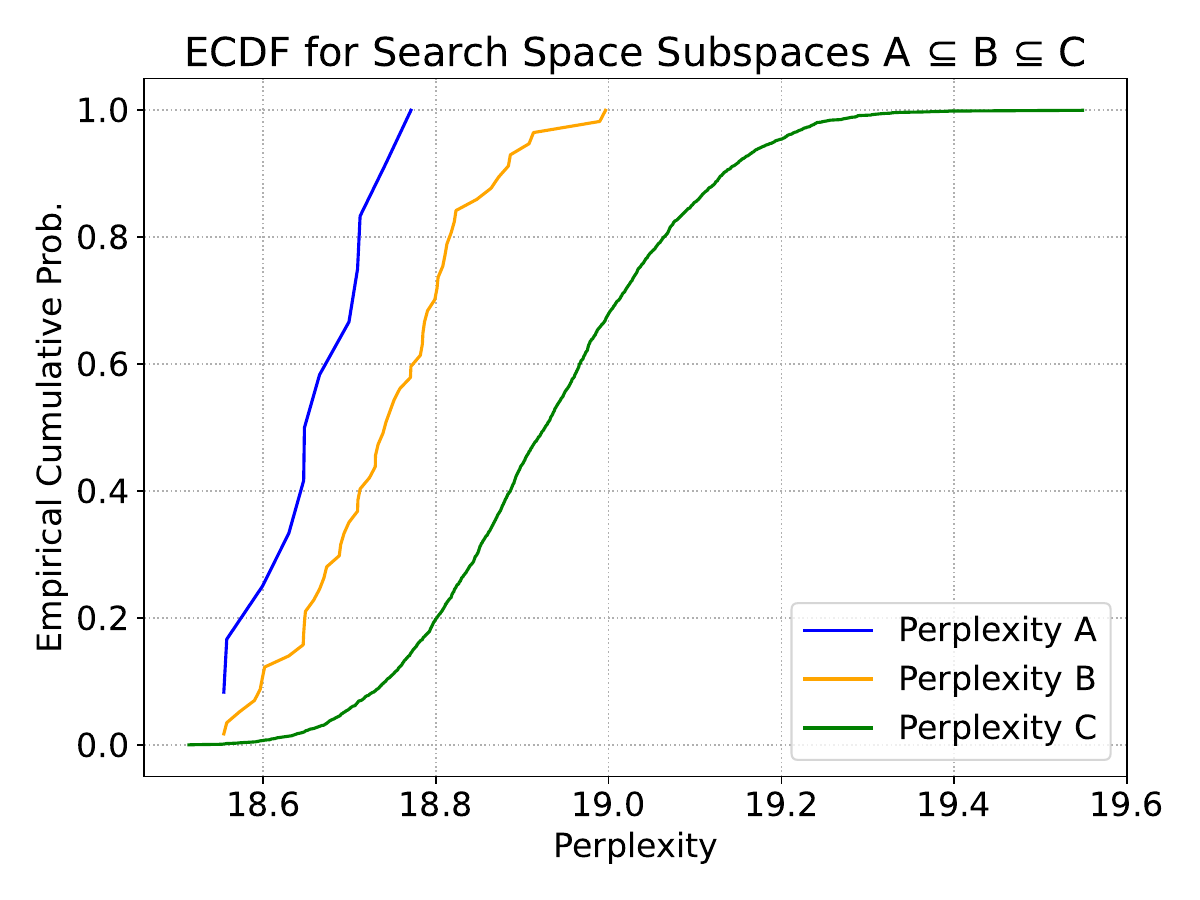}
        \caption{Embedding dimension 1280}
        \label{fig:gptl1280}
    \end{subfigure}
    \caption{ECDF plots for GPT-L.}
    \label{fig:gptlecdf}
\end{figure}

In this section, we present the ECDF plots of perplexity on different search space scales, computed using the 10k ground truth observations from the supernetwork. The largest set, $C$, contains all architectures for a fixed embedding dimension size. $B$, a subset of $C$, contains all architectures that, in addition to the fixed embedding dimension $e$, have the number of layers set to the largest possible value $l=l_3$, namely $12,\ 24, \text{ and } 36 $ for GPT-S, -M and -L, respectively. $A$, a subset of $B$, contains all architectures that, in addition to the number of layers set to largest possible values, have the average MLP ratio and number of heads greater than a fixed threshold. We show results for all 3 Transformer scales: GPT-S, -M and -L, in Figures~\ref{fig:gptsecdf}, \ref{fig:gptmecdf} and \ref{fig:gptlecdf}, respectively.
\vspace{-10mm}
\vspace{-10mm}



\section{HW-GPT-Bench API Examples}
\label{sec:moreexamples}
\vspace{4ex}

\begin{figure}[H]
\vspace{-5ex}
\centering
\begin{lstlisting}[caption={Hardware agnostic metric using the HW-GPT-Bench API.},label=fig:query-flops-params-memory,language=Python,numbers=none,basicstyle=\ttfamily\tiny]
from hwgpt.api import HWGPT 
api = HWGPT(search_space="m") # initialize API for GPT-M
random_arch = api.sample_arch() # sample random arch
api.set_arch(random_arch) # set  arch
flops = api.query(metric="flops") # query flops for the architecture
params = api.query(metric="params") # query params for the architecture
float16_memory = api.query(metric="f16mem") # query float16 memory for the architecture
bfloat16_memory = api.query(metric="bf16mem") # query bfloat16 memory for the architecture
\end{lstlisting}
\end{figure}

\vspace{4ex}
\begin{figure}[H]
\vspace{-8ex}
\centering
\begin{lstlisting}[caption={Ground truth observations using the HW-GPT-Bench API.},label=fig:query-gt,language=Python,numbers=none,basicstyle=\ttfamily\tiny]
from hwgpt.api import HWGPT 
api = HWGPT(search_space="m") # initialize API for GPT-M
random_arch = api.sample_arch_gt() # sample random arch from amongst the 10k ground truth archs
api.set_arch(random_arch) # set  arch
results = api.query(gt=True) # get all ground truth results for the architecture
energy = api.query(metric="energy",gt=True)  # get ground truth energy observations for all architectures
rtx2080 = api.query(device="rtx2080",gt=True) # get all hw metrics for rtx2080 device
\end{lstlisting}
\end{figure}

\vspace{4ex}
\begin{figure}[H]
\vspace{-8ex}
\centering
\begin{lstlisting}[caption={Running MOO with 3 objectives using the HW-GPT-Bench API.},label=fig:moo-3d,language=Python,numbers=none,basicstyle=\ttfamily\tiny]
from hwgpt.api import HWGPT 
api = HWGPT(search_space="m") # initialize API for GPT-M
nsga2_results_2d = api.run_baseline(method="nsga2", device="h100", metrics=["energy","perplexity"],ppl_predictor="mlp") # nsga-2d 
nsga2_results_3d = api.run_baseline_3d(method="nsga2", device="h100", metrics=["energy","perplexity","latency"],ppl_predictor="mlp") #nsga-3d 
\end{lstlisting}
\end{figure}

\section{HW-GPT-Bench Release and Maintainance}
HW-GPT-Bench will be distributed under the Apache 2.0 License, tailored explicitly for academic research. The Apache 2.0 License is chosen for its permissive characteristics within the open-source community, permitting users to freely utilize, modify, and distribute the software under the condition of proper attribution and adherence to Apache 2.0 stipulations. This licensing strategy is pivotal in fostering broad adoption of HW-GPT-Bench.

In addition to its release, we are committed to fostering community engagement with the benchmark. We will actively monitor and respond to inquiries, issues, and suggestions related to HW-GPT-Bench, thereby cultivating a collaborative environment conducive to ongoing improvement and innovation.

Looking forward, our development roadmap includes plans to expand HW-GPT-Bench to encompass a wider range of devices and language model spaces. This expansion aims to bolster the benchmark's utility and relevance, accommodating emerging research demands and technological advancements in the field

\section{Limitations and Future Work}
While our work is the first one to efficiently benchmark different decoder-only architectures on a variety of gpu and cpu devices, there are several possible enhancements possible, which we leave to future work. Firstly, currently the benchmark is limited to decoder only models and we believe it would be interesting to extend to encoder-decoder models and state-space models. Secondly, currently the benchmark trains supernetworks from scratch and scaling to very large models (eg: Llama 3.1 405b), would require expensive retraining. Initializing from pretrained models  and exploring parameter-efficient finetuning methods for supernet finetuning, is important to avoid retraining and make most efficient use of available compute. Furthermore, since the benchmark is developed primarily in an academic setting, we couldn’t profile the architectures on edge devices and specifically edge devices which are optimized for LLM inference. However, provide a plug and play framework by releasing all our profiling scripts and hope for community contributions to enhance the benchmark for newer hardware devices.
\newpage
\newpage
\section*{NeurIPS Paper Checklist}

\begin{enumerate}

\item {\bf Claims}
    \item[] Question: Do the main claims made in the abstract and introduction accurately reflect the paper's contributions and scope?
    \item[] Answer: \answerYes{} 
    \item[] Justification: The claims made in the abstract and intro accurately reflect the contributions and scope of our paper
    \item[] Guidelines:
    \begin{itemize}
        \item The answer NA means that the abstract and introduction do not include the claims made in the paper.
        \item The abstract and/or introduction should clearly state the claims made, including the contributions made in the paper and important assumptions and limitations. A No or NA answer to this question will not be perceived well by the reviewers. 
        \item The claims made should match theoretical and experimental results, and reflect how much the results can be expected to generalize to other settings. 
        \item It is fine to include aspirational goals as motivation as long as it is clear that these goals are not attained by the paper. 
    \end{itemize}

\item {\bf Limitations}
    \item[] Question: Does the paper discuss the limitations of the work performed by the authors?
    \item[] Answer: \answerYes{} 
    \item[] Justification: We discuss the potential limitations and braoder impact of our work. 
    \item[] Guidelines:
    \begin{itemize}
        \item The answer NA means that the paper has no limitation while the answer No means that the paper has limitations, but those are not discussed in the paper. 
        \item The authors are encouraged to create a separate "Limitations" section in their paper.
        \item The paper should point out any strong assumptions and how robust the results are to violations of these assumptions (e.g., independence assumptions, noiseless settings, model well-specification, asymptotic approximations only holding locally). The authors should reflect on how these assumptions might be violated in practice and what the implications would be.
        \item The authors should reflect on the scope of the claims made, e.g., if the approach was only tested on a few datasets or with a few runs. In general, empirical results often depend on implicit assumptions, which should be articulated.
        \item The authors should reflect on the factors that influence the performance of the approach. For example, a facial recognition algorithm may perform poorly when image resolution is low or images are taken in low lighting. Or a speech-to-text system might not be used reliably to provide closed captions for online lectures because it fails to handle technical jargon.
        \item The authors should discuss the computational efficiency of the proposed algorithms and how they scale with dataset size.
        \item If applicable, the authors should discuss possible limitations of their approach to address problems of privacy and fairness.
        \item While the authors might fear that complete honesty about limitations might be used by reviewers as grounds for rejection, a worse outcome might be that reviewers discover limitations that aren't acknowledged in the paper. The authors should use their best judgment and recognize that individual actions in favor of transparency play an important role in developing norms that preserve the integrity of the community. Reviewers will be specifically instructed to not penalize honesty concerning limitations.
    \end{itemize}

\item {\bf Theory Assumptions and Proofs}
    \item[] Question: For each theoretical result, does the paper provide the full set of assumptions and a complete (and correct) proof?
    \item[] Answer: \answerNA{} 
    \item[] Justification: We do not present any theoritical results in our paper. 
    \item[] Guidelines:
    \begin{itemize}
        \item The answer NA means that the paper does not include theoretical results. 
        \item All the theorems, formulas, and proofs in the paper should be numbered and cross-referenced.
        \item All assumptions should be clearly stated or referenced in the statement of any theorems.
        \item The proofs can either appear in the main paper or the supplemental material, but if they appear in the supplemental material, the authors are encouraged to provide a short proof sketch to provide intuition. 
        \item Inversely, any informal proof provided in the core of the paper should be complemented by formal proofs provided in appendix or supplemental material.
        \item Theorems and Lemmas that the proof relies upon should be properly referenced. 
    \end{itemize}

    \item {\bf Experimental Result Reproducibility}
    \item[] Question: Does the paper fully disclose all the information needed to reproduce the main experimental results of the paper to the extent that it affects the main claims and/or conclusions of the paper (regardless of whether the code and data are provided or not)?
    \item[] Answer: \answerYes{}{} 
    \item[] Justification: We release the hyperparameters used in our paper, our code, raw datasets, pretrained models and a open-source api to ensure our benchmark results are reproducible. 
    \item[] Guidelines:
    \begin{itemize}
        \item The answer NA means that the paper does not include experiments.
        \item If the paper includes experiments, a No answer to this question will not be perceived well by the reviewers: Making the paper reproducible is important, regardless of whether the code and data are provided or not.
        \item If the contribution is a dataset and/or model, the authors should describe the steps taken to make their results reproducible or verifiable. 
        \item Depending on the contribution, reproducibility can be accomplished in various ways. For example, if the contribution is a novel architecture, describing the architecture fully might suffice, or if the contribution is a specific model and empirical evaluation, it may be necessary to either make it possible for others to replicate the model with the same dataset, or provide access to the model. In general. releasing code and data is often one good way to accomplish this, but reproducibility can also be provided via detailed instructions for how to replicate the results, access to a hosted model (e.g., in the case of a large language model), releasing of a model checkpoint, or other means that are appropriate to the research performed.
        \item While NeurIPS does not require releasing code, the conference does require all submissions to provide some reasonable avenue for reproducibility, which may depend on the nature of the contribution. For example
        \begin{enumerate}
            \item If the contribution is primarily a new algorithm, the paper should make it clear how to reproduce that algorithm.
            \item If the contribution is primarily a new model architecture, the paper should describe the architecture clearly and fully.
            \item If the contribution is a new model (e.g., a large language model), then there should either be a way to access this model for reproducing the results or a way to reproduce the model (e.g., with an open-source dataset or instructions for how to construct the dataset).
            \item We recognize that reproducibility may be tricky in some cases, in which case authors are welcome to describe the particular way they provide for reproducibility. In the case of closed-source models, it may be that access to the model is limited in some way (e.g., to registered users), but it should be possible for other researchers to have some path to reproducing or verifying the results.
        \end{enumerate}
    \end{itemize}

\item {\bf Open access to data and code}
    \item[] Question: Does the paper provide open access to the data and code, with sufficient instructions to faithfully reproduce the main experimental results, as described in supplemental material?
    \item[] Answer: \answerYes{} 
    \item[] Justification: We release our code, the hyperparameters used and pretrained model checkpoints, raw result files etc.
    \item[] Guidelines:
    \begin{itemize}
        \item The answer NA means that paper does not include experiments requiring code.
        \item Please see the NeurIPS code and data submission guidelines (\url{https://nips.cc/public/guides/CodeSubmissionPolicy}) for more details.
        \item While we encourage the release of code and data, we understand that this might not be possible, so “No” is an acceptable answer. Papers cannot be rejected simply for not including code, unless this is central to the contribution (e.g., for a new open-source benchmark).
        \item The instructions should contain the exact command and environment needed to run to reproduce the results. See the NeurIPS code and data submission guidelines (\url{https://nips.cc/public/guides/CodeSubmissionPolicy}) for more details.
        \item The authors should provide instructions on data access and preparation, including how to access the raw data, preprocessed data, intermediate data, and generated data, etc.
        \item The authors should provide scripts to reproduce all experimental results for the new proposed method and baselines. If only a subset of experiments are reproducible, they should state which ones are omitted from the script and why.
        \item At submission time, to preserve anonymity, the authors should release anonymized versions (if applicable).
        \item Providing as much information as possible in supplemental material (appended to the paper) is recommended, but including URLs to data and code is permitted.
    \end{itemize}

\item {\bf Experimental Setting/Details}
    \item[] Question: Does the paper specify all the training and test details (e.g., data splits, hyperparameters, how they were chosen, type of optimizer, etc.) necessary to understand the results?
    \item[] Answer: \answerYes{} 
    \item[] Justification: In our code and in the appendix of the paper we present the dataset splits used, the hyperparameters used. 
    \item[] Guidelines:
    \begin{itemize}
        \item The answer NA means that the paper does not include experiments.
        \item The experimental setting should be presented in the core of the paper to a level of detail that is necessary to appreciate the results and make sense of them.
        \item The full details can be provided either with the code, in appendix, or as supplemental material.
    \end{itemize}

\item {\bf Experiment Statistical Significance}
    \item[] Question: Does the paper report error bars suitably and correctly defined or other appropriate information about the statistical significance of the experiments?
    \item[] Answer: \answerYes{}{} 
    \item[] Justification: We perform latency and energy profiling across multiple evaluations, but surrogates which incorporate uncertainties directly. We also perform search using the baselines on our benchmark on multiple seeds. 
    \item[] Guidelines:
    \begin{itemize}
        \item The answer NA means that the paper does not include experiments.
        \item The authors should answer "Yes" if the results are accompanied by error bars, confidence intervals, or statistical significance tests, at least for the experiments that support the main claims of the paper.
        \item The factors of variability that the error bars are capturing should be clearly stated (for example, train/test split, initialization, random drawing of some parameter, or overall run with given experimental conditions).
        \item The method for calculating the error bars should be explained (closed form formula, call to a library function, bootstrap, etc.)
        \item The assumptions made should be given (e.g., Normally distributed errors).
        \item It should be clear whether the error bar is the standard deviation or the standard error of the mean.
        \item It is OK to report 1-sigma error bars, but one should state it. The authors should preferably report a 2-sigma error bar than state that they have a 96\% CI, if the hypothesis of Normality of errors is not verified.
        \item For asymmetric distributions, the authors should be careful not to show in tables or figures symmetric error bars that would yield results that are out of range (e.g. negative error rates).
        \item If error bars are reported in tables or plots, The authors should explain in the text how they were calculated and reference the corresponding figures or tables in the text.
    \end{itemize}

\item {\bf Experiments Compute Resources}
    \item[] Question: For each experiment, does the paper provide sufficient information on the computer resources (type of compute workers, memory, time of execution) needed to reproduce the experiments?
    \item[] Answer: \answerYes{} 
    \item[] Justification: Yes we provide details on the hardware used and a detailed table with search times in the appendix of the paper. 
    \item[] Guidelines:
    \begin{itemize}
        \item The answer NA means that the paper does not include experiments.
        \item The paper should indicate the type of compute workers CPU or GPU, internal cluster, or cloud provider, including relevant memory and storage.
        \item The paper should provide the amount of compute required for each of the individual experimental runs as well as estimate the total compute. 
        \item The paper should disclose whether the full research project required more compute than the experiments reported in the paper (e.g., preliminary or failed experiments that didn't make it into the paper). 
    \end{itemize}
    
\item {\bf Code Of Ethics}
    \item[] Question: Does the research conducted in the paper conform, in every respect, with the NeurIPS Code of Ethics \url{https://neurips.cc/public/EthicsGuidelines}?
    \item[] Answer: \answerYes{} 
    \item[] Justification: Our research conforms to the NeurIPS Code of Ethics and we make sure to preserve the anonymity of our submission. 
    \item[] Guidelines:
    \begin{itemize}
        \item The answer NA means that the authors have not reviewed the NeurIPS Code of Ethics.
        \item If the authors answer No, they should explain the special circumstances that require a deviation from the Code of Ethics.
        \item The authors should make sure to preserve anonymity (e.g., if there is a special consideration due to laws or regulations in their jurisdiction).
    \end{itemize}

\item {\bf Broader Impacts}
    \item[] Question: Does the paper discuss both potential positive societal impacts and negative societal impacts of the work performed?
    \item[] Answer: \answerYes{} 
    \item[] Justification: Yes, we discuss the potential positive and negative impacts of the work in the "Conclusions, Broader Impact and Implications" section \ref{sec:conclusion}. 
    \item[] Guidelines:
    \begin{itemize}
        \item The answer NA means that there is no societal impact of the work performed.
        \item If the authors answer NA or No, they should explain why their work has no societal impact or why the paper does not address societal impact.
        \item Examples of negative societal impacts include potential malicious or unintended uses (e.g., disinformation, generating fake profiles, surveillance), fairness considerations (e.g., deployment of technologies that could make decisions that unfairly impact specific groups), privacy considerations, and security considerations.
        \item The conference expects that many papers will be foundational research and not tied to particular applications, let alone deployments. However, if there is a direct path to any negative applications, the authors should point it out. For example, it is legitimate to point out that an improvement in the quality of generative models could be used to generate deepfakes for disinformation. On the other hand, it is not needed to point out that a generic algorithm for optimizing neural networks could enable people to train models that generate Deepfakes faster.
        \item The authors should consider possible harms that could arise when the technology is being used as intended and functioning correctly, harms that could arise when the technology is being used as intended but gives incorrect results, and harms following from (intentional or unintentional) misuse of the technology.
        \item If there are negative societal impacts, the authors could also discuss possible mitigation strategies (e.g., gated release of models, providing defenses in addition to attacks, mechanisms for monitoring misuse, mechanisms to monitor how a system learns from feedback over time, improving the efficiency and accessibility of ML).
    \end{itemize}
    
\item {\bf Safeguards}
    \item[] Question: Does the paper describe safeguards that have been put in place for responsible release of data or models that have a high risk for misuse (e.g., pretrained language models, image generators, or scraped datasets)?
    \item[] Answer: \answerYes{}{} 
    \item[] Justification: We use and release datasets which are open-source and released under the Apache 2.0 license. These models are trained on open-source datasets and we intend the use of these models only for research purposes. 
    \item[] Guidelines:
    \begin{itemize}
        \item The answer NA means that the paper poses no such risks.
        \item Released models that have a high risk for misuse or dual-use should be released with necessary safeguards to allow for controlled use of the model, for example by requiring that users adhere to usage guidelines or restrictions to access the model or implementing safety filters. 
        \item Datasets that have been scraped from the Internet could pose safety risks. The authors should describe how they avoided releasing unsafe images.
        \item We recognize that providing effective safeguards is challenging, and many papers do not require this, but we encourage authors to take this into account and make a best faith effort.
    \end{itemize}

\item {\bf Licenses for existing assets}
    \item[] Question: Are the creators or original owners of assets (e.g., code, data, models), used in the paper, properly credited and are the license and terms of use explicitly mentioned and properly respected?
    \item[] Answer: \answerYes{}{} 
    \item[] Justification: We release our own code and cite appropriately in cases where we use code and pretrained models from other repositories.  
    \item[] Guidelines:
    \begin{itemize}
        \item The answer NA means that the paper does not use existing assets.
        \item The authors should cite the original paper that produced the code package or dataset.
        \item The authors should state which version of the asset is used and, if possible, include a URL.
        \item The name of the license (e.g., CC-BY 4.0) should be included for each asset.
        \item For scraped data from a particular source (e.g., website), the copyright and terms of service of that source should be provided.
        \item If assets are released, the license, copyright information, and terms of use in the package should be provided. For popular datasets, \url{paperswithcode.com/datasets} has curated licenses for some datasets. Their licensing guide can help determine the license of a dataset.
        \item For existing datasets that are re-packaged, both the original license and the license of the derived asset (if it has changed) should be provided.
        \item If this information is not available online, the authors are encouraged to reach out to the asset's creators.
    \end{itemize}

\item {\bf New Assets}
    \item[] Question: Are new assets introduced in the paper well documented and is the documentation provided alongside the assets?
    \item[] Answer: \answerYes{} 
    \item[] Justification: Yes we release our code and models under the Apache 2.0 license and our code and models are well documented. 
    \item[] Guidelines:
    \begin{itemize}
        \item The answer NA means that the paper does not release new assets.
        \item Researchers should communicate the details of the dataset/code/model as part of their submissions via structured templates. This includes details about training, license, limitations, etc. 
        \item The paper should discuss whether and how consent was obtained from people whose asset is used.
        \item At submission time, remember to anonymize your assets (if applicable). You can either create an anonymized URL or include an anonymized zip file.
    \end{itemize}

\item {\bf Crowdsourcing and Research with Human Subjects}
    \item[] Question: For crowdsourcing experiments and research with human subjects, does the paper include the full text of instructions given to participants and screenshots, if applicable, as well as details about compensation (if any)? 
    \item[] Answer: \answerNA{} 
    \item[] Justification: Our research is not dependent on/based on human subjects. 
    \item[] Guidelines:
    \begin{itemize}
        \item The answer NA means that the paper does not involve crowdsourcing nor research with human subjects.
        \item Including this information in the supplemental material is fine, but if the main contribution of the paper involves human subjects, then as much detail as possible should be included in the main paper. 
        \item According to the NeurIPS Code of Ethics, workers involved in data collection, curation, or other labor should be paid at least the minimum wage in the country of the data collector. 
    \end{itemize}

\item {\bf Institutional Review Board (IRB) Approvals or Equivalent for Research with Human Subjects}
    \item[] Question: Does the paper describe potential risks incurred by study participants, whether such risks were disclosed to the subjects, and whether Institutional Review Board (IRB) approvals (or an equivalent approval/review based on the requirements of your country or institution) were obtained?
    \item[] Answer: \answerNA{} 
    \item[] Justification:  We do not collect data on human subjects for our research. 
    \item[] Guidelines:
    \begin{itemize}
        \item The answer NA means that the paper does not involve crowdsourcing nor research with human subjects.
        \item Depending on the country in which research is conducted, IRB approval (or equivalent) may be required for any human subjects research. If you obtained IRB approval, you should clearly state this in the paper. 
        \item We recognize that the procedures for this may vary significantly between institutions and locations, and we expect authors to adhere to the NeurIPS Code of Ethics and the guidelines for their institution. 
        \item For initial submissions, do not include any information that would break anonymity (if applicable), such as the institution conducting the review.
    \end{itemize}

\end{enumerate}

\end{document}